\documentclass[review,11pt]{elsarticle}

\usepackage{amsmath}
\usepackage{bbm}
\usepackage{amsfonts}
\usepackage{amssymb}
\usepackage{mathdots}
\usepackage{mathdots}
\usepackage{lipsum}
\usepackage[utf8]{inputenc}
\usepackage{enumerate}
\usepackage{amssymb}
\usepackage{centernot}
\usepackage[mathscr]{euscript}
\usepackage{url}
\usepackage{hyphenat} % \usepackage[none]{hyphenat}
\usepackage{bm}
\usepackage{verbatim}
\usepackage{accents}
\usepackage{graphicx}
\usepackage{subcaption}
\usepackage{listings}
\usepackage{xcolor}
\usepackage{graphicx}
\usepackage{grffile}
\usepackage{algorithm,algorithmicx}
\usepackage{algpseudocode}
\algnewcommand\algorithmicinput{\textbf{Input: }}
\algnewcommand\algorithmicoutput{\textbf{Output: }}
\usepackage{afterpage}
\usepackage{epstopdf}
\usepackage{widetext}
\usepackage{cuted}
\everymath{\displaystyle}

\DeclareMathOperator*{\argmax}{argmax}

\newcommand\blue[1]{\textcolor{blue}{#1}}
\newcommand\E[1]{{\mathbb{E}\left[#1\right]}}

\usepackage{hyperref}

\begin{document}

\begin{frontmatter}

\title{aphBO-2GP-3B: A budgeted asynchronous parallel multi-acquisition functions for constrained Bayesian optimization on high-performing computing architecture}
% \tnotetext[mytitlenote]{Fully documented templates are available in the elsarticle package on \href{http://www.ctan.org/tex-archive/macros/latex/contrib/elsarticle}{CTAN}.}

% \address{Radarweg 29, Amsterdam}
% \fntext[myfootnote]{Since 1880.}

\author[add4]{Anh Tran\corref{mycorrespondingauthor}}
\author[add4]{Mike Eldred}
\author[add4]{Tim Wildey}
\author[add5]{Scott McCann}
\author[add3]{Jing Sun}
% \cortext[mycorrespondingauthor]{Corresponding author}
% \ead[url]{George Woodruff School of Mechanical Engineering, Georgia Institute of Technology, Atlanta, GA 30332}
\author[add2]{Robert J. Visintainer}
% \author[add2]{Krishnan V. Pagalthivarthi}
% \author[add1]{Yan Wang\corref{mycorrespondingauthor}}
\cortext[mycorrespondingauthor]{Corresponding author: anhtran@sandia.gov}

% \address[add1]{George Woodruff School of Mechanical Engineering, Georgia Institute of Technology, Atlanta, GA 30332}
\address[add4]{Optimization and Uncertainty Quantification, Sandia National Laboratories, Albuquerque, NM 87123}
\address[add5]{Xilinx Inc., 2100 All Programmable Dr., San Jose, CA 95124}
\address[add3]{Medpace Inc., Cincinnati, OH 45227}
\address[add2]{GIW Industries Inc., Grovetown, GA 30813}

\begin{abstract}
% % \textit{\lipsum[1]}
High-fidelity complex engineering simulations are often predictive, but also computationally expensive and often require substantial computational efforts. 
The mitigation of computational burden is usually enabled through parallelism in high-performance cluster (HPC) architecture. 
Optimization problems associated with these applications is a challenging problem due to the high computational cost of the high-fidelity simulations. 
In this paper, an asynchronous parallel constrained Bayesian optimization method is proposed to efficiently solve the computationally-expensive simulation-based optimization problems on the HPC platform, with a budgeted computational resource, where the maximum number of simulations is a constant. 
The advantage of this method are three-fold. 
First, the efficiency of the Bayesian optimization is improved, where multiple input locations are evaluated massively parallel in an asynchronous manner to accelerate the optimization convergence with respect to physical runtime. 
This efficiency feature is further improved so that when each of the inputs is finished, another input is queried without waiting for the whole batch to complete. 
Second, the proposed method can handle both known and unknown constraints. 
% The known constraints are formulated as inequality constraints, which are incorporated by penalizing the acquisition function. 
% The unknown constraints, which cannot be accessed without evaluating the objective function, are coupled to the aphBO-2GP-3B framework using a binary classifier to distinguish feasible and infeasible regions. 
% One of the most common unknown constraints is when the objective function is missing due to ill-conditioned problems, mesh irregularities. 
% The proposed aphBO-2GP-3B framework efficiently solves the missing problem by interpolating on the objective GP and improving the binary classifier. 
% , by incorporating another binary probabilistic classifier to learn and distinguish between feasible and infeasible regions. 
Third, the proposed method samples several acquisition functions based on their rewards using a modified GP-Hedge scheme. 
The proposed framework is termed aphBO-2GP-3B, which means \textbf{a}synchronous \textbf{p}arallel \textbf{h}edge \textbf{B}ayesian \textbf{o}ptimization with two Gaussian processes and three batches. 
The numerical performance of the proposed framework aphBO-2GP-3B is comprehensively benchmarked using 16 numerical examples, compared against other 6 parallel Bayesian optimization variants and 1 parallel Monte Carlo as a baseline, and demonstrated using two real-world high-fidelity expensive industrial applications. The first engineering application is based on finite element analysis (FEA) and the second one is based on computational fluid dynamics (CFD) simulations. 
\end{abstract}


\end{frontmatter}

% \linenumbers

\textit{In memory of John Michael Furlan.}

\section{Introduction}

% - what is the problem you are solving
Bayesian optimization (BO) is a well-known effective surrogate-based optimization methodology for high-fidelity complex simulations \cite{jones1998efficient}. 
The efficiency is mainly achieved through a Gaussian process (GP) surrogate model to approximate the response surface, as the optimization process advances. 
The GP model incorporates search history and updates sequentially as soon as new information is available. 
The traditional BO approach relies on an underlying Gaussian process (GP) to model the response surface and utilizes an acquisition function to locate the most valuable input point for the next sampling location, simply by maximizing the acquisition function. 
However, like any other sequential optimization method, the traditional sequential BO approach suffers from the computational cost of the simulation, in which high-fidelity simulations typically correspond to high computational cost. 
Furthermore, in practical settings, the simulation does not always return a definite output. 
Such situation can occur in any real-world engineering application if the mesh is irregular, the solver is ill-conditioned, or the simulation is hung indefinitely. 
These problems, which are common but have not been adequately addressed, pose a challenge for any applications of BO on two fronts: parallel optimization and optimization under unknown constraints. 

% - why is this problem important 
Direct applications of BO are usually limited in terms of efficiency and robustness. 
Enabling parallelism in BO is critical in improving efficiency of the traditional BO approach, because better optimization solution can be obtained within a shorter amount of physical time. 
The parallelism in optimization further enhances the parallelism of complex simulations on a high-performance computing (HPC) platform, because it is independent of the parallelism of the simulation packages, thus the speedup is improved linearly in theory. 
The main reason is that with more samples, the underlying GP can converge faster in approximating the response and this can accelerate the optimization process. 
Na\"{i}ve applications of the traditional BO method often faces a challenge when the simulation does not return the output, because the underlying GP model requires both inputs and outputs to construct the response surface. 
% - what people have done so far to solve this problem
% A straightforward extension in parallel BO is the sequential batch parallelization, which samples concurrently a fixed number of samples in a batch, where each batch is executed sequentially. 
% Extension of the traditional BO approach is reviewed more carefully in Section \ref{sec:LiteratureReview}. 
Due to the broad scope of BO methods, in Section \ref{sec:LiteratureReview}, the literature review for the traditional BO method and its variants are discussed, including a brief introduction to BO, application to constrained problems, and its parallel extension. 
% In this section, the literature review related to BO method is discussed. 
% Section \ref{subsec:GP} provides a brief formulation on the GP formulation. 
% Section \ref{subsec:acquisitionFunction} describes several commonly used acquisition functions in BO method. 
% Section \ref{subsec:knownConstraints} and Section \ref{subsec:unknownConstraints} discuss currently existing approaches in BO literature to handle known and unknown constraints, respectively. 
% Section \ref{subsec:parallelization} presents the batch parallel approach to parallelize BO.
% In particular, Section \ref{subsubsec:sequentialBatchParallel} and Section \ref{subsubsec:asynchronousBatchParallel} describe the batch-sequential and batch asynchronous parallelization approaches, respectively. 

% - what approach do you take to solve the problem (an overview)
In this work, the aphBO-2GP-3B framework is developed as an extension to our previous pBO-2GP-3B framework \cite{tran2019pbo}, to efficiently solve a simulation-based constrained optimization on HPC platforms, where the maximum number of simulations is fixed as a constant. 
Similar to the pBO-2GP-3B, the aphBO-2GP-3B framework relies on three batches, where priorities are assigned differently depending on the batch, with more focus emphasized on the batch that optimizes the problem. 
Supporting the aphBO-2GP-3B framework are two distinct GP models, where the first GP corresponds to the objective function, whereas the second GP corresponds to the unknown constraints of the simulations. 
The known constraints are formulated as inequality constraints, which are then penalized directly into the acquisition function when searching for the next sampling location. 
Compared to the pBO-2GP-3B approach, the proposed aphBO-2GP-3B framework supersedes in two aspects. 
First, the parallelization is implemented in an asynchronous manner, which completely relaxes the synchronous batch-sequential parallel conditions, to further improve the efficiency of the parallel BO method. 
Second, multiple acquisition functions are considered in the main batch and sampled based on its reward according to the GP-Hedge scheme \cite{hoffman2011portfolio}.

% - why is your approach unique / what is your technical contribution / what is new
The aphBO-2GP-3B framework presents a practical implementation of parallel constrained BO method that is highly applicable for many computationally expensive high-fidelity engineering simulations on the HPC platform. 
The advantage of the current approach is the improvement of both efficiency and robustness of the traditional BO method. 
The efficiency of the aphBO-2GP-3B is enabled through several approaches within the framework, including sampling multiple acquisition functions and the implementation of highly asynchronous parallel feature. 
The robustness of the aphBO-2GP-3B is achieved through coupling with a probabilistic binary classifier to distinguish feasible and infeasible regions. 
In this paper, we consider a general constrained optimization problem with focus given to problems with computationally expensive high-fidelity engineering simulation. 

% paper outline

The remainder of the paper is organized as follows. 
Section \ref{sec:LiteratureReview} provides the literature review for GP and BO with different relevant extensions, including known constrained, unknown constrained, synchronous batch-sequential parallel, and asynchronous parallel. 
In Section \ref{sec:aphBO-2GP-3B}, the proposed aphBO-2GP-3B framework is described and discussed in details. 
Section \ref{sec:FCBGA} presents the first application of the aphBO-2GP-3B framework for thermo-mechanical flip-chip design optimization based on finite element analysis (FEA) simulation. 
Section \ref{sec:3dCasingWear} presents the second application of the aphBO-2GP-3B framework for slurry pump casing design optimization using multiphase 3D computational fluid dynamics (CFD) simulation. 
Section \ref{sec:Discussion} discusses and Section \ref{sec:Conclusion} concludes the paper, respectively.

\section{Related works}
\label{sec:LiteratureReview}

In this section, the literature review related to BO method is discussed. 
Section \ref{subsec:GP} provides a brief formulation on the GP formulation. 
Section \ref{subsec:acquisitionFunction} describes several commonly used acquisition functions in BO method. 
Section \ref{subsec:knownConstraints} and Section \ref{subsec:unknownConstraints} discuss currently existing approaches in BO literature to handle known and unknown constraints, respectively. 
Section \ref{subsec:parallelization} presents the batch parallel approach to parallelize BO.
In particular, Section \ref{subsubsec:sequentialBatchParallel} and Section \ref{subsubsec:asynchronousBatchParallel} describe the synchronous batch-sequential parallel and asynchronous parallel approaches, respectively.

\subsection{Gaussian process \& Bayesian optimization}
\label{subsec:GP}

% \redbf{paraphrase this whole subsection}

Comprehensive and critical review studies are provided by Brochu et al. \cite{brochu2010tutorial}, Shahriari et al. \cite{shahriari2016taking}, Frazier \cite{frazier2018tutorial}, and Jones et al. \cite{jones1998efficient} for BO method and its variants. 
A brief introduction to GP is discussed, as the GP surrogate model plays a critical role in BO methods. 
We adopt the notation from Shahriari et al. \cite{shahriari2016taking} and Tran et al. \cite{tran2020smfbo2cogp,tran2020weargp,tran2019sbfbo2cogp,tran2019pbo,tran2019constrained,tran2018efficient,tran2018weargp} for its clarity and consistency. In this formulation, we treat the optimization problem in the maximization settings,
\begin{equation}
\argmax_{x\in\mathcal{X}} f(x)
\end{equation}
subject to a set of nonlinear inequalities constraints
\begin{equation}
\lambda_j (\bm{x}) \leq 0, \quad 1 \leq j \leq J.
\end{equation}

Assume that $f$ is a function of $\bm{x}$, where $\bm{x} \in \mathcal{X}$ is the $d$-dimensional input. A $\mathcal{GP}(\mu_0,k)$ is a nonparametric model over functions $f$,
which is fully characterized by the prior mean functions $\mu_0(x): \mathcal{X} \mapsto \mathbb{R}$ and the positive-definite kernel, or covariance function $k:\mathcal{X} \times \mathcal{X} \mapsto \mathbb{R}$. 
In GP regression, it is assumed that $\bm{f} = f_{1:n}$ is jointly Gaussian, and the observation $y$ is normally distributed given $f$, leading to 
\begin{equation}
\label{eq:prior}
\bm{f} | \bm{x} \sim \mathcal{N}(\bm{m},\bm{K}),
\end{equation}
\begin{equation}
\bm{y} | \bm{f},\sigma^2 \sim \mathcal{N}(\bm{f},\sigma^2 \bm{I}),
\end{equation}
where $m_i := \mu(\bm{x}_i)$, and $K_{i,j} := k(\bm{x}_i,\bm{x_j})$. Equation \ref{eq:prior} describes the prior distribution induced by the GP.

The covariance kernel $k(\cdot, \cdot)$ is a choice of modeling covariance between inputs, depending on the smoothness assumption of $f$. The family of Mat{\'e}rn kernels is arguably one of the most popular choices for kernels, offering a broad class for stationary kernels which are controlled by a smoothness parameter $\nu>0$ (cf. Section 4.2, \cite{rasmussen2006gaussian}), including the square-exponential ($\nu \to \infty$) and exponential $(\nu = 1/2)$ kernels. The Mat{\'e}rn kernels are described as
\begin{equation}
\bm{K}_{i,j} = k(\bm{x}_i, \bm{x}_j) = \theta_0^2 \frac{2^{1-\nu}}{\Gamma(\nu)} (\sqrt{2\nu} r)^{\nu} K_{\nu}(\sqrt{2\nu}r), 
\end{equation}
where $K_\nu$ is a modified Bessel fuction of the second kind and order $\nu$. 
Common kernels for GP include~\cite{shahriari2016taking}
\begin{itemize}
\item $\nu = 1/2: k_{\text{Mat{\'e}rn}1} (\bm{x}, \bm{x'}) = \theta_0^2 \exp{(-r)}$ (also known as exponential kernel), 
\item $\nu = 3/2: k_{\text{Mat{\'e}rn}3} (\bm{x}, \bm{x'}) = \theta_0^2 \exp{(-\sqrt{3}r)} (1+\sqrt{3} r)$, 
\item $\nu = 5/2: k_{\text{Mat{\'e}rn}5} (\bm{x}, \bm{x'}) = \theta_0^2 \exp{(-\sqrt{5}r)} \left( 1 + \sqrt{5}r + \frac{5}{3}r^2 \right)$, 
\item $\nu \to \infty: k_{\text{sq-exp}} (\bm{x}, \bm{x'}) = \theta_0^2 \exp{\left(-\frac{r^2}{2} \right)}$ (also known as square exponential or automatic relevance determination kernel),
\end{itemize}
where $r^2 = (\bm{x} - \bm{x}')\bm{\Lambda}(\bm{x} - \bm{x}')$, and $\bm{\Lambda}$ is a diagonal matrix of $d$ squared length scale $\theta_i$. 

% One of the most widely used kernels is the squared exponential kernel (also known as automatic relevance determination kernel), which can be described as 
% \begin{equation}
% \bm{K}_{i,j} = k(\bm{x}_i, \bm{x}_j) = \theta_0^2 \exp \left( - \frac{r^2}{2} \right),
% \end{equation}
% where $r^2 = (\bm{x} - \bm{x}')\bm{\Lambda}(\bm{x} - \bm{x}')$, and $\bm{\Lambda}$ is a diagonal matrix of $d$ squared length scale $\theta_i$. 
% The exponential kernel is also commonly used, which can be described as
% \begin{equation}
% \bm{K}_{i,j} = k(\bm{x}_i, \bm{x}_j) = \theta_0^2 \exp \left( - r \right). 
% \end{equation}
% It turns out that they are special cases of 
% Mat{\'e}rn-3/2 $(\nu = 3/2)$ and Mat{\'e}rn-5/2 $(\nu = 5/2)$ are also used widely in the literature.

Let the dataset $\mathcal{D}={(\bm{x}_i,y_i)}_{i=1}^n$ denote a collection of $n$ noisy observations and $\bm{x}$ denote an arbitrary input of dimension $d$. 
Under the formulation of GP, given the dataset $\mathcal{D}_n$, the prediction for an unknown arbitrary point is characterized by the posterior Gaussian distribution, which can be described by the posterior mean and posterior variance functions, respectively as 
\begin{equation}
\mu_n(\bm{x}) = \mu_0(\bm{x}) + \bm{k}(\bm{x})^T (\bm{K} + \sigma^2 \bm{I})^{-1} (\bm{y} - \bm{m}),
\end{equation}
and
\begin{equation}
\sigma_n^2 = k(\bm{x}, \bm{x}) - \bm{k}(\bm{x})^T (\bm{K}  + \sigma^2 \bm{I})^{-1} \bm{k}(\bm{x}),
\end{equation}
where $\bm{k}(\bm{x})$ is the covariance vector between the query point $\bm{x}$ and $\bm{x}_{1:n}$. 
% Estimating hyper-parameters $\theta$ of the objective GP involves optimization of the likelihood and computation of the determinant and the inverse of the covariance matrix, resulting in $\mathcal{O}(n^3)$ algorithmic complexity. 
% One of the significant drawbacks for the classical BO is the computational bottleneck in optimizing the maximum likelihood function as the number of observations $n$ increases. 
The main drawback of GP formulation is its scalability $\mathcal{O}(n^3)$ that originates from the computation of the inverse of the covariance matrix $\bm{K}$. 
Connection from GP to convolution neural network has been proposed where it is proved to be theoretically equivalent to single layer with infinite width \cite{lee2017deep} or infinite convolutional filters \cite{garriga2018deep}.

\subsection{Acquisition function}
\label{subsec:acquisitionFunction}

In the traditional BO method, which is sequential, the GP model is constructed for the objective function, and the next sampling location is determined by maximizing the acquisition function based on the constructed GP. 
This acquisition function is evaluated based on the underlying GP surrogate model, thus converting the cost of evaluating the real simulation to the cost of evaluating on the GP model. 
Compared between the two, the later is much more computationally appealing because it is semi-analytical. 
The acquisition function must balance between the exploitation and exploration flavors of the BO method. 
Too much exploitation would drive the numerical solution to a local minima, whereas too much exploration would make BO an inefficient optimization method. 
We review three main acquisition functions that are typically used in the literature: the probability of improvement (PI), the expected improvement (EI), and the upper-confident bounds (UCB). 

Denote $\mu(\bm{x})$, $\sigma^2(\bm{x})$, and $\theta(\bm{x})$ as the posterior mean, the posterior variance, and the hyper-parameters of the objective GP model, respectively. 
$\theta(\bm{x})$ is obtained by maximizing the log likelihood estimation over a plausible chosen range. 
Let $\phi(\cdot)$ and $\Phi(\cdot)$ be the standard normal probability distribution function and cumulative distribution function, respectively, and $\bm{x}_{\text{best}} = \argmax_{1\leq i \leq n} f(\bm{x}_i)$ be the best-so-far sample. 
Rigorously, the acquisition function should be written as $a(\bm{x};\{\bm{x}_i,y_i \}_{i=1}^n,\theta)$, but for the sake of simplicity, we drop the dependence on the observations and simply write as $a(\bm{x})$ and $\mathbb{E}(\cdot)$ is implicitly understood as $\mathbb{E}_{y \sim p(y | \mathcal{D}_n, \bm{x})} (\cdot)$ unless specified otherwise.

The PI acquisition function \cite{kushner1964new} is defined as 
  \begin{equation}
  a_{\text{PI}}(\bm{x}) = \text{Pr}(y > f(\bm{x}_\text{best})) = \E{\mathbbm{1}_{y>f(\bm{x}_\text{best})}} = \Phi(\gamma(\bm{x})),
  \end{equation}
where
  \begin{equation}
  \label{eq:normalizedZ}
  \gamma(\bm{x}) = \frac{\mu(\bm{x}) - f(\bm{x}_{\text{best}})}{\sigma(\bm{x})},
  \end{equation}
indicates the deviation away from the best sample. 
The PI acquisition function is constructed based on the idea of binary utility function, where a unit reward is received if a new best-so-far sample is found, and zero otherwise. 
% , i.e.
% \begin{equation}
% a_{\text{PI}}(\bm{x}) 
% \end{equation}

The EI acquisition function \cite{mockus1975bayesian,mockus1982bayesian,bull2011convergence,snoek2012practical} is defined as 
\begin{equation}
a_{\text{EI}}(\bm{x}) = \sigma(\bm{x}) \cdot (\gamma(\bm{x}) \Phi(\gamma(\bm{x})) + \phi(\gamma(\bm{x})).
\end{equation}
The EI acquisition is constructed based on an improvement utility function, where the reward is the relative difference if a new best-so-far sample is found, and zero otherwise. 
A closely related generalization of the EI acquisition function, called knowledge-gradient (KG) acquisition function, has been suggested in \cite{scott2011correlated}. Under the assumptions of noise-free and the sampling function is restricted, the EI acquisition function is recovered from the KG acquisition function. 
If the EI acquisition function is rewritten as 
\begin{equation}
a_\text{EI}(\bm{x}) = \E{\max\left(y, f(\bm{x}_\text{best})\right) - f(\bm{x}_\text{best})} = \E{\max(y - f(\bm{x}_\text{best}, 0)} = \E{(y - f(\bm{x}_\text{best})^+},
\end{equation} 
then the KG acquisition function is expressed as 
\begin{equation}
a_\text{KG}(\bm{x}) = \E{ \max \mu_{n+1}(\bm{x}) | \bm{x}_{n+1} = \bm{x}} - \max(\mu_n(\bm{x}))
\end{equation}
for one-step look-ahead acquisition function. 

The UCB acquisition function \cite{auer2002using,srinivas2009gaussian,srinivas2012information} is defined as
\begin{equation}
a_{\text{UCB}}(\bm{x}) = \mu(\bm{x}) + \kappa \sigma(\bm{x}),
\end{equation}
where $\kappa$ is a hyper-parameter describing the acquisition exploitation-exploration balance. 
% There are at least three ways to obtain the $\kappa$ parameter. 
Here, we adopt the $\kappa$ computation from Daniel et al. \cite{daniel2014active}, where
\begin{equation}
\kappa = \sqrt{\nu \gamma_n},\quad \nu = 1, \quad \gamma_n = 2\log{\left(\frac{n^{d/2 + 2}\pi^2}{3\delta} \right)},
\end{equation}
and $d$ is the dimensionality of the problem, and $\delta \in (0,1)$ \cite{srinivas2012information}. 

Another type acquisition function is entropy-based, such as GP-PES \cite{hernandez2014predictive,hernandez2015predictive,hernandez2016predictive,hernandez2016general}, GP-ES \cite{hennig2012entropy}, GP-MES \cite{wang2017max}. Since GP is collectively a distribution of functions, the distribution of the global optimum $\bm{x}^*$ can be estimated as well from sampling the GP posterior. The predictive-entropy-search (PES) can be written as
\begin{equation}
% \begin{array}{lll}
a_\text{PES}(\bm{x}) = H[p(\bm{x}^* | \mathcal{D}_n)] - \mathbb{E} \left[ H [p(\bm{x}^* | \mathcal{D} \cup \{ (\bm{x},y) \}) ]  \right] 
= H[p(y | \mathcal{D}_n, \bm{x})] - \mathbb{E}_{p(\bm{x}^* | \mathcal{D}_n)} [H(p(y | \mathcal{D}_n, \bm{x}, \bm{x}^*))] 
% \end{array}
\end{equation}
Wang et al. \cite{wang2017max} proposed GP-MES acquisition function, which effectively suggests sampling $y^*$ from the 1-dimensional output space $\mathbb{R}$ using Gumbel distribution instead of $\bm{x}$ in the high-dimensional input space $\mathcal{X}$, i.e.
\begin{equation}
a_\text{MES}(\bm{x}) = H[p(y | \mathcal{D}_n, \bm{x})] - \mathbb{E} [H(p(y | \mathcal{D}_n, \bm{x}, y^*))],
\end{equation}

Last but not least, since the $f(\bm{x})$ is Gaussian, the differential entropy can be simplified to a function of posterior variance $\sigma^2(\bm{x})$, i.e.
\begin{equation}
H[p(y | \mathcal{D}_n, \bm{x})] = 0.5 \log\left[ 2\pi e (\sigma^2(\bm{x}) + \sigma^2) \right],
\end{equation} 
and thus pure exploration search which targets $\bm{x}$ with maximum posterior variance $\sigma^2 (\bm{x})$, i.e. $a_\text{PE}(\bm{x}) = \sigma^2(\bm{x})$ is theoretically identical with the maximum differential entropy. 
Interestingly, Wilson et al. \cite{wilson2018maximizing} proposed a reparameterization of most commonly used acquisition functions in terms of deep learning convolution kernels. 
Connection from GP to convolution neural network has been proposed where it is proved to be theoretically equivalent to single layer with infinite width \cite{lee2017deep} or infinite convolutional filters \cite{garriga2018deep}.

\subsection{Constraints}

Digabel and Wild \cite{digabel2015taxonomy} proposed a QRAK taxonomy to classify constrained optimization problems. 
Conceptually speaking, there are two types of constraints: known and unknown. 
% The main difference between known constraints and unknown constraints are as follows. 
The known constraints can be imposed beforehand without running the simulation, and thus can be handled using a separate constraint evaluator. Usually, the known constraint evaluator is computationally cheap compared to the simulation. 
On the contrary, the unknown constraints cannot be accessed without running the simulations, and thus are only evaluated at the same time when the simulation is performed. The unknown constraints are more expensive to obtain and require more careful attention. 
A review and comparison study is performed by Parr et al. \cite{parr2012infill} for different schemes to handle constraints using both synthetic and real-world applications. 
Many previous works discussed below in the literature prefer to couple constraint satisfaction problems with the EI acquisition due to its consistent numerical performance.

\subsubsection{Known constraints}
\label{subsec:knownConstraints}

Known-constrained optimization problems are well studied in the literature and mostly formulated as a set of inequality constraints. 
Gramacy and Lee \cite{gramacy2010optimization} introduced an acquisition function based on the integrated expected conditional improvement by integrating the EI acquisition function and the probabilities of satisfying constraints over the whole space. 
Gramacy et al. \cite{gramacy2016modeling} combined the EI acquisition function and augmented Lagrangian framework to convert a constrained to an unconstrained optimization problem. 
Zhou et al. \cite{zhou2017robust} also converted a constrained multi-fidelity optimization problem to an unconstrained multi-fidelity optimization problem by adding penalty function accounting for the constraint violations. 
Schonlau et al. \cite{schonlau1998global} proposed a constrained EI acquisition function by multiplying the EI acquisition with the probabilities to satisfy for each individual constraint. 
Gardner et al. \cite{gardner2014bayesian} penalized the known constraints directly by assigning zero improvement to all infeasible points through a binary feasibility indicator function. 
Letham et al. \cite{letham2018constrained} extended the methods of Gardner et al. \cite{gardner2014bayesian} and Gelbart et al. \cite{gelbart2014bayesian} in a noisy experiment settings at Facebook.

\subsubsection{Unknown constraints}
\label{subsec:unknownConstraints}

Unknown-constrained problems are harder and did not receive enough research attention until recently. 
Gelbart et al. \cite{gelbart2014bayesian} proposed to threshold the probability of constraint satisfaction, where the threshold is a user-defined parameter. 
Hern\'{a}ndez-Lobato et al. \cite{hernandez2015predictive} proposed an alternative acquisition function based on GP-PES \cite{hernandez2014predictive,hernandez2016general} for constrained problems by choosing the sampling location that maximizes the expected reduction of differential entropy of the posterior. 
Basudhar et al. \cite{basudhar2012constrained} used support vector machines to calculate the probability of satisfying unknown constraints and combined with the EI acquisition function. 
Sacher et al. \cite{sacher2018classification} extended the method of Basudhar et al. \cite{basudhar2012constrained} and combined with augmented Lagrangian approach for handling both known and unknown constraints. 
Lee et al. \cite{lee2011optimization} coupled the random forest classifier with the EI acquisition function by multiplying the EI acquisition function with the feasible probability.

\subsection{Parallelization}
\label{subsec:parallelization}

To accelerate the optimization process for computationally expensive applications, there are mainly two approaches. 
The first approach is to parallelize the applications on the HPC platform, by exploiting multi-core architecture and decomposing the domain accordingly. The limitation of this approach is that in most applications, there is always a theoretical speedup that can only be achieved theoretically by Amdahl's law \cite{hill2008amdahl}, posing a computational threshold for accelerating the process. 
Furthermore, this approach is not efficient due to the diminishing return in increasing the number of processors used on the HPC platform. 
The second approach is to parallelize the optimizer itself by concurrently running at multiple sampling locations. For the BO method, this is typically achieved by batch parallel approach, which a batch of sampling points are evaluated, as opposed to one single sampling location. 
To search for new sampling locations, one can rely on the Gaussian posterior of the outputs and condition the parallel acquisition based on the posterior.

\subsubsection{Sequential batch parallel}
\label{subsubsec:sequentialBatchParallel}

Similar to the constrained BO problem, the EI acquisition function has been used extensively in the literature. 
Ginsbourger et al. \cite{ginsbourger2008multi,ginsbourger2010kriging,chevalier2013fast}, Roustant et al. \cite{roustant2012dicekriging} proposed the q-EI framework to select multiple points based on the EI acquisition function. 
Analytical formula is available for q-EI with the batch size $q$ of two, but as $q$ grows, it requires a numerical computation of high-dimensional integration, which can be achieved by MC. 
Marmin et al. \cite{marmin2015differentiating,marmin2016efficient} proposed a analytical simplification for q-EI to avoid high-dimensional integration in large batch based on gradient-based ascent algorithms.
Wang et al. \cite{wang2016parallel} also employed a gradient-based approach but relying on infinitesimal perturbation analysis to construct a stochastic gradient estimator to simplify the high-dimensional integration in the original q-EI framework. 
Wang et al. \cite{wang2016parallel} and Wu and Frazier \cite{wu2016parallel} (q-KG) proposed a parallel version for KG acquisition function \cite{frazier2009knowledge}. 
Rontsis et al. \cite{rontsis2017distributionally} proposed another acquisition function GP-OEI to reduce the computational burden of q-EI on high-dimensional space, where the lower and upper bounds are computationally tractable in high-dimensional space and showed its numerical robustness over q-EI. 
Snoek et al. \cite{snoek2012practical} proposed an integrated EI acquisition function based on Monte Carlo (MC) sampling, called GP-EI-MCMC to construct a batch. 
Azimi et al. \cite{azimi2010batch,azimi2012batch,azimi2012hybrid} proposed a simulation matching scheme GP-SM \cite{azimi2010batch}, coordinated matching scheme GP-BCM \cite{azimi2012batch}, and a hybrid sequential-parallel \cite{azimi2012hybrid} to recalibrate the batch behavior to the sequential BO after a number of steps. 
Shah and Ghahramani \cite{shah2015parallel} proposed a batch parallel GP-PES approach based on entropy acquisition function. 
Desautels et al. \cite{desautels2014parallelizing} proposed a batch parallel hallucination scheme, called GP-BUCB-ACUB, that combines both UCB acquisition function and kriging believer heuristic in Ginsbourger et al \cite{ginsbourger2008multi,ginsbourger2010kriging}. 
Contal et al. \cite{contal2013parallel} extended the method of Desautels et al. \cite{desautels2014parallelizing} by promoting exploration within a batch while leaving only one sampling point for exploitation. 
Kathuria et al. \cite{kathuria2016batched} and Wang et al. \cite{wang2017batched} employed the determinantal point process to propose a batch selection policy GP-DPP, and proved the expected regret bound of DPP-SAMPLE is less than the regret bound of GP-UCB-PE \cite{contal2013parallel}. 
Daxberger and Low \cite{daxberger2017distributed} proposed a distributed batch GP-UCB, called DP-GP-UCB  to jointly optimize a batch of inputs, by formulating the batch input selection as a multi-agent distributed constraint optimization problem, as opposed to selecting the inputs of a batch one at a time, while preserving the scalability in the batch size. 
Gonz{\'a}lez et al. \cite{gonzalez2016batch} proposed a local penalization GP-LP method to penalize the acquisition function locally with an estimated Lipschitz constant to construct a batch. 
% Kathuria et al. \cite{kathuria2016batched} showed that GP-UCB-PE is a special case as DPP-MAX, where the maximization rule is done via a greedy selection rule, and suggested that GP-PPES \cite{shah2015parallel} performs better than GP-BUCB \cite{desautels2014parallelizing} and GP-UCB-PE \cite{contal2013parallel} approach, and that GP-UCB-PE \cite{contal2013parallel} performs better than GP-SM approach \cite{azimi2010batch}. There are also concerns that GP-UCB-PE and GP-BUCB are too greedy in the batch selection process, and thus prone to be non-optimal with respect to the "immediate overconfidence" measure \cite{kathuria2016batched}. 
Nguyen et al. \cite{nguyen2016budgeted} proposed a budgeted batch BO, termed GP-B3O, which samples the acquisition functions, then equates the number of peaks as the batch sizes and assigns sampling locations at the sampled peaks. 
However, most of the above-mentioned approaches are synchronous batch-sequential parallel, in the sense that at the end of each batch, all the simulations must be completed in order to move forward.

\subsubsection{Asynchronous parallel}
\label{subsubsec:asynchronousBatchParallel}

% Le Riche et al. \cite{le2012study} proposed a joint EI criterion to locate the next sampling locations based on q-EI framework. 

Ginsbourger et al. \cite{ginsbourger2011dealing} discussed and proposed the expected EI by conditioning the EI acquisition on the busy sampling points. 
Janusevskis et al. \cite{janusevskis2012expected} proposed an asynchronous multi-points EI acquisition function by estimating the progress of the available and incoming sampling locations. 
Letham et al. \cite{letham2018constrained} also employed the q-EI framework to asynchronous parallelize in noisy environment. 
GP-EI-MCMC \cite{snoek2012practical} can also be extended to accommodate the asynchronous parallel feature in BO. 
Kami{\'n}ski and Szufel \cite{kaminski2018parallel} extended the q-KG framework for asynchronous parallel BO method. 
Kotthaus et al. \cite{kotthaus2017rambo} proposed a resource-aware framework with different priorities, called RAMBO, that aim to reduce idle time on the computational workers through knapsack problem formulation. 
Kandasamy et al. \cite{kandasamy2017asynchronous} proposed an asynchronous parallel BO based on Thompson sampling approach \cite{thompson1933likelihood}. 
Pokuri et al. \cite{pokuri2018paryopt} implemented PARyOpt package with an asynchronous function evaluator where users can impose a threshold for a batch to finish before moving to a next batch.
Alvi et al. \cite{alvi2019asynchronous} further extended a locally penalization scheme to asynchronously parallel BO \cite{gonzalez2016batch}.

\section{Methodology}
\label{sec:aphBO-2GP-3B}

Here, we consider the optimization problem 
\begin{equation}
\bm{x}^* = \argmax_{\bm{x} \in \mathcal{X}} f(\bm{x}),
\end{equation}
subject to 
\begin{equation}
\lambda_j(\bm{x}) \leq g_j, \quad j=1,\dots,J, 
\end{equation}
where $f$ is evaluated through a computationally expensive engineering simulation, where the output $f(\bm{x})$ may not exist on some subspace of the input domain $\mathcal{X}$. $\lambda_j(\bm{x}), 1\leq j \leq J$, is a known constraint, formulated as a set of nonlinear inequality, and is cheap to evaluate. 
The input domain therefore can be decomposed into two disjoint regions, feasible and infeasible, as $\mathcal{X} = \mathcal{X}_{\text{infeasible}} \cup \mathcal{X}_{\text{feasible}}$, where the feasible region is a collection of input $\bm{x}$, where none of the nonlinear inequality constraints is violated, and $f(\bm{x}) \in \mathbb{R}$ exists, and the infeasible region is the complement of the feasible region. 

We further decompose the infeasible regions as 
% $\mathcal{X}_{\text{infeasible}} = \mathcal{X}^{\text{known}}_{\text{infeasible}} \cup \mathcal{X}^{\text{unknown}}_{\text{infeasible}}$ are 
the union of two subspaces, 
where 
%$\mathcal{X}^{\text{known}}_{\text{infeasible}} = 
$ \{ \bm{x} \in \mathcal{X} \quad | \quad \exists j: \quad \lambda_j(\bm{x}) > g_j \}$ is the subspace where at least one \textit{known} constraint is violated, and 
% $\mathcal{X}^{\text{unknown}}_{\text{infeasible}} = \{ \bm{x} \in \mathcal{X} \quad | \quad \text{infeasible for unknown reasons} \}$ 
the complement is the subspace where $\bm{x}$ is infeasible for some \textit{unknown} reasons, such as the functional evaluator $f(\bm{x})$ fails unexpectedly, numerical divergence in the solver is detected, or mesh creation is failed uncontrollably at the particular input $\bm{x}$. 
It is worthy to note that these two subspaces are not necessarily disjoint. 

In our previous work \cite{tran2019pbo}, we proposed a pBO-2GP-3B framework to handle known and unknown constrained in a batch-sequential parallel manner. 
The current method aphBO-2GP-3B is the natural extension of the pBO-2GP-3B framework in two main directions: the implementation of the asynchronous feature, and the implementation of multi-acquisition function under the GP-Hedge scheme, as described in Hoffman et al \cite{hoffman2011portfolio}. 
Compared to other methods in literature, the aphBO-2GP-3B approach is more practical in several aspects. 

First, the aphBO-2GP-3B approach provides a flexibility by relaxing the probabilistic binary classifier in distinguishing feasible and infeasible regions. 
There is no restriction in choosing the binary classifier, even though some classifiers tend to outperform others. 
The binary classifier can be $k$NN \cite{bentley1975multidimensional}, AdaBoost \cite{hastie2009multi}, RandomForest \cite{breiman2001random}, support vector machine \cite{hearst1998support} (SVM), least squares support vector machine (LSSVM) \cite{suykens1999least}, GP \cite{rasmussen2004gaussian}, and convolutional neural network \cite{lecun2015deep}. 
Similar to the pBO-2GP-3B, we adopted the method of Gardner et al. \cite{gardner2014bayesian} to penalize the acquisition function where the known constraints are violated. 

Second, multiple acquisition functions are considered and statistically sampled based on their performance measured by rewards. The reward handling scheme is only valid for feasible samples. This approach unifies and leverages the advantage of different acquisition functions individually, including PI, EI, and UCB. 

Third, it takes advantage of HPC parallelism by parallelizing asynchronously to reduce the wait time. This is more efficient than the synchronous batch-sequential parallel BO approach because the synchronous time requirement within a batch is removed, allowing more simulations to be completed. 
The problem is considered in a HPC environment where the number of cores used for optimization are limited. 

Similar to the GP-UCB framework \cite{desautels2014parallelizing} and kriging believer heuristic \cite{ginsbourger2008multi,ginsbourger2010kriging}, our approach aphBO-2GP-3B hallucinates the objective GP by assuming that the posterior means at the sampling locations are the observations until the actual observations are available. The term "hallucination", coined by Desautels in GP-UCB \cite{desautels2014parallelizing}, describes the procedure to temporarily assign the posterior mean of the objective GP as the actual observation, in order to sample the next location. 
The temporary assignment of the objective function is removed once the observation at the particular sampling point is made. 
In the current approach, we also generalize the hallucination module for the feasibility, by temporarily assuming that the sampling location is feasible. 
% Because the objective GP is temporarily "hallucinated" until the observation is obtained, 

% The feasibility of the querying sampling locations is assumed to be feasible until the actual feasibility is obtained. 
For the GP interpolation process, at the infeasible sampling points where the outputs do not exist, an interpolation process is employed to interpolate for the objective GP to reflect the true posterior variance $\sigma^2_{\text{objective}} = 0$ at these infeasible locations $\bm{x}_{\text{infeasible}}$. 
Similar the the GP hallucination process described above, this interpolation process updates the objective GP at the infeasible locations by assigning the posterior means as the observation and re-train the objective GP; however, the infeasibility of these locations remain unchanged. 
% This interpolation process updates whenever an observation is made. 
Without the GP interpolation process, the objective GP does not reflect the true uncertainty at these infeasible sampling locations, i.e. a large posterior variance is attributed to these locations, whereas a small posterior variance should be attributed, indeed. 
In other words, the $\sigma^2_{\text{objective}}$ is available at $\bm{x}_{\text{infeasible}}$, while the $\mu_{\text{objective}}$ is not at these infeasible sampling locations. 
Therefore, we must update $\mu_{\text{objective}}$ to reflect our knowledge, by the GP interpolation process. 
The GP interpolation process is achieved by three steps. 
In the first step, the objective GP is fitted only using feasible sampling locations. 
In the second step, the posterior mean of the objective GP is obtained by the fitted objective GP prediction. 
In the third step, these posterior means are used as the actual observations to fit the objective GP again. 

The aphBO-2GP-3B framework is constructed based on the extension of GP-UCB \cite{desautels2014parallelizing} and GP-UCB-PE \cite{contal2013parallel} using three different batches, where each batch is assigned a different job. 
The first batch is called the acquisition batch $\mathcal{B}_{\text{acquisition}}$, where a chosen acquisition is maximized to locate the next sampling point. 
The main job of the first batch is to optimize the objective function. 
Therefore, most of the computational effort and priority is reserved for the first batch. 
The second batch is called the exploration batch for the objective GP $\mathcal{B}_{\text{explore}}$, aiming at exploring the objective GP in the most uncertain regions, where $\sigma^2_{\text{objective}}$ is high. 
The third batch is called the exploration batch for the classification GP $\mathcal{B}_{\text{exploreClassif}}$, aiming at exploring the most uncertain regions for the GP classifier. 
GP is one of a few binary classifiers where uncertainty associated with the prediction is quantified. For other classifiers where uncertainty is not quantified, the third batch can be simply ignored.

\subsection{Constraints}

The aphBO-2GP-3B framework supports both known and unknown constraints as described above. 

\subsubsection{Known constraints}

The known constraints $\lambda_j(\bm{x}) \leq g_j$ are evaluated before the functional evaluator $f$ is revoked. Consequently, we can define an indicator function for constraints satisfaction as
\begin{equation}
\mathcal{I}(\bm{x}) = \begin{cases}
0, \quad \text{ if } \exists j: \lambda_j(\bm{x}) > g_j, \\ 
1, \quad \text{ if } \forall j: \lambda_j(\bm{x}) \leq g_j. \\ 
\end{cases}
\end{equation}
This function is also written as $\mathcal{I} \left(\lambda(\bm{x}) \leq \bm{g}\right)$. 
Since the known constraints can be evaluated before evaluating the function $f$, the acquisition function $a(\bm{x})$ is penalized as zero by multiplying with the constraint indicator function $\mathcal{I}(\bm{x})$. 
The next sampling point is continuously searched until a sampling point with a non-zero acquisition function value that does not violate the known constraints is found.

\subsubsection{Unknown constraints}
The unknown constraints in general are harder to handle, because it requires an approximation within the BO. 
To distinguish between feasible and infeasible regions adaptively, we propose to couple an external binary classifier that adaptively learns as the optimization advances. 
The choice of the external binary classifier is left to users, but GP is recommended to utilize the third batch. 
Denote this classifier as $\text{clf}(\bm{x})$. 
For each point in the input space $\bm{x} \in \mathcal{X}$, the classifier predicts a probability mass function for the feasibility of $\bm{x}$ as $Pr(\text{clf} (\bm{x}) = 1)$ if $\bm{x}$ is feasible, and $Pr(\text{clf} (\bm{x}) = 0)$ if $\bm{x}$ is infeasible. 
It is obvious that $Pr(\text{clf} (\bm{x}) = 0) + Pr(\text{clf} (\bm{x}) = 1) = 1$. 
This probability mass function can be used to condition on the acquisition function to drive the sampling points away from infeasible regions to feasible regions.

\subsubsection{Combining both known and unknown constraints}

The acquisition function $a(\bm{x})$ can be modified to reflect both known and unknown constraints. 
First, we incorporate the known constraints to the original acquisition function by multiplying them together as
\begin{equation}
a^{**}(\bm{x}) = a(\bm{x}) \cdot \mathcal{I}(\lambda(\bm{x}) \leq \bm{g}).
\end{equation}
Second, we condition the new acquisition function $a^{**}(\bm{x})$ on the probability mass function for the unknown function as 
\begin{equation}
a^{***}(\bm{x}) = 
\begin{cases}
a^{**}(\bm{x}), & \text{ if } \text{clf}(\bm{x}) = 1, \\
0 , & \text{ if } \text{clf}(\bm{x}) = 0. 
\end{cases}
\end{equation}

Taking the expectation of $a^{***}(\bm{x})$ on the probability mass function from the binary classifier yields
\begin{equation}
\begin{array}{lll}
a^*(\bm{x}) &=& \mathbb{E}[a^{***}(\bm{x})] \\
&=& a^{**}(\bm{x}) \cdot Pr(\text{clf}(\bm{x}) = 1) + 0 \cdot Pr(\text{clf}(\bm{x}) = 0) \\ 
&=& a^{**}(\bm{x}) \cdot Pr(\text{clf}(\bm{x}) = 1) \\ 
&=& a(\bm{x}) \cdot \mathcal{I}(\lambda(\bm{x}) \leq \bm{g}) \cdot Pr(\text{clf}(\bm{x}) = 1), \\ 
\end{array}
\label{eq:modAcqFunc}
\end{equation}

The new acquisition function $a^{*}(\bm{x})$ is our constrained acquisition function, which is constructed by combining the original acquisition function $a(\bm{x})$ with the penalization for known and unknown constraints.

\subsection{Asynchronous parallel}

% \begin{figure}[!htbp]
% \centering
% \includegraphics[width=0.65\textwidth, keepaspectratio]{figsAsyncBO/cropped.asyncParBOScheme}

% \end{figure}

\begin{figure}[!htbp]
\centering
\subcaptionbox{Batch-sequential parallel.
\label{fig:batchSequentialParallel}
}
  [.495\linewidth]{\includegraphics[width=0.495\textwidth, keepaspectratio]{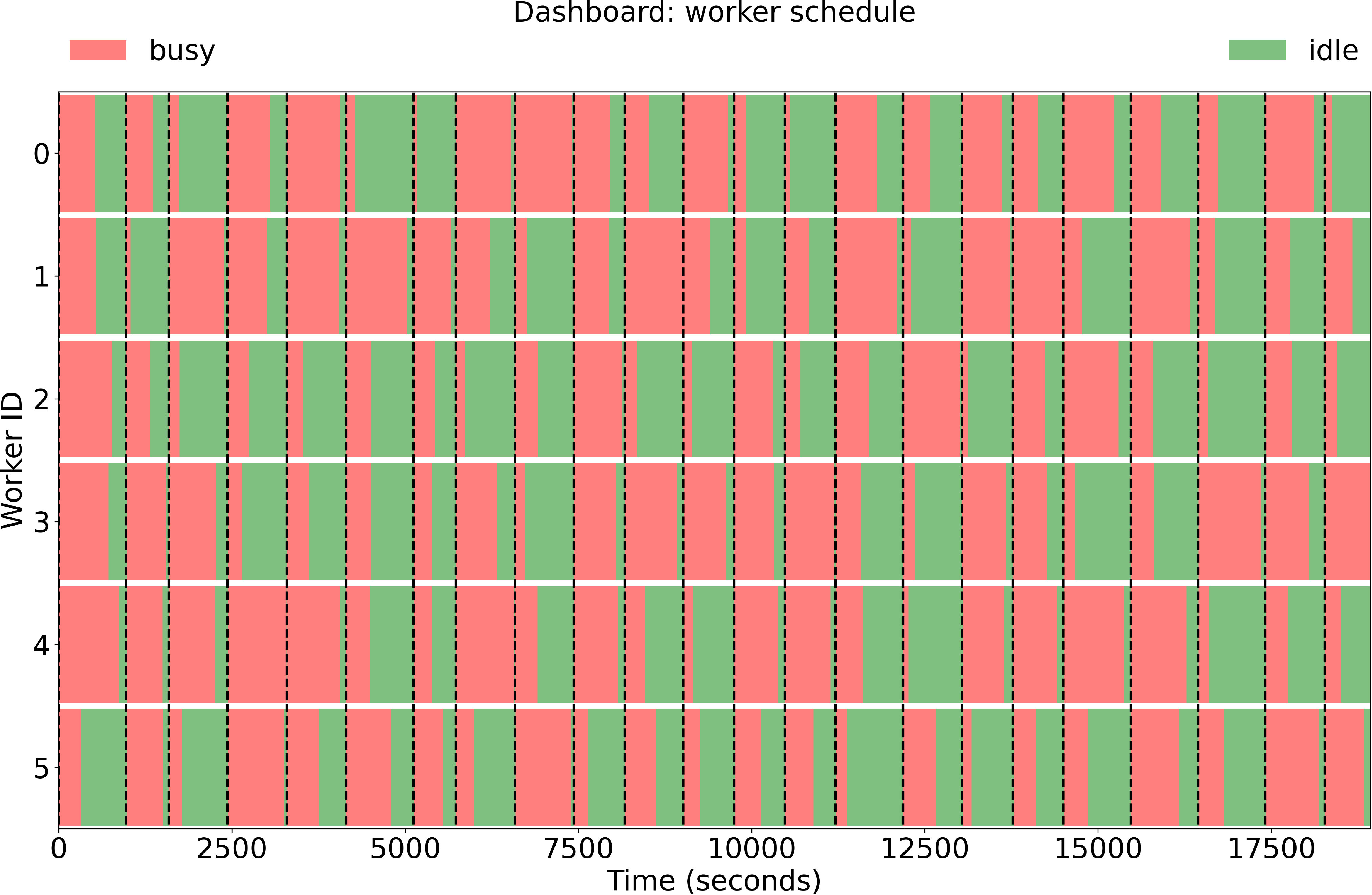}}
\hfill
\subcaptionbox{Asynchronous parallel.
\label{fig:asynchronousParallel}
}
  [.495\linewidth]{\includegraphics[width=0.495\textwidth, keepaspectratio]{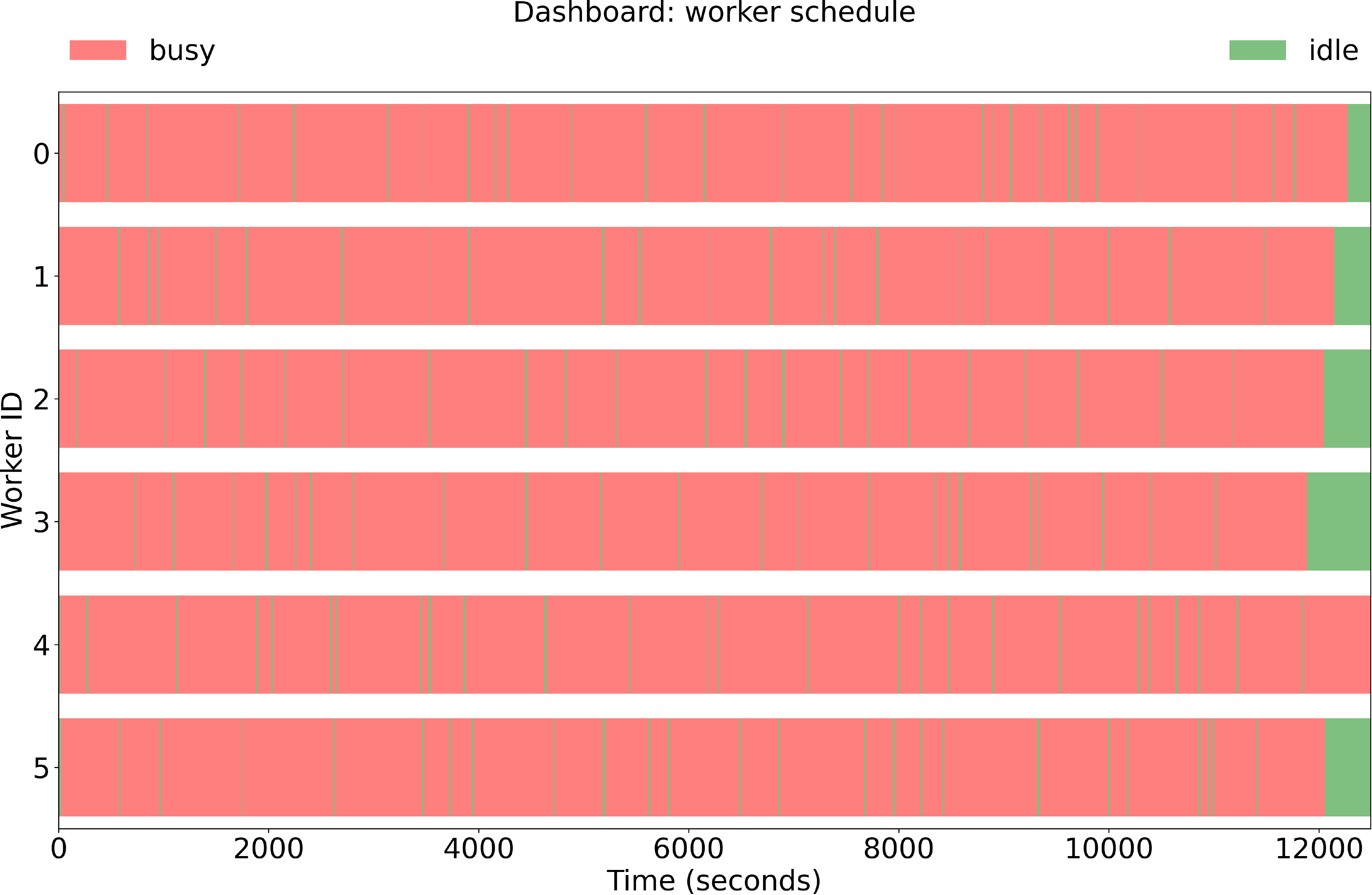}}
\caption{Difference between batch-sequential and asynchronous parallelizations. 
In batch-sequential parallelization (Figure \ref{fig:batchSequentialParallel}), all workers need to wait until the batch is finished before moving on to the next batch. 
In asynchronous parallelization (Figure \ref{fig:asynchronousParallel}), all the workers receive the most up-to-date information and work independently with each other. 
Asynchronous parallelization is more efficient than batch-sequential parallelization because it reduces idle time for workers. 
% where each batch is assigned a different priority and simulated concurrently without synchronizing barrier (as in sequential batch parallelization).
}
\label{fig:cropped.asyncParBOScheme}
\end{figure}

The asynchronous parallel feature is constructed based on the priority of the batch, as
\begin{equation}
\label{eq:priority}
\mathcal{B}_{\text{acquisition}} > \mathcal{B}_{\text{explore}} > \mathcal{B}_{\text{exploreClassif}},
\end{equation}
because the first batch $\mathcal{B}_{\text{acquisition}}$ is the most important one among three batches, whenever the first batch is under filled, then the computational effort is reserved for filling the first batch. 
If the first batch is filled, then the second batch and the third batch is considered, respectively, according to the priority listed in Equation \ref{eq:priority}. 

The asynchronous parallel feature is implemented by periodically checking if these batches are under filled. If the batches are full, then the aphBO-2GP-BO optimizer will hold until a pending sampling point is finished and some batch is under filled as a result.

\subsection{Multi-acquisition function}

Unified approaches considering multiple acquisition functions have been considered in the literature. 
For example, Shahriari proposed an entropy-based approach, called GP-EPS \cite{shahriari2014entropy}, for searching the best acquisition function within the same portfolio. 
In the current aphBO-2GP-3B framework, we adopt and modify the GP-Hedge scheme \cite{hoffman2011portfolio} where multiple acquisition functions are considered simultaneously for the first batch $\mathcal{B}_{\text{acquisition}}$. 
De Palma et al. \cite{de2019sampling} proposed to sample the acquisition function portfolio using Thompson sampling. 
After the initial sampling stage, the reward for each acquisition function is reset to zero. 
To locate a next sampling point in the batch $\mathcal{B}_{\text{acquisition}}$, first, the acquisition function is determined by sampling based on a probability mass function, where the parameters of the probability mass function correspond to their rewards, or more precisely, their numerical performance. 
Second, when the acquisition function is determined, the new sampling point is determined by maximizing the chosen acquisition function. 
Third, when a sampling point is finished querying, if the sampling point is determined to be feasible, then the reward is added into the cumulative reward of that acquisition function. The cumulative rewards are tracked throughout the optimization process. 
Hoffman et al. \cite{hoffman2011portfolio} proposed to choose $\eta_t = \sqrt{\frac{8\ln{k}}{n}}$ as a time-varying parameter based on Cesa-Bianchi and Lugosi \cite{cesa2006prediction} (Section 2.3), where $k$ is the number of acquisition functions considered in the GP-Hedge scheme. 
% Here we slightly change to $\eta_N = \sqrt{\frac{8\ln{n}}{n}}$ as the dataset grows. 

\begin{algorithm}
\caption{Improved GP-Hedge for aphBO-2GP-3B: Inverse sampling multiple acquisition functions \blue{(modification highlighted in blue)}.}
\label{alg:gpHedge}

\algorithmicinput observations $(y_i, c_i)_{i=1}^n$; $n^*$ acquisition functions $\{a_j\}_{j=1}^{n^*}$

\algorithmicoutput acquisition function $a_{j^*}$ sampled at iteration $n+1$ based on gains and rewards and the next sampling point $\bm{x}^*$

\begin{algorithmic}[1]
\State $\eta_n \gets \sqrt{\frac{8\ln{k}}{n}}$ \Comment{set scaling parameters for rewards (cf. \cite{hoffman2011portfolio}): $\eta \in \mathbb{R}^+$}
\State $g_0^j \gets 0$ for $1 \leq j \leq n^*$ \Comment{initialize; $n^*$: number of acquisition functions}
\For{$i=1,2,\dots,n$}
% \State \quad augment the data $\mathcal{D}_{1:i} = \{\mathcal{D}_{1:-i}, (\bm{x}_i, y_t, c_t) \} $
% \State \quad Receive award $r^i_t = \mu(\bm{x}^i_t)$ from the updated GP
\If{ \blue{ $\bm{x}_i$ feasible, $y_i$ is best-so-far }} \Comment{\blue{modification for constrained problems}}
	\State \quad \blue{ $r^j_i \gets 1$ for $a_j$ at iteration $i$ } \Comment{ \blue {receive reward} (original \cite{hoffman2011portfolio}: $r^i_t \gets \mu(\bm{x}_i)$)}
\Else
	\State \quad \blue{$r^j_i \gets 0$} \Comment{\blue{no rewards if not feasible or not best-so-far}}
\EndIf
\State \quad $g^j_i \gets g^j_{i-1} + r^j_i$ \Comment{update gains}
\EndFor
\For{$1 \leq j \leq n^*$}
\State $p_{n+1}(j) \gets \frac{\exp(\eta g^j_{n})}{\sum_{j=1}^{n^*} \exp(\eta g^j_{n})}$ \Comment{normalize p.m.f. of $a_j$'s}
\EndFor
% \For{$1 \leq j \leq n^*$} \Comment{draw acquisition function $a_{j^*} \gets a_k$ with probability} 
% \State \blue{draw $u \sim \mathcal{U}(0,1)$}
% \State \blue{find $j^*$ s.t. $ \sum_{j=1}^{j^*-1} p_{n+1}(j)  \leq u \leq \sum_{j=1}^{j^*} p_{n+1}(j) $} \Comment{ \blue {determine the acquisition function $a_{j^*}$ using inverse transform sampling} }
\State determine the acquisition function $a_{j^*}$
% \EndFor
\State locate the next sampling point $\bm{x}^* \gets \argmax_{ \bm{x}} a_{j^*}(\bm{x} | \mathcal{D}_{n})$ \Comment{locate $\bm{x}^*$}
% \State sample the objective function $y_t \gets f(\bm{x}_i)$
\end{algorithmic}
\end{algorithm}

\subsection{Implementation and Summary}

\begin{figure}[!htbp]
\centering
\includegraphics[width=0.80\textwidth, keepaspectratio]{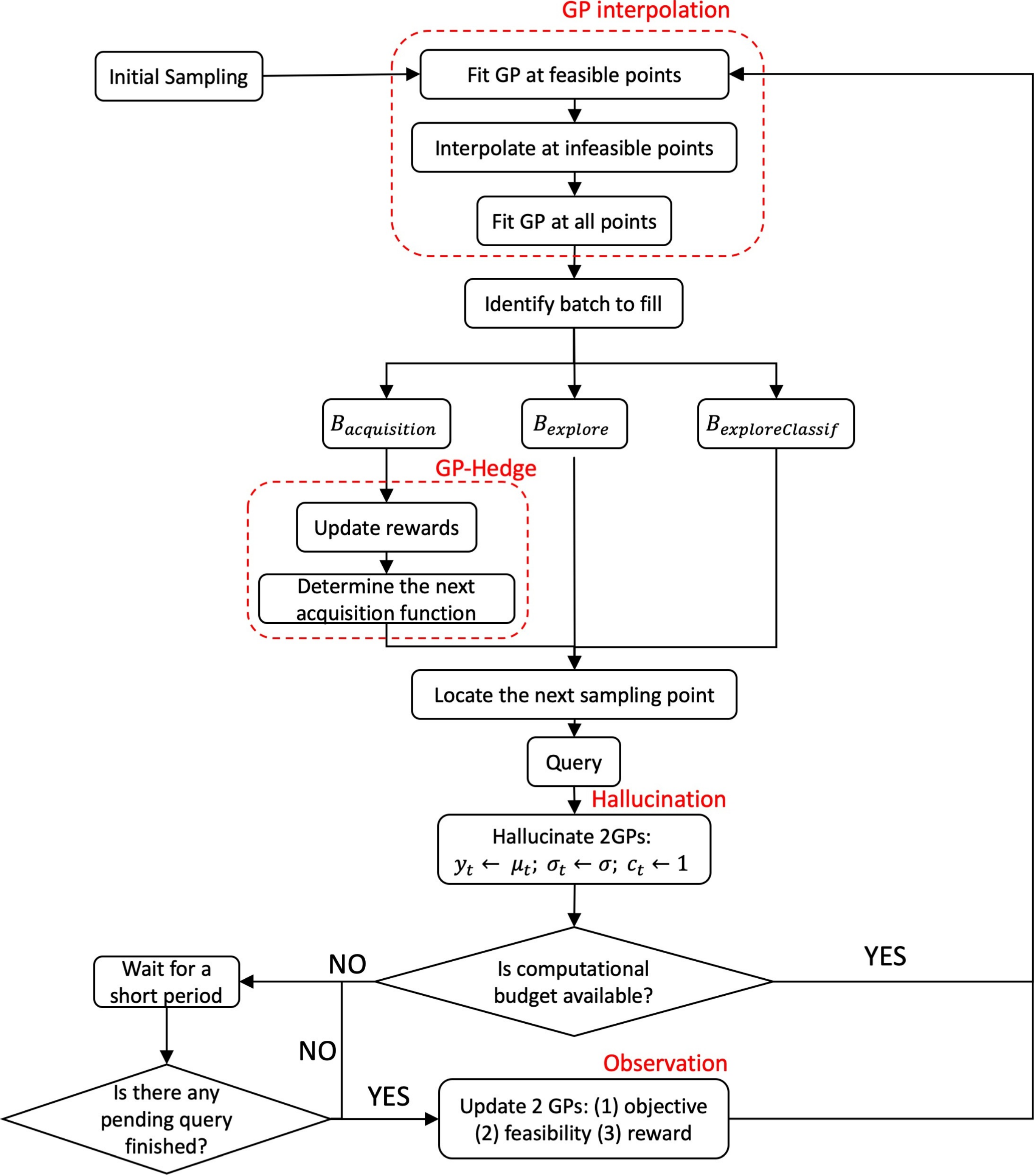}
\caption{The aphBO-2GP-3BO flow chart. At any given time if the computational resource is available, the GP interpolation module updates the objective GP $\mathcal{GP}_\text{objective}$ by hallucinating at points that are either (a) being queried or (b) have been queried but are infeasible. Once $\mathcal{GP}_\text{objective}$ finishes updating, one sampling point is assigned to the batch according to its priority, with descending priority given to $\mathcal{B}_{\text{acquisition}}$, to $\mathcal{B}_{\text{explore}}$, to $\mathcal{B}_{\text{exploreClassif}}$. For the batch $\mathcal{B}_{\text{acquisition}}$, the rewards is updated to sample the acquisition function, which can be PI, EI, or UCB by GP-Hedge scheme. Once the acquisition function is determined, the next sampling point is located and then queried. $\mathcal{GP}_\text{objective}$ is then hallucinated again at the queried sampling point. 
If there is available computational budget, the loop returns to GP interpolation module; otherwise, the algorithm waits at least one query is finished. 
}
\label{fig:cropped.asyncParBOflowchart}
\end{figure}

The aphBO-2GP-3B framework is summarized in Algorithm \ref{alg:aphbo2gp3b}. 
The current GP implementation is built upon modified MATLAB DACE \cite{nielsen2002dace} and ooDACE \cite{gorissen2010surrogate,couckuyt2012blind,couckuyt2013oodace,couckuyt2014oodace} toolboxes. 
The CMA-ES method is employed \cite{hansen2001completely,hansen2003reducing} to maximize the acquisition function and locate the next sampling point. 
Python scripts are devised as an interface between the aphBO-2GP-3BO in MATLAB and the actual application. 
After a sampling location is queried, the optimizer checks if all the batches have been filled or if there is any pending query has been completed. 
If there is any batch that is not filled completely, then it will be filled according to the order described in Equation \ref{eq:priority}. 
The aphBO-2GP optimizer then moves forward, assuming the posterior mean is the observation and continues to search for new sampling locations. 
If all the batches are full, then the optimizer simply waits. 
If there is any pending query that has been completed, then the results including objective function, feasibility, reward, are updated simultaneously, followed by the GP interpolation module before locating a new sampling point. 
% If there is no pending query that has been completed, i.e. all sampling points are being , the optimizer continues to search for a new sampling location according to the priority and query. 
% The loop repeats until the computational budget is met. Then the aphBO-2GP-3B optimizer simply halts and waits. 
Figure \ref{fig:cropped.asyncParBOflowchart} presents the flowchart for aphBO-2GP-3BO framework. 
The GP-Hedge scheme, the GP interpolation module, and the hallucination module are highlighted to emphasize their importance.

\begin{algorithm}
\caption{aphBO-2GP-3B algorithm.}
\label{alg:aphbo2gp3b}
\algorithmicinput dataset $\mathcal{D}_n$ consisting of input, observation, feasibility $(\bm{x}, y_i, c_i)_{i=1}^n$ 

\algorithmicinput objective $\mathcal{GP} (\bm{x}, y_i)_{i=1}^n$, and classification $\mathcal{GP} (\bm{x}, c_i)_{i=1}^n$
\begin{algorithmic}[1]
% \State Set remainder $r = l$
% \For{$i=n+1,\dots$}
\While{convergence criteria not met}
\While{number of queries $\not < \mathcal{B}_{\text{budget}}$} \Comment{wait indefinitely for available resource}
\State check periodically if there is available computational resource to use
\EndWhile

% \If{number of queries $< \mathcal{B}_{\text{budget}}$} \Comment{if there is available resource}

\State update input/output/feasible/complete for all cases \Comment{update; if not complete then hallucinate, i.e. $y_i \gets \mu(\bm{x}_i), \sigma_i^2 \gets \sigma^2, c_i \gets$ feasible}
\State update/augment dataset $\mathcal{D}_{n+1} = \{ \mathcal{D}_n , (\bm{x}^{(B)}_{n+1}, y^{(B)}_{n+1}, c^{(B)}_{n+1}) \}$ \Comment{update}
\State interpolate 2GPs \Comment{update/hallucinate 2GPs}
\State \quad construct $\mathcal{GP}_\text{objective}$ \Comment{hallucinate $\mathcal{GP}_\text{objective}$}
% \footnotesize
% \State \quad \quad collect feasible data subset $(\bm{x}_i, y_i, c_i=\text{feasible}) \}_{i=1}^n$
\State \quad \quad construct $\mathcal{GP}_{\text{objective}}(\bm{x}_i, y_i | c_i=\text{feasible})$ for feasible points 
\State \quad \quad hallucinate $\mathcal{GP}_\text{objective}$, i.e. $y_i \gets \mu_i, \sigma_i^2 \gets \sigma^2$, at infeasible points $c_i=\text{infeasible}$
\State \quad \quad reconstruct $\mathcal{GP}_\text{objective}$ using both feasible and infeasible points 
% \normalsize
\State \quad construct $\mathcal{GP}_{\text{classification}}(\bm{x}_i,c_i)$ \Comment{hallucinate $\mathcal{GP}_{\text{classification}}$}

\State identify which batch needs filling \Comment{get batch type}
\State locate the next sampling location for the selected batch \Comment{find sampling location}
% \footnotesize
\State update/hallucinate 2GPs
\State \quad \textit{acquisition}: 
\State \quad \quad compute rewards for each acquisition function \Comment{GP-Hedge (Alg. \ref{alg:gpHedge})}
\State \quad \quad sample acquisition function portfolio \Comment{GP-Hedge (Alg. \ref{alg:gpHedge})}
\State \quad \quad locate the next sampling point with the sampled acquisition function
\State \quad \textit{exploration}: 
% \State \quad \quad update/hallucinate 2GPs 
\State \quad \quad locate the next sampling location where $\sigma_{\text{objective}}^2$ is maximized
\State \quad \textit{classification}
% \State \quad \quad update/hallucinate 2GPs 
\State \quad \quad locate the next sampling location where $\sigma_{\text{classification}}^2$ is maximized
% \normalsize
\State query objective function for objective $y^{(B)}_{n+1}$ and feasibility $c^{(B)}_{n+1}$ \Comment{query}
\State update number of queries += 1 \Comment{increase the number of busy workers}

% \EndIf
% \EndFor
\EndWhile
\end{algorithmic}
\end{algorithm}

\section{Numerical benchmark}

In this section, we benchmark the aphBO-2GP-3B framework by comparing its with two of its main variants: the synchronous batch-sequential parallel pBO-2GP-3B \cite{tran2019pbo} with PI, EI, UCB acquisition functions and the non-GP-Hedge asynchronous parallel apBO-2GP-3B with PI, EI, UCB acquisition functions. 
To augment the comparison, we also provide the numerical performance of the sequential BO and sequential Monte Carlo random search. 
The implementations of the testing functions used are publicly available \cite{surjanovic2016virtual} in both MATLAB and R, where R scripts are used in our benchmark. 
We investigate their performance with respect to both the physical wall-clock time and the number of evaluations. 
To simulate the random behaviors of real-world applications in terms of computational time, we append a uniformly distributed random waiting time after the functional evaluation is completed, i.e. $t^* \sim \mathcal{U}(\underline{t}, \overline{t})$, where $\underline{t} = 30$ s and $\overline{t} = 900$ s. 
% In this benchmark study, the R implementation of benchmark functions are used, and the waiting times are set as $\underline{t} = 2$s, $\overline{t} = 120$s. 
We compare 6 different variants of parallel BO methods, including
\begin{itemize}
% \item classical BO with UCB, EI, and PI acquisition functions,
\item pBO-2GP-3BO with UCB, EI, and PI acquisition functions \cite{tran2019pbo} (the synchronous batch-sequential parallel),
\item apBO-2GP-3BO -- without GP-Hedge, with UCB, EI, and PI acquisition functions (the asynchronous parallel without GP-Hedge),
\item aphBO-2GP-3BO -- with GP-Hedge (the asynchronous parallel with GP-Hedge),
\item asynchronous parallel Monte Carlo random search as a baseline,
\end{itemize}
where each method is repeated 5 times. 
% \redbf{details asuslocal lscpu here}
To be fair, all benchmark runs are performed on the same workstation with Ubuntu 18.04.4 LTS, with AMD A10-6700 APU CPU and 16GB of memory. 
Furthermore, in order to avoid the complication of benchmarking on a relatively less capable workstation, the waiting time is relatively long so that running the optimizer does not have a significant impact on the CPU as well as the physical memory. 
The list of name, dimensionality, batch sizes, function form, and global minimum of benchmark functions \footnote{Full documentation and implementation are available online at \href{https://www.sfu.ca/~ssurjano/index.html}{https://www.sfu.ca/$\sim$ssurjano/index.html}} are as follows. 
% For the sake of completeness, the analytical benchmarking functions\footnote{Also documented online at \href{https://www.sfu.ca/~ssurjano/index.html}{https://www.sfu.ca/$\sim$ssurjano/index.html}} are included as follows.

\begin{itemize}
\item eggholder (2d): batch size = (2,2,0): max iteration = 80
\begin{equation}
f(\bm{x}) = -(x_2 + 47) \sin\left( \left| x_2 + \frac{x_1}{2} + 47 \right| \right) - x_1 \sin\left( \left| x_1 - (x_2+47) \right| \right)
\end{equation}
on $x_i \in [-512, 512]$ for $i=1,2$. Global minimum $f(\bm{x^*}) = -959.6407$ at $\bm{x^*} = (512, 404.2319)$. 

\item three-hump camel (2d): batch size = (2,2,0): max iteration = 80
\begin{equation}
f(\bm{x}) = 2x_1^2 - 1.05 x_1^4 + \frac{x_1^6}{6} + x_1 x_2 + x_2^2
\end{equation}
on the domain of $x_i \in [-5,5]$ for $i=1,2$. Global minimum $f(\bm{x^*}) = 0$ at $\bm{x^*} = (0,0)$.

\item six-hump camel (2d): batch size = (3,1,0): max iteration = 80
\begin{equation}
f(\bm{x}) = \left( 4 - 2.1 x_1^2 + \frac{x_1^4}{3} \right)x_1^2 + x_1 x_2 + (-4 + 4x_2^2)x_2^2
\end{equation}
on the domain of $\bm{x} \in [-3,3] \times [-2,2]$. Global minimum $f(\bm{x^*}) = -1.0316$ at $\bm{x^*} = (-0.0898,0.7126)$.

\item hartmann (3d): batch size = (3,3,0): max iteration = 150
\begin{equation}
f(\bm{x}) = - \sum_{i=1}^4 \alpha_i \exp\left( - \sum_{j=1}^3 A_{ij} (x_j - P_{ij})^2 \right),
\end{equation}
where $\bm{A} = \begin{pmatrix} 3.0 & 10 & 30 \\ 0.1 & 10 & 35 \\ 3.0 & 10 & 30 \\ 0.1 & 10 & 35 \end{pmatrix}$, $\bm{P} = 10^{-4} \begin{pmatrix} 3689 & 1170 & 2673 \\ 4699 & 4387 & 7470 \\ 1091 & 8732 & 5547 \\ 381 & 5743 & 8828 \end{pmatrix}$, and $\alpha = (1.0, 1.2, 3.0, 3.2)^T$ on the domain of $[0,1]^3$. Global minimum $f(\bm{x^*}) = -3.86278$ at $\bm{x^*} = (0.114614, 0.555649, 0.852547)$.

\item hartmann (4d): batch size = (4,4,0): max iteration = 160
\begin{equation}
f(\bm{x}) = \frac{1}{0.839} \left[ 1.1 - \sum_{i=1}^4 \alpha_i \exp\left( - \sum_{j=4}^3 A_{ij} (x_j - P_{ij})^2 \right) \right],
\end{equation}
where $\bm{A} = \begin{pmatrix} 10 & 3 & 17 & 3.50 & 1.7 & 8 \\ 0.05 & 10 & 17 & 0.1 & 8 & 14 \\ 3 & 3.5 & 1.7 & 10 & 17 & 8 \\ 17 & 8 & 0.05 & 10 & 0.1 & 14 \end{pmatrix}$, $\bm{P} = 10^{-4} \begin{pmatrix} 1312 & 1696 & 5569 & 124 & 8283 & 5886 \\ 2329 & 4135 & 8307 & 3736 & 1004 & 9991 \\ 2348 & 1451 & 3522 & 2883 & 3047 & 6650 \\ 4047 & 8828 & 8732 & 5743 & 1091 & 381 \end{pmatrix}$, and $\alpha = (1.0, 1.2, 3.0, 3.2)^T$ on the domain of $[0,1]^4$.

\item ackley (free d) (5d): batch size = (6,4,0): max iteration = 200
\begin{equation}
f(\bm{x}) = -a \exp \left( - b\sqrt{\frac{1}{d} \sum_{i=1}^d x_i^2} \right) - \exp\left( \frac{1}{d} \sum_{i=1}^d \cos(cx_i) \right) + a + \exp(1),
\end{equation}
where $a=20, b= 0.2, c = 2\pi$, on $[-32.768, +32.768]^d$. Global minimum $f(\bm{x^*}) = 0$ at $\bm{x^*} = (0, \dots, 0)$. 

\item hartmann (6d): batch size = (5,5,0): max iteration = 300
\sloppy
\begin{equation}
f(\bm{x}) =  - \sum_{i=1}^4 \alpha_i \exp\left( - \sum_{j=6}^3 A_{ij} (x_j - P_{ij})^2 \right) ,
\end{equation}
where $\bm{A} = \begin{pmatrix} 10 & 3 & 17 & 3.50 & 1.7 & 8 \\ 0.05 & 10 & 17 & 0.1 & 8 & 14 \\ 3 & 3.5 & 1.7 & 10 & 17 & 8 \\ 17 & 8 & 0.05 & 10 & 0.1 & 14 \end{pmatrix}$, $\bm{P} = 10^{-4} \begin{pmatrix} 1312 & 1696 & 5569 & 124 & 8283 & 5886 \\ 2329 & 4135 & 8307 & 3736 & 1004 & 9991 \\ 2348 & 1451 & 3522 & 2883 & 3047 & 6650 \\ 4047 & 8828 & 8732 & 5743 & 1091 & 381 \end{pmatrix}$, and $\alpha = (1.0, 1.2, 3.0, 3.2)^T$ on the domain of $[0,1]^4$. Global minimum $f(\bm{x^*}) = -3.32237$ at $\bm{x^*} = (0.20169, 0.150011, 0.476874, 0.275332, 0.311652, 0.6573)$.

\item michalewicz (free d) (10d): batch size = (5,5,0): max iteration = 400
\begin{equation}
f(\bm{x}) = - \sum_{i=1}^d \sin(x_i) \sin^{2m}\left( \frac{i x_i^2}{\pi} \right),
\end{equation}
on the domain of $[0,\pi]^d$.

\item perm0db (free d) (80d): batch size = (6,4,0): max iteration = 400
\begin{equation}
f(\bm{x}) = \sum_{i=1}^d \left( \sum_{j=1}^d (j+\beta) \left( x_j^i - \frac{1}{j^i} \right) \right)^2,
\end{equation}
on the domain of $[-d,d]^d$. Global minimum $f(\bm{x^*}) = 0$ at $\bm{x^*} = \left(\frac{1}{1}, \frac{1}{2}, \dots, \frac{1}{d} \right)$.

\item rosenbrock (free d) (20d): batch size = (6,4,0): max iteration = 400
\begin{equation}
f(\bm{x}) = \sum_{i=1}^{d-1} \left[ 100(x_{i+1} - x_i^2)^2 + (x_i - 1)^2 \right],
\end{equation}
on the domain of $[-5,10]^d$. Global minimum $f(\bm{x^*}) = 0$ at $\bm{x^*} = (1,\dots,1)$.

\item dixon-price (free d) (25d): batch size = (7,7,0): max iteration = 480
\begin{equation}
f(\bm{x}) = (x_1 - 1)^2 + \sum_{i=2}^{d} i(2 x_i^2 - x_{i-1})^2
\end{equation}
on the domain of $[-10,10]^d$. Global minimum $f(\bm{x^*}) = 0$ at $x_i = 2^{-\frac{2^i - 2}{2^i} }$ for all $1 \leq i \leq d$.

\item trid (free d) (30d): batch size = (6,4,0): max iteration = 400
\begin{equation}
f(\bm{x}) = \sum_{i=1}^d (x_i - 1)^2 - \sum_{i=2}^d x_i x_{i-1},
\end{equation}
on the domain of $[-d^2,d^2]$. Global minimum $f(\bm{x^*}) = -d (d+4)(d-1) / 6$ at $x_i = i(d + 1 - i)$ for all $1 \leq i \leq d$.

\item sumsqu (free d) (40d): batch size = (6,4,0): max iteration = 400
\begin{equation}
f(\bm{x}) = \sum_{i=1} i x_i^2,
\end{equation}
on the domain of $[-5.12, 5.12]^d$. Global minimum $f(\bm{x^*}) = 0$ at $\bm{x^*} = (0, \dots, 0)$. 

\item sumpow (free d) (50d): batch size = (6,4,0): max iteration = 400
\begin{equation}
f(\bm{x}) = \sum_{i=1}^d |x_i|^{i+1},
\end{equation}
on the domain of $[-1,1]^d$. Global minimum $f(\bm{x^*}) = 0$ at $\bm{x^*} = (0, \dots, 0)$. 

\item spheref (free d) (60d): batch size = (6,4,0): max iteration = 400
\begin{equation}
f(\bm{x}) = \sum_{i=1}^d x_i^2,
\end{equation}
on the domain of $[-5.12,5.12]^d$. Global minimum $f(\bm{x^*}) = 0$ at $\bm{x^*} = (0, \dots, 0)$. 

\item rothyp (free d) (70d): batch size = (6,4,0): max iteration = 400
\begin{equation}
f(\bm{x}) = \sum_{i=1}^d \sum_{j=1}^i x_j^2,
\end{equation}
on the domain of $[-65.536,65.536]^d$. Global minimum $f(\bm{x^*}) = 0$ at $\bm{x^*} = (0, \dots, 0)$. 

\end{itemize}

% Show:
% \begin{enumerate}
% \item convergence w.r.t. time
% \item convergence w.r.t. iteration
% \item acquisition portfolio
% \item number of acquisition w.r.t. iteration
% \item dashboard: track worker schedule
% \item dashboard: track worker acquisition
% \end{enumerate}
% % compareConvergence_ByTime_bench.py
% % compareConvergence_ByIter_bench.py
% % plotAcquisition.py
% % trackWorkerSchedule_asyncPar.py / trackWorkerSchedule_syncPar.py
% % trackWorkerAcquisition_asyncPar.py

% \subsection{eggholder (2d)}
\begin{figure}[!htbp]
\centering
\subcaptionbox{Benchmark by physical wall-clock time.
\label{fig:benchByTime_egg}
}
  [0.425\linewidth]{\includegraphics[width=0.425\textwidth, keepaspectratio]{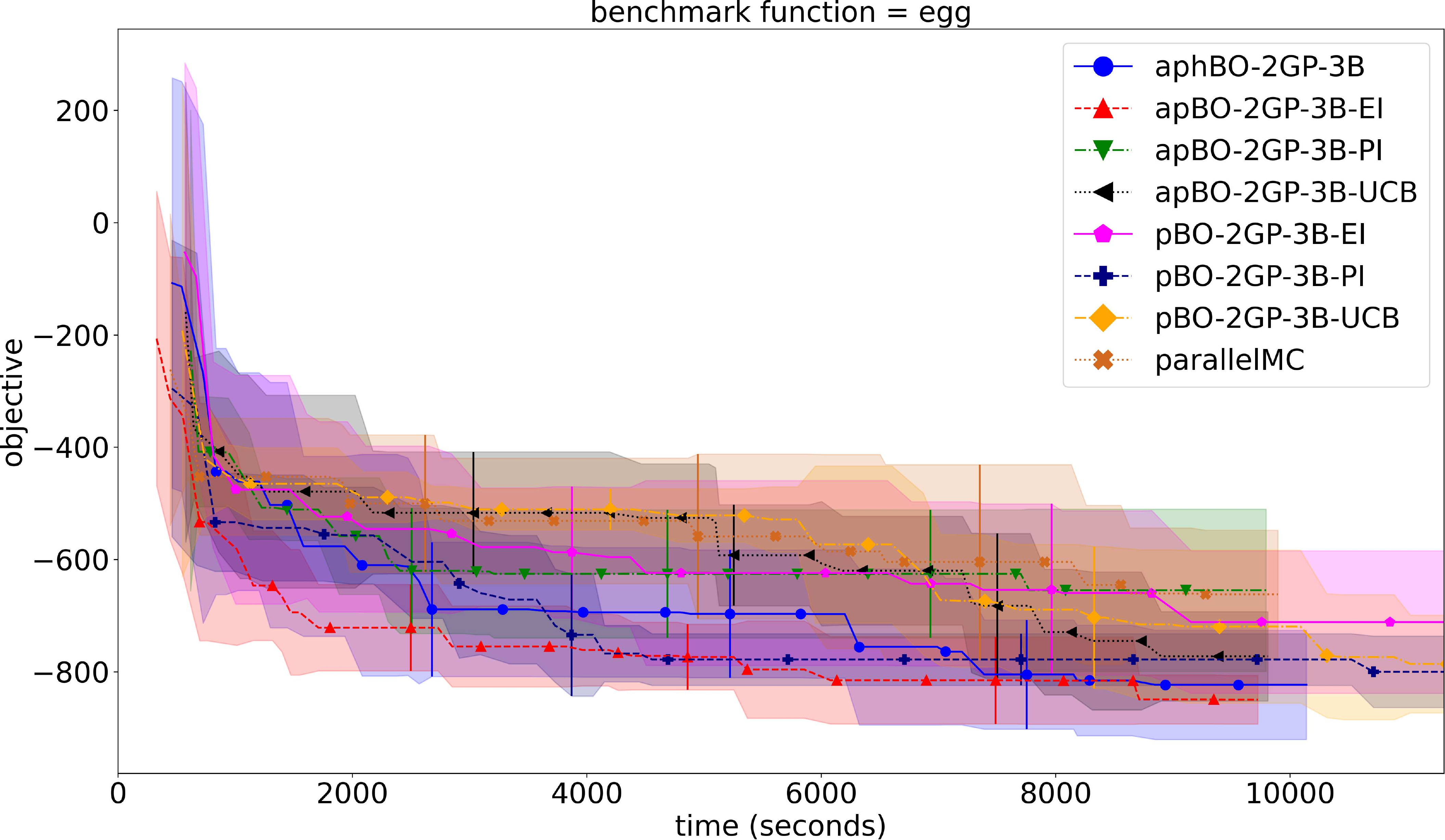}}
\hfill
\subcaptionbox{Benchmark by iteration.
\label{fig:benchByIter_egg}
}
  [0.425\linewidth]{\includegraphics[width=0.425\textwidth, keepaspectratio]{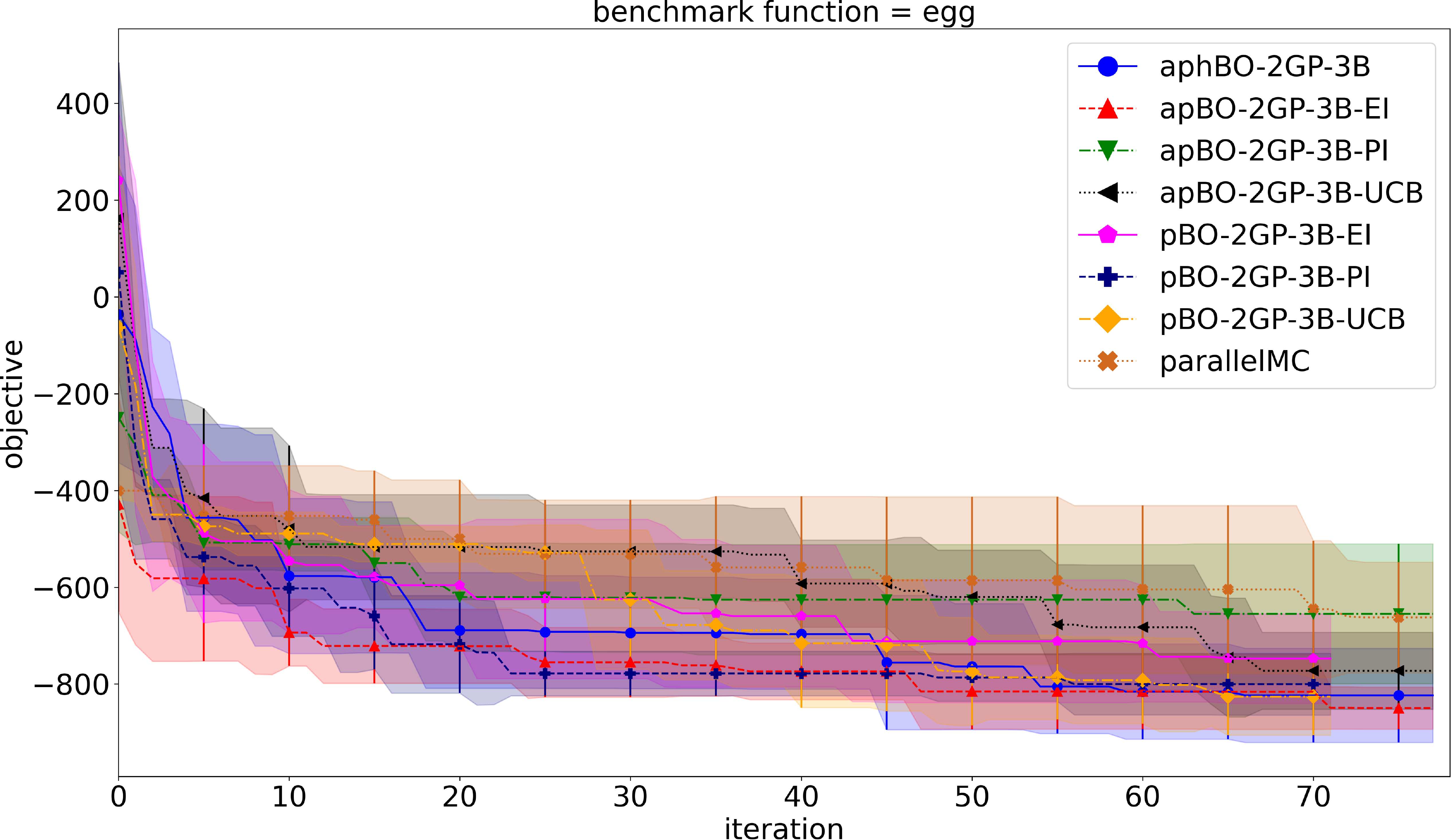}}
% \caption{Benchmark for egg.}
% \label{fig:bench_egg}
% \end{figure}
% \vspace{-2.50cm}
\medskip

% \subsection{three-hump camel (2d)}
% \begin{figure}[!htbp]
\centering
\subcaptionbox{Benchmark by physical wall-clock time.
\label{fig:benchByTime_camel3}
}
  [0.425\linewidth]{\includegraphics[width=0.425\textwidth, keepaspectratio]{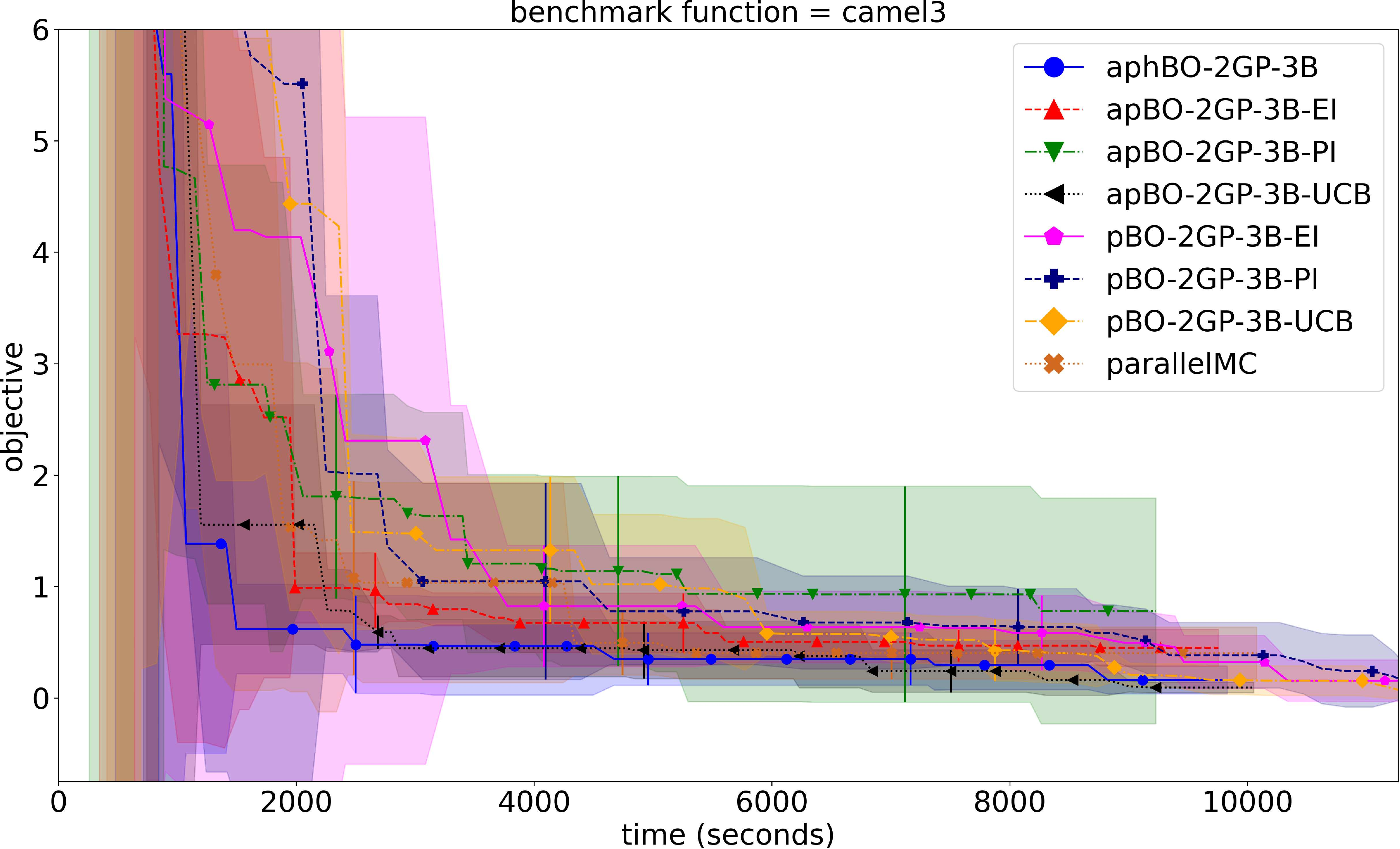}}
\hfill
\subcaptionbox{Benchmark by iteration.
\label{fig:benchByIter_camel3}
}
  [0.425\linewidth]{\includegraphics[width=0.425\textwidth, keepaspectratio]{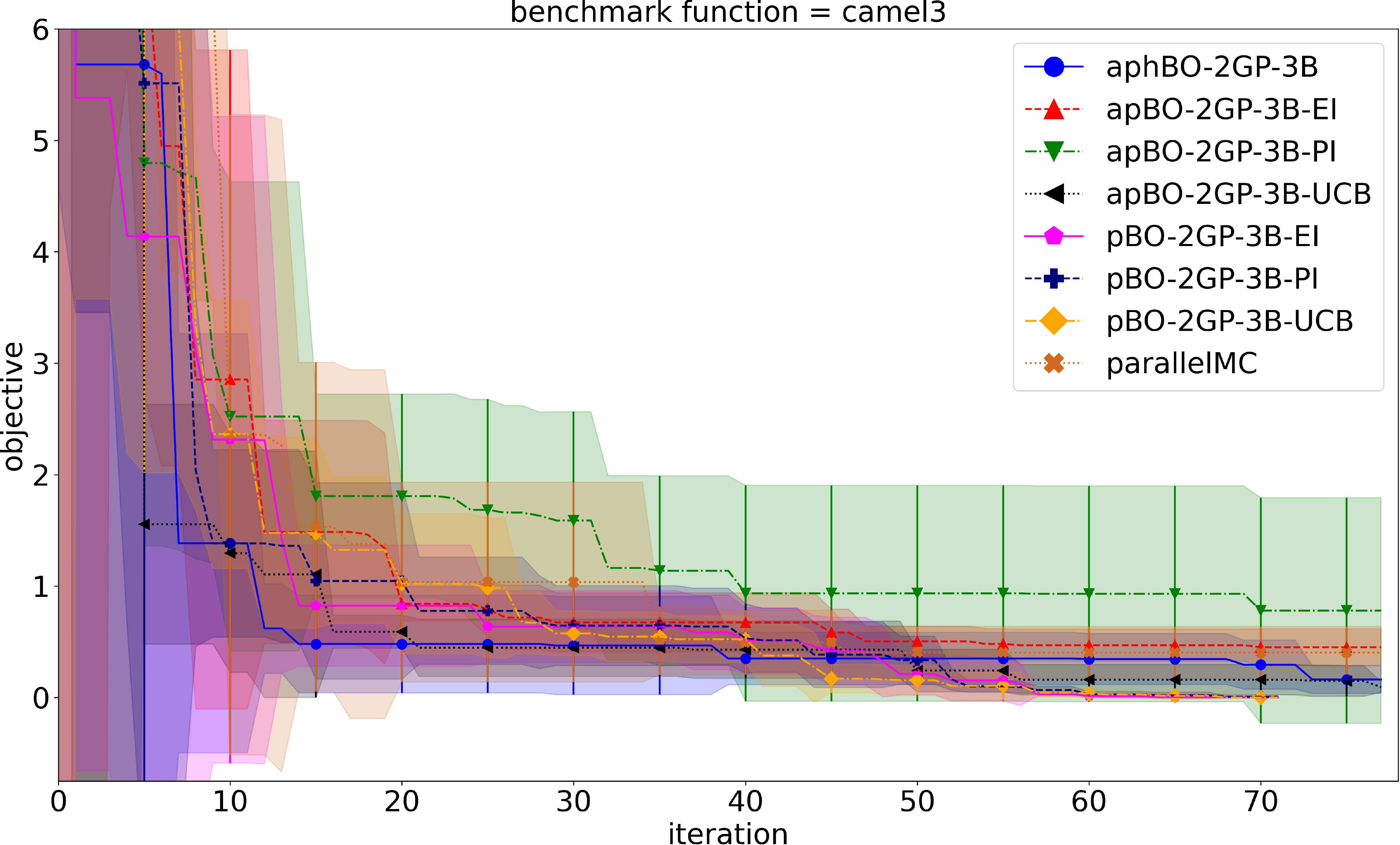}}
% \caption{Benchmark for camel3.}
% \label{fig:bench_camel3}
% \end{figure}
% \vspace{-2.50cm}
\medskip

% \subsection{six-hump camel (2d)}
% \begin{figure}[!htbp]
\centering
\subcaptionbox{Benchmark by physical wall-clock time.
\label{fig:benchByTime_camel6}
}
  [0.425\linewidth]{\includegraphics[width=0.425\textwidth, keepaspectratio]{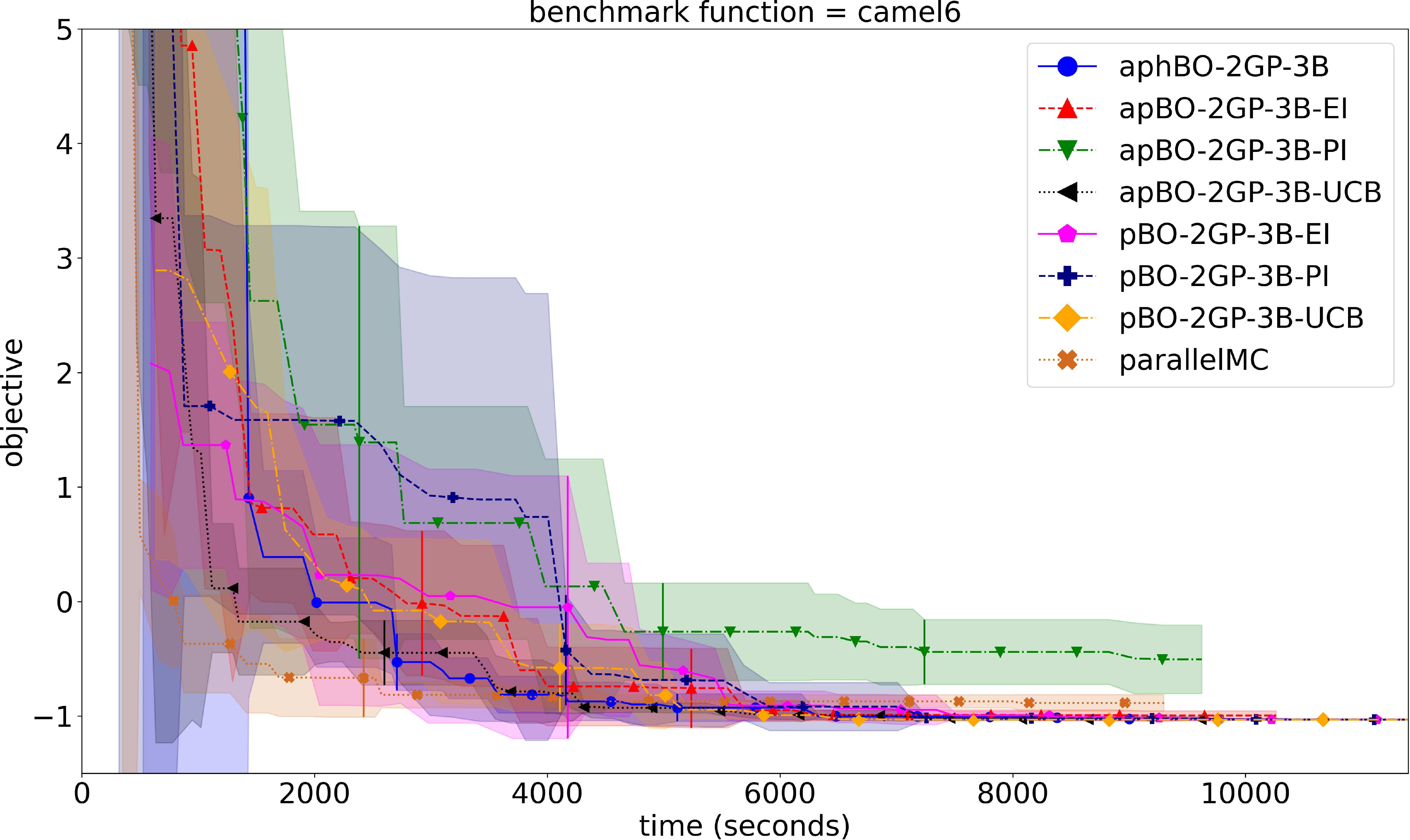}}
\hfill
\subcaptionbox{Benchmark by iteration.
\label{fig:benchByIter_camel6}
}
  [0.425\linewidth]{\includegraphics[width=0.425\textwidth, keepaspectratio]{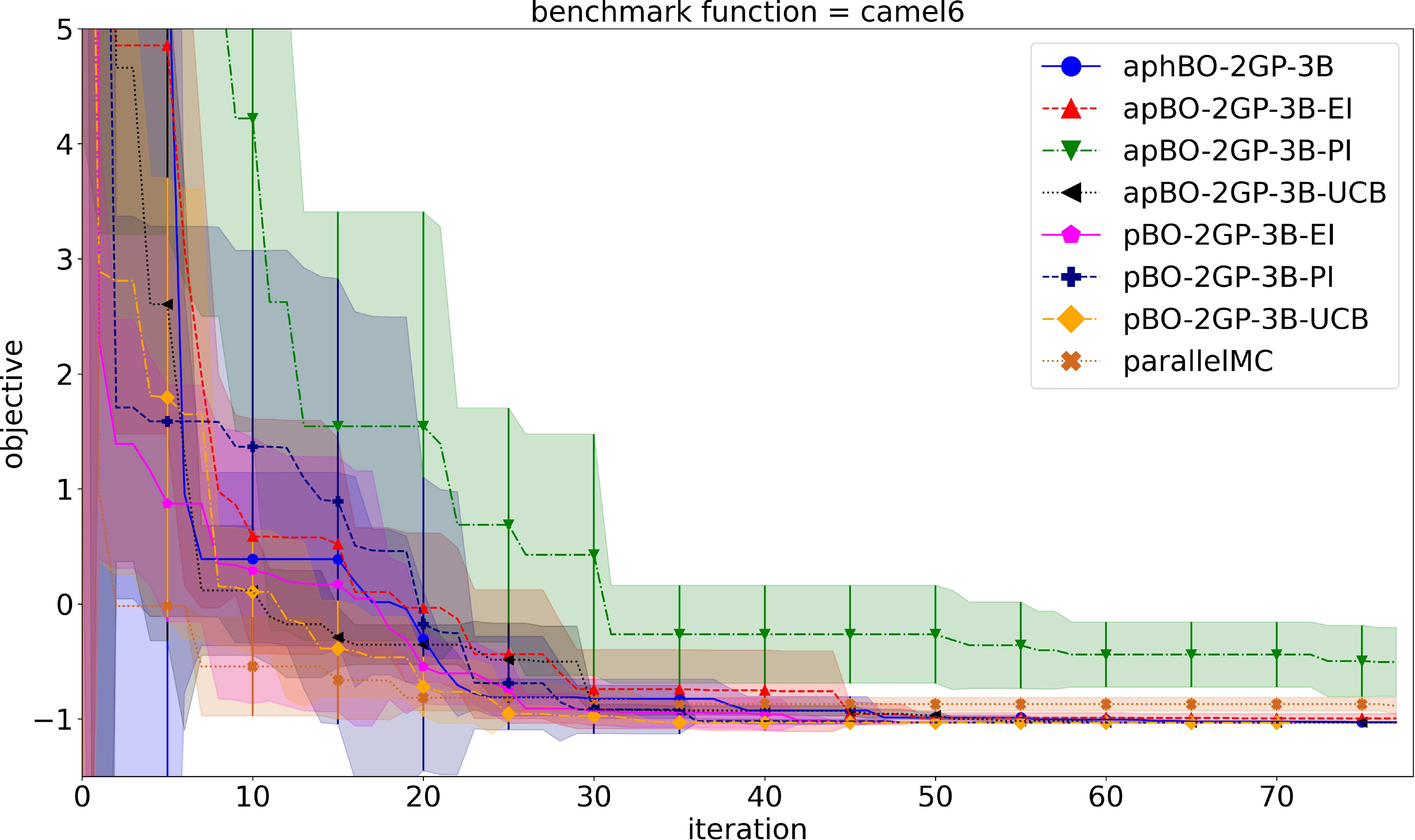}}
% \caption{Benchmark for camel6.}
% \label{fig:bench_camel6}
% \end{figure}
% \vspace{-2.50cm}
\medskip

% \subsection{hartmann (3d)}
% \begin{figure}[!htbp]
\centering
\subcaptionbox{Benchmark by physical wall-clock time.
\label{fig:benchByTime_hart3}
}
  [0.425\linewidth]{\includegraphics[width=0.425\textwidth, keepaspectratio]{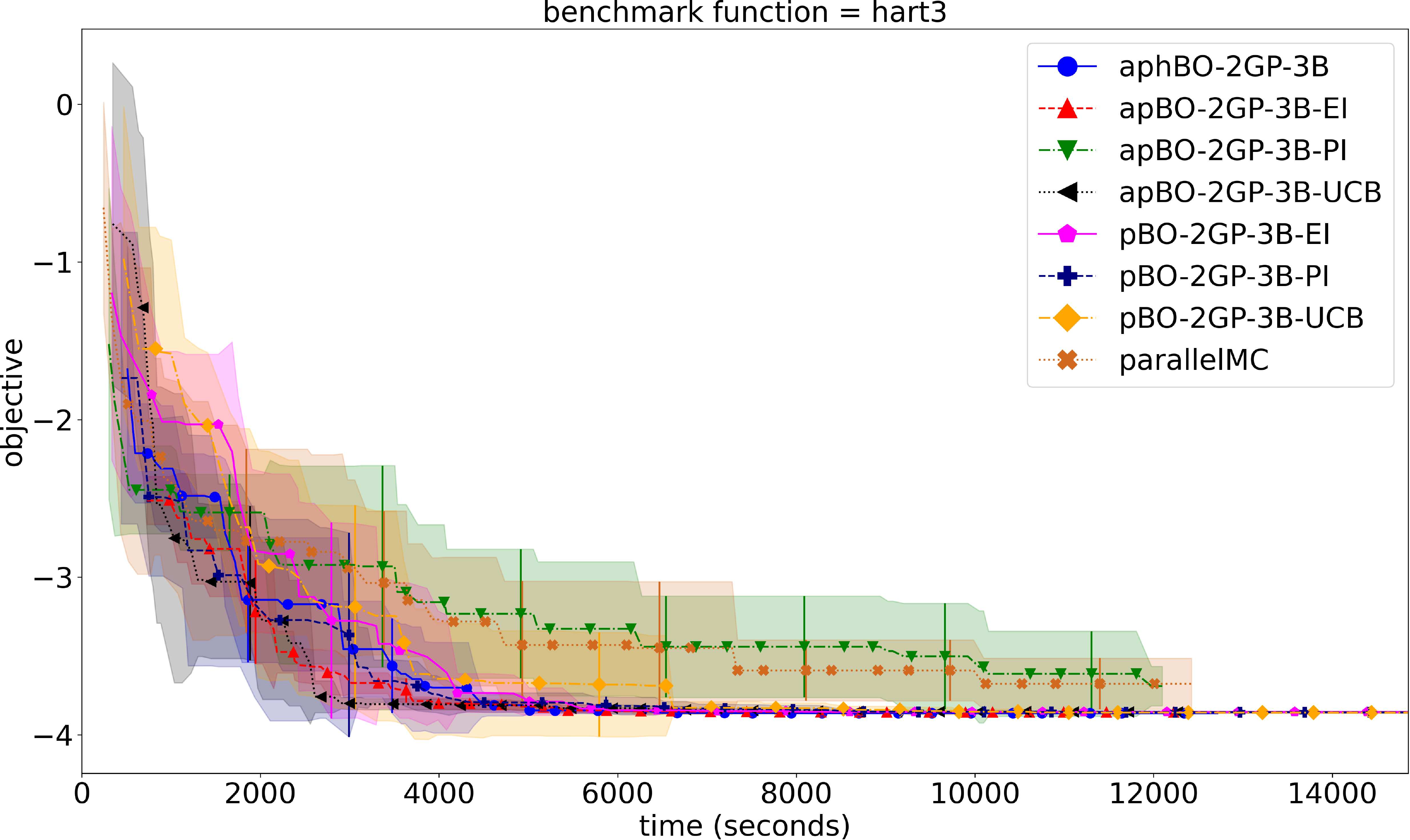}}
\hfill
\subcaptionbox{Benchmark by iteration.
\label{fig:benchByIter_hart3}
}
  [0.425\linewidth]{\includegraphics[width=0.425\textwidth, keepaspectratio]{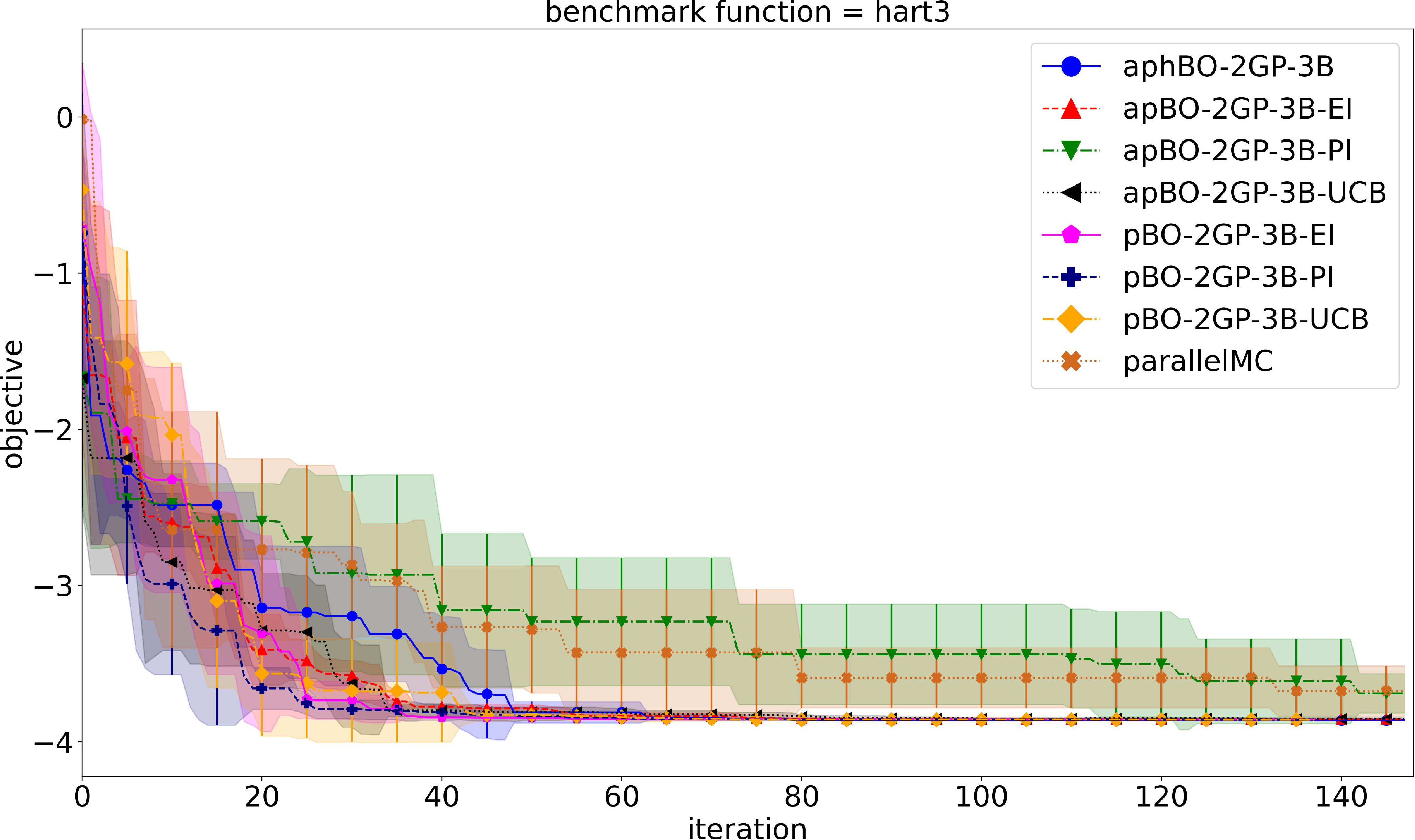}}
% \caption{Benchmark for hart3.}
% \label{fig:bench_hart3}
\end{figure}
% \vspace{-2.50cm}
\medskip

% \subsection{hartmann (4d)}
\begin{figure}[!htbp]\ContinuedFloat
\centering
\subcaptionbox{Benchmark by physical wall-clock time.
\label{fig:benchByTime_hart4}
}
  [0.425\linewidth]{\includegraphics[width=0.425\textwidth, keepaspectratio]{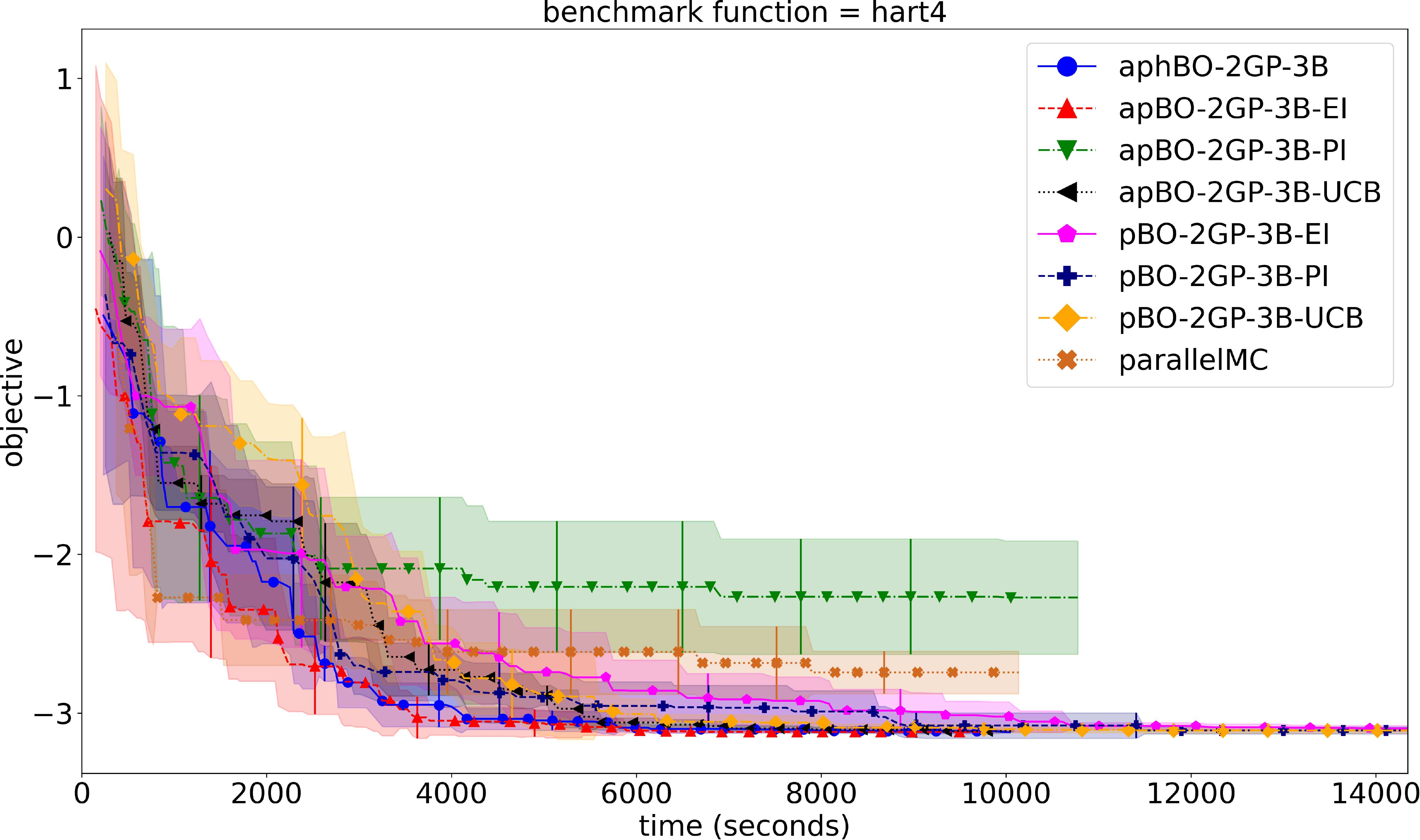}}
\hfill
\subcaptionbox{Benchmark by iteration.
\label{fig:benchByIter_hart4}
}
  [0.425\linewidth]{\includegraphics[width=0.425\textwidth, keepaspectratio]{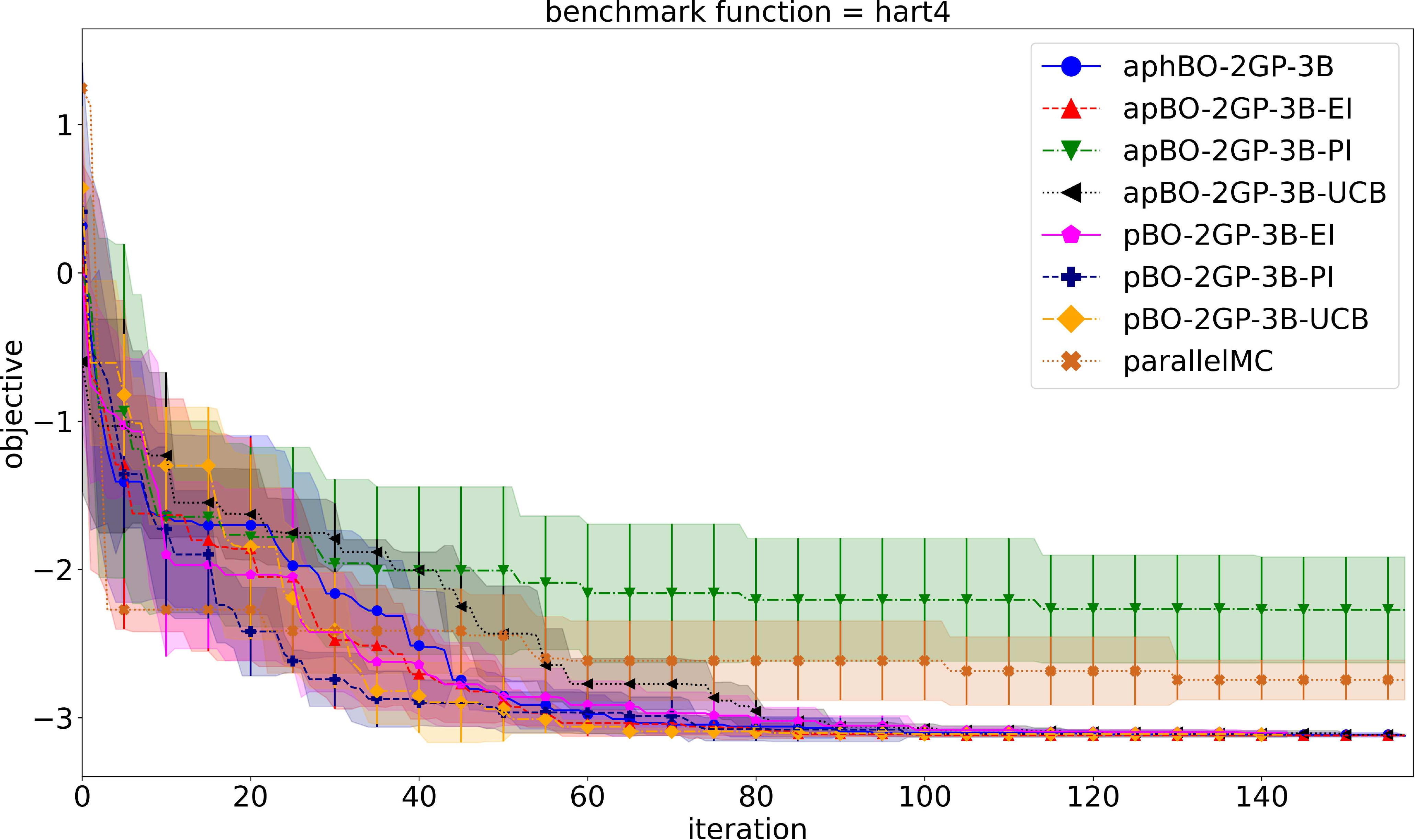}}
% \caption{Benchmark for hart4.}
% \label{fig:bench_hart4}
% \end{figure}
% \vspace{-2.50cm}
\medskip

% \subsection{ackley (free d) (5d)}
% \begin{figure}[!htbp]
\centering
\subcaptionbox{Benchmark by physical wall-clock time.
\label{fig:benchByTime_ackley}
}
  [0.425\linewidth]{\includegraphics[width=0.425\textwidth, keepaspectratio]{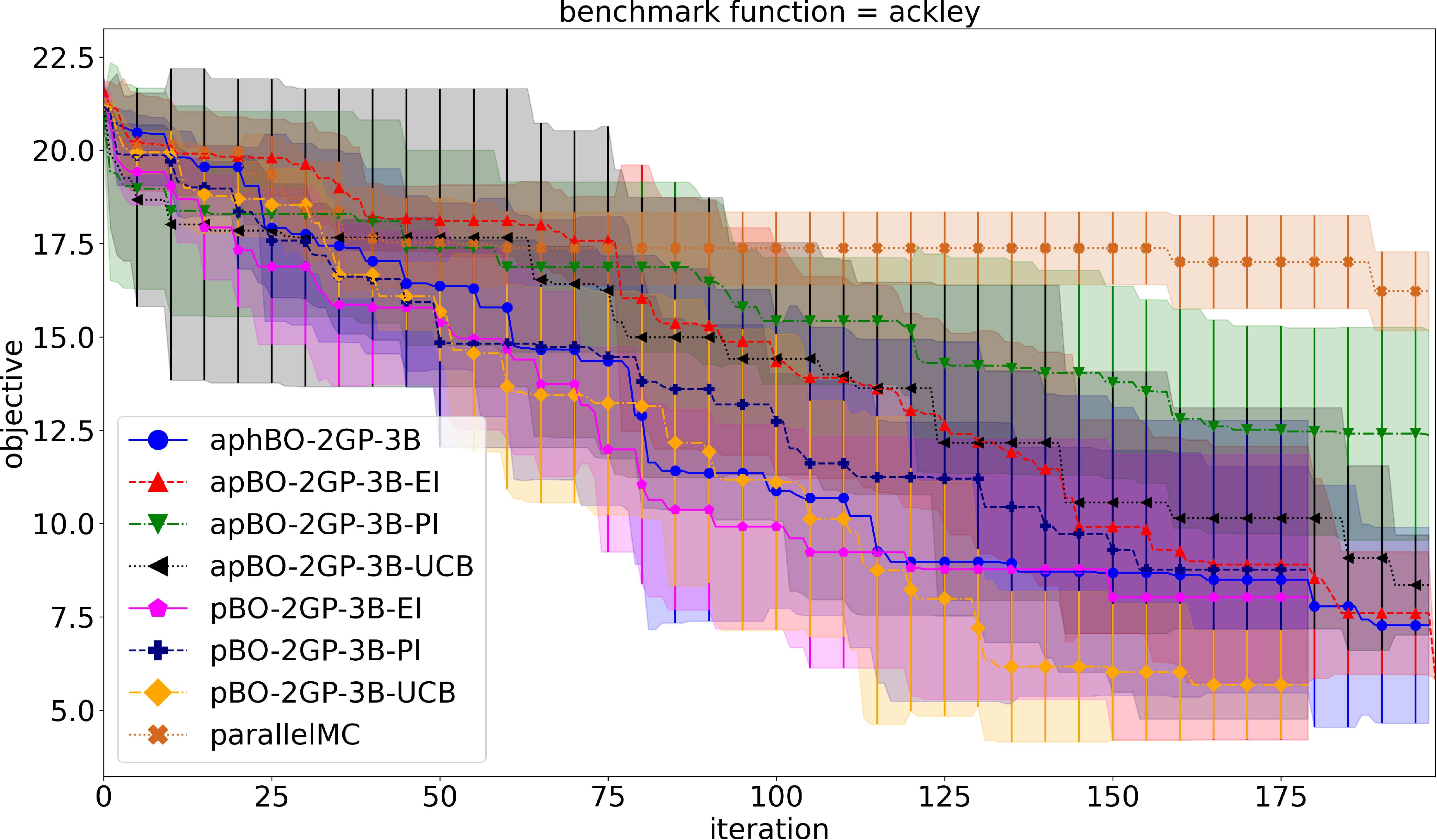}}
\hfill
\subcaptionbox{Benchmark by iteration.
\label{fig:benchByIter_ackley}
}
  [0.425\linewidth]{\includegraphics[width=0.425\textwidth, keepaspectratio]{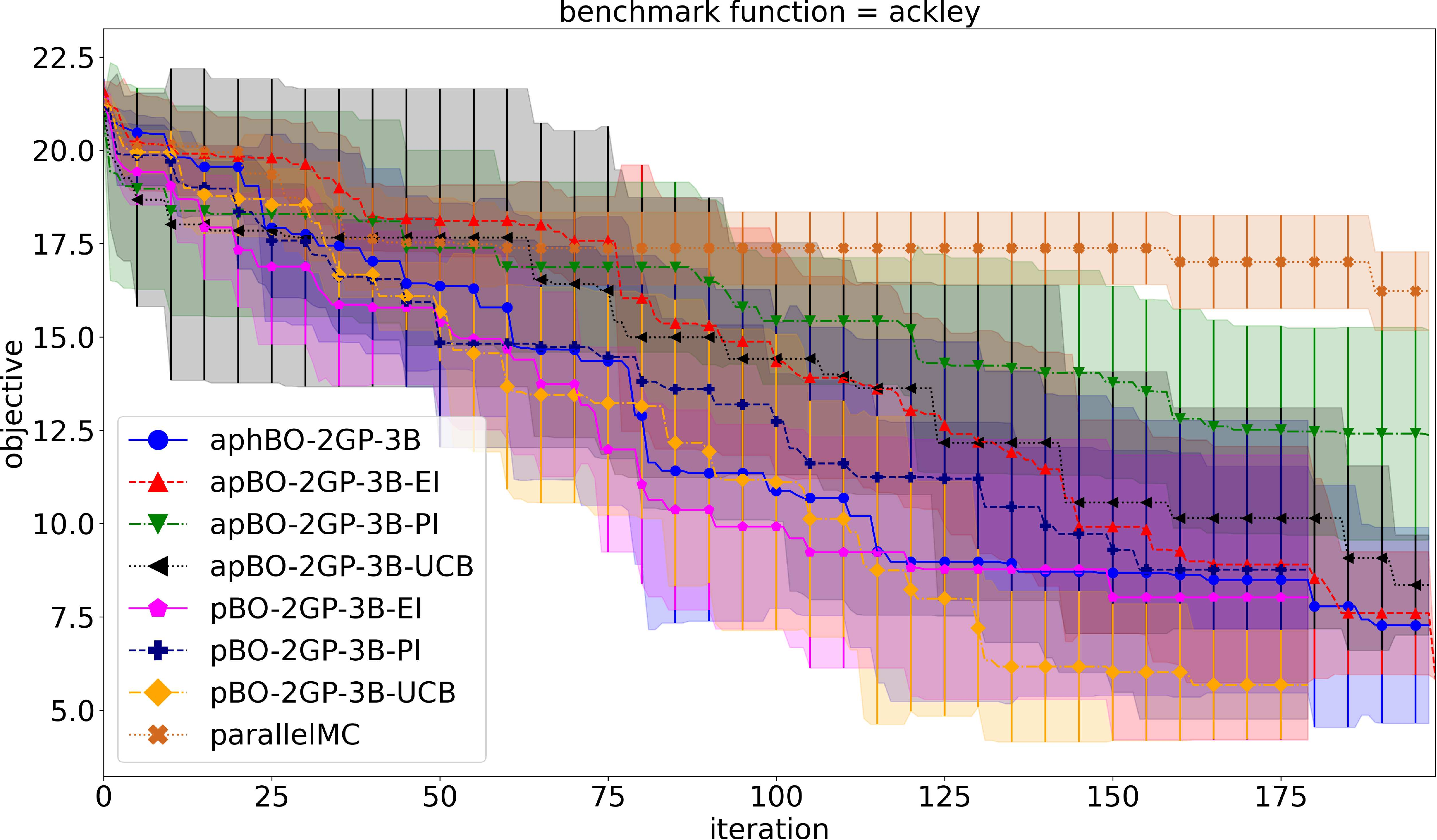}}
% \caption{Benchmark for ackley.}
% \label{fig:bench_ackley}
% \end{figure}
% \vspace{-2.50cm}
\medskip

% \subsection{hartmann (6d)}
% \begin{figure}[!htbp]
\centering
\subcaptionbox{Benchmark by physical wall-clock time.
\label{fig:benchByTime_hart6}
}
  [0.425\linewidth]{\includegraphics[width=0.425\textwidth, keepaspectratio]{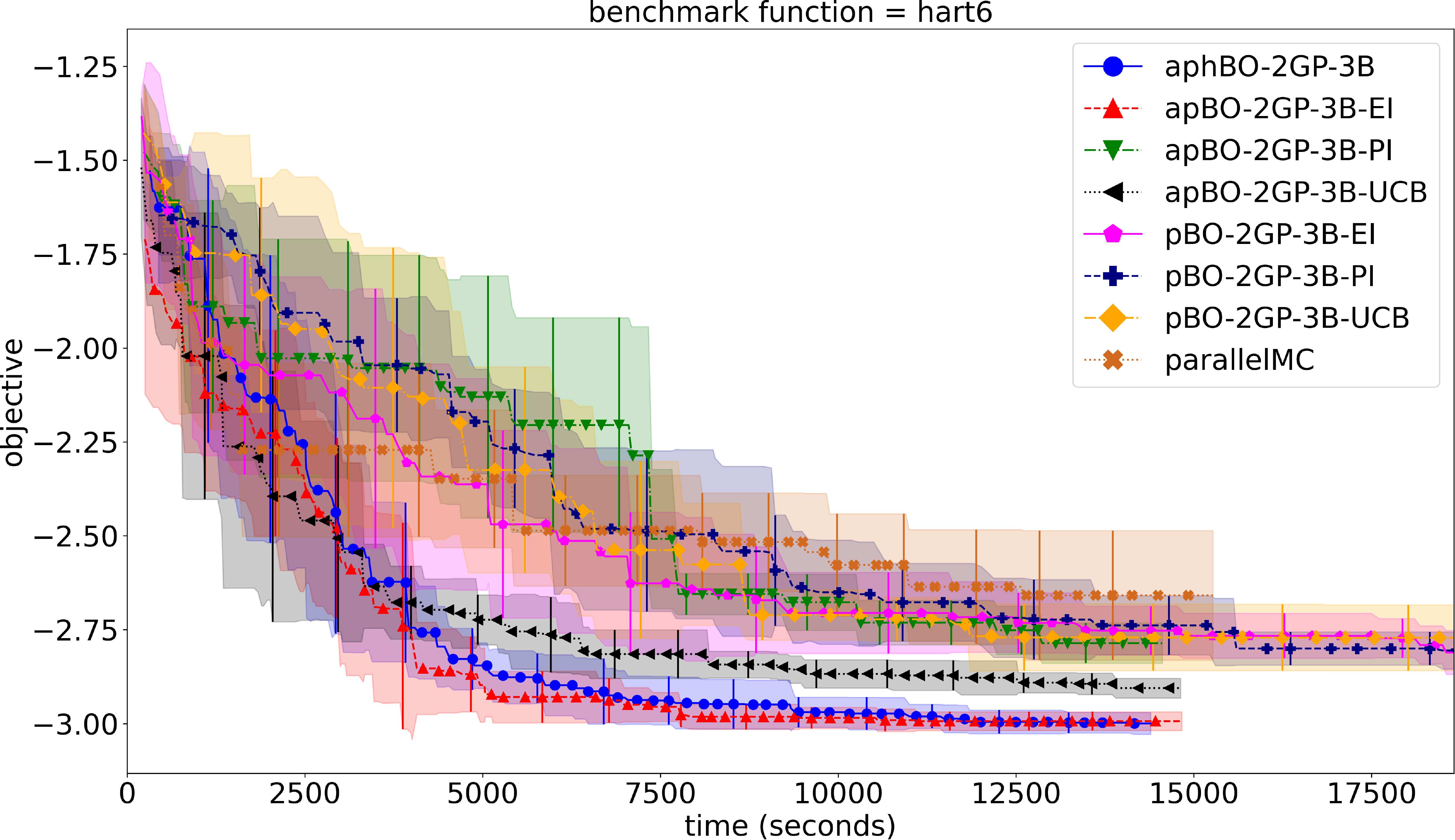}}
\hfill
\subcaptionbox{Benchmark by iteration.
\label{fig:benchByIter_hart6}
}
  [0.425\linewidth]{\includegraphics[width=0.425\textwidth, keepaspectratio]{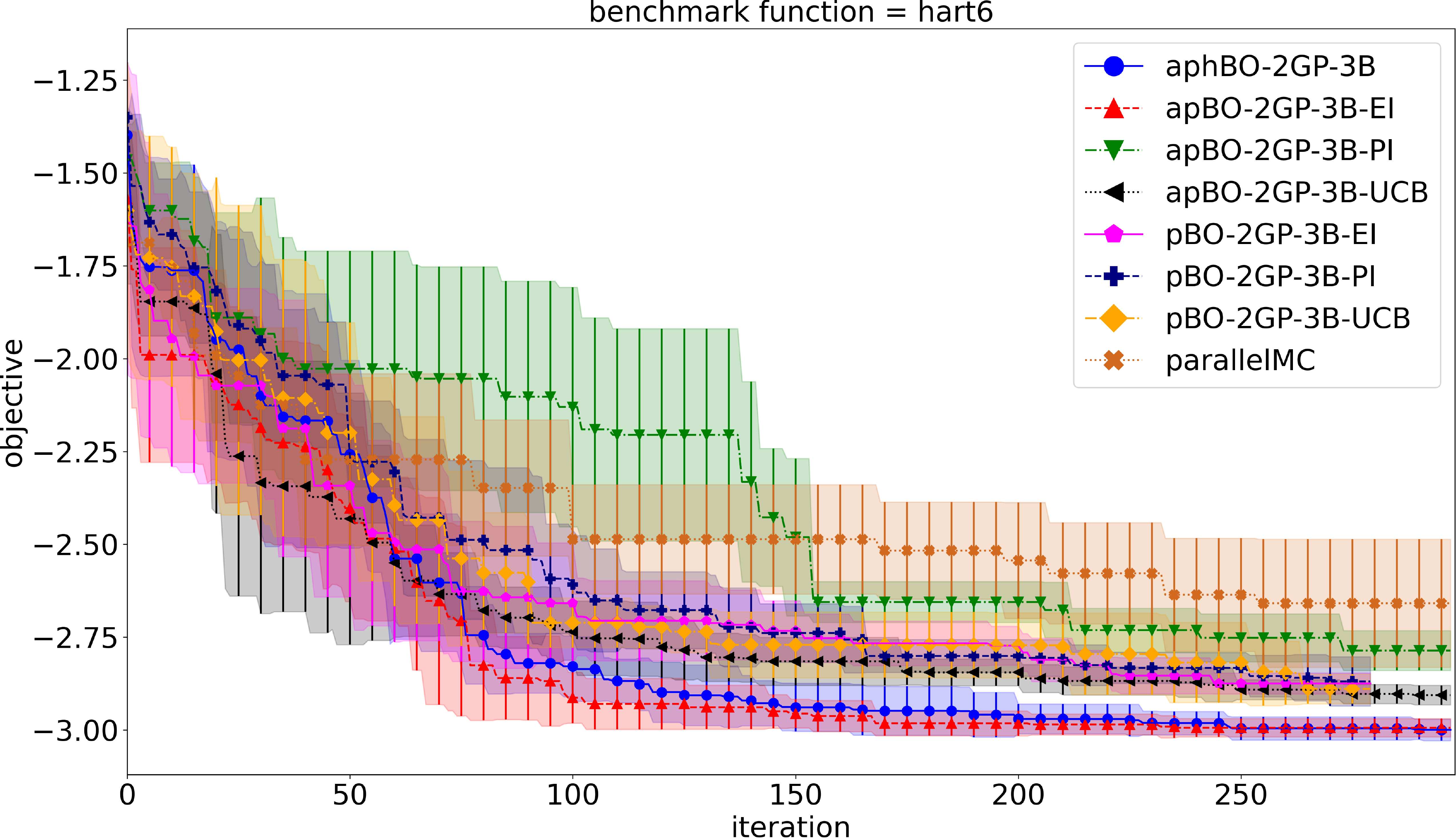}}
% \caption{Benchmark for hart6.}
% \label{fig:bench_hart6}
% \end{figure}
% \vspace{-2.50cm}
\medskip

% \subsection{michalewicz (free d) (10d)}
% \begin{figure}[!htbp]
\centering
\subcaptionbox{Benchmark by physical wall-clock time.
\label{fig:benchByTime_michal}
}
  [0.425\linewidth]{\includegraphics[width=0.425\textwidth, keepaspectratio]{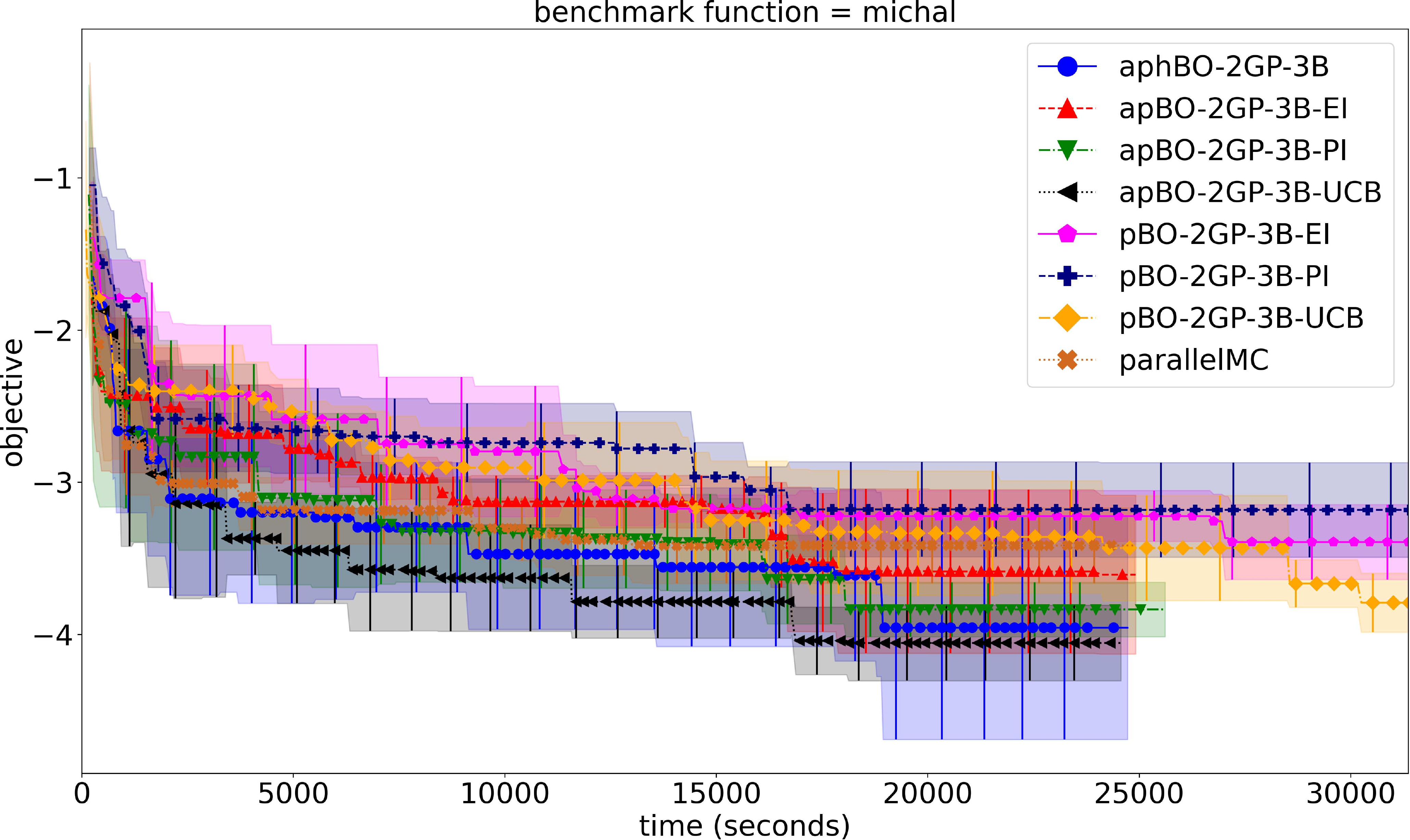}}
\hfill
\subcaptionbox{Benchmark by iteration.
\label{fig:benchByIter_michal}
}
  [0.425\linewidth]{\includegraphics[width=0.425\textwidth, keepaspectratio]{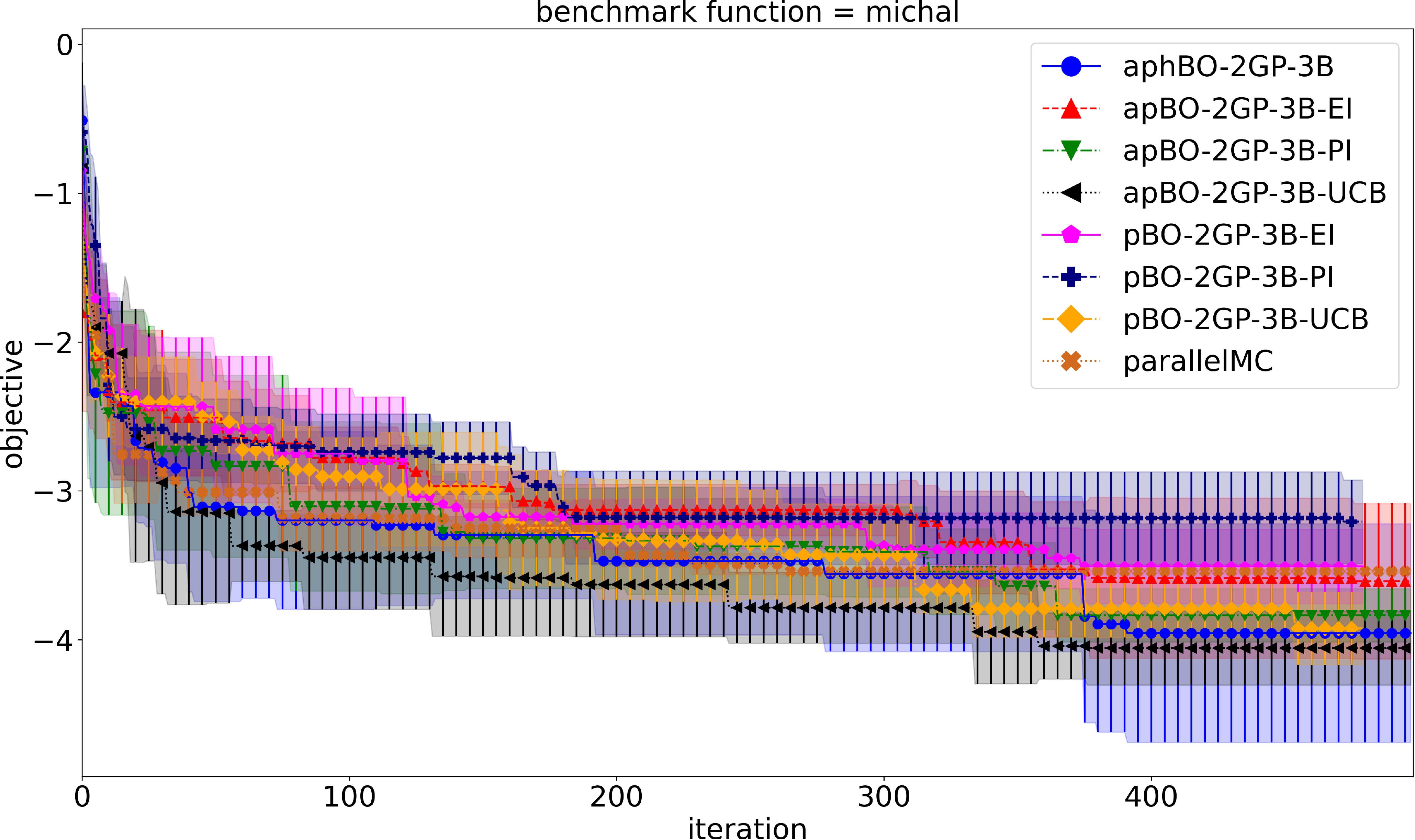}}
\caption{Benchmark for michal.}
\label{fig:bench_michal}
\end{figure}
% \vspace{-2.50cm}

% \subsection{perm0db (free d) (80d)}
\begin{figure}[!htbp]\ContinuedFloat
\centering
\subcaptionbox{Benchmark by physical wall-clock time.
\label{fig:benchByTime_perm0db}
}
  [0.425\linewidth]{\includegraphics[width=0.425\textwidth, keepaspectratio]{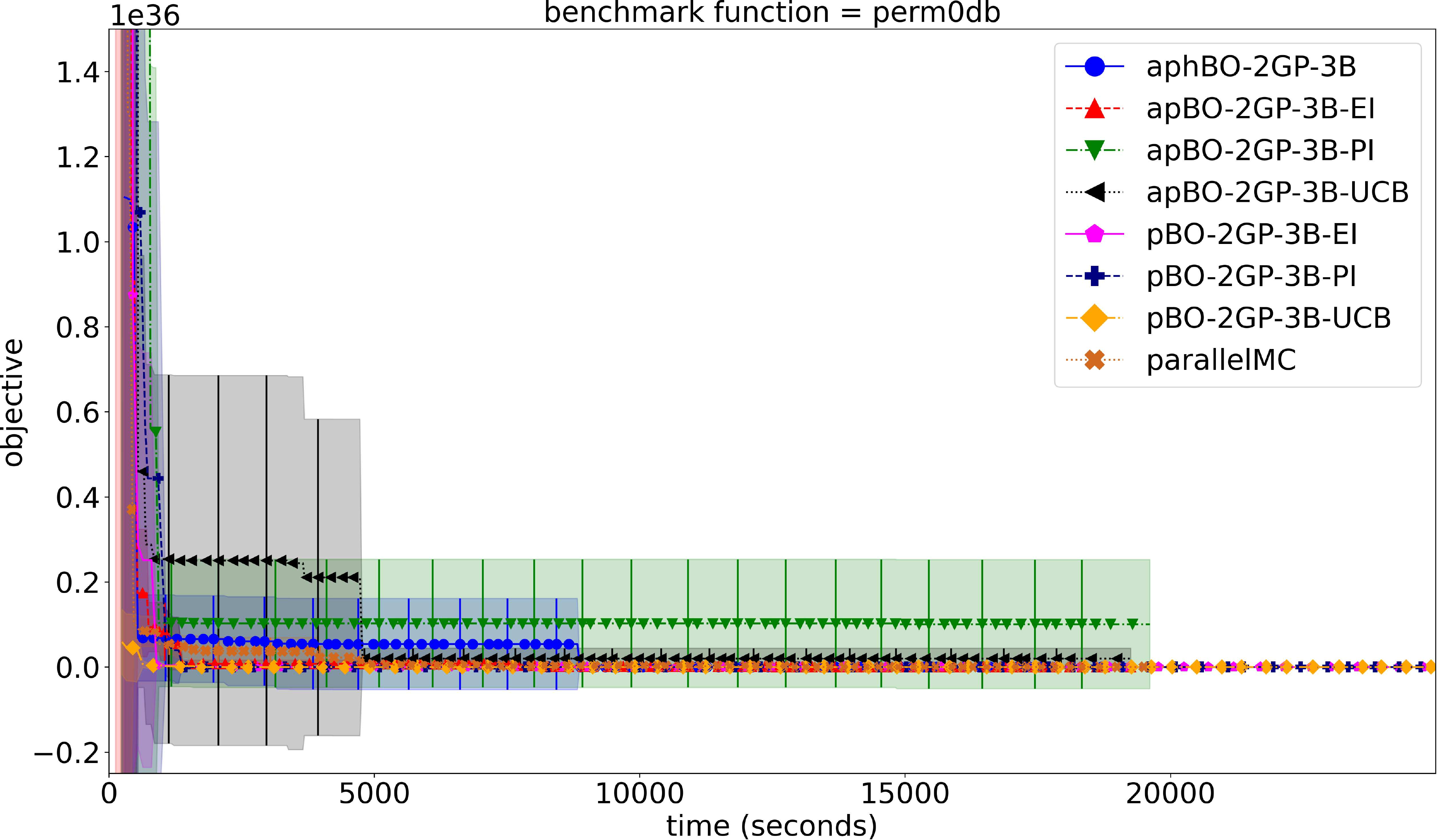}}
\hfill
\subcaptionbox{Benchmark by iteration.
\label{fig:benchByIter_perm0db}
}
  [0.425\linewidth]{\includegraphics[width=0.425\textwidth, keepaspectratio]{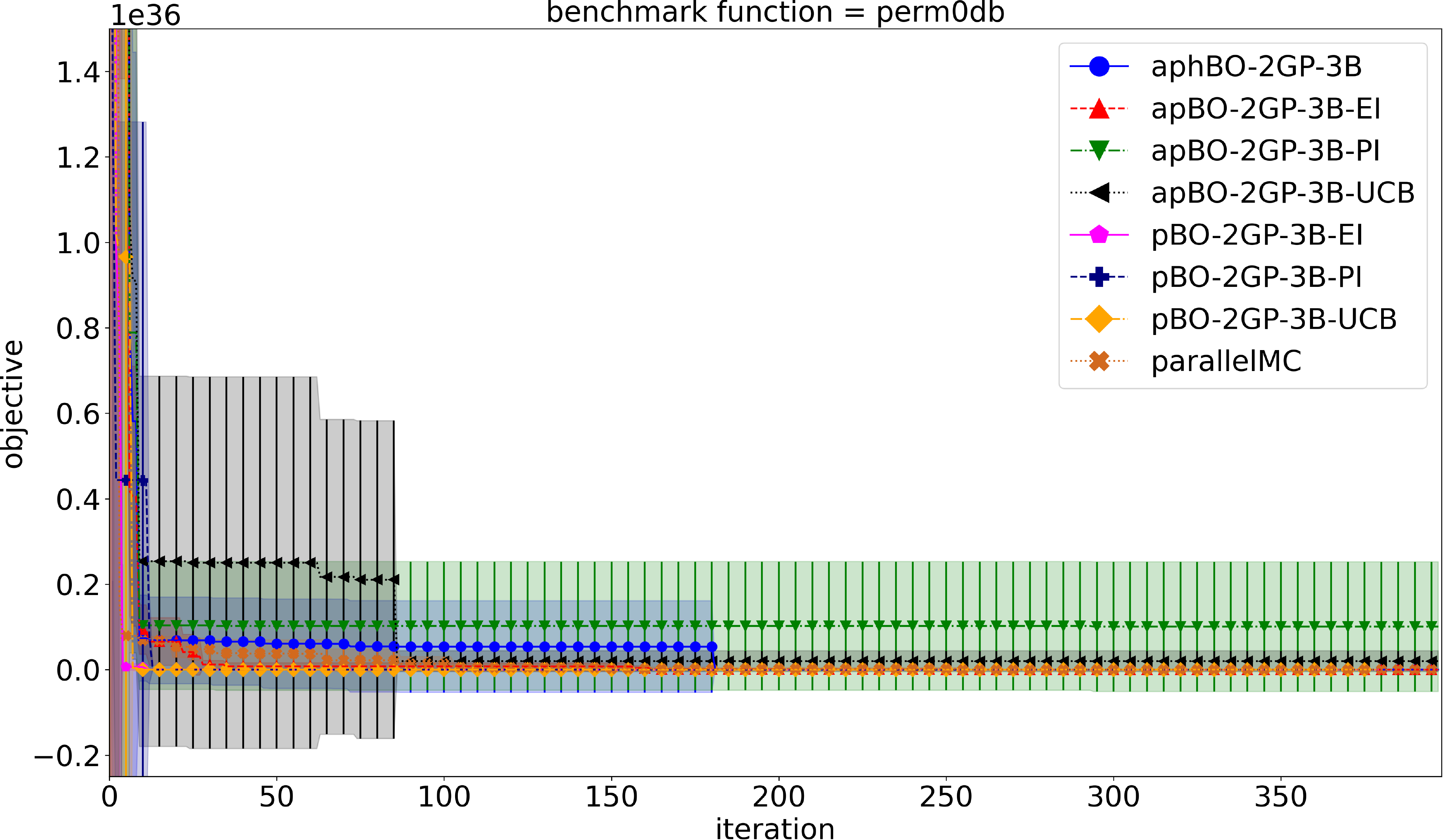}}
% \caption{Benchmark for perm0db.}
% \label{fig:bench_perm0db}
% \end{figure}
% \vspace{-2.50cm}
\medskip

% \subsection{rosenbrock (20d)}
% \begin{figure}[!htbp]
\centering
\subcaptionbox{Benchmark by physical wall-clock time.
\label{fig:benchByTime_rosen}
}
  [0.425\linewidth]{\includegraphics[width=0.425\textwidth, keepaspectratio]{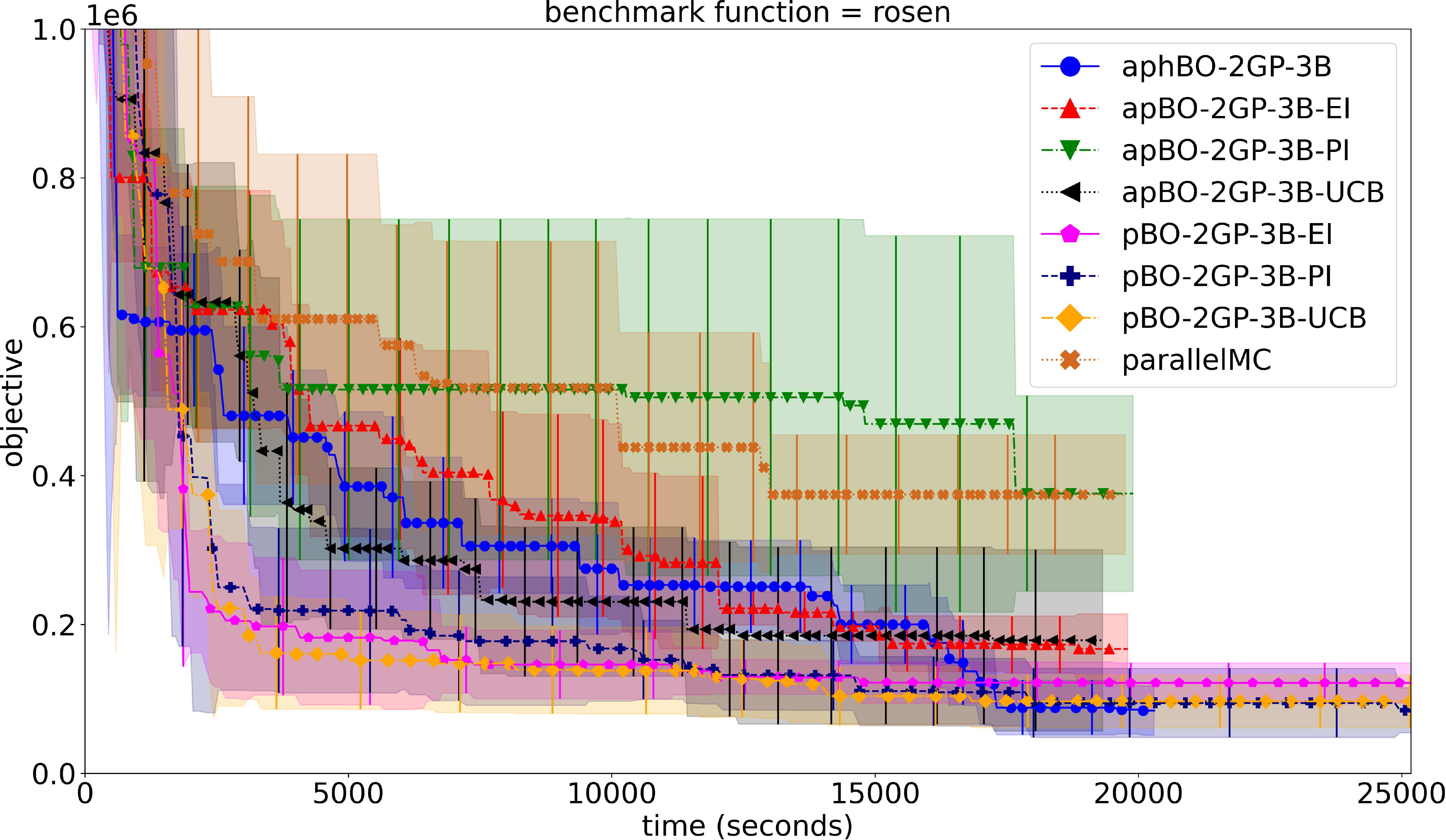}}
\hfill
\subcaptionbox{Benchmark by iteration.
\label{fig:benchByIter_rosen}
}
  [0.425\linewidth]{\includegraphics[width=0.425\textwidth, keepaspectratio]{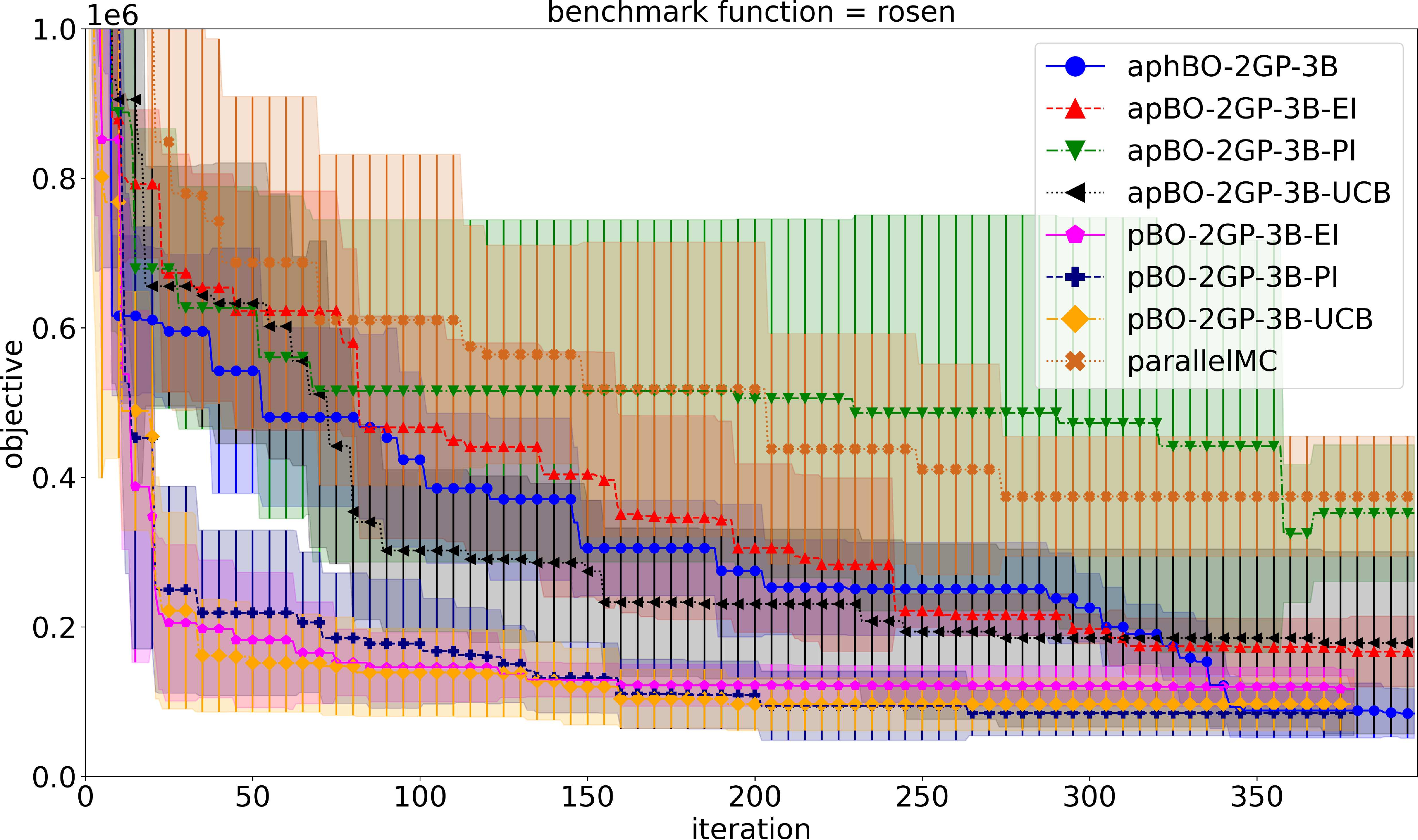}}
% \caption{Benchmark for rosen.}
% \label{fig:bench_rosen}
% \end{figure}
% \vspace{-2.50cm}
\medskip

% \subsection{dixon-price (free d) (25d)}
% \begin{figure}[!htbp]
\centering
\subcaptionbox{Benchmark by physical wall-clock time.
\label{fig:benchByTime_dixonpr}
}
  [0.425\linewidth]{\includegraphics[width=0.425\textwidth, keepaspectratio]{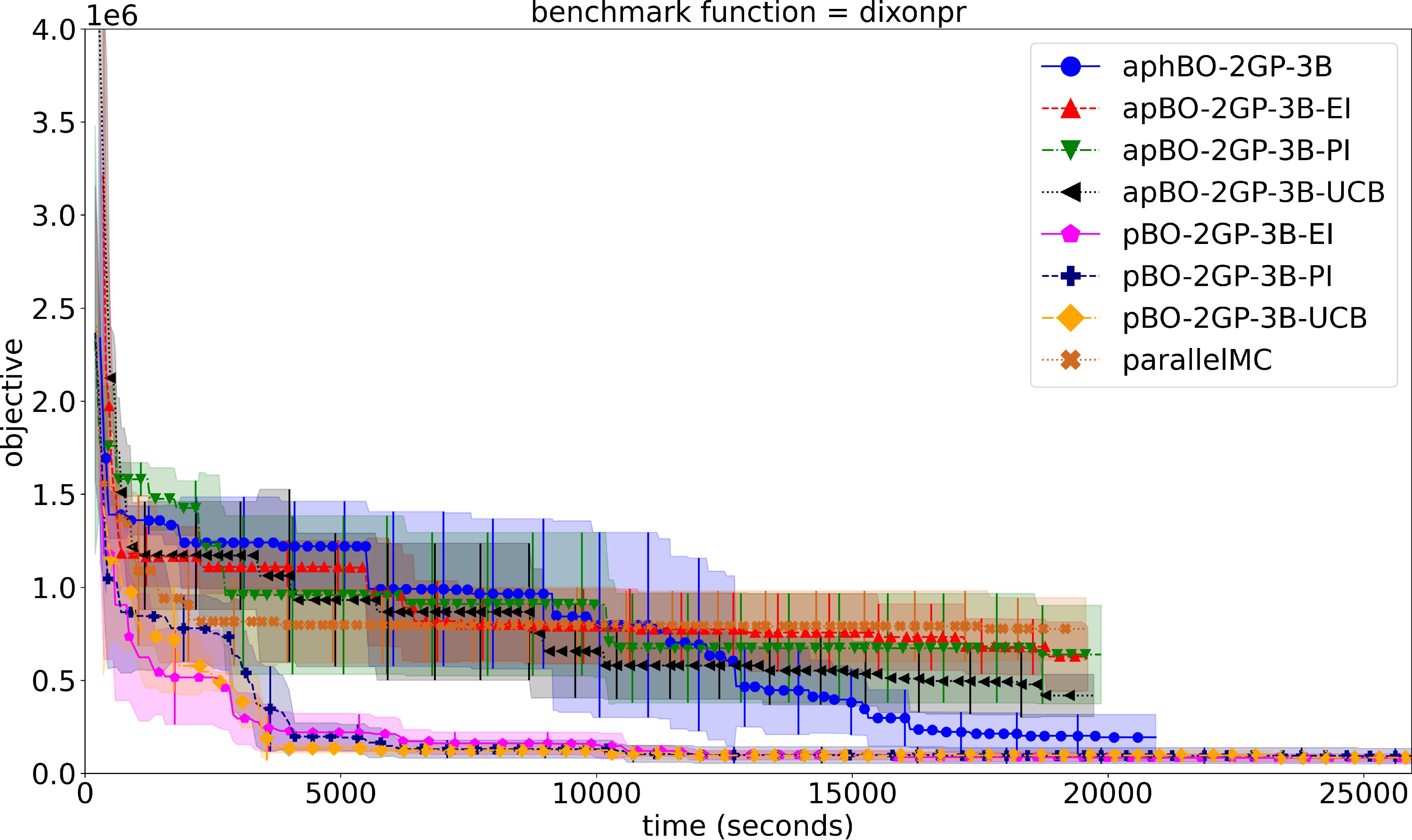}}
\hfill
\subcaptionbox{Benchmark by iteration.
\label{fig:benchByIter_dixonpr}
}
  [0.425\linewidth]{\includegraphics[width=0.425\textwidth, keepaspectratio]{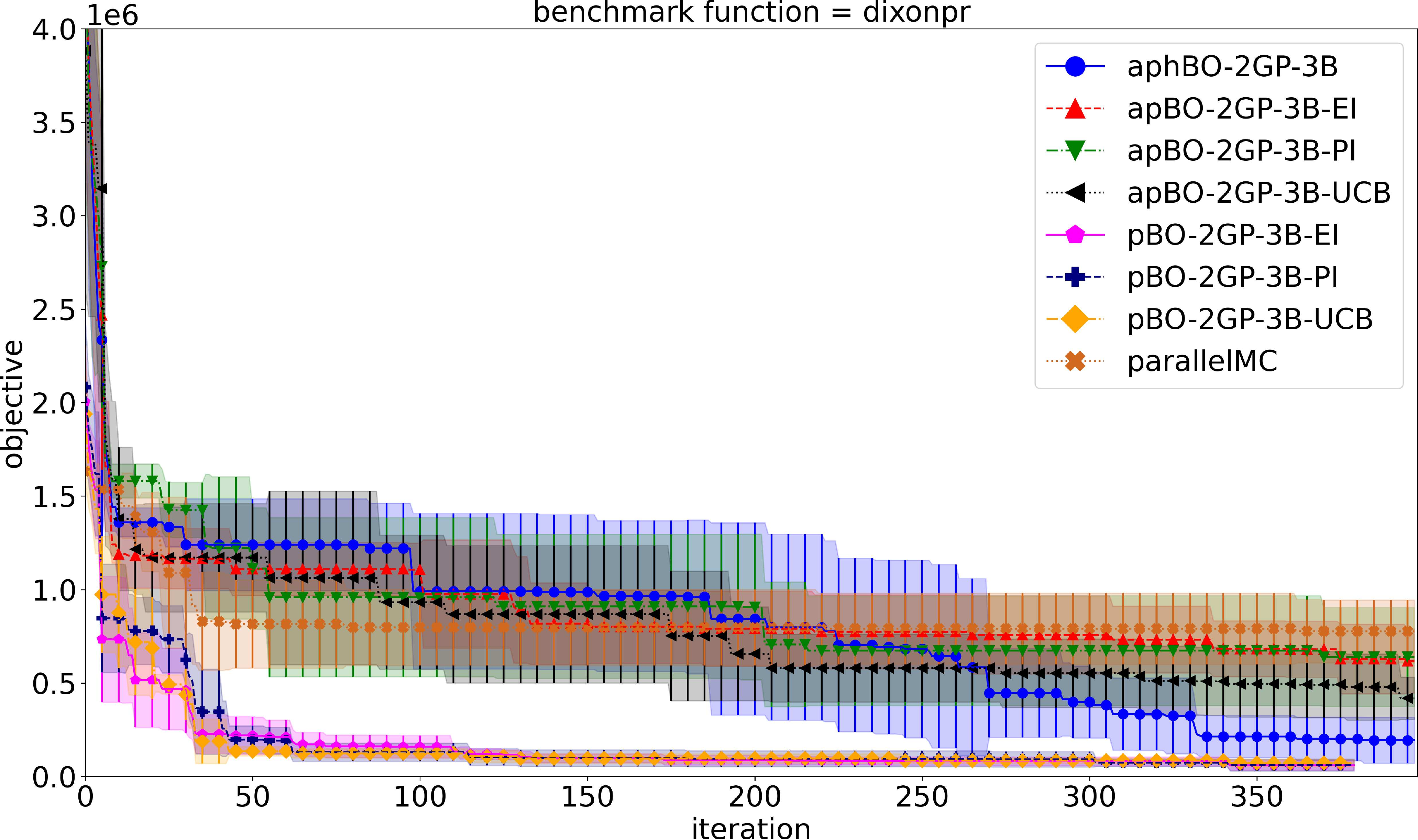}}
% \caption{Benchmark for dixonpr.}
% \label{fig:bench_dixonpr}
% \end{figure}
% \vspace{-2.50cm}
\medskip

% \subsection{trid (free d) (30d)}
% \begin{figure}[!htbp]
\centering
\subcaptionbox{Benchmark by physical wall-clock time.
\label{fig:benchByTime_trid}
}
  [0.425\linewidth]{\includegraphics[width=0.425\textwidth, keepaspectratio]{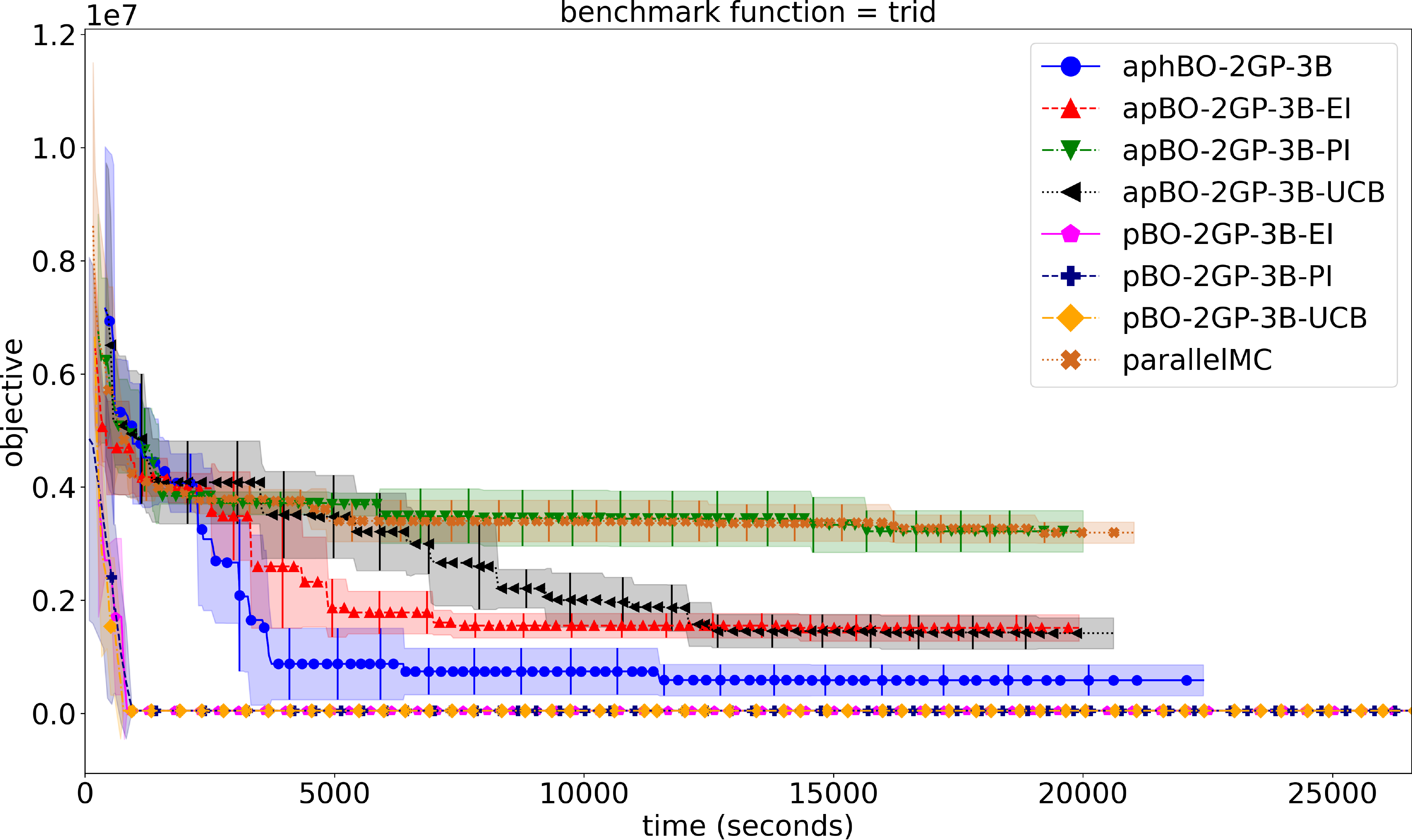}}
\hfill
\subcaptionbox{Benchmark by iteration.
\label{fig:benchByIter_trid}
}
  [0.425\linewidth]{\includegraphics[width=0.425\textwidth, keepaspectratio]{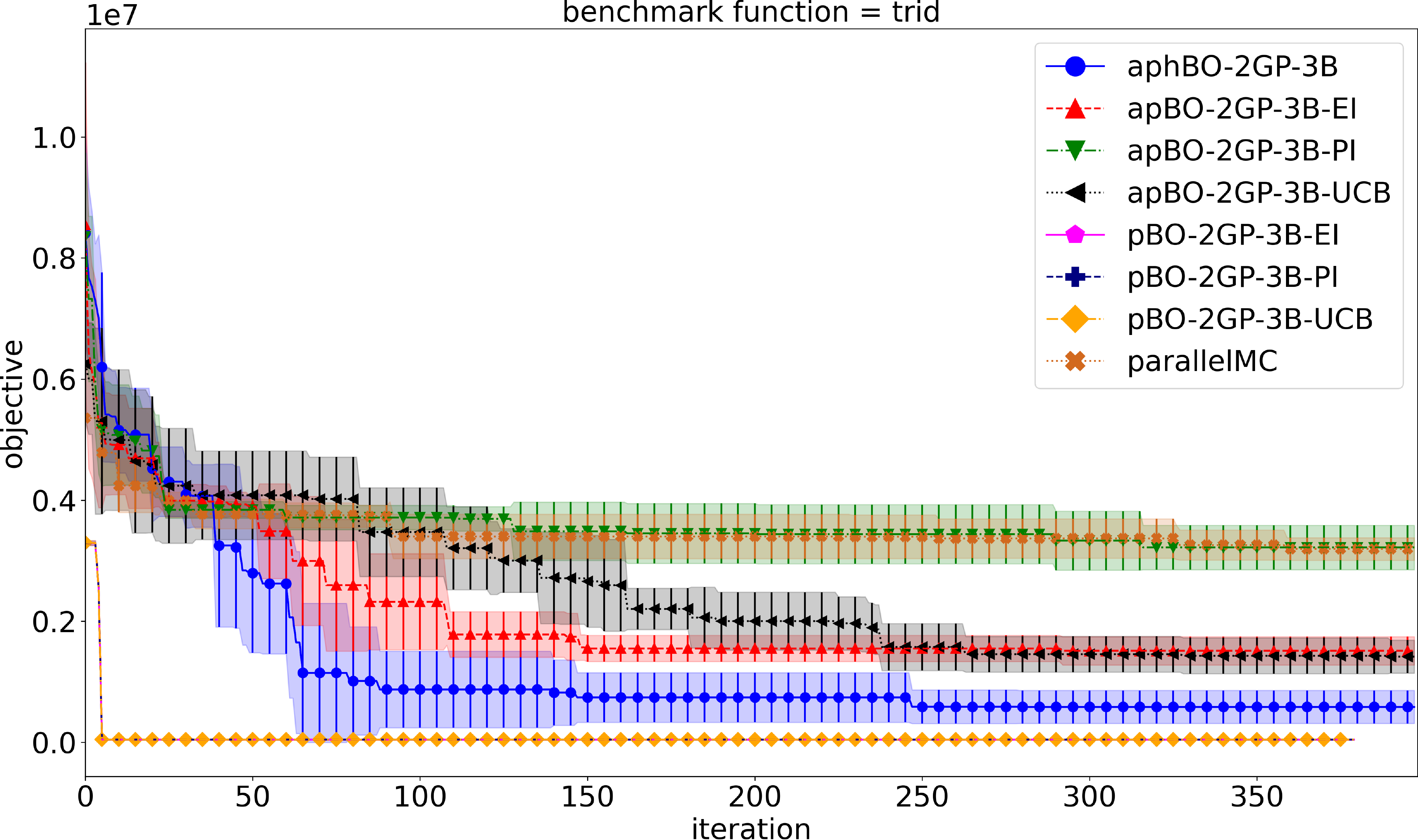}}
\caption{Benchmark for trid.}
\label{fig:bench_trid}
\end{figure}
% \vspace{-2.50cm}

% \subsection{sumsqu (free d) (40d)}
\begin{figure}[!htbp]
\centering
\subcaptionbox{Benchmark by physical wall-clock time.
\label{fig:benchByTime_sumsqu}
}
  [0.425\linewidth]{\includegraphics[width=0.425\textwidth, keepaspectratio]{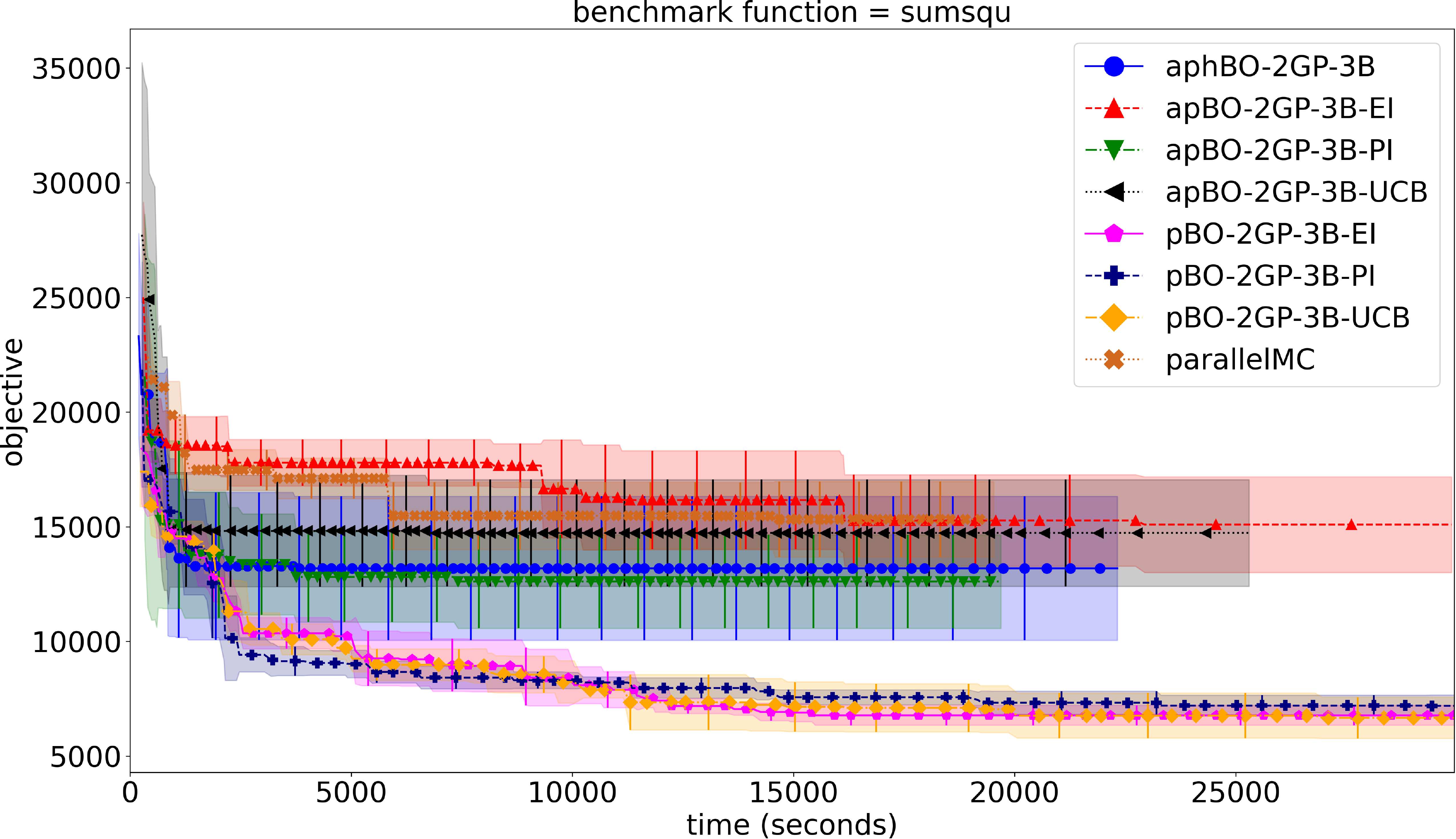}}
\hfill
\subcaptionbox{Benchmark by iteration.
\label{fig:benchByIter_sumsqu}
}
  [0.425\linewidth]{\includegraphics[width=0.425\textwidth, keepaspectratio]{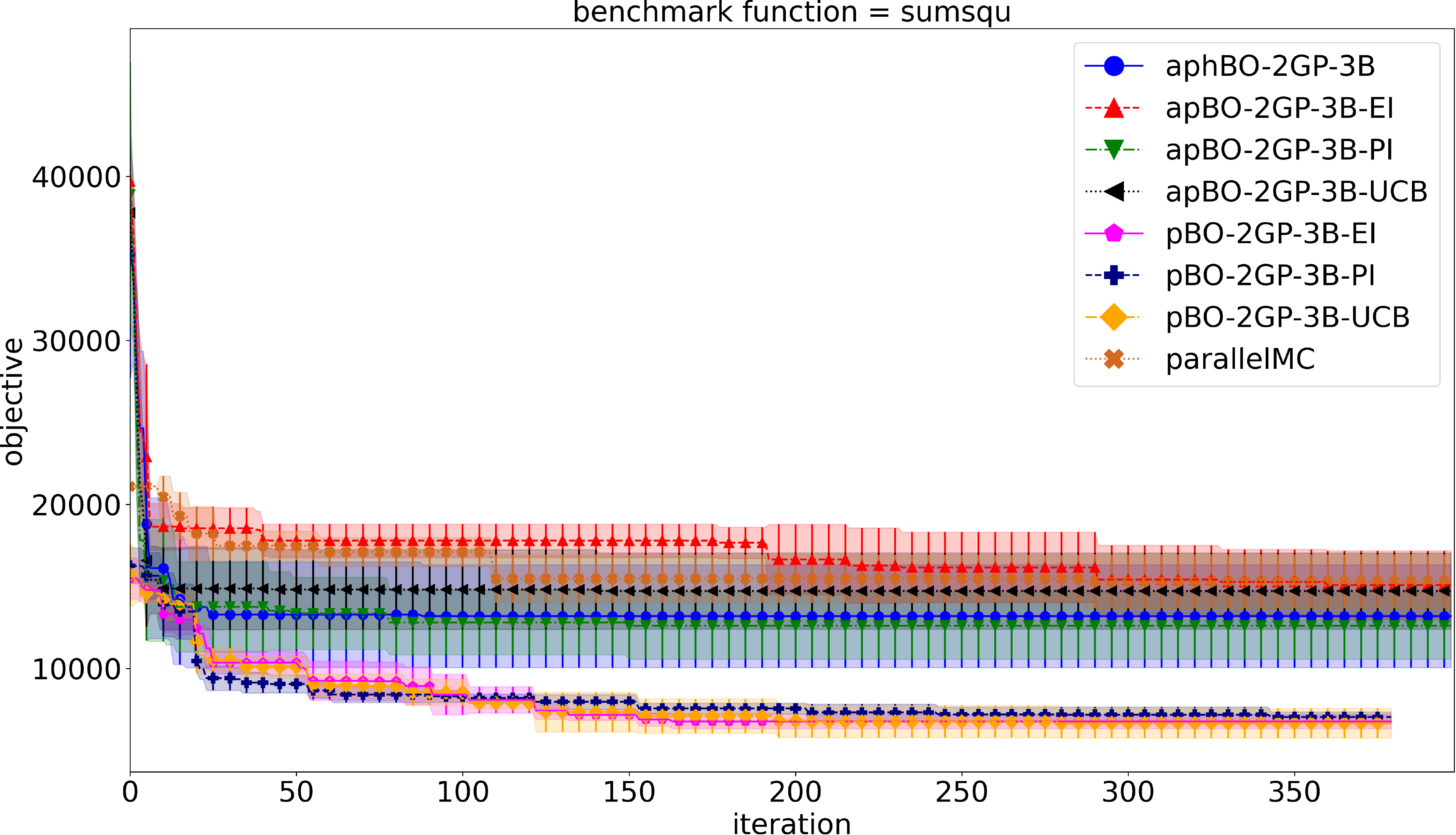}}
% \caption{Benchmark for sumsqu.}
% \label{fig:bench_sumsqu}
% \end{figure}
% \vspace{-2.50cm}
\medskip

% \subsection{sumpow (free d) (50d)}
% \begin{figure}[!htbp]
\centering
\subcaptionbox{Benchmark by physical wall-clock time.
\label{fig:benchByTime_sumpow}
}
  [0.425\linewidth]{\includegraphics[width=0.425\textwidth, keepaspectratio]{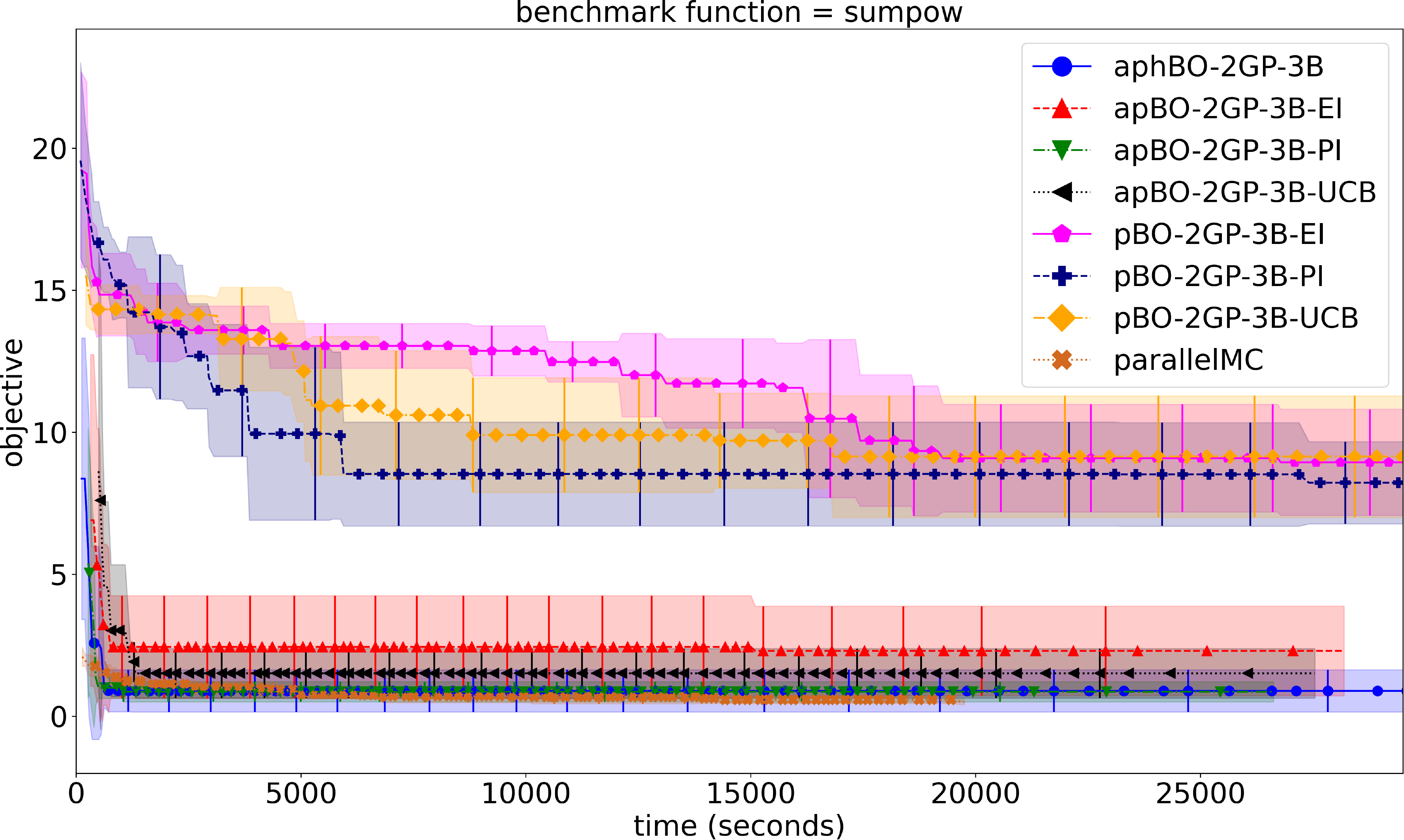}}
\hfill
\subcaptionbox{Benchmark by iteration.
\label{fig:benchByIter_sumpow}
}
  [0.425\linewidth]{\includegraphics[width=0.425\textwidth, keepaspectratio]{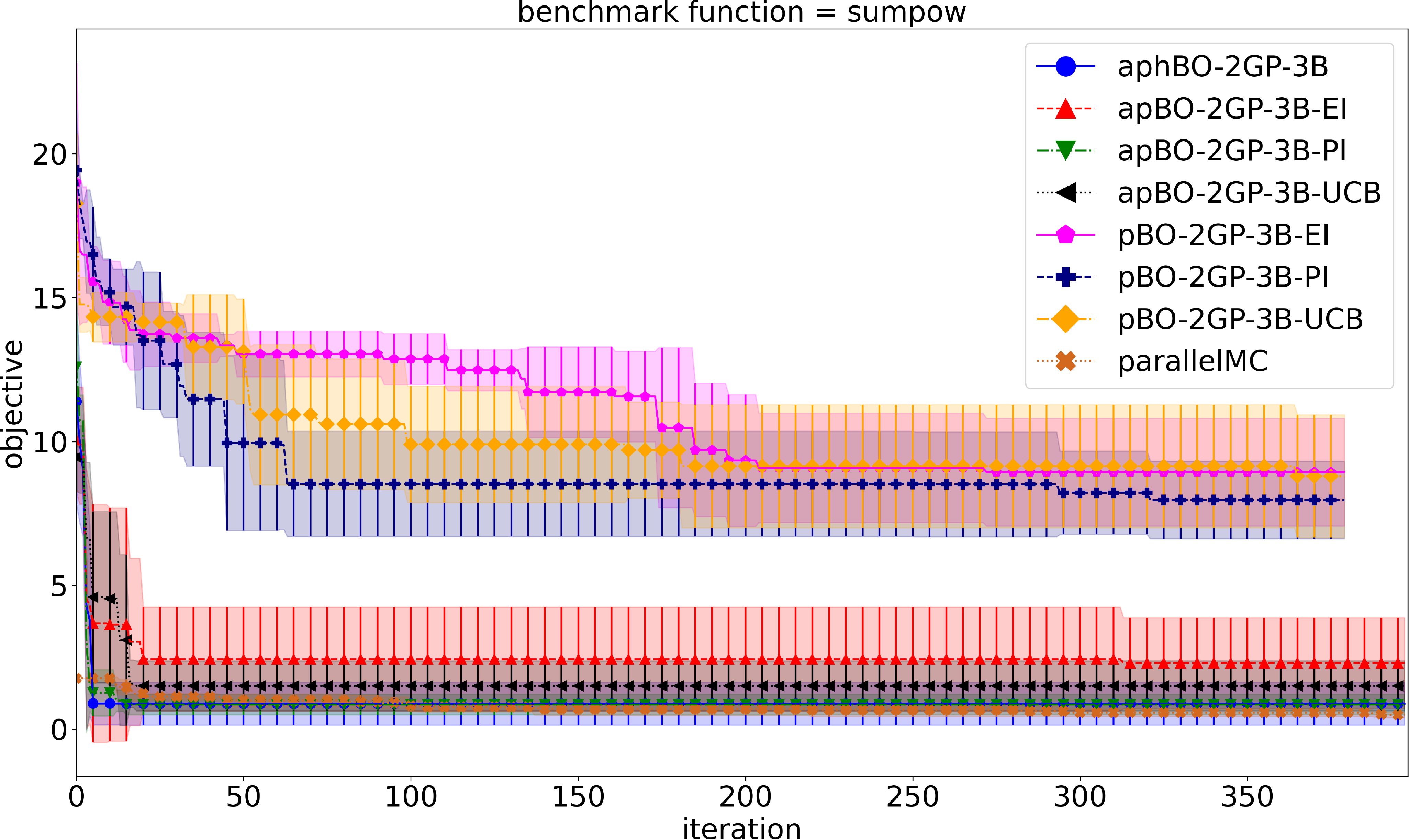}}
% \caption{Benchmark for sumpow.}
% \label{fig:bench_sumpow}
% \end{figure}
% \vspace{-2.50cm}
\medskip

% \subsection{spheref (free d) (60d)}
% \begin{figure}[!htbp]
\centering
\subcaptionbox{Benchmark by physical wall-clock time.
\label{fig:benchByTime_spheref}
}
  [0.425\linewidth]{\includegraphics[width=0.425\textwidth, keepaspectratio]{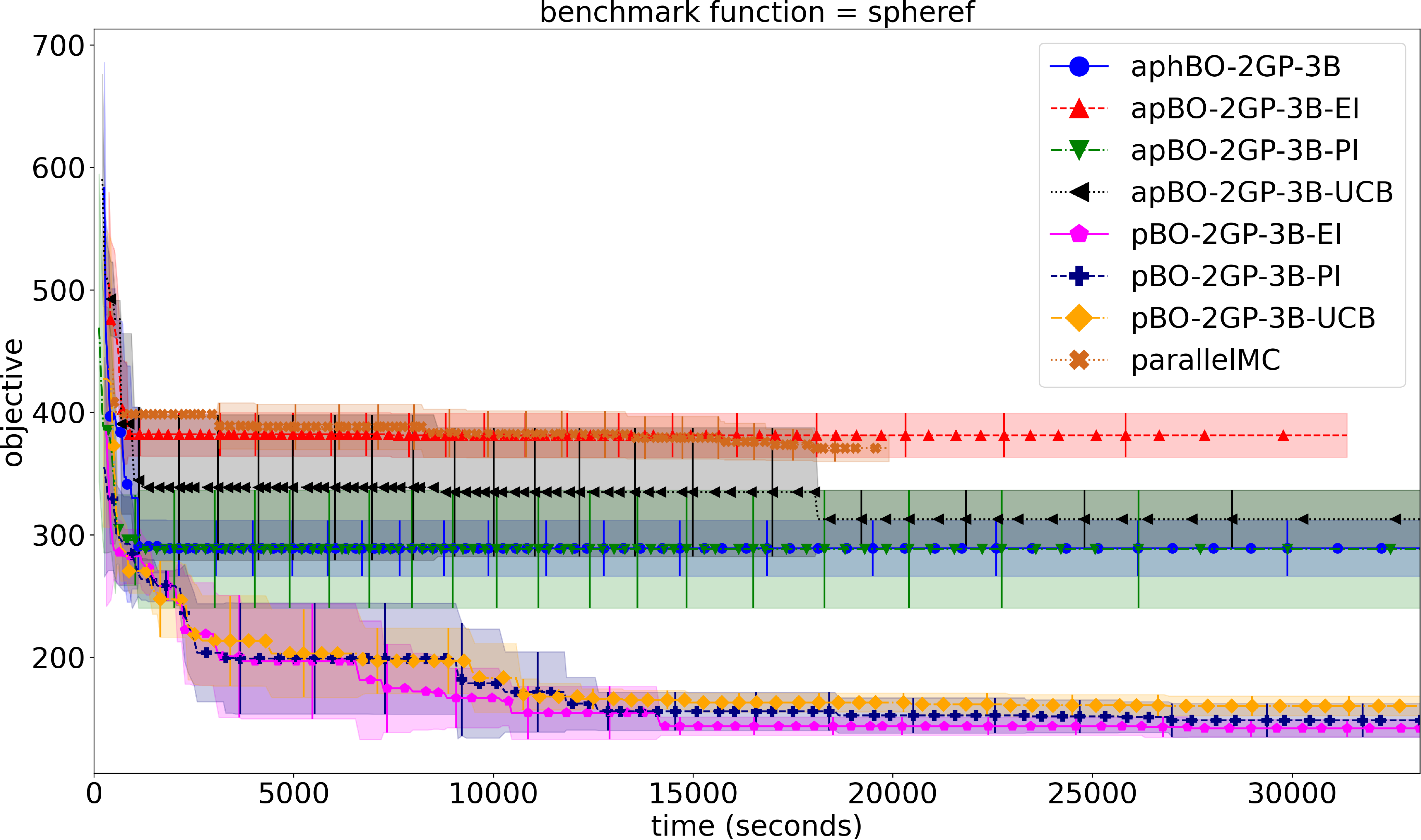}}
\hfill
\subcaptionbox{Benchmark by iteration.
\label{fig:benchByIter_spheref}
}
  [0.425\linewidth]{\includegraphics[width=0.425\textwidth, keepaspectratio]{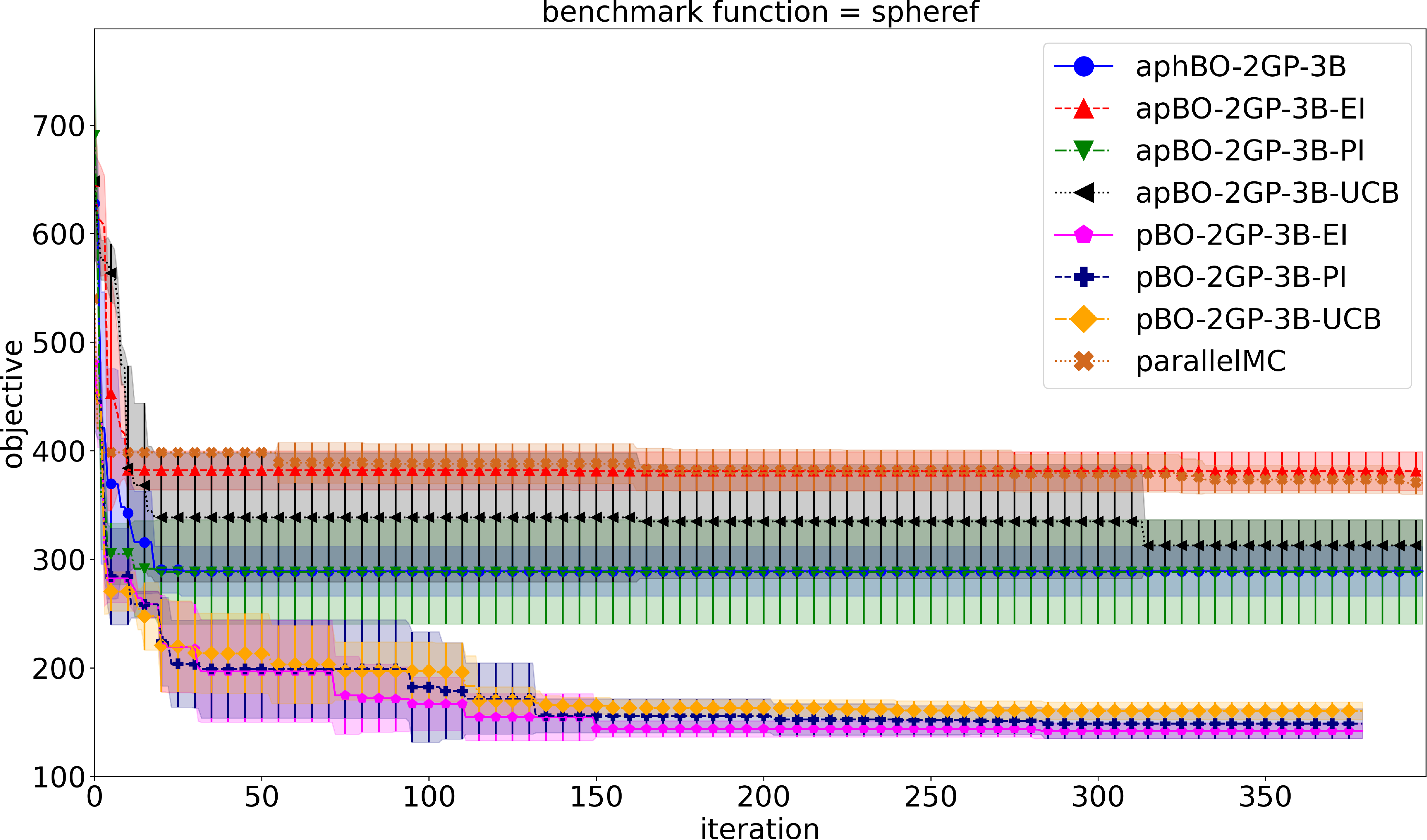}}
% \caption{Benchmark for spheref.}
% \label{fig:bench_spheref}
% \end{figure}
% \vspace{-2.50cm}
\medskip

% \subsection{rothyp (free d) (70d)}
% \begin{figure}[!htbp]
\centering
\subcaptionbox{Benchmark by physical wall-clock time.
\label{fig:benchByTime_rothyp}
}
  [0.425\linewidth]{\includegraphics[width=0.425\textwidth, keepaspectratio]{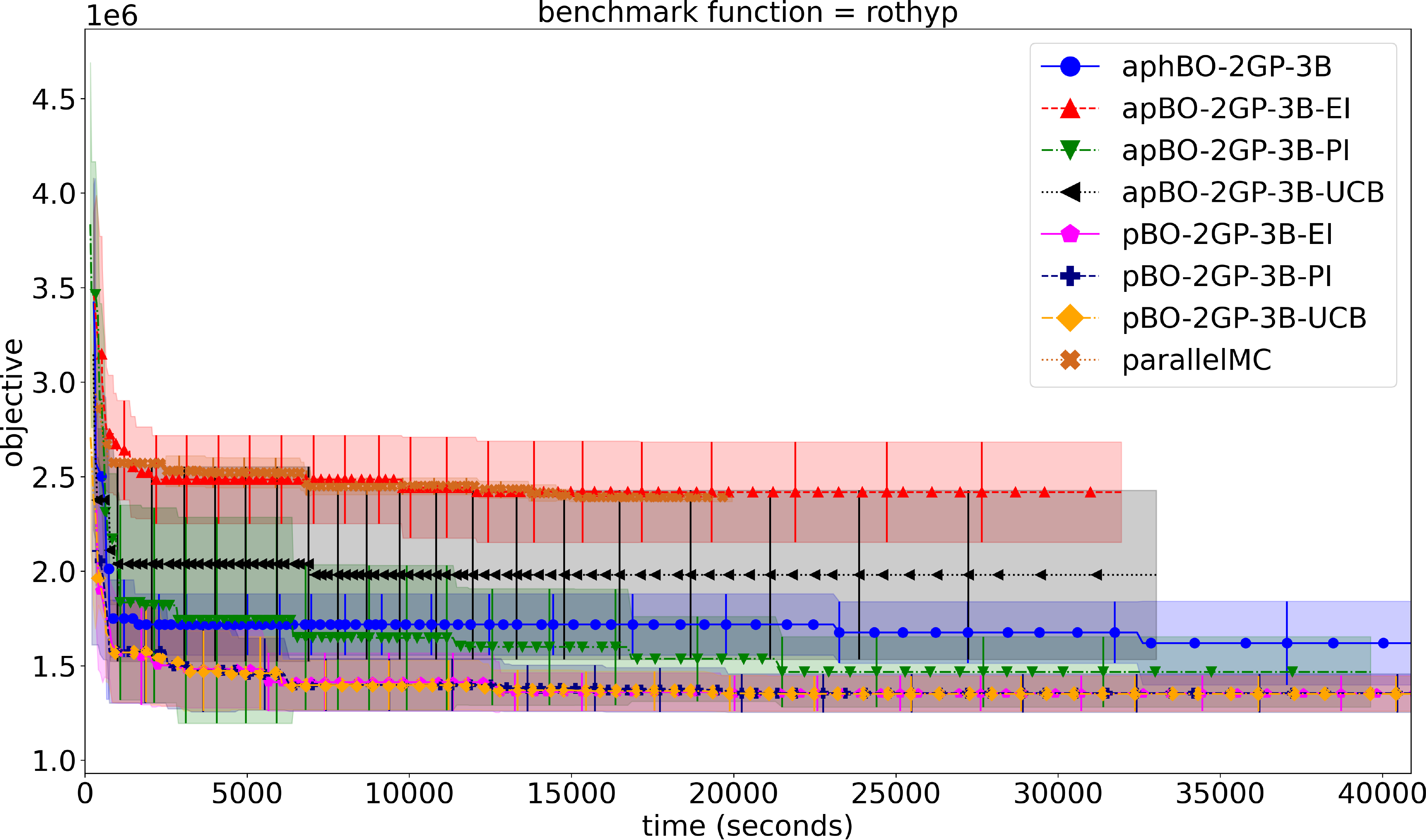}}
\hfill
\subcaptionbox{Benchmark by iteration.
\label{fig:benchByIter_rothyp}
}
  [0.425\linewidth]{\includegraphics[width=0.425\textwidth, keepaspectratio]{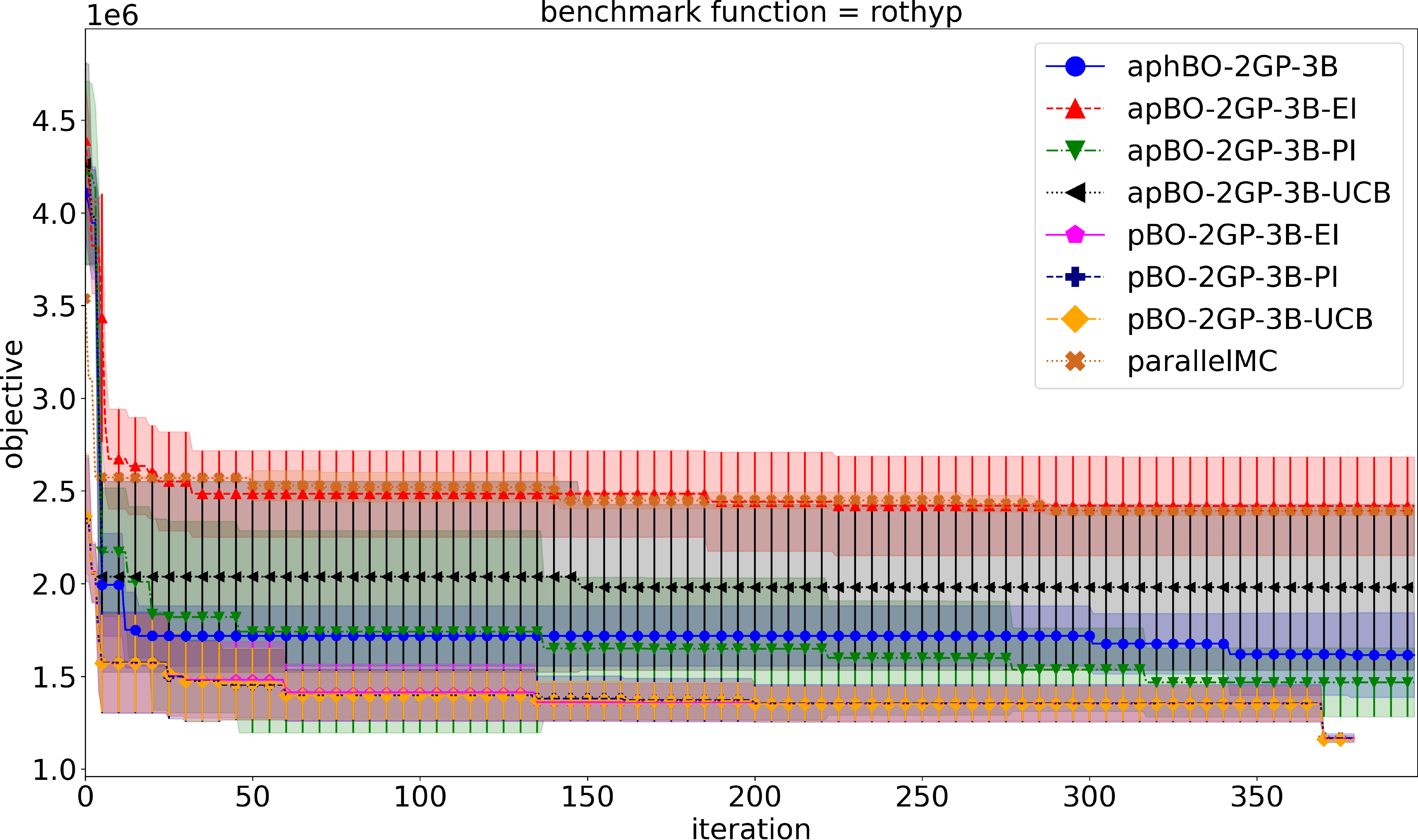}}
\caption{Benchmark for rothyp.}
\label{fig:bench_rothyp}
\end{figure}
% \vspace{-2.50cm}

Because no constraints are involved in these benchmark functions, the size of the third batch $\mathcal{B}_{\text{exploreClassif}}$ is reduced to zero. 
Based on the results of Desautels et al. \cite{desautels2014parallelizing}, Contal et al. \cite{contal2013parallel}, and Tran et al. \cite{tran2019pbo}, it has been proved theoretically and shown numerically that parallel BO methods supersede sequential BO methods in term of both convergences in time and in iterations. 
Therefore, we omit the inclusion of classical BO methods, which are sequential, with different acquisition functions due to its inferiority in numerical performance compared to parallel BO methods. 
To enhance a fair comparison, we use the same initial points, the same settings for the auxiliary optimizer, as well as the pinging time for checking results periodically, for all methods. 
It is noted that benchmarking parallel BO methods are very challenging in a shared HPC platform, because the numerical performance may be dependent on the availability of the computational resource. 

The results show a significant speedup in low-dimensional problems $(d<10)$, while somewhat on par on high-dimensional problems with the asynchronous parallel feature. Overall, UCB acquisition function is very competitive, compared to both EI and PI acquisition functions. While it is not always the best performer in some benchmark functions, the performance of the proposed aphBO-2GP-3B is stable and nearly optimal across most of the function, which demonstrates its robustness and resiliency. Furthermore, the computational time for asynchronous parallel BO is shorter, while performing on par with the batch-sequential parallel approaches, which makes the former ones more computationally appealing.

\section{Engineering example 1: Flip-chip package}
\label{sec:FCBGA}

In this section, the application of the proposed aphBO-2GP-3B framework to design flip-chip package is demonstrated, where the objective is to minimize the strain energy density, which is an accurate indicator for its fatigue life. 

\subsection{Thermo-mechanical finite element model}

As a possible application of the aphBO-2GP-3B framework, a lidless flip-chip package containing a monolithic silicon die (FCBGA) mounted on a printed circuit board (PCB) with a stiffener ring was considered, such as for a commercial field programmable gate array. The mechanical design space for FCBGAs and PCBs, which represents the dimensions and material choices, was chosen because it is a complex engineering problem with several variables. 
A parametrized thermo-mechanical FEA model of an FCBGA on PCB was constructed using ANSYS 19.1 using ANSYS Parametric Design Language (APDL) \cite{mccann2016board}. 
% \redbf{Details on Anh’s algorithm $\leftrightarrow$ ADPL text file $\leftrightarrow$ ANSYS if desired.} 

\begin{table}[!htbp]
\caption{Material properties}
\label{tab:ansysFCBGAmatlProperty}
\centering
\begin{tabular}{|l|l|l|} \hline
\textbf{Material} & \textbf{CTE (ppm/$^{\circ}$C)} & \textbf{Modulus (GPa)} \\ \hline 
Silicon         & 2.6-3.7                      & 169 (x), 130 (y) \\ 
SAC305 Solder   & 22                           & 50               \\ 
Underfill       &  30 ($T<T_g$), 100 ($T>T_g$) & 3                \\ 
Substrate       & Variable                     & 30               \\ 
PCB             & Variable                     & 30               \\ 
Ring Adhesive   & 120                          & 0.04             \\ 
Ring            & Variable                     & 190              \\ 
Copper          & 17                           & 117              \\ \hline
\end{tabular}
\end{table}

\begin{figure}[!htbp]
\centering
\includegraphics[width=\textwidth, keepaspectratio]{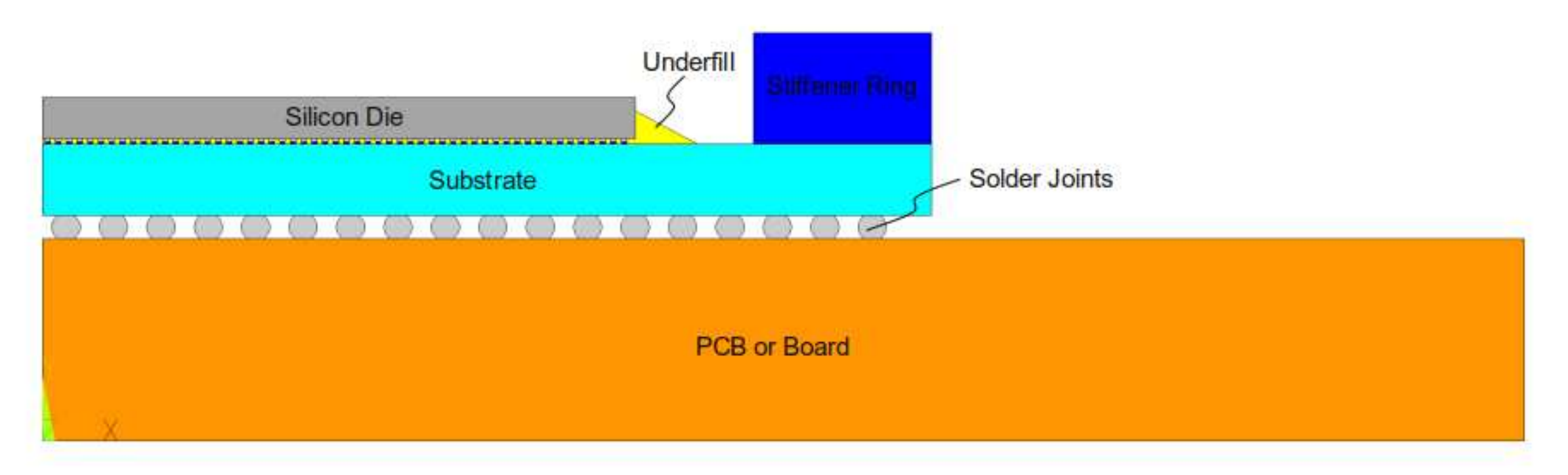}
\caption{Finite element model geometry.}
\label{fig:cropped.feModelGeom}
\end{figure}

Several choices were fixed to facilitate model construction and to represent a realistic design space. For example, only material properties of currently available materials were chosen and interconnect pitches were chosen based on 7nm generation design rules (circa 2020). The variable parameters include die dimensions, substrate dimensions, substrate coefficient of thermal expansion (CTE), ring dimensions, ring CTE, underfill fillet size, and board CTE. The material models are given in Table 1; solder was represented with Anand’s model \cite{anand1985constitutive}. A two-dimensional, half symmetry model was chosen to reduce computation time due to the large number of simulations required. While this may lower the numerical accuracy of the model, the relative outputs are not expected to change. 
The FE model geometry is presented in Figure \ref{fig:cropped.feModelGeom}. 
No validation of the FEA simulation could be performed because no chip was built, however, the authors have previously validated similar models with experimental results \cite{mccann2017use,mccann2018warpage}. 
Table \ref{tab:ansysFCBGAmatlProperty} presents the material properties used in the FCBGA thermo-mechanical FEA simulation. 

As an output, after the model is solved, the component warpage at 20$^\circ$C, the component warpage at 200$^\circ$C, and the strain energy density in the outermost solder joint from the third thermal cycle of -40 to 125$^\circ$C \cite{darveaux2000effect, standard2009package}, are calculated. The component warpage at 20$^\circ$C is a commonly required customer metric, must be below JEDEC specifications \cite{standard2009package}, as well as component warpage at 200$^\circ$C must be below JEITA \cite{standard2007measurement}. The strain energy density has been identified as an accurate way to predict fatigue life of solder joints during thermal cycling \cite{darveaux2000effect}. 
% SED should be minimized to maximize to fatigue life.
In this problem, the constraints are imposed that the component warpage at 20$^{\circ}$C and 200$^{\circ}$C must be below a 300$\mu$m and 75$\mu$m, respectively. 
It is noted that the component warpages at different temperatures are only quantifiable if the model converges appropriately. 
Thus, regression on the constraints is not always possible.

% 3 – JEDEC JESD22-B112B https://www.jedec.org/standards-documents/docs/jesd-22-b112, https://www.jedec.org/system/files/docs/22B112B.pdf
% 4 – JIETA_ED7306 Warpage Specification https://home.jeita.or.jp/tsc/std-pdf/ED-7306_E.pdf 
% 5 - L. Anand. "Constitutive Equations for Hot-Working of Metals". International Journal of Plasticity. Vol. 1. 213-231. 1985.
% 6 – https://ieeexplore.ieee.org/document/8008859
% 7 - https://ieeexplore.ieee.org/document/8429866

\subsection{Optimization results}

% results are located at /home/anhvt89/Documents/bayesOpt/flip-chipBGA/flip-chipBGA_17Jan19-2
% /media/anhvt89/seagateRepo/bayesOpt/flip-chipBGA-sMfConsBo/flip-chipBGA_17Jan19-2

Table \ref{tab:ansysFCBGAdesignVar} list the design variables and its associated parts, as well as its lower and upper bounds in this case study. 
% The aphBO-2GP-3B framework is deployed on the Georgia Tech PACE HPC system, where more than 50,000 processors are available on a RHEL 6.7 operating system. 
The pool of workers used in the optimization is fixed at 16 due to a limited number of licenses. 
The FEA simulation typically takes about 18-21 minutes on a single processor. 
% To avoid the diminishing return, the number of processors used is always set at one. 
For the sake of demonstration, we set the number of processors to one.

\begin{table}[!htbp]
\centering
\caption{Design variables for the FCBGA design optimization.}
\label{tab:ansysFCBGAdesignVar}
\begin{tabular}{|l|l|l|l|l|} \hline
\textbf{Variable}   & \textbf{Design part}            & \textbf{Lower bound}         & \textbf{Upper bound}            &  \textbf{Optimal value}           \\ \hline
$x_1$      & die                    & 20000               & 30000                  &  20702                   \\ 
$x_2$      & die                    & 300                 & 750                    &  320                     \\ 
$x_3$      & substrate              & 30000               & 40000                  &  35539                   \\ 
$x_4$      & substrate              & 100                 & 1800                   &  1614                    \\ 
$x_5$      & substrate              & $10\cdot 10^{-6}$   & $17\cdot 10^{-6}$      &  $17\cdot 10^{-6}$       \\ 
$x_6$      & stiffener ring         & 2000                & 6000                   &  4126                    \\ 
$x_7$      & stiffener ring         & 100                 & 2500                   &  1646                    \\ 
$x_8$      & stiffener ring         & $8\cdot 10^{-6}$    & $25 \cdot 10^{-6}$     &  $8.94\cdot 10^{-6}$     \\ 
$x_9$      & underfill              & 1.0                 & 3.0                    &  1.52                    \\ 
$x_{10}$   & underfill              & 0.5                 & 1.0                    &  0.804                   \\ 
$x_{11}$   & PCB board              & $12.0\cdot 10^{-6}$ & $16.7\cdot 10^{-6}$    &  $16.7\cdot 10^{-6}$     \\ \hline
\end{tabular}
\end{table}

Nine initial sampling points are used to initialize the aphBO-2GP-3BO. 
The interface between the aphBO-2GP-3B framework and the FCBGA application includes a Python and a Shell scripts. 
A Python script is devised to obtain the output and feasibility of the application, whereas a Shell script is devised to modify the input script for the application and query the script on the HPC platform. 

\begin{figure}[!htbp]
\centering
\includegraphics[width=0.75\textwidth, keepaspectratio]{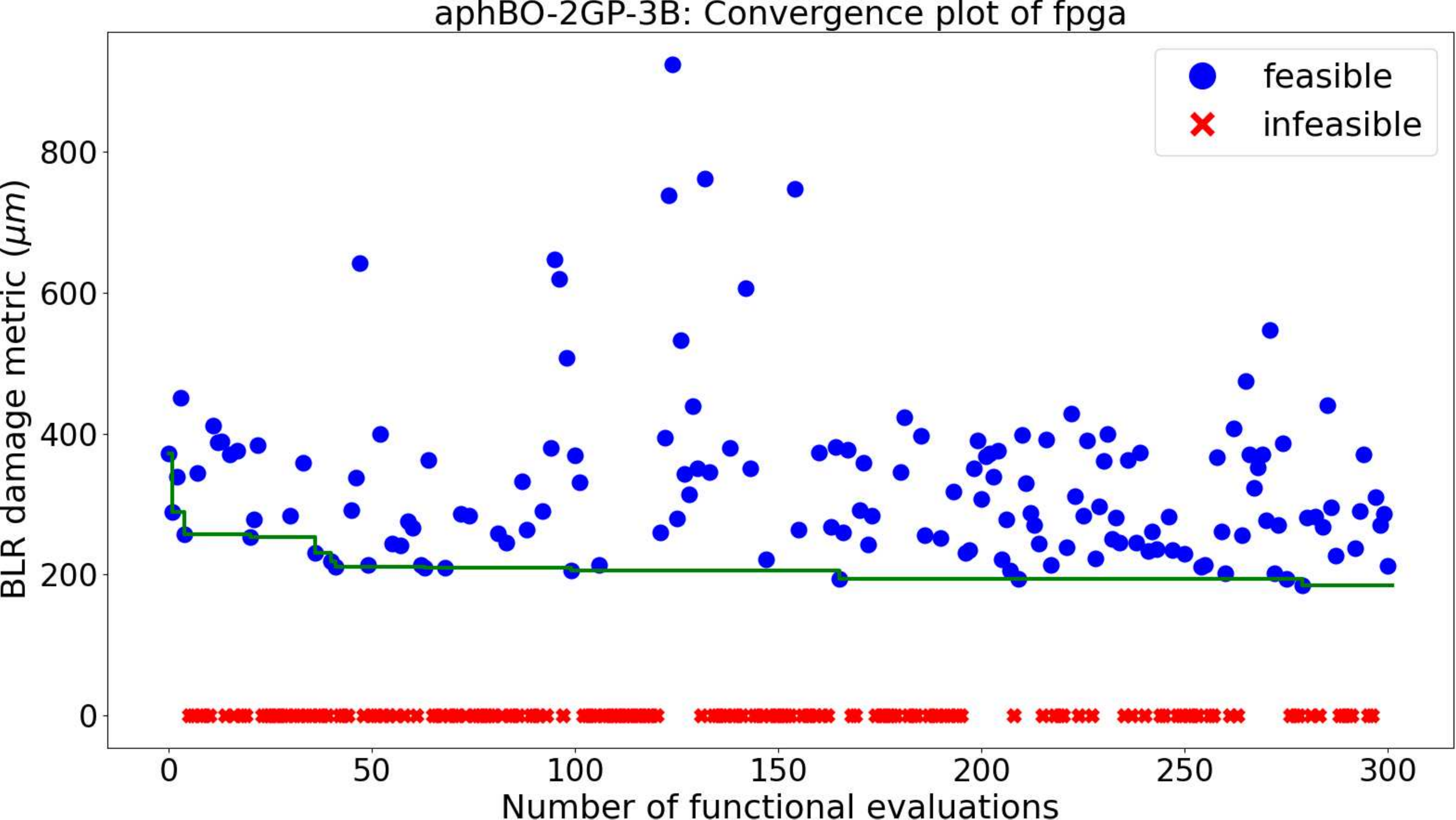}
\caption{Convergence plot of the aphBO-2GP-3B framework for the FCBGA design optimization application.}
\label{fig:cropped.ansysFcbgaConvPlot}
\end{figure}

Figure \ref{fig:cropped.ansysFcbgaConvPlot} presents the convergence plot of the FCBGA design optimization using the aphBO-2GP-3B framework in 24 hours, where the feasible sampling points are plotted as blue circles, whereas the infeasible sampling points are plotted as red crosses. 
In total, 1016 simulations are obtained within 24 hours. 
This is to compare with a traditional BO method, where the same amount of simulations would take 14.11 days, as opposed to 1 day with the aphBO-2GP-3BO framework. 
The factor of 14.11 in time difference is attributed to a number of factors, mainly the waiting time when submitting a job on the shared HPC.
This issue of waiting time in the HPC queue is very common in academic or government environment settings, but not in an industrial setting. 
It is also shown in Figure \ref{fig:cropped.ansysFcbgaConvPlot} that the FCBGA application is highly constraint, where most of the input space is infeasible. 
Thus, the effectiveness of the GP binary classifier in the aphBO-2GP-3BO is demonstrated.

\begin{figure}[!htbp]
\centering
\subcaptionbox{Original design in $xy$-plane.
\label{fig:fcbgaOriginalDesign}
}
  [.45\linewidth]{\includegraphics[width=0.45\textwidth, keepaspectratio]{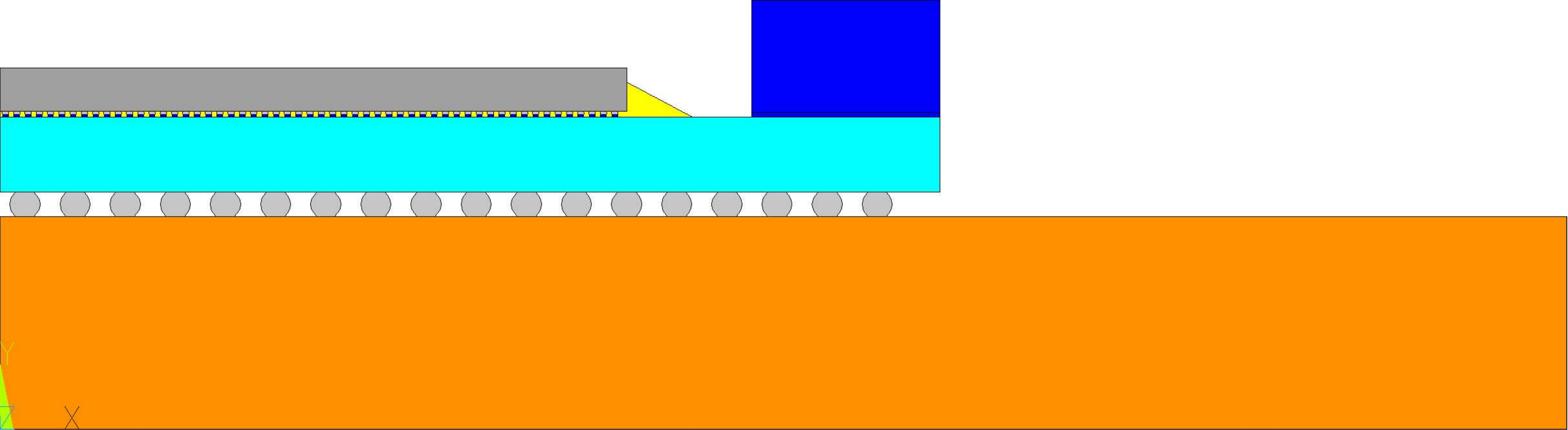}}
\hfill
\subcaptionbox{Optimal design in $xy$-plane.
\label{fig:fcbgaOptimalDesign}
}
  [.45\linewidth]{\includegraphics[width=0.45\textwidth, keepaspectratio]{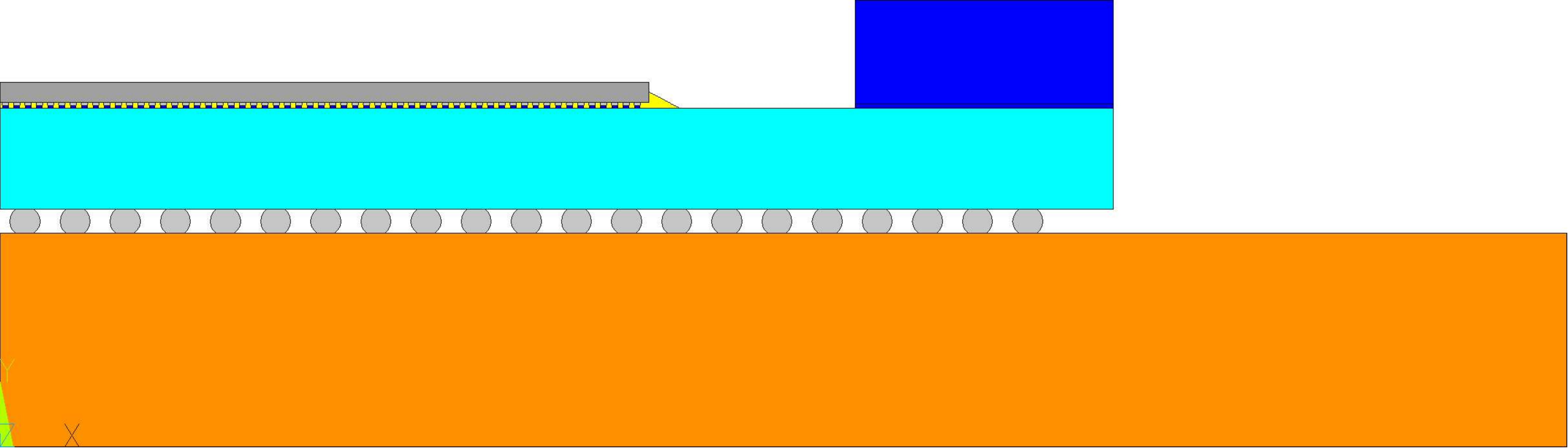}}
\caption{Comparison between original design (Figure \ref{fig:fcbgaOriginalDesign}) and optimal design (Figure \ref{fig:fcbgaOptimalDesign}) of flip-chip package.}
\label{fig:fcbgaDesignComparison2}
\end{figure}

\begin{figure}[!htbp]
\centering
\subcaptionbox{Component warpage evaluation at -40$^{\circ}$C.
\label{fig:cropped.package001-optimizedpackagewarpageat-40C}
}
  [0.30\linewidth]{\includegraphics[width=0.30\textwidth, keepaspectratio]{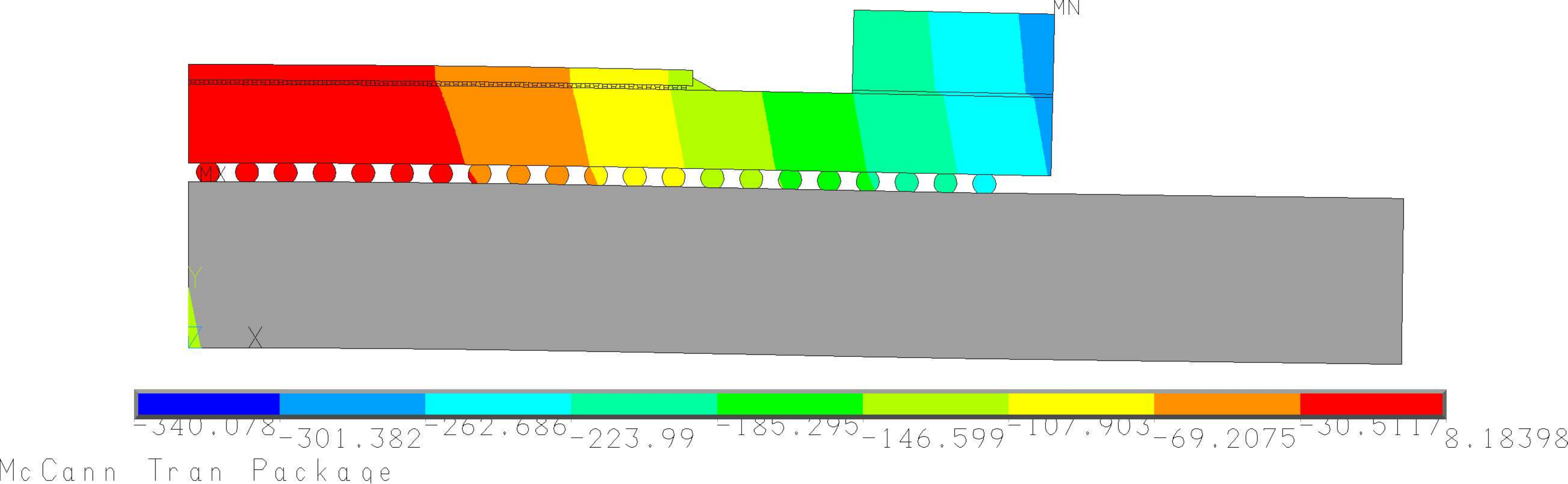}}
\hfill
\subcaptionbox{Component warpage evaluation at 20$^{\circ}$C.
\label{fig:cropped.package001-packagewarpageat20C}
}
  [0.30\linewidth]{\includegraphics[width=0.30\textwidth, keepaspectratio]{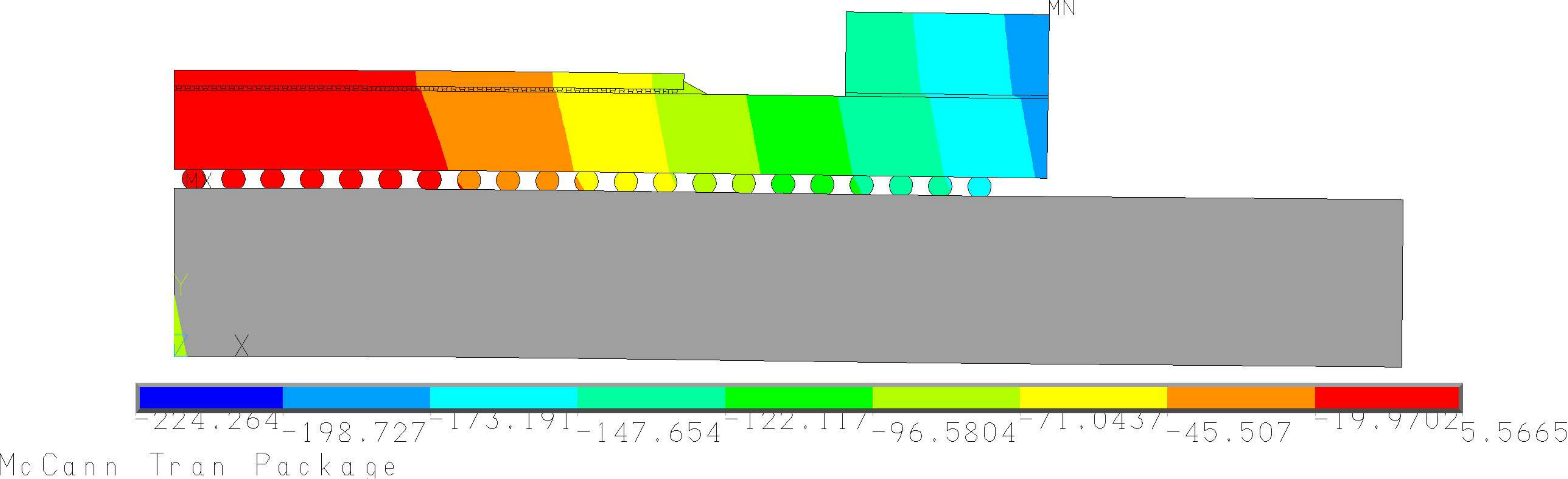}}
\hfill
\subcaptionbox{Component warpage evaluation at 200$^{\circ}$C.
\label{fig:cropped.package001-packagewarpageat200C}
}
  [0.30\linewidth]{\includegraphics[width=0.30\textwidth, keepaspectratio]{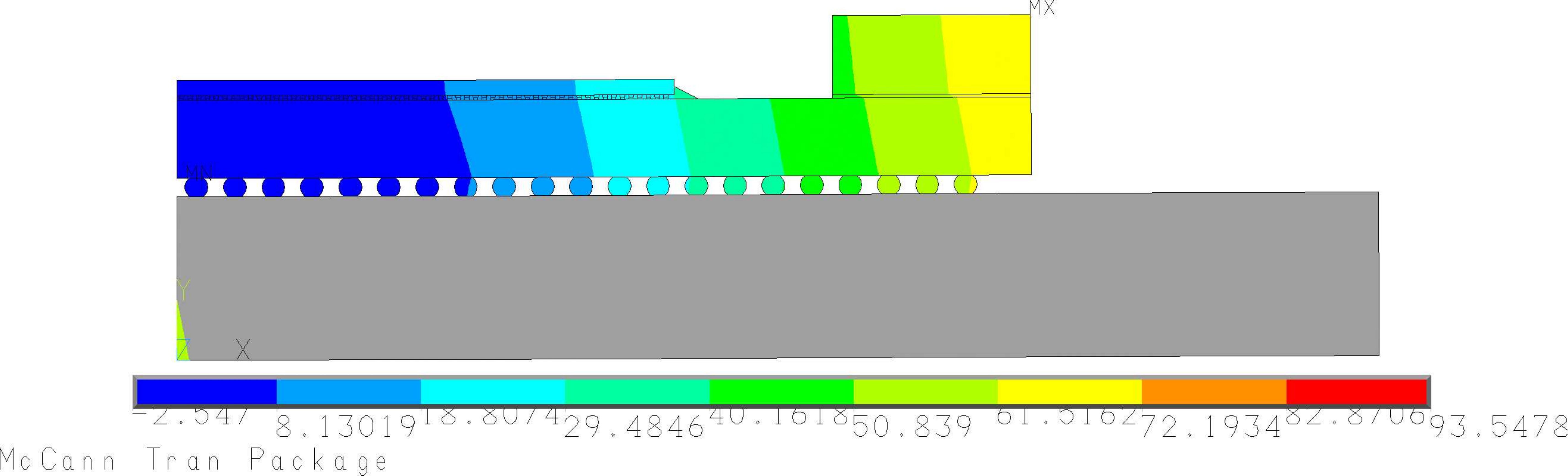}}
% \subcaptionbox{First principal stress contour of the optimal flip-chip package at -40$^{\circ}$C.
% \label{fig:cropped.package001-optimizedS1at-40C}
% }
%   [0.30\linewidth]{\includegraphics[width=0.30\textwidth, keepaspectratio]{figsAsyncBO/cropped.package001-optimizedS1at-40C}}
% \hfill
% \subcaptionbox{First principal stress contour of the optimal flip-chip package at 200$^{\circ}$C.
% \label{fig:cropped.package001-optimizedS1at200C}
% }
%   [0.30\linewidth]{\includegraphics[width=0.30\textwidth, keepaspectratio]{figsAsyncBO/cropped.package001-optimizedS1at200C}}

\caption{Warpage contour of the optimal flip-chip package at different temperatures, -40$^{\circ}$C, 20$^{\circ}$C, and 200$^{\circ}$C. 
}
\label{fig:fcbgaOptimalDesignEvaluation}
\end{figure}

The performance of the optimal package is evaluated at different temperatures to further examine its robustness for a range of temperature, from -40$^{\circ}$C to 200$^{\circ}$C. 
% Figure \ref{fig:cropped.package001-optimizedS1at-40C} and Figure \ref{fig:cropped.package001-optimizedS1at200C} show the first principal stress of the optimal design flip-chip package at -40$^{\circ}$C and 200$^{\circ}$C, respectively. 
Figures \ref{fig:cropped.package001-optimizedpackagewarpageat-40C}, \ref{fig:cropped.package001-packagewarpageat20C}, and \ref{fig:cropped.package001-packagewarpageat200C} show the contours of the predicted component warpage at the temperatures of -40$^{\circ}$C, 20$^{\circ}$C, and 200$^{\circ}$C, respectively.

Figure \ref{fig:fcbgaOriginalDesign} shows the original design of the flip-chip package, whereas Figure \ref{fig:fcbgaOptimalDesign} shows the optimal design of the flip-chip package as the result of applying the aphBO-2GP-3BO framework. 
The aphBO-2GP-3BO optimized solution is shown in Table \ref{tab:ansysFCBGAdesignVar}. 
The values are relatively close to the designs used in the microelectronics packaging industry, indicating that the commercial package design is relatively well optimized and validating the aphBO-2GP-3BO result. 
Specifically, the die is relatively thin and small and the substrate is thick to minimize the component warpage. 
The substrate CTE is close to the PCB board CTE to improve board level reliability. 
The substrate size may increase with little impact to the constraints monitored here. 
The underfill fillet size and ring dimensions are close to industry designs. 
The stiffener ring CTE is the one notable excursion from industry design, although industry is constrained by actual material selection, which includes cost, manufacturability, and availability, which are not reflected in aphBO-2GP-3BO's constraints. 
The BLR damage metric, which is the objective of this case study, is reduced from 372 $\mu m$ to 185 $\mu m$, representing 50.27\% improvement.

% Convergence folderName:
% Iter 0: fpga_Iter1: 372.00000000
% Iter 1: fpga_Iter2: 289.00000000
% Iter 4: fpga_Iter5: 257.00000000
% Iter 20: fpga_Iter21: 253.00000000
% Iter 36: fpga_Iter37: 231.00000000
% Iter 40: fpga_Iter41: 219.00000000
% Iter 41: fpga_Iter42: 211.00000000
% Iter 63: fpga_Iter64: 210.00000000
% Iter 99: fpga_Iter100: 205.00000000
% Iter 165: fpga_Iter166: 194.00000000
% Iter 279: fpga_Iter280: 185.00000000
% [0, 1, 4, 20, 36, 40, 41, 63, 99, 165, 279, 301]

\section{Engineering example 2: 3D CFD slurry pump casing design optimization}
\label{sec:3dCasingWear}

In this section, we demonstrate the aphBO-2GP-3B framework using a 3D multiphase CFD simulation. An in-house multiphase CFD wear code is utilized to predict the wear rate of different slurry pump casing. 
In order to optimize the slurry pump casing performance, the predicted maximum wear rate of the casing is considered as the objective function. 

\subsection{3D CFD casing wear model}

The CFD simulation assumes a constant particle size (or mono-size) and thus the number of species in the particle size distribution is simplified to one. 
A 14-dimensional input $\bm{x}$ is formed for each CFD simulation to model the geometry of the slurry pump casing. The optimization procedure is carried out for pump assembly Z0534, at the input operating conditions of $Q=89637.900$ gpm, $H=50$ m, $n = 849.000$ RPM, $\eta = 82.400$, $d_{50} = 300\mu$m, $d_{85} = 690\mu$m, $d_{\text{eff}} = 495\mu$m, $C_v=20\%$, and \%BEPQ = 99.6\%, where $Q$ is the volumetric flow rate, $N$ is the impeller angular speed, $\eta$ is the hydraulic efficiency. $d_{50}$, $d_{85}$ are the 50th and 85th percentile of the particle size distribution. 
$d_{\text{eff}}= 495\mu$m is the effective particle size, which is calculated as the average of the $d_{50}$ and $d_{85}$ and used as an input for mono-size species in the CFD simulation. 
$\%BEPQ$ is the percentage of best efficiency point flow rate. 
The design impeller vane diameter is 1.7018 m, the shroud diameter is 1.7780 m, the suction diameter is 0.6604 m, and the discharge diameter is 0.6096 m. 
The pump specific speed $N_s$ in US units in 1425.6.

\subsection{CFD casing wear model}
% \label{sec:3dCasingWear}

% results at /media/anhvt89/ExtHD3TB/asyncPar_pBO-2GP-3B_8Nov18_Batch8

\begin{figure}[!htbp]
\centering
\subcaptionbox{3D pump casing with mesh and its boundary conditions.
\label{fig:3dCasSchematic3}
}
  [.55\linewidth]{\includegraphics[height=220px, keepaspectratio]{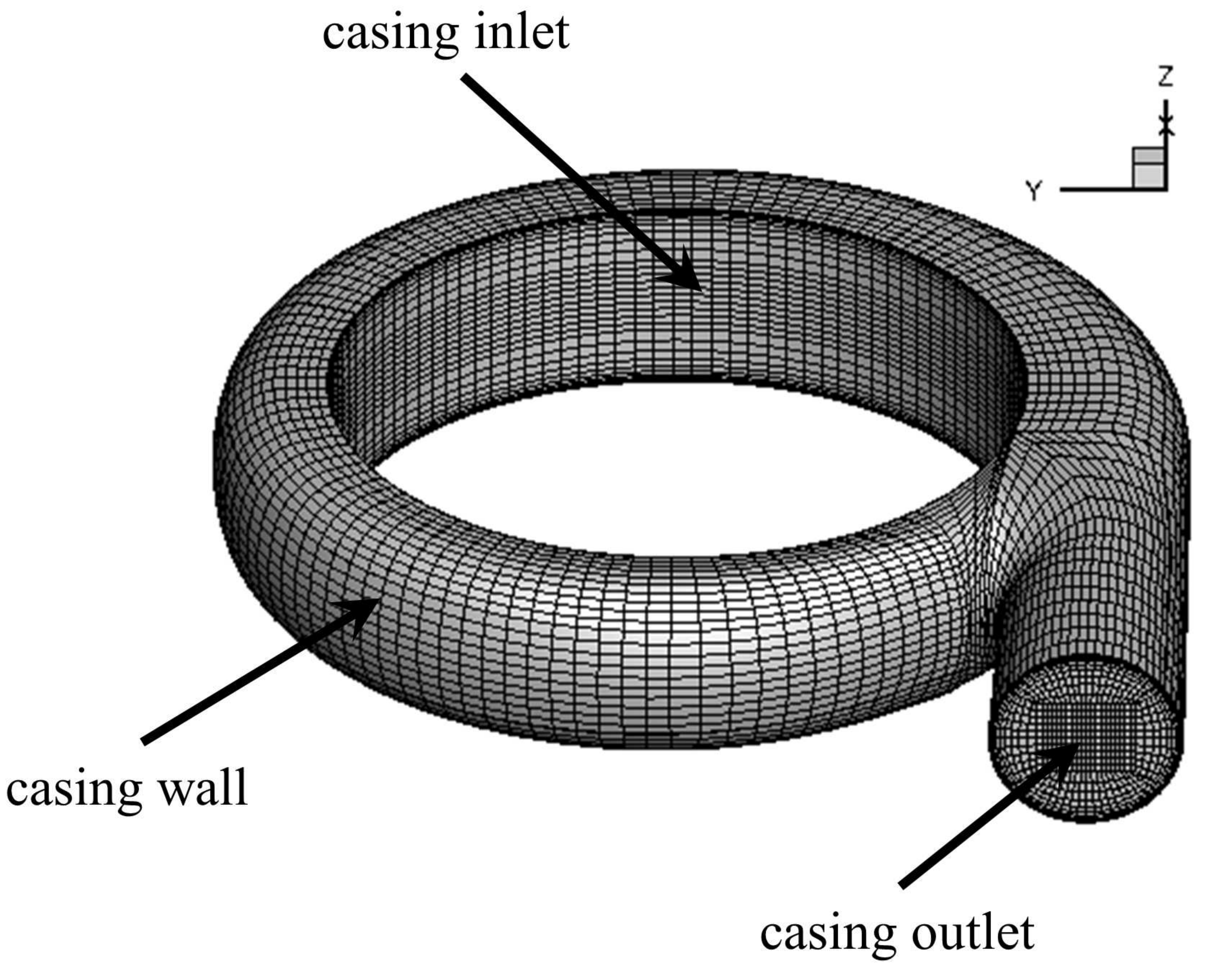}}
\hfill
\subcaptionbox{Sections of casing inlet. 
\label{fig:sectionsOfCasingInlet4}
}
  [.40\linewidth]{\includegraphics[height=150px, keepaspectratio]{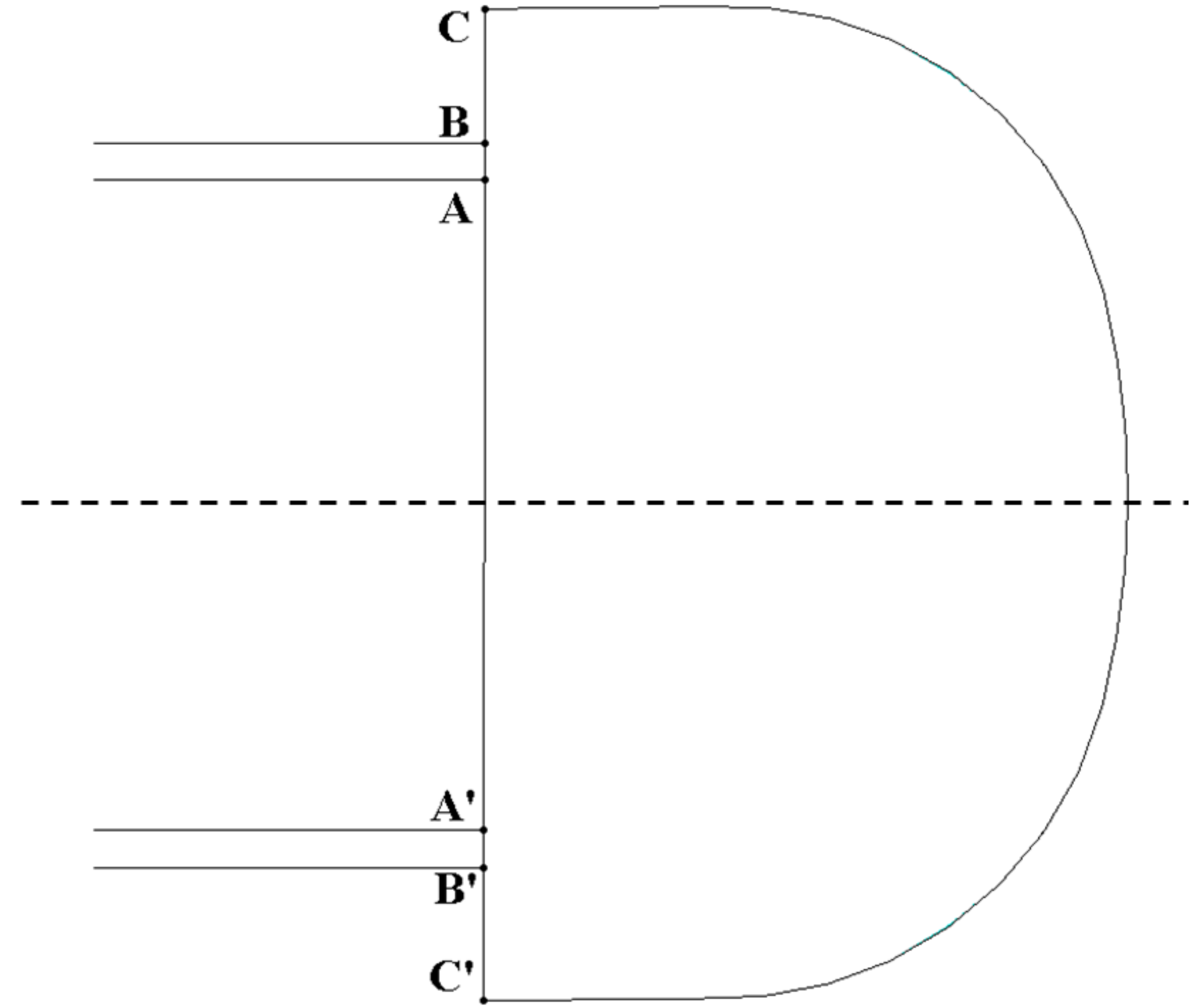}}

\caption{Mesh and boundary condition for the 3D CFD casing pump design optimization.}
\label{fig:3dCasing}
\end{figure}

% The CFD casing wear model used to predict erosive wear in the centrifugal slurry pump casing is the co-authors' previous work \cite{pagalthivarthi2009solid,pagalthivarthi2015finite}. 
The multiphase CFD casing wear model to predict erosive wear in the slurry centrifugal casing is the co-authors' previous work \cite{pagalthivarthi2009solid,pagalthivarthi2015finite}. 
A short summary of the simulation is provided for the sake of completeness. 
% Here a brief description of 3D CFD casing wear model is provided as application background. 
The computational domain of the 3D CFD casing wear model is presented in Figure \ref{fig:3dCasSchematic3}. 
Using volume and time averaged governing equations, the continuity and momentum equations of the mixture and different solids species is derived based on an Eulerian-Eulerian framework. 
The particle size distribution can be discretized into finitely many subclasses, where each of them is treated using the Eulerian approach described above. 
% The casing inlet is divided into three sections, as shown in Figure \ref{fig:sectionsOfCasingInlet4}. 
Figure \ref{fig:sectionsOfCasingInlet4} presents three sections of the casing inlet. 
Different inlet velocity boundary conditions are applied on the region AA', BC, and B'C'. 
Based on the head $H$ and flow rate, $Q$, of the slurry pump, the radial and tangential velocities are imposed independently. 
The regions AB and A'B' are treated as impermeable walls, where Spalding wall functions \cite{white1991viscous} are applied. 
Similar to the CFD impeller wear model \cite{pagalthivarthi2013wear}, in CFD casing wear model, the Spalart-Allmaras turbulence model \cite{spalart1994one} is used. 
The set of nonlinear governing equations in the finite element problem are solved iteratively until a steady numerical solution with a tolerable residual level is obtained. 
Based on the steady-state CFD solutions, the constitutive model for wear prediction is applied with the empirically determined wear coefficients \cite{pagalthivarthi2009solid,pagalthivarthi2015finite} to predict the sliding wear and impact wear as a function of concentration, solids density, velocity magnitude, tangential velocity, shear stress, impingement angle, and the particle size. 
The total wear is simply the sum of the calculated sliding wear and impact wear.

\subsection{Optimization results}

In this example, the aphBO-2GP-3B framework is combined with the 3D multiphase CFD simulation described above on a Intel Xeon Silver 4110 x2 @2.10GHz machine with 16 cores, 32 threads, 256GB of memory where Ubuntu 18.04 LTS is the operating system. 
% CPU: Xeon silver 4410 x2 (total 16 cores, 32 threads. more info here)
% MEM: 256 GB

% https://ark.intel.com/content/www/us/en/ark/products/123547/intel-xeon-silver-4110-processor-11m-cache-2-10-ghz.html
Two Python scripts are devised to write the input script and read the output for the CFD application. 
The computational runtime varies between 6.8 to 13.83 hours, depending on the convergence of the design. 
The optimization case study is performed for approximately 168 hours, which is equivalent to approximately 1176 CPU hours. 
The batch sizes are set at (5,3,1) for $(\mathcal{B}_{\text{acquisition}}, \mathcal{B}_{\text{explore}}, \mathcal{B}_{\text{exploreClassif}})$, respectively. 

\begin{figure}[!ht]
\centering
\includegraphics[width=0.75\textwidth, keepaspectratio]{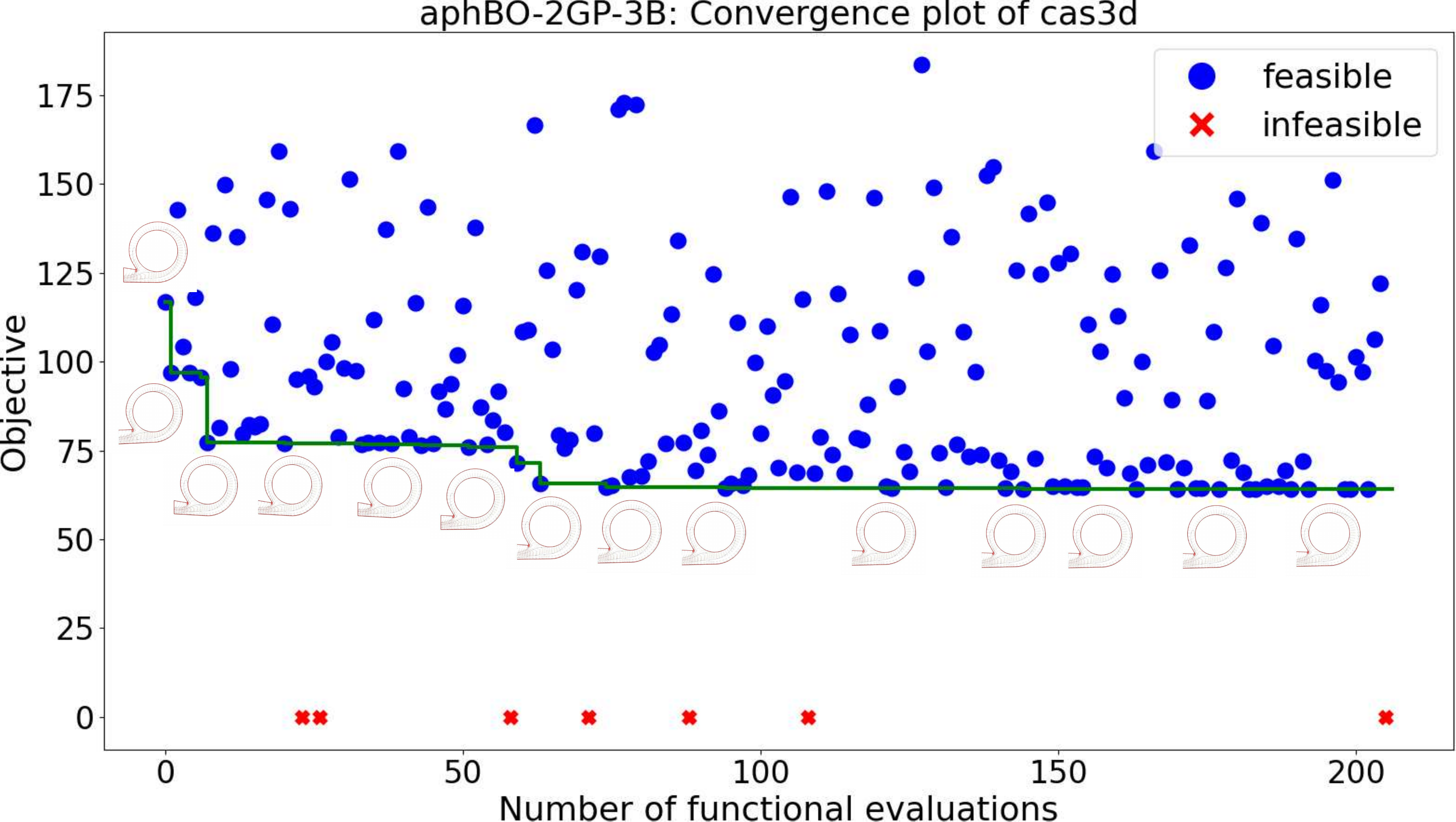}
\caption{Convergence plot of the aphBO-2GP-3B framework for the 3D CFD slurry pump casing design optimization application with evolutionary designs.}
\label{fig:cropped.3dCasingWear}
\end{figure}

Figure \ref{fig:cropped.3dCasingWear} shows the convergence plot of the aphBO-2GP-3B framework for the CFD application, where the first functional evaluation is the original design. 
The maximum wear rate in the original design is calculated as 116.7958$\mu m/hr$, whereas in the optimal design, the maximum wear rate is calculated as 64.0584$\mu m/hr$, showing an improvement of 45.915\% reduction in wear. 

% Do you want to overwrite current postproc.*.dat by re-running utils/buildLogs.sh? [y/n or 1/0]: n

% By default, BO solves for [MAXMIMIZATION] constrained optimization problems. 
% What settings do you want, min or max? [max or 1/min or 0]: 1
% Convergence folderName:
% Iter 0: cas3d_Iter1: 116.79580000
% Iter 1: cas3d_Iter2: 96.95786000
% Iter 4: cas3d_Iter5: 96.84860000
% Iter 6: cas3d_Iter7: 95.65457000
% Iter 7: cas3d_Iter8: 77.39253000
% Iter 20: cas3d_Iter21: 77.10147000
% Iter 33: cas3d_Iter34: 76.62708000
% Iter 43: cas3d_Iter44: 76.51635000
% Iter 51: cas3d_Iter52: 75.91136000
% Iter 59: cas3d_Iter60: 71.51360000
% Iter 63: cas3d_Iter64: 65.78669000
% Iter 74: cas3d_Iter75: 64.80942000
% Iter 94: cas3d_Iter95: 64.38196000
% Iter 122: cas3d_Iter123: 64.36025000
% Iter 144: cas3d_Iter145: 64.18984000
% Iter 163: cas3d_Iter164: 64.06123000
% Iter 177: cas3d_Iter178: 64.06050000
% Iter 192: cas3d_Iter193: 64.05842000
% [0, 1, 4, 6, 7, 20, 33, 43, 51, 59, 63, 74, 94, 122, 144, 163, 177, 192, 206]

\begin{figure}[!htbp]
\centering
\subcaptionbox{Original design.
\label{fig:comparisonMixtureVelocity:original:CrossSection}
}
  [.45\linewidth]{\includegraphics[width=0.4001\textwidth, keepaspectratio]{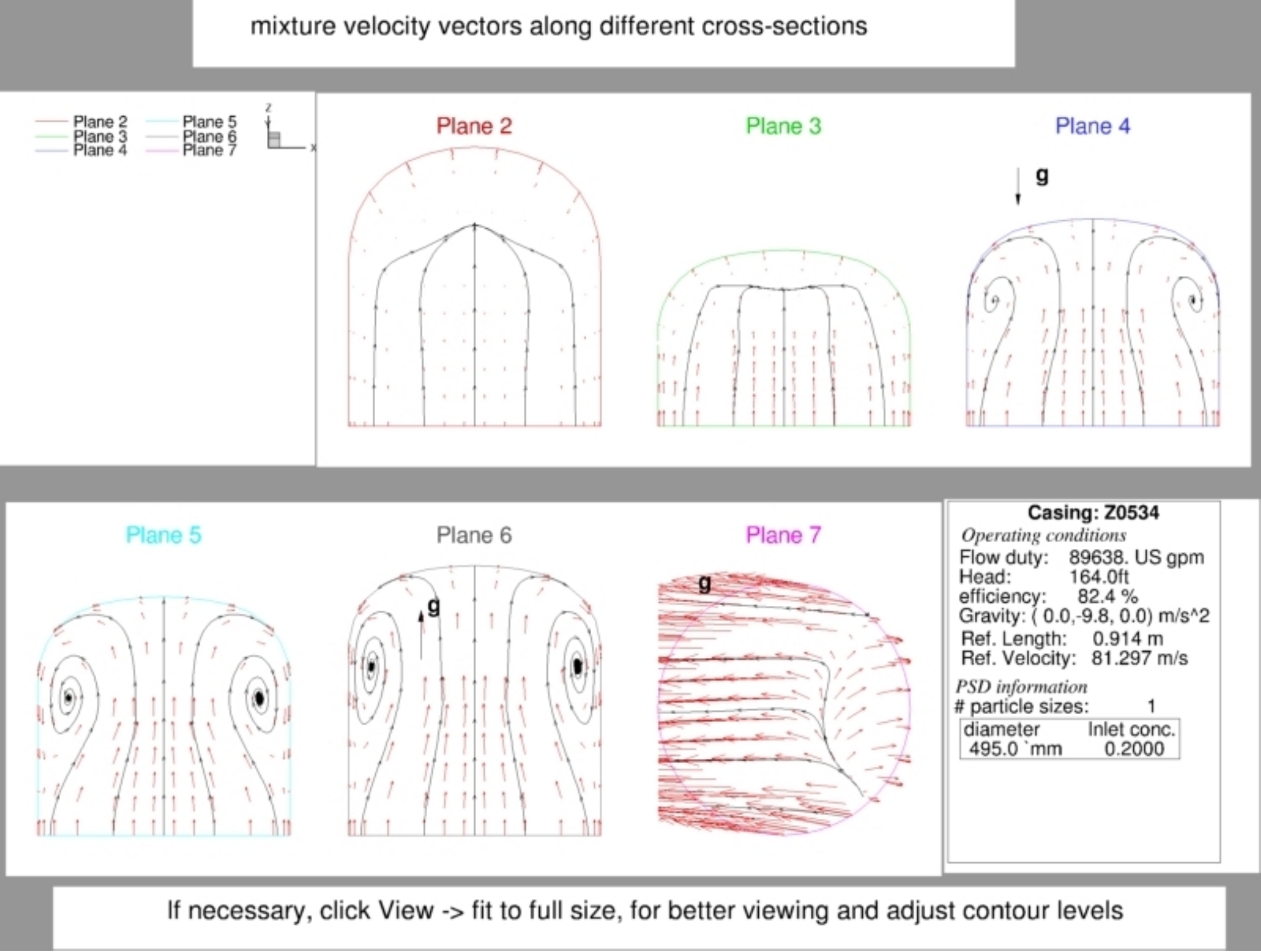}}
\hfill
\subcaptionbox{Optimal design.
\label{fig:comparisonMixtureVelocity:optimal:CrossSection}
}
  [.45\linewidth]{\includegraphics[width=0.4001\textwidth, keepaspectratio]{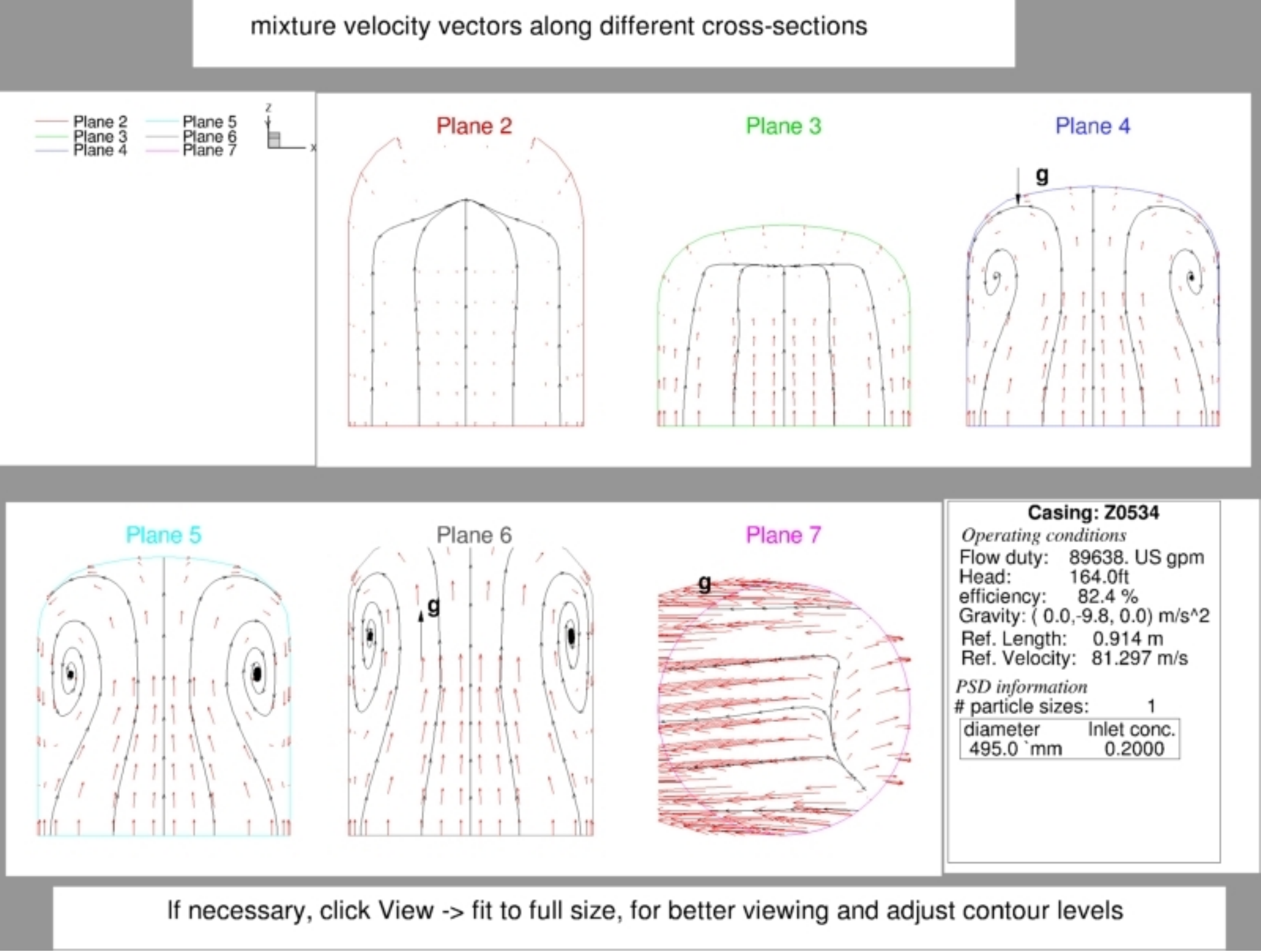}}

\centering
\subcaptionbox{Original design.
\label{fig:comparisonMixtureVelocity:original:MidPlane}
}
  [.45\linewidth]{\includegraphics[width=0.4001\textwidth, keepaspectratio]{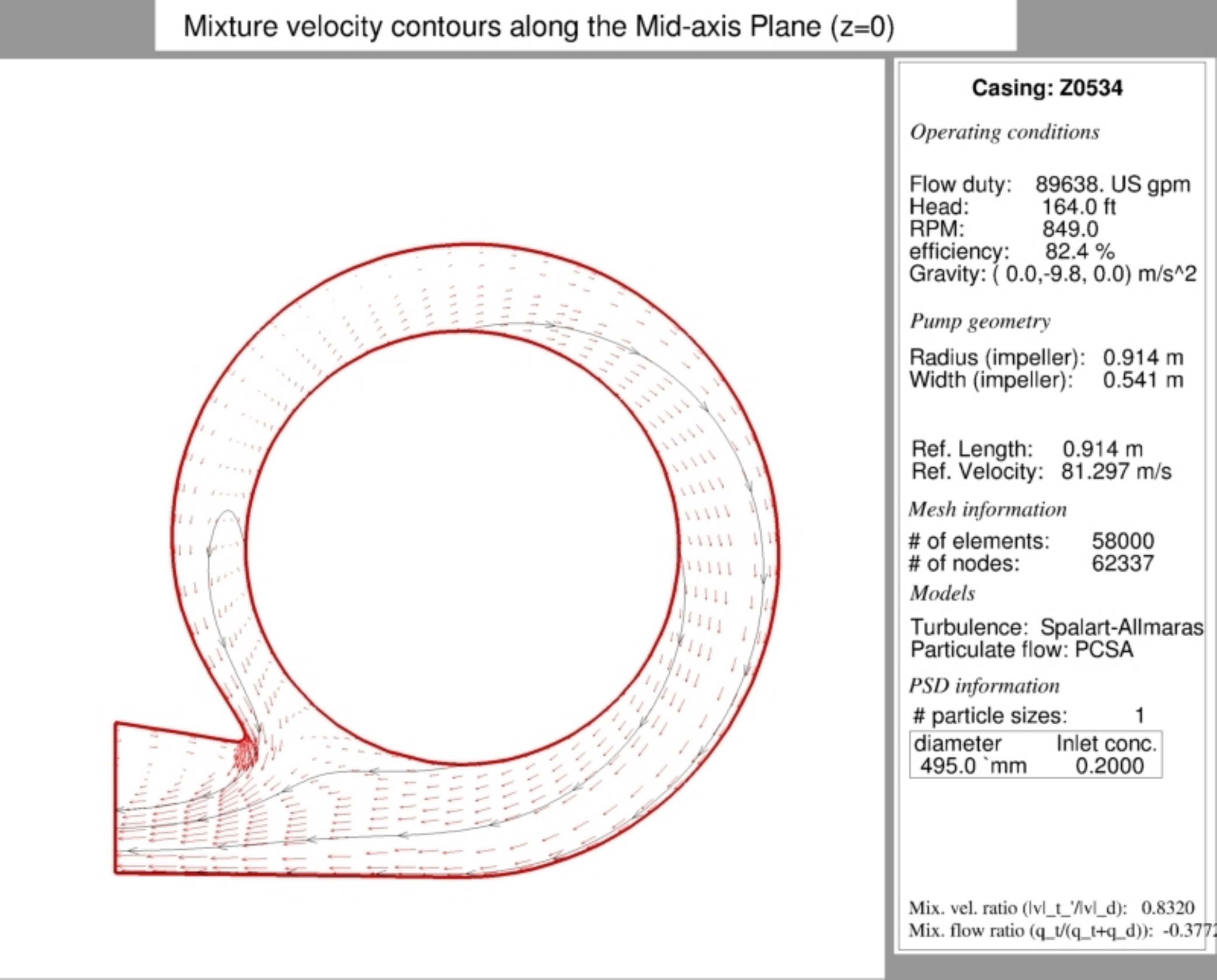}}
\hfill
\subcaptionbox{Optimal design.
\label{fig:comparisonMixtureVelocity:optimal:MidPlane}
}
  [.45\linewidth]{\includegraphics[width=0.4001\textwidth, keepaspectratio]{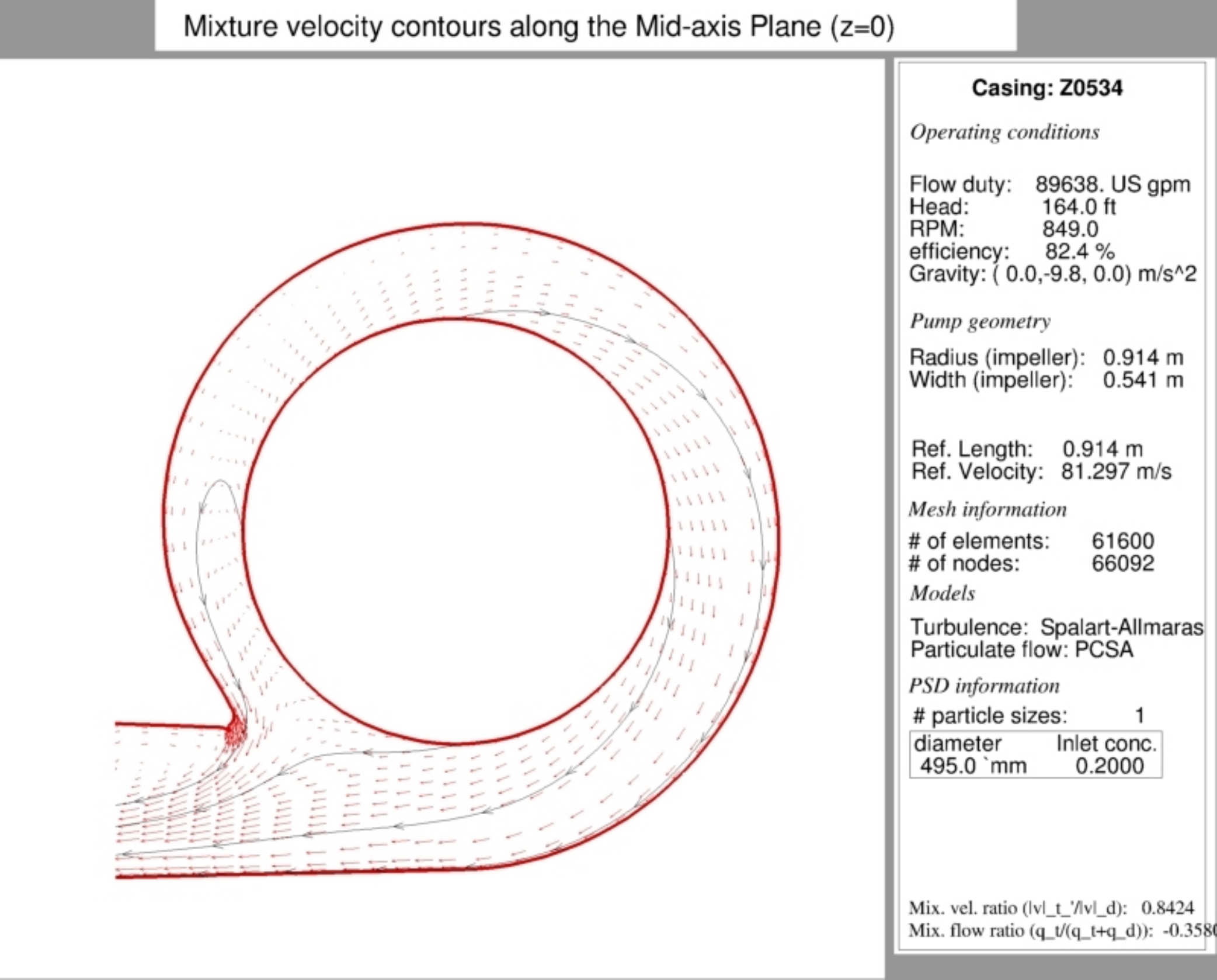}}
\caption{Comparison of mixture velocity.}
\label{fig:comparisonMixtureVelocity}
\end{figure}

\begin{figure}[!htbp]
\centering
\subcaptionbox{Original design.
\label{fig:comparisonOverallConcentration:original:CrossSection}
}
  [.45\linewidth]{\includegraphics[width=0.4001\textwidth, keepaspectratio]{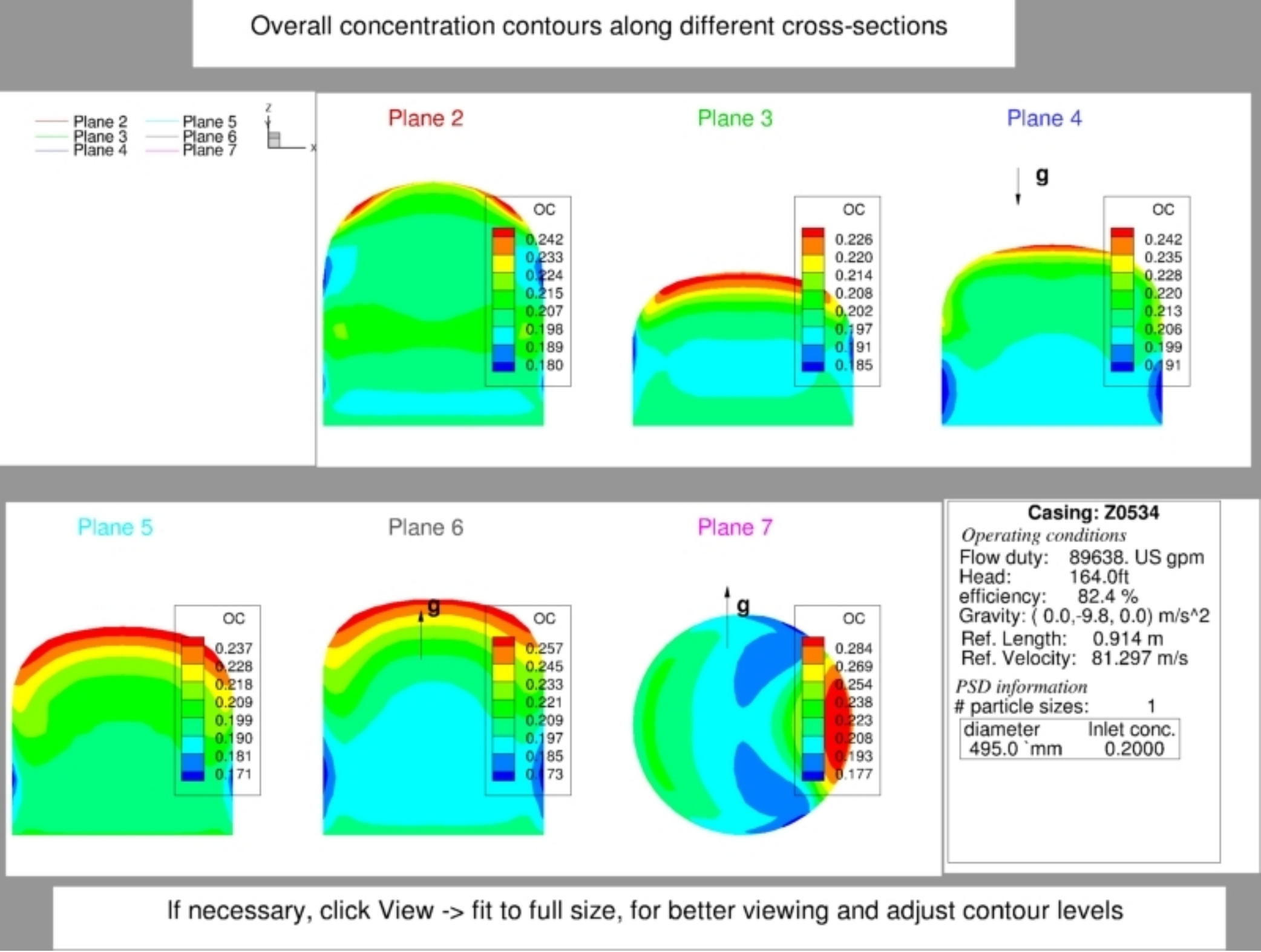}}
\hfill
\subcaptionbox{Optimal design.
\label{fig:comparisonOverallConcentration:optimal:CrossSection}
}
  [.45\linewidth]{\includegraphics[width=0.4001\textwidth, keepaspectratio]{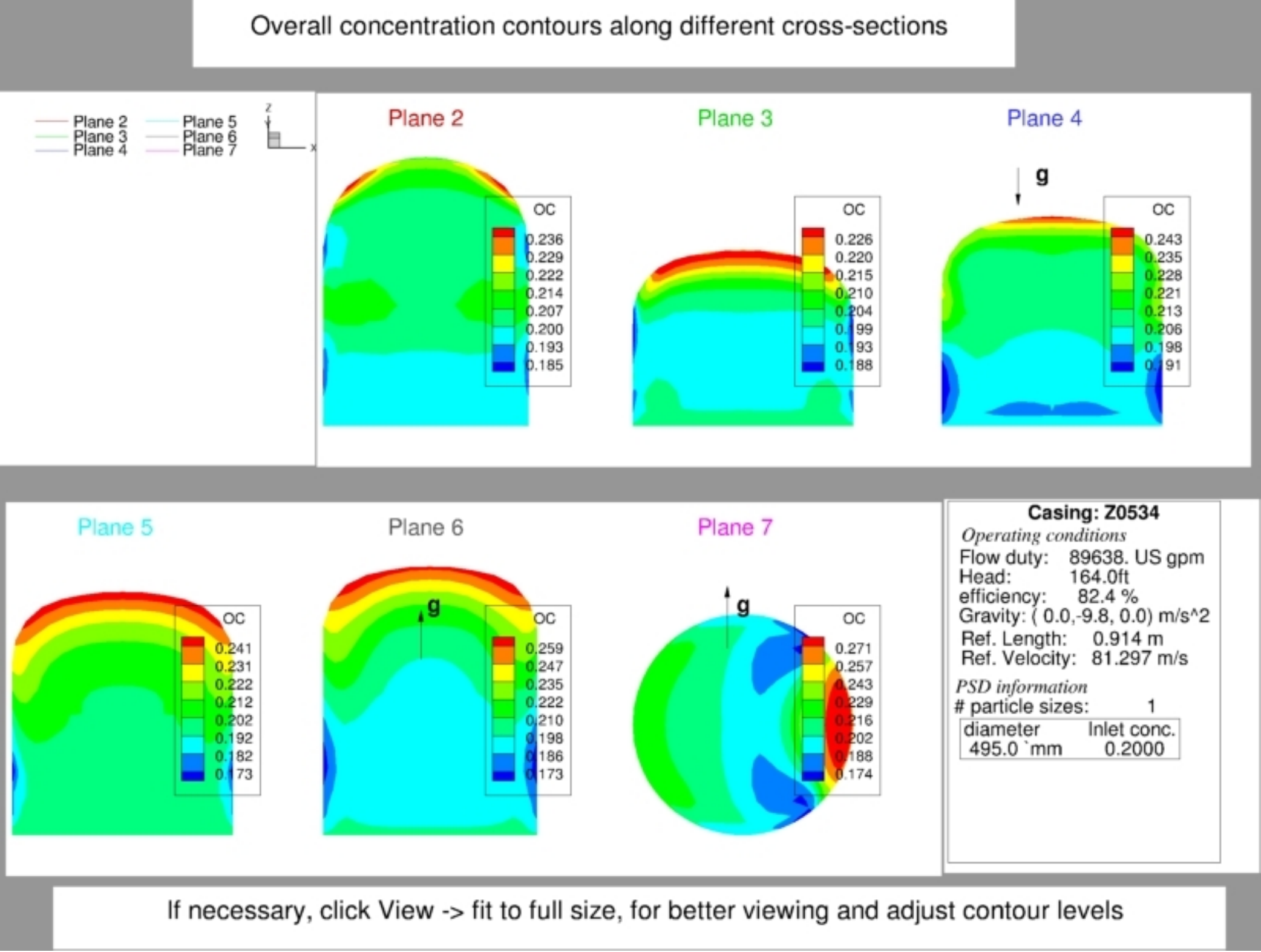}}

\centering
\subcaptionbox{Original design.
\label{fig:comparisonOverallConcentration:original:MidPlane}
}
  [.45\linewidth]{\includegraphics[width=0.4001\textwidth, keepaspectratio]{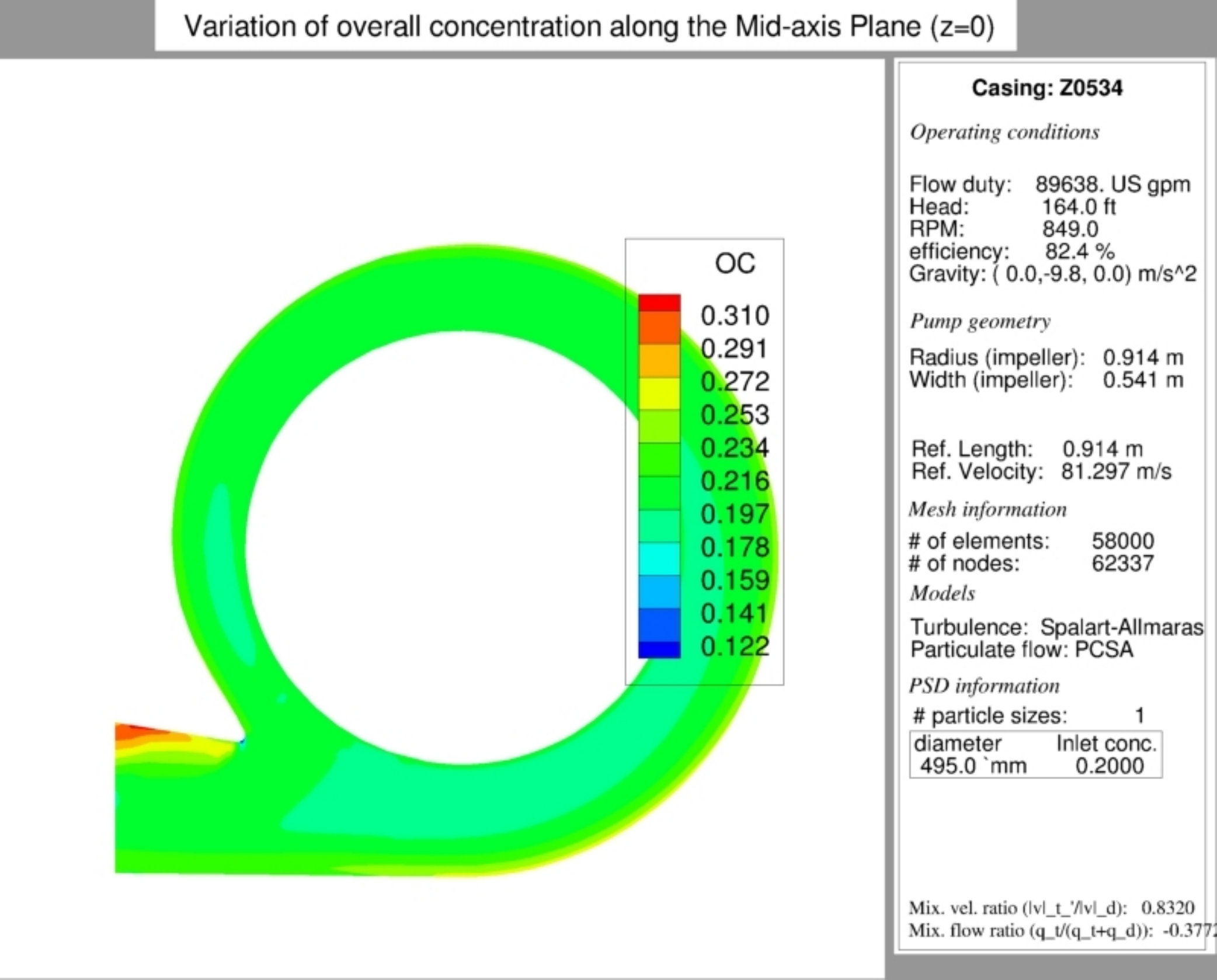}}
\hfill
\subcaptionbox{Optimal design.
\label{fig:comparisonOverallConcentration:optimal:MidPlane}
}
  [.45\linewidth]{\includegraphics[width=0.4001\textwidth, keepaspectratio]{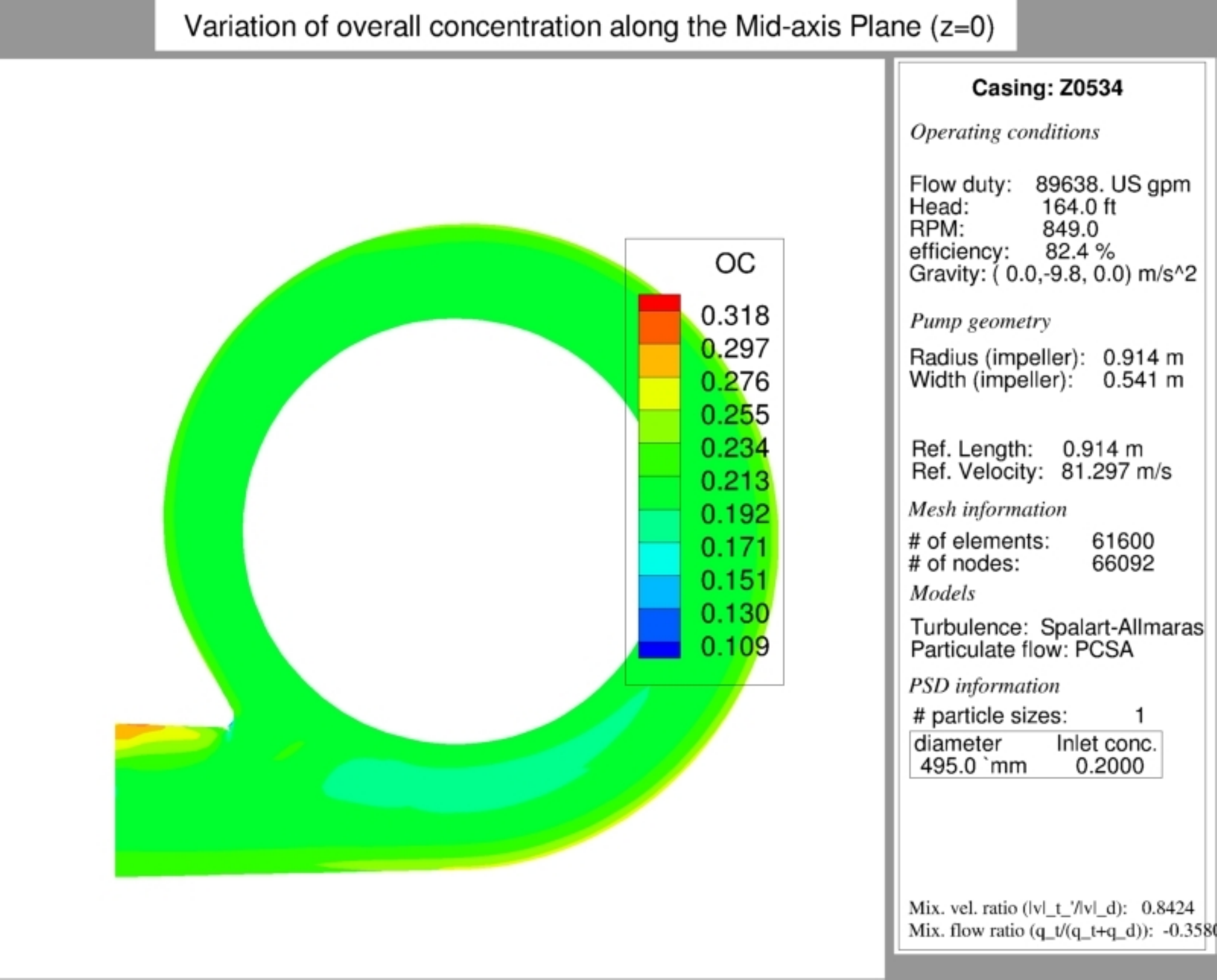}}
\caption{Comparison of overall concentration.}
\label{fig:comparisonOverallConcentration}
\end{figure}

% \begin{figure}[!htbp]
% \centering
% \subcaptionbox{Original design.
% \label{fig:original}
% }
%   [.45\linewidth]{\includegraphics[width=0.4001\textwidth, keepaspectratio]{figsAsyncBO/cas3d_Iter1/cropped_2d_pm_pl2-7}}
% \hfill
% \subcaptionbox{Optimal design.
% \label{fig:optimal}
% }
%   [.45\linewidth]{\includegraphics[width=0.4001\textwidth, keepaspectratio]{figsAsyncBO/cas3d_Iter193/cropped_2d_pm_pl2-7}}

% \centering
% \subcaptionbox{Original design.
% \label{fig:original}
% }
%   [.45\linewidth]{\includegraphics[width=0.4001\textwidth, keepaspectratio]{figsAsyncBO/cas3d_Iter1/cropped_2d_pm_z=0}}
% \hfill
% \subcaptionbox{Optimal design.
% \label{fig:optimal}
% }
%   [.45\linewidth]{\includegraphics[width=0.4001\textwidth, keepaspectratio]{figsAsyncBO/cas3d_Iter193/cropped_2d_pm_z=0}}
% \caption{Comparison of pressure mixture.}
% \label{fig:comparison}
% \end{figure}

\begin{figure}[!htbp]
\centering
\subcaptionbox{Original design.
\label{fig:comparisonWear:original:ImpactWear}
}
  [.45\linewidth]{\includegraphics[width=0.4001\textwidth, keepaspectratio]{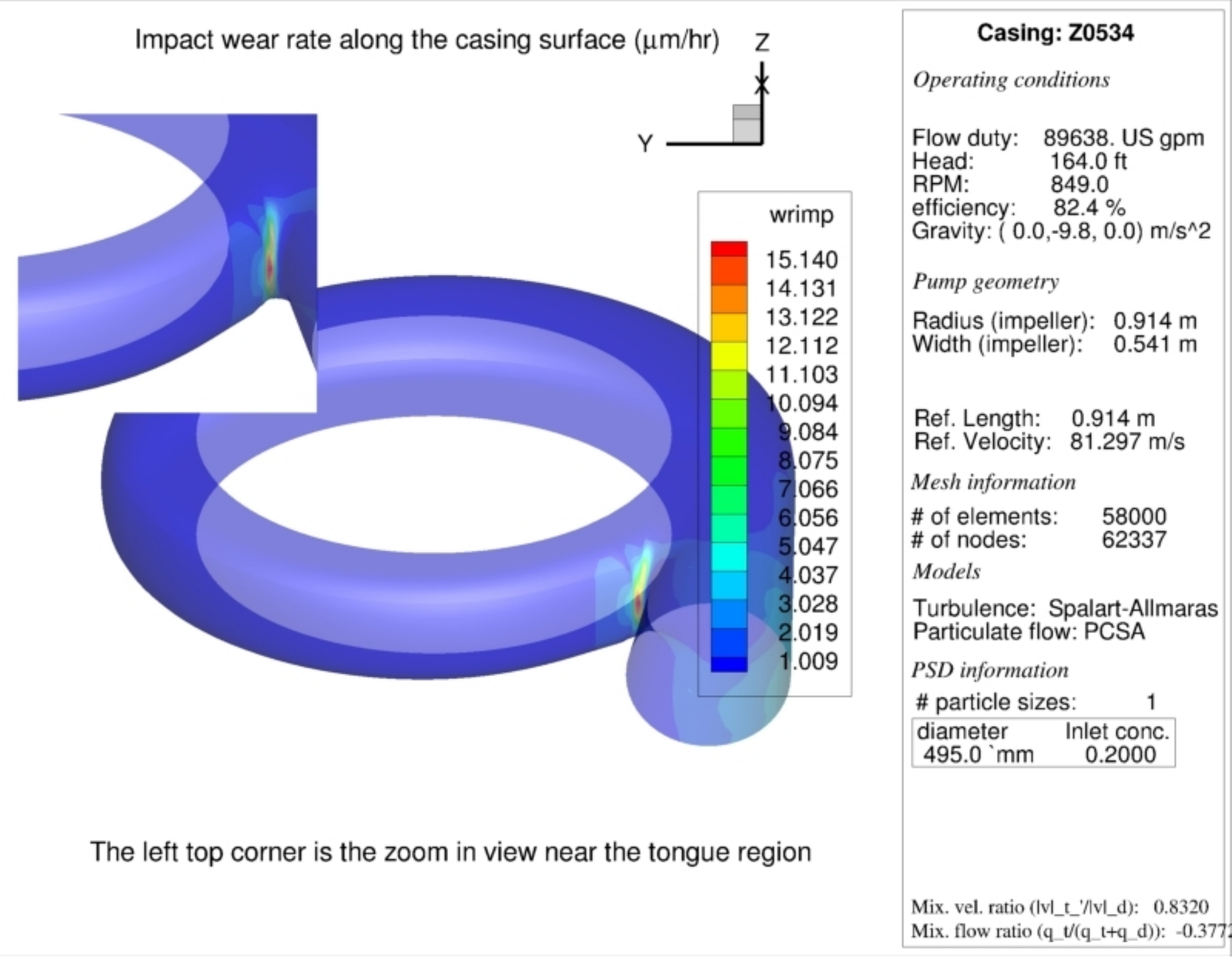}}
\hfill
\subcaptionbox{Optimal design.
\label{fig:comparisonWear:optimal:ImpactWear}
}
  [.45\linewidth]{\includegraphics[width=0.4001\textwidth, keepaspectratio]{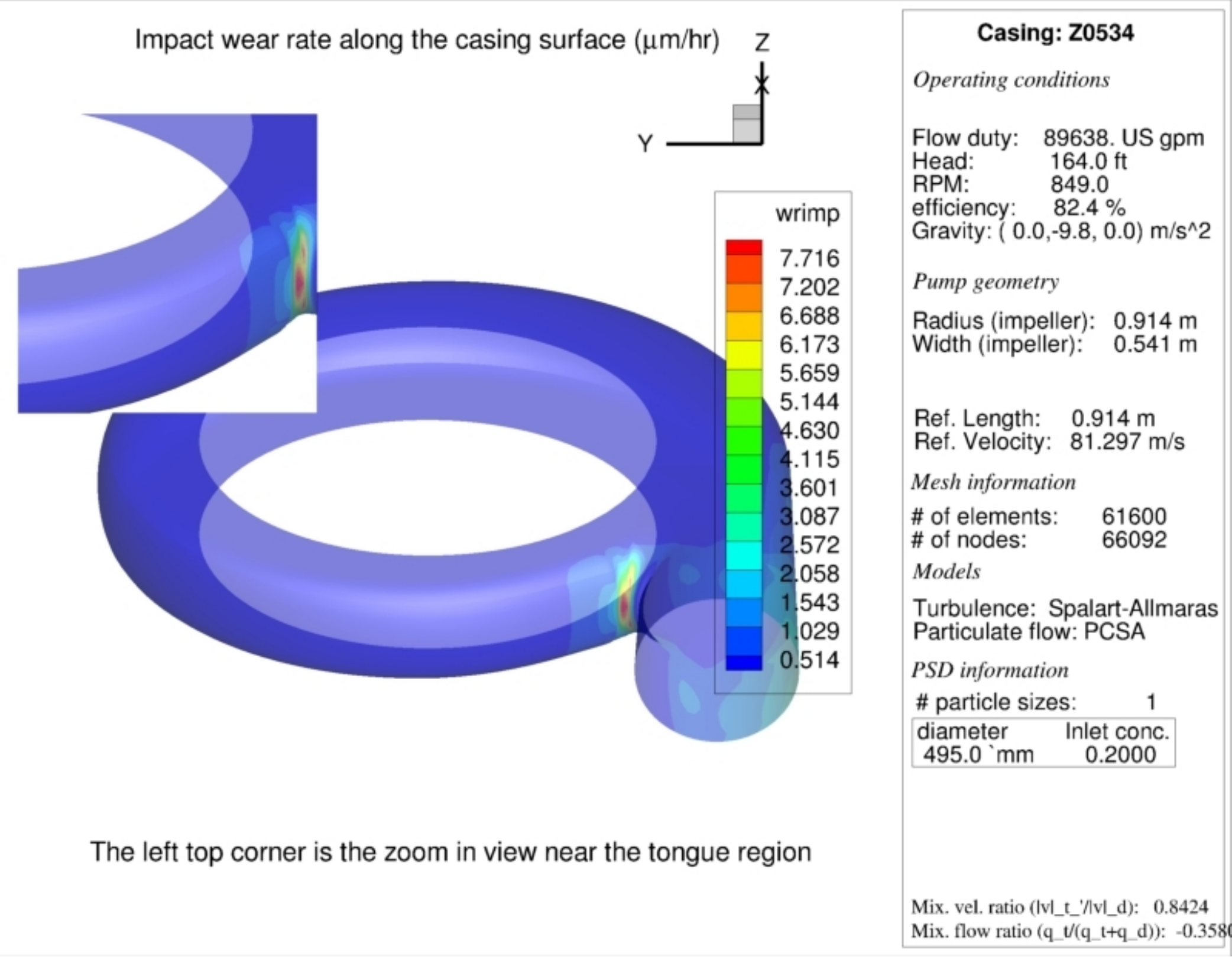}}

\centering
\subcaptionbox{Original design.
\label{fig:comparisonWear:original:SlidingWear}
}
  [.45\linewidth]{\includegraphics[width=0.4001\textwidth, keepaspectratio]{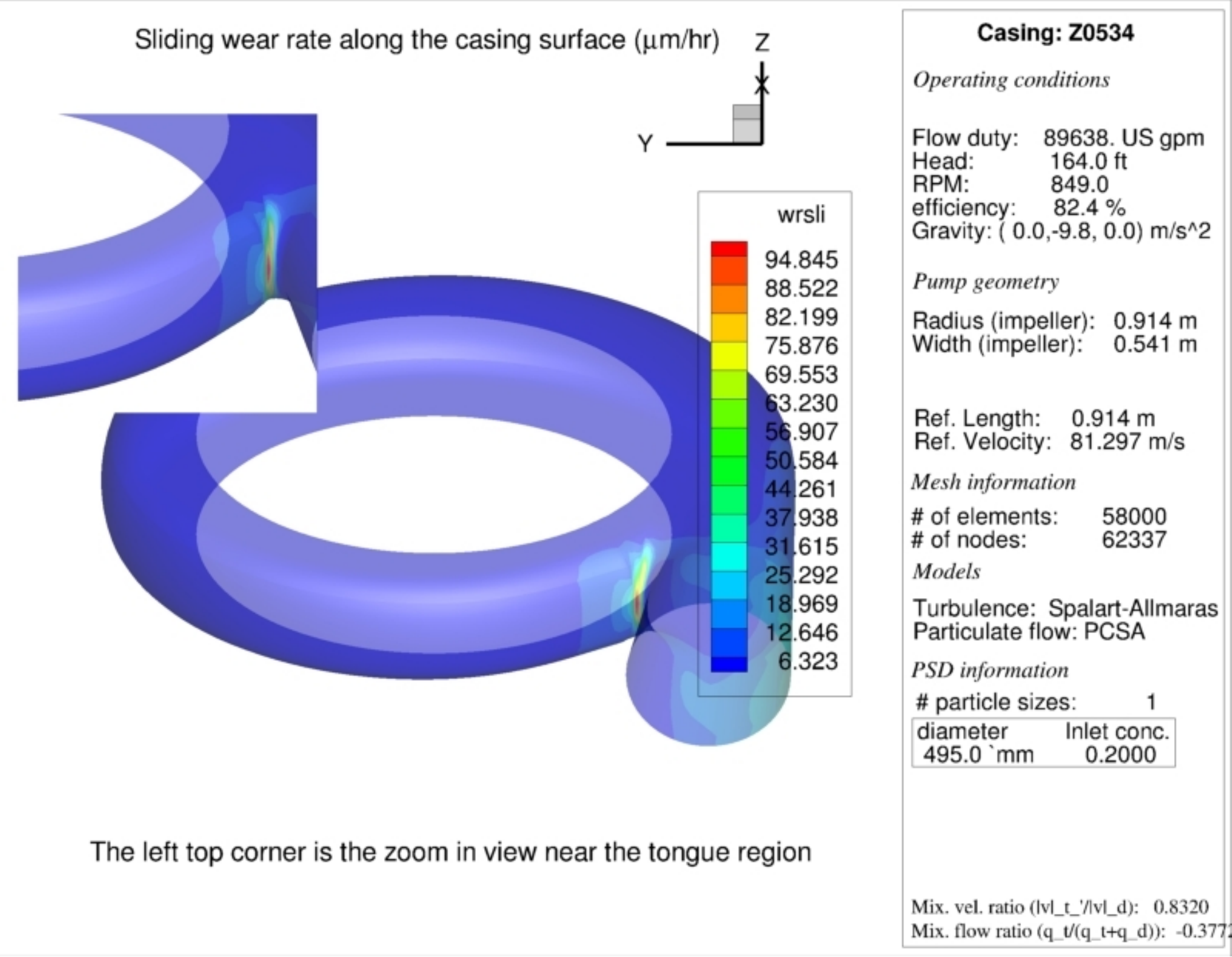}}
\hfill
\subcaptionbox{Optimal design.
\label{fig:comparisonWear:optimal:SlidingWear}
}
  [.45\linewidth]{\includegraphics[width=0.4001\textwidth, keepaspectratio]{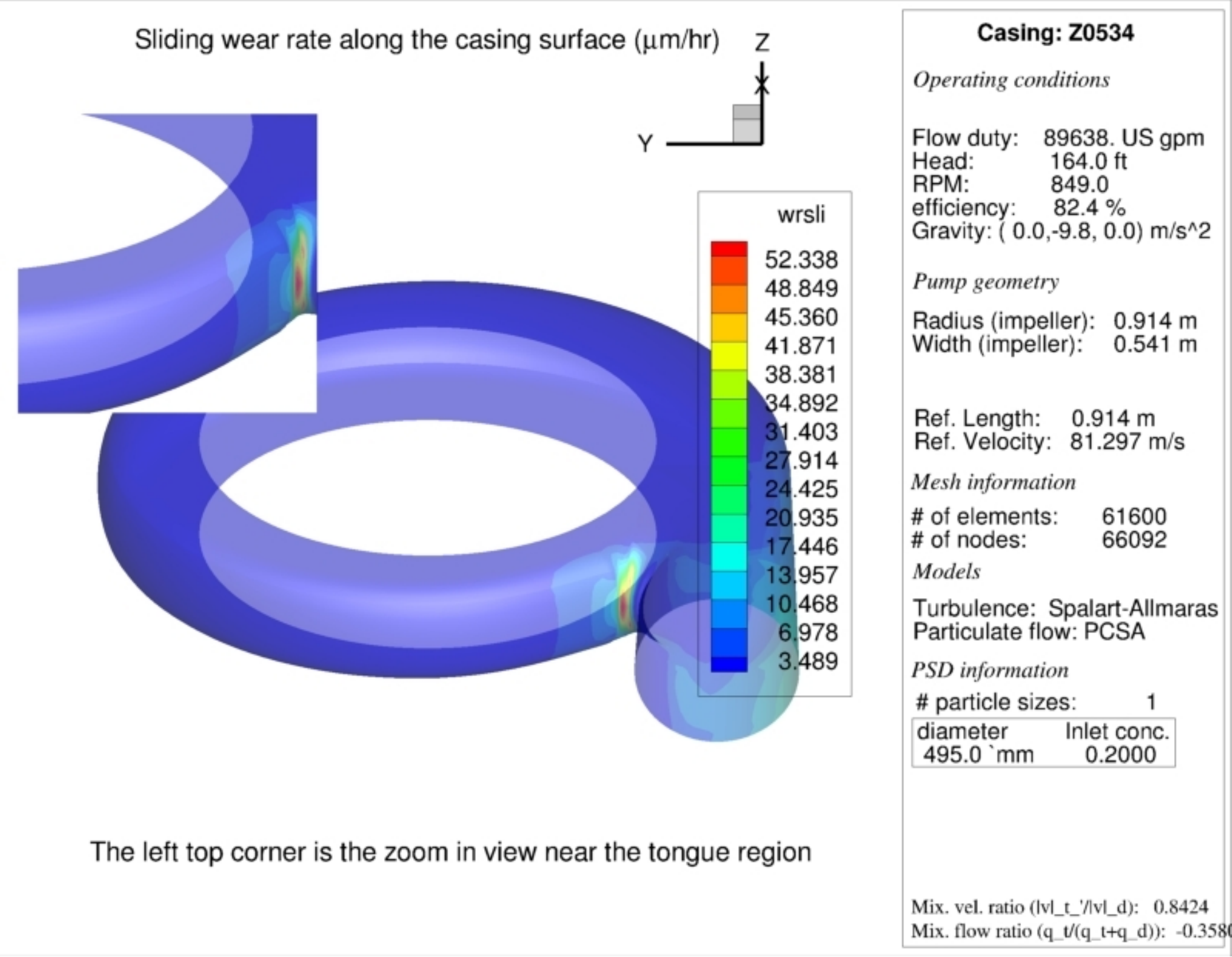}}

\centering
\subcaptionbox{Original design.
\label{fig:comparisonWear:original:TotalWear}
}
  [.45\linewidth]{\includegraphics[width=0.4001\textwidth, keepaspectratio]{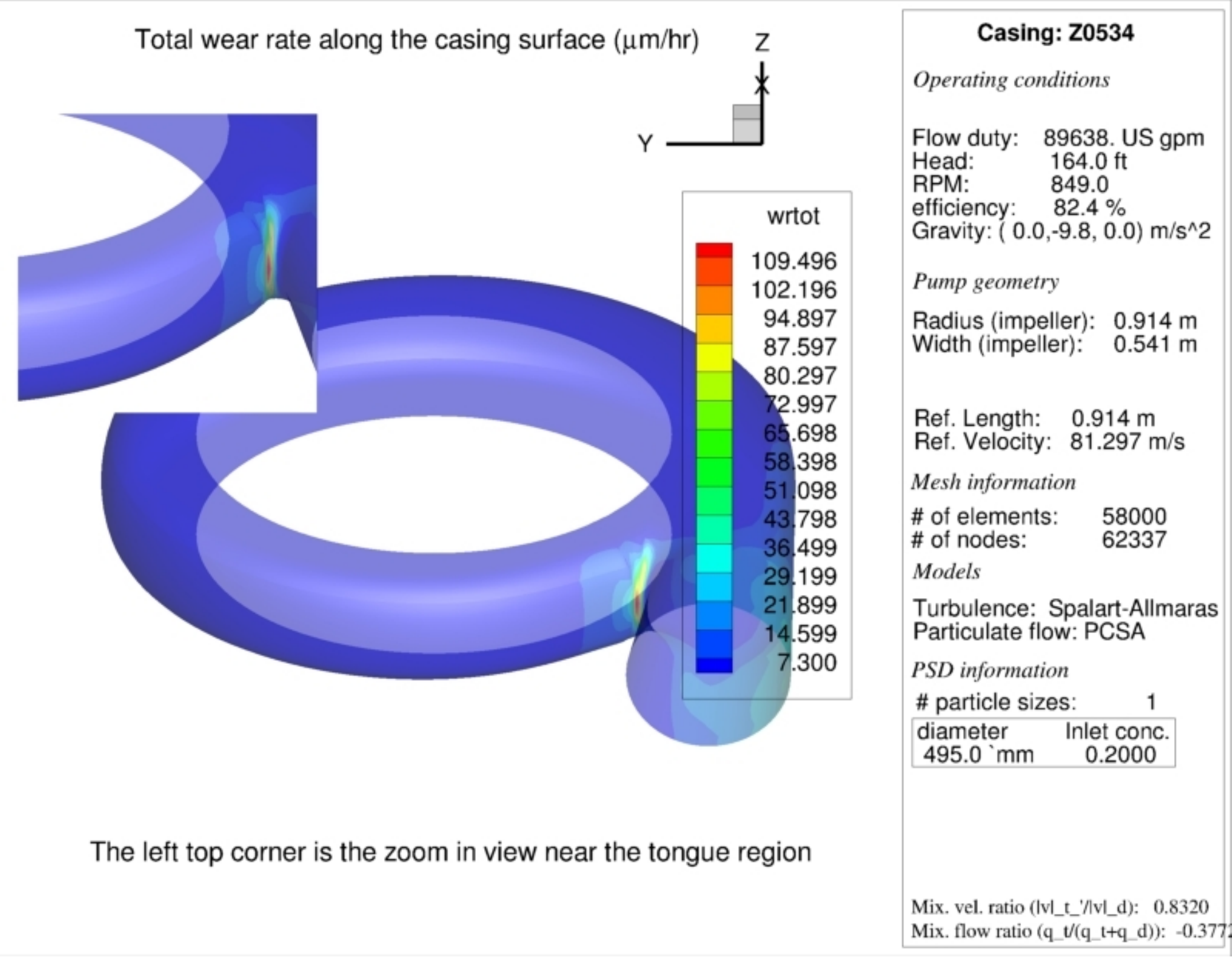}}
\hfill
\subcaptionbox{Optimal design.
\label{fig:comparisonWear:optimal:TotalWear}
}
  [.45\linewidth]{\includegraphics[width=0.4001\textwidth, keepaspectratio]{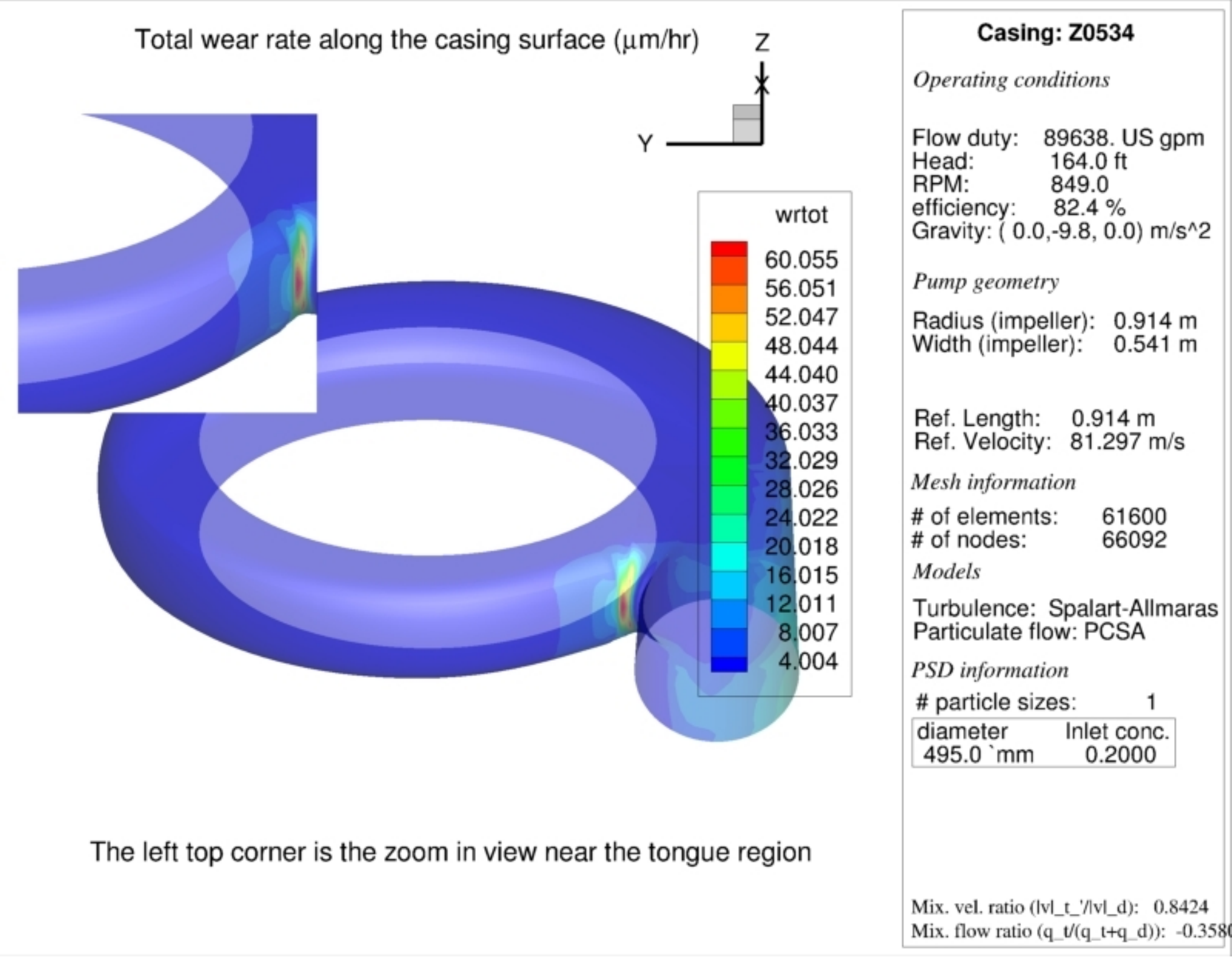}}
\caption{Comparison of wear.}
\label{fig:comparisonWear}
\end{figure}

Figure \ref{fig:comparisonMixtureVelocity} compares the velocity fields in original and optimal design, where Figure \ref{fig:comparisonMixtureVelocity:original:CrossSection} and Figure \ref{fig:comparisonMixtureVelocity:optimal:CrossSection} show the mixture velocity fields of the original and optimal designs, respectively, at different cross sections. It is observed that secondary flows significantly increase in the optimal design, which may help reducing the wear rate on the casing wall. The magnitude of the velocity fields are comparable to each other. Figure \ref{fig:comparisonMixtureVelocity:original:MidPlane} and Figure \ref{fig:comparisonMixtureVelocity:optimal:MidPlane} show the mixture velocity contour in the symmetric midplane of the slurry pump casing. Even though the geometry models are slightly different, the velocity fields in the midplane are nearly identical to each other, in terms of both direction and magnitude. 

Figure \ref{fig:comparisonOverallConcentration} compares the concentration fields between the original and optimal designs. Figure \ref{fig:comparisonOverallConcentration:original:CrossSection} and Figure \ref{fig:comparisonOverallConcentration:optimal:CrossSection} show the overall concentration fields in original and optimal designs, respectively. The concentration field of the optimal design is slightly higher than that of the original design. Differences in geometry models between the original and optimal designs at various cross sections are shown. The height of the cross sections are significantly increased, whereas the width of the cross sections are slightly decreased. The pattern of the concentration fields are also similar between the original and optimal designs. 

Figure \ref{fig:comparisonWear} compares the wear, which are the main concern in this case study, between the original and optimal designs. Figure \ref{fig:comparisonWear:original:ImpactWear} and Figure \ref{fig:comparisonWear:optimal:ImpactWear} show the impact wear in the casing between the original and optimal designs, respectively, where the optimal design shows significant reduction in terms of magnitude. Near the tongue region of the casing, the hot spot area slightly increases. Figure \ref{fig:comparisonWear:original:SlidingWear} and Figure \ref{fig:comparisonWear:optimal:SlidingWear} show the sliding wear between  original and optimal designs, respectively. Again, significant reduction in sliding wear is observed, which subsequently lead to a significant reduction in total wear, shown in Figure \ref{fig:comparisonWear:original:TotalWear} and Figure \ref{fig:comparisonWear:optimal:TotalWear} for original and optimal designs, respectively. 
For a comprehensive comparison between original and optimal casing designs, we refer the readers to \ref{app:CFDcomparison}.

% Figure \ref{fig:3dCasingWearOriginal} presents the total wear of the original slurry pump casing design, whereas Figure \ref{fig:3dCasingWearOptimal} presents the total wear of the optimal design. 
% In both cases, the peak wear appears at the casing tongue, which is very common in slurry pump casings. 
% The pressure and velocity fields slight change, as a result of the change in the pump casing geometry. 
% Figure \ref{fig:3dCasingOriginalDesign} shows the original design, whereas Figure \ref{fig:3dCasingOptimalDesign} shows the optimal design. 
% Both figures are shown with total wear contour in $\mu m/hr$. 
% In the optimal design, the diameter of the pump casing discharge slightly reduces. The inner radius of the pump casing also reduces, however, the outer radius of the pump casing increases, resulting in a deeper optimized casing. 

\section{Discussion}
\label{sec:Discussion}

One important feature of asynchronous BO is that the users can be more patient with the cases that are difficult to converge (i.e. converge after a long time), particularly for applications where the computational runtime varies widely. 
% In particular, it is very useful for applications with a wide range of computational run time. 
The main reason is that the HPC budget is used efficiently in the aphBO-2GP-3B framework, thus increasing the aggressiveness by reducing the waiting time will not improve the efficiency much. 
Compared to other synchronous batch-sequential parallel approaches, in which the whole batch might halt due to one ill-conditioned simulation, the asynchronous parallel feature relaxes the constraints of thresholding the computational time for the simulation. 
However, it is recommended to choose the cutoff computational time appropriately depending on the application.

Multiple acquisition functions are considered using the GP-Hedge \cite{hoffman2011portfolio} approach, resulting in further increases in computational efficiency. Compared to our previous pBO-2GP-3B approach, the aphBO-2GP-3B framework is improved in terms of computational efficiency. This is achieved by firstly reducing the waiting time of other computational workers, and secondly by considering multiple acquisition functions and promoting the one which corresponds to better performance.

One of the main drawbacks of BO is the scalability of the surrogate GP model, which prevents the traditional BO method from observing more than $10^4$ sample points.   
However, this issue is usually not critical for applications with high-fidelity expensive simulations. 
Because the computational time for one simulation is already computationally substantial, $10^4$ simulations would be significantly computationally expensive. 
However, scalable GP methods exist to cope with its scalability drawback. For example, a few well-known methods are subset of data \cite{tran2018efficient}, the subset of regressor, the deterministic training conditional, and the partially and fully independent training conditions approximations \cite{quinonero2007approximation,chalupka2013framework}.

The aphBO-2GP-3B framework can also be extended to solve equality constraints, for example, by adopting the augmented Lagrangian approach, as well as optimization under uncertainty where noisy evaluation is common. 
Such problems are more challenging due to its noisy nature. For noisy problems, one can consider Letham et al. \cite{letham2017constrained} approach or stochastic kriging \cite{ankenman2010stochastic} to further improve the current framework.

In this paper, the batch sizes are assumed to be constant and are user-defined parameters. More adaptive approaches can be used to further improve the computational efficiency of the proposed aphBO-2GP-3B framework, where more exploration is promoted at the beginning of the optimization process, and more exploitation is promoted later on. 
This can be achieved by generalizing the GP-Hedge scheme to include more acquisition functions. 
However, the inner parameters of GP-Hedge may need to be calibrated again, as the objectives of batches differ greatly.

One may can also consider to integrate a dynamic resource allocation on the top of the current aphBO-2GP-3B framework to reduce the computational time. 
In HPC settings, the average waiting time and the number of available processors are insightful information that can be used to change the batch sizes adaptively.

We propose to learn the feasible and infeasible regions adaptively by borrowing different binary classifiers in machine learning context. 
There is no restriction in choosing the binary probabilistic classifier. 
However, it is noted that the numerical performance of aphBO-2GP-3BO depends on the performance of the binary classifier, as the probability is used to locate the next sampling point. 
Also, the third batch in the aphBO-2GP-3B framework is designed to force the classifier to learn in its most uncertain regions. 
This feature is only available for a few binary classifier, such as GP, where the uncertainty can be quantified.

However, the current aphBO-2GP-3B framework does not rigorously solve the dynamic resource allocation problem on the HPC platform, where the computational resource is typically shared where computational constraints are also involved. 
This opens up the opportunity for further research in the future to consider the BO method in the context of online dynamic HPC resources.

\section{Conclusion}
\label{sec:Conclusion}

In this paper, we present an asynchronous parallel BO method that supports known and unknown constraints in the context of HPC platform. 
We show that the proposed aphBO-2GP-3B framework can be easily applied for computationally expensive high-fidelity applications in industrial settings. 
The aphBO-2GP-3B framework is constructed based on two GPs associated with three distinct batches. 
The first GP corresponds to the objective function, as in the traditional BO method. 
The first GP is associated with the first and second batch, where the first batch aims to optimize and the second batch aims to explore most uncertain regions. 
The second GP corresponds to the binary classifier, where uncertainty is measured, and the GP classifier is forced to learn according to the third batch. 
The hallucination module, which is similar to the kriging heuristic liar approach, allows to construct the GP at the sampling locations where observations are not made yet. 
This allows the BO framework to move forward and parallelize the optimization, while temporarily disintegrating with the application and waiting for its feedback on completion later on. 

The aphBO-2GP-3B approach is demonstrated using two industrial applications. 
The first application is concerned with the design of flip-chip package, where a thermo-mechanical FEA model is used to predict the fatigue life. 
The second application is concerned with the design of slurry pump casing, where a 3D multiphase CFD simulation is used to predict the wear rate of the pump casing. 
It has been shown the aphBO-2GP-3B can highly parallelize across different multi-core HPCs, thus demonstrating the effectiveness of the proposed framework.

\section*{Acknowledgment}

AT thanks Aaron Cutright (GIW Industries) for his kind technical support, as well as Krishnan V. Pagalthivarthi for his CFD implementation used in this paper. 
% This research was supported in part through research cyberinfrastructure resources and services provided by the Partnership for an Advanced Computing Environment (PACE) at the Georgia Institute of Technology, Atlanta, Georgia, USA. \cite{PACE}. 
The views expressed in the article do not necessarily represent the views of the U.S. Department of Energy or the United States Government. 
Sandia National Laboratories is a multimission laboratory managed and operated by National Technology and Engineering Solutions of Sandia, LLC., a wholly owned subsidiary of Honeywell International, Inc., for the U.S. Department of Energy's National Nuclear Security Administration under contract DE-NA-0003525.

% \bibliography{mybibfile}
\bibliographystyle{elsarticle-num}
\bibliography{lib}

\appendix
\clearpage
\section{Numerical benchmark results}

In this appendix, we provide more benchmark results for the proposed asynchronous parallel method, aphBO-2GP-3B, particularly on the portfolio of the acquisition function sampled. 
\begin{itemize}
\item eggholder (2d): batch size = (2,2,0): max iteration = 80
\item three-hump camel (2d): batch size = (2,2,0): max iteration = 80
\item six-hump camel (2d): batch size = (3,1,0): max iteration = 80
\item hartmann (3d): batch size = (3,3,0): max iteration = 150
\item hartmann (4d): batch size = (4,4,0): max iteration = 160
\item ackley (free d) (5d): batch size = (6,4,0): max iteration = 200
\item hartmann (6d): batch size = (5,5,0): max iteration = 300
\item michalewicz (free d) (10d): batch size = (5,5,0): max iteration = 400
\item perm0db (free d) (80d): batch size = (6,4,0): max iteration = 400
\item rosenbrock (free d) (20d): batch size = (6,4,0): max iteration = 400
\item dixon-price (free d) (25d): batch size = (7,7,0): max iteration = 480
\item trid (free d) (30d): batch size = (6,4,0): max iteration = 400
\item sumsqu (free d) (40d): batch size = (6,4,0): max iteration = 400
\item sumpow (free d) (50d): batch size = (6,4,0): max iteration = 400
\item spheref (free d) (60d): batch size = (6,4,0): max iteration = 400
\item rothyp (free d) (70d): batch size = (6,4,0): max iteration = 400
\end{itemize}

\begin{figure}[!htbp]
\centering
\subcaptionbox{egg: acquisition portfolio.
}
  [0.30\linewidth]{\includegraphics[width=0.30\textwidth, keepaspectratio]{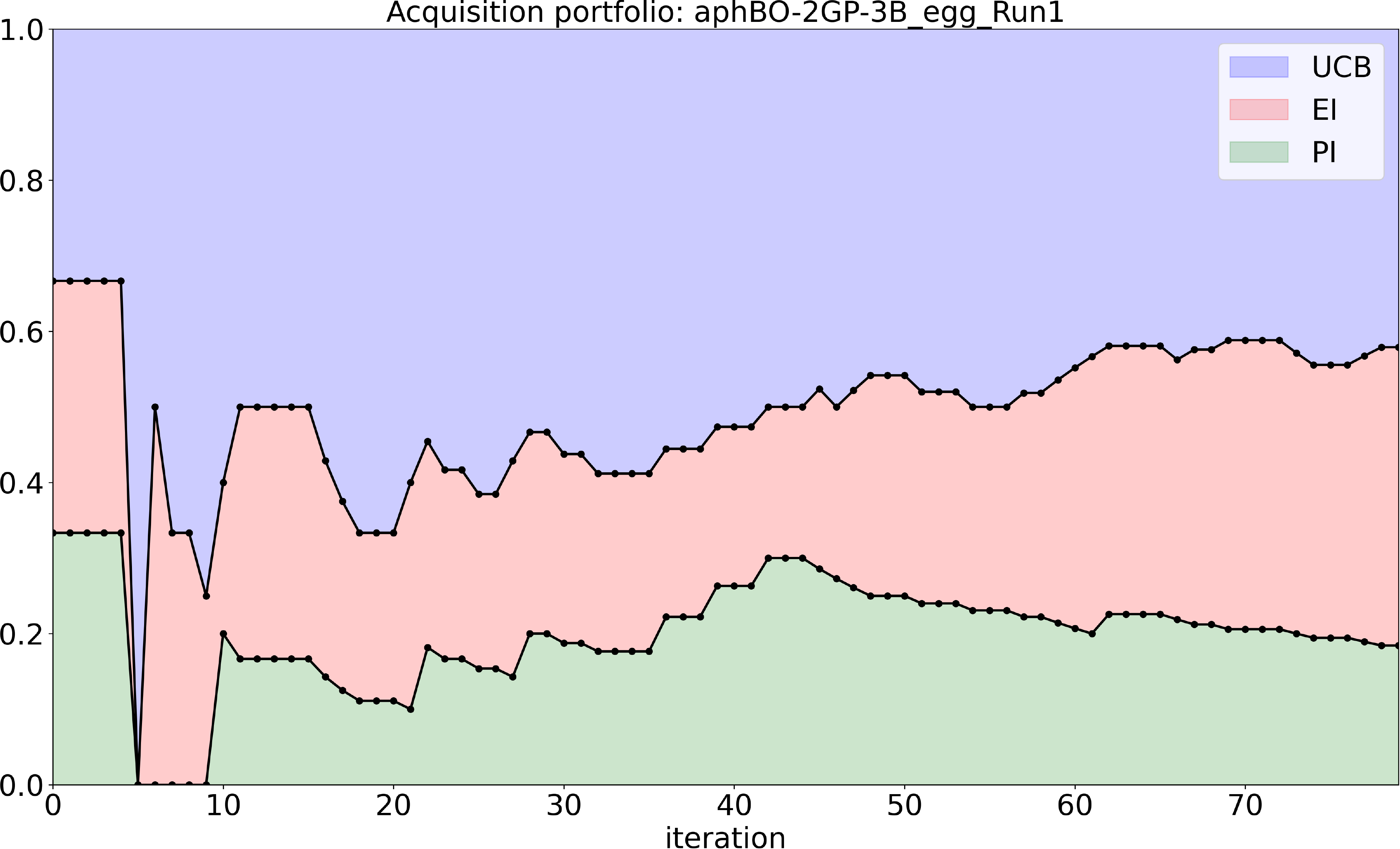}}
\hfill
\subcaptionbox{egg: number of acquisition functions.
}
  [0.30\linewidth]{\includegraphics[width=0.30\textwidth, keepaspectratio]{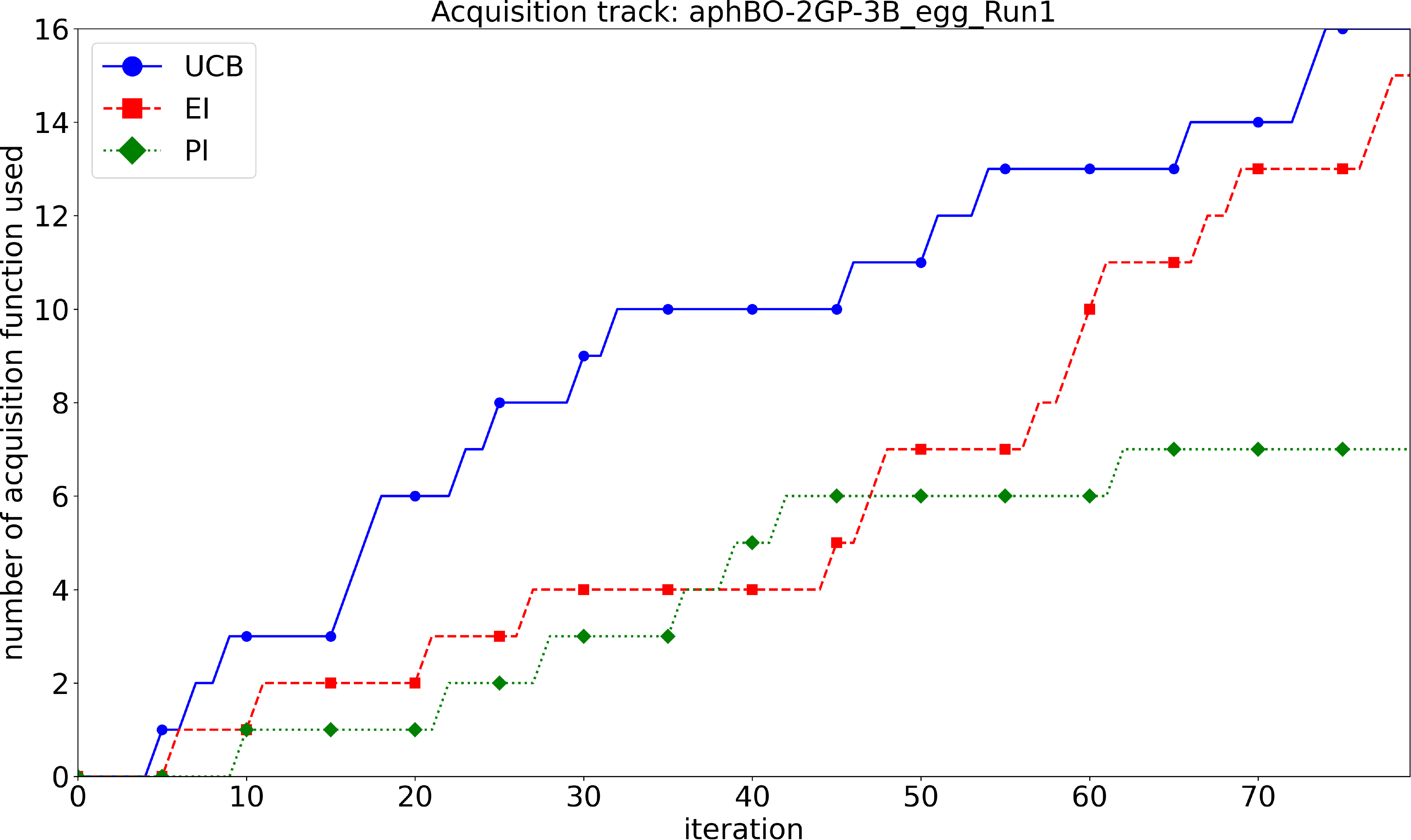}}
\hfill
\subcaptionbox{egg: scheduler dashboard.
}
  [0.30\linewidth]{\includegraphics[width=0.30\textwidth, keepaspectratio]{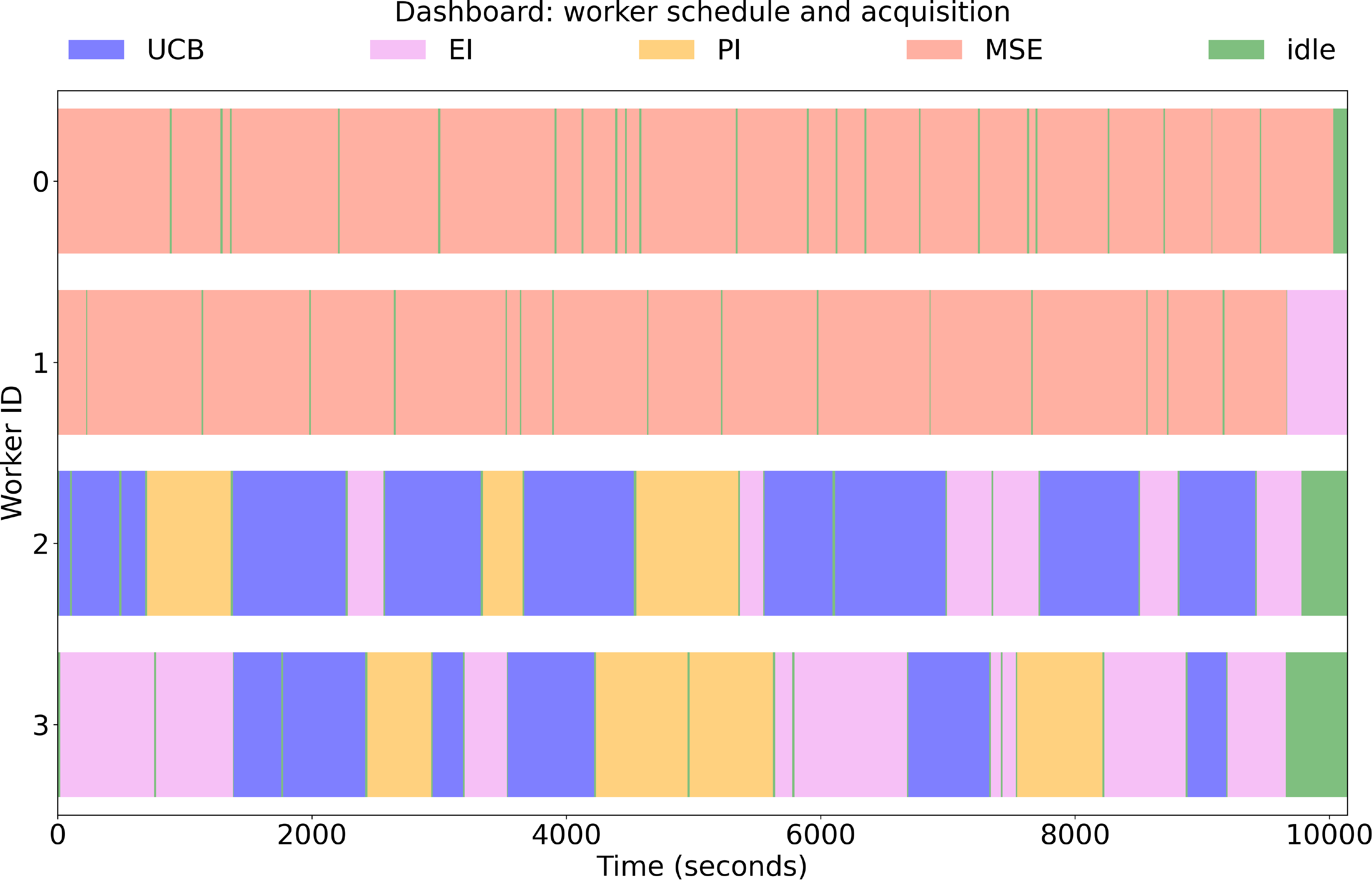}}
% \caption{Acquisition portfolio and scheduler.}
% \label{fig:portfolio_egg}
% \end{figure}
\medskip
% \begin{figure}[!htbp]
\centering
\subcaptionbox{camel6: acquisition portfolio.
}
  [0.30\linewidth]{\includegraphics[width=0.30\textwidth, keepaspectratio]{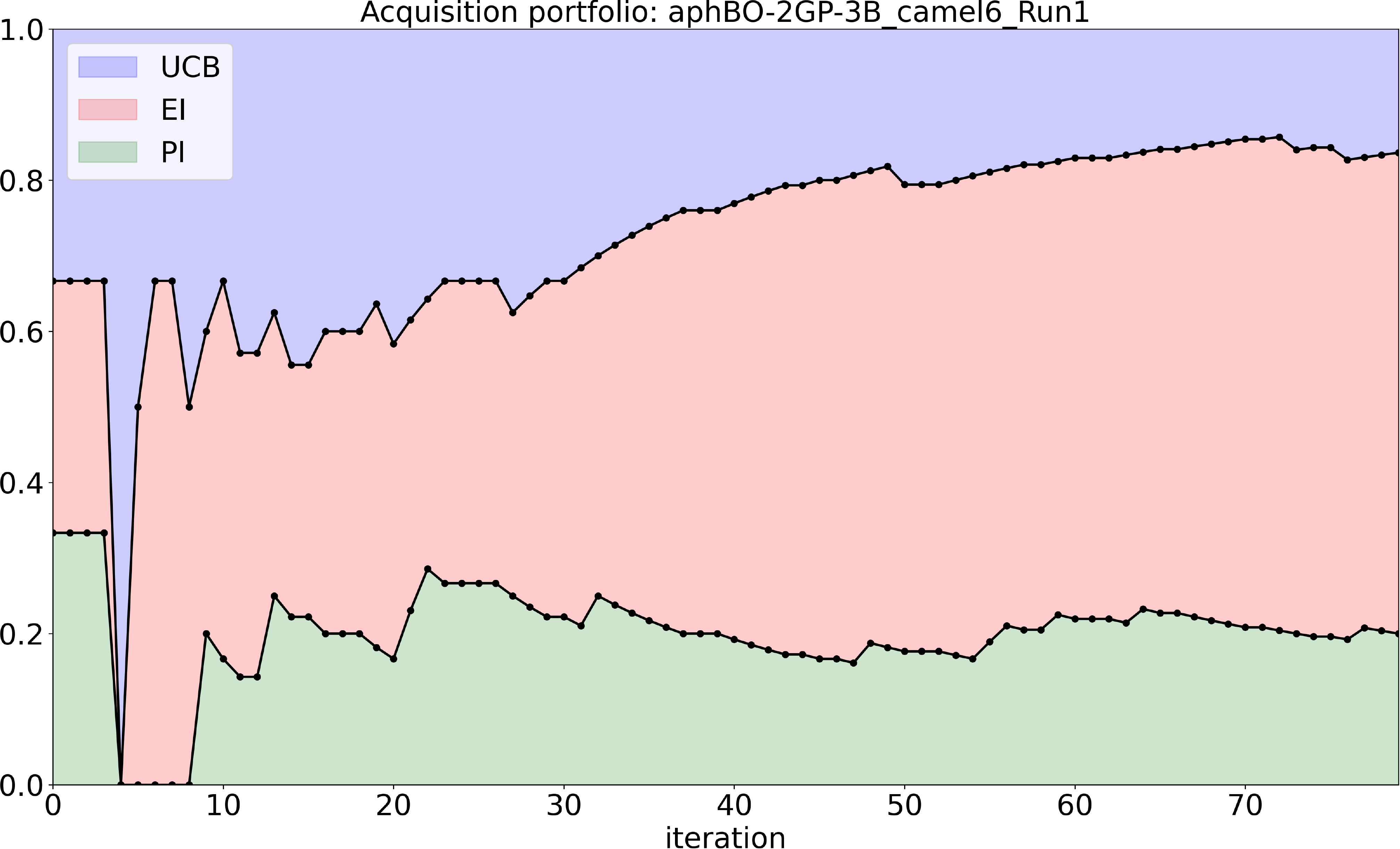}}
\hfill
\subcaptionbox{camel6: number of acquisition functions.
}
  [0.30\linewidth]{\includegraphics[width=0.30\textwidth, keepaspectratio]{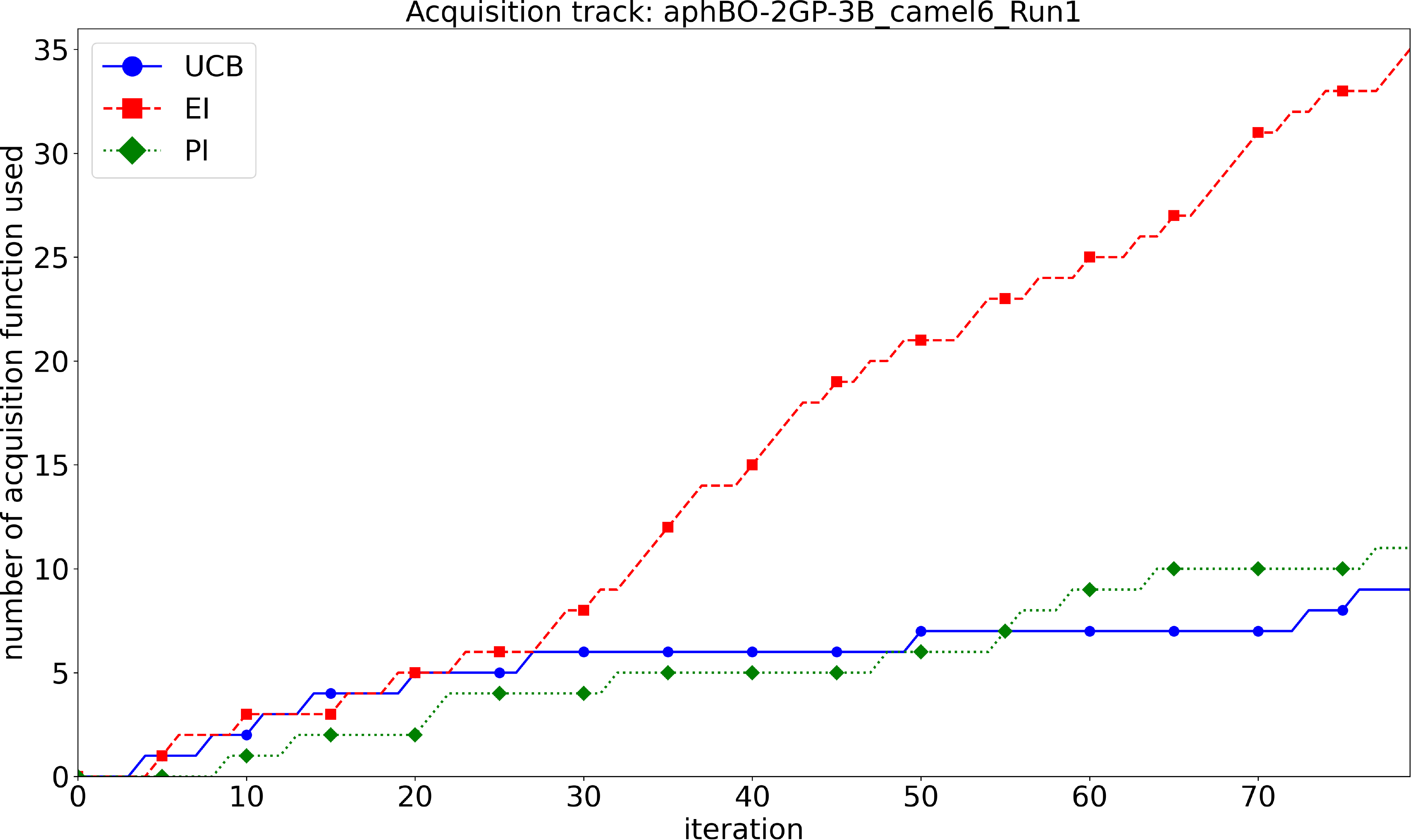}}
\hfill
\subcaptionbox{camel6: scheduler dashboard.
}
  [0.30\linewidth]{\includegraphics[width=0.30\textwidth, keepaspectratio]{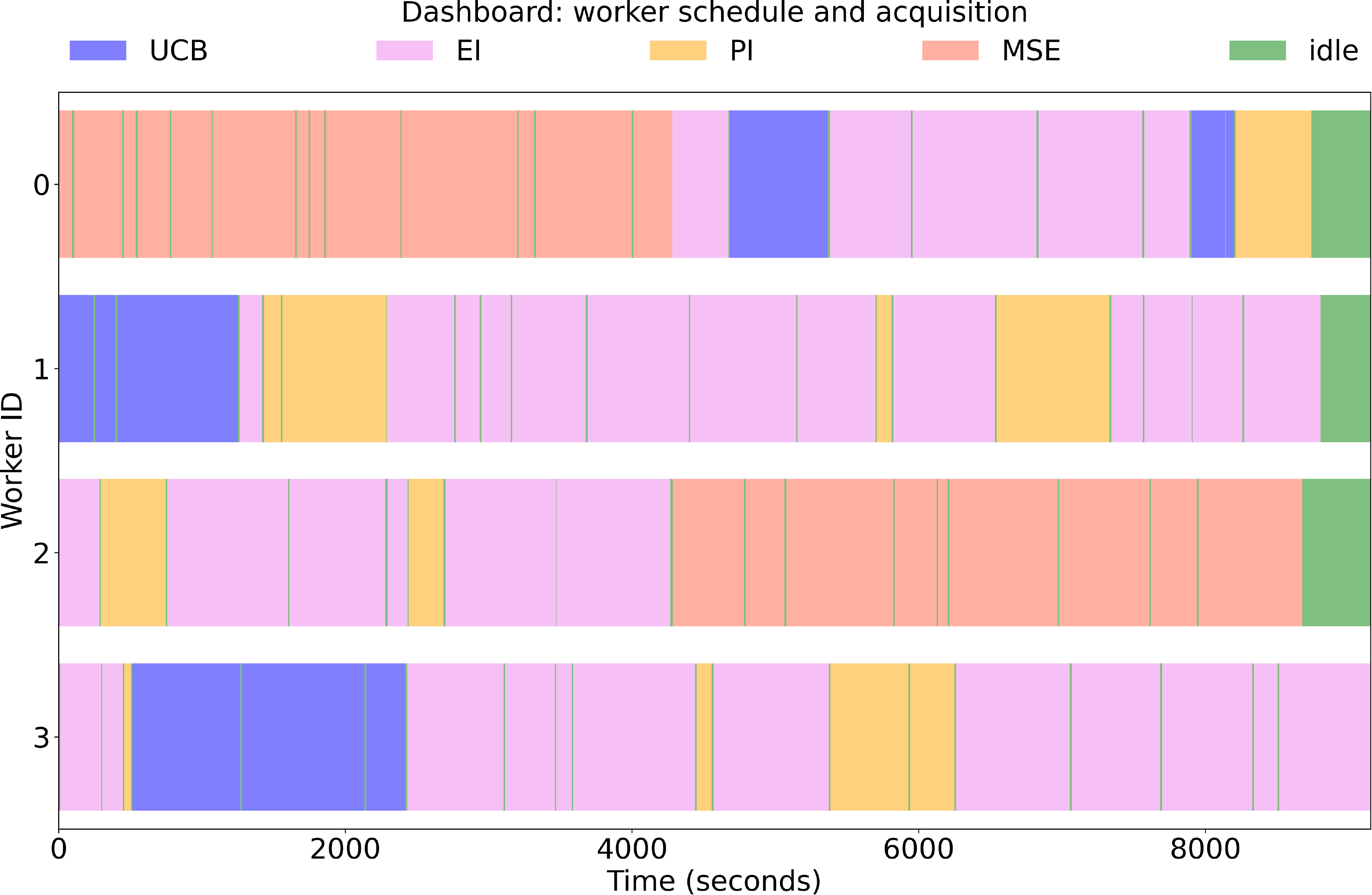}}
% \caption{Acquisition portfolio and scheduler.}
% \label{fig:portfolio_camel6}
% \end{figure}
\medskip
% \begin{figure}[!htbp]
\centering
\subcaptionbox{hart3: acquisition portfolio.
}
  [0.30\linewidth]{\includegraphics[width=0.30\textwidth, keepaspectratio]{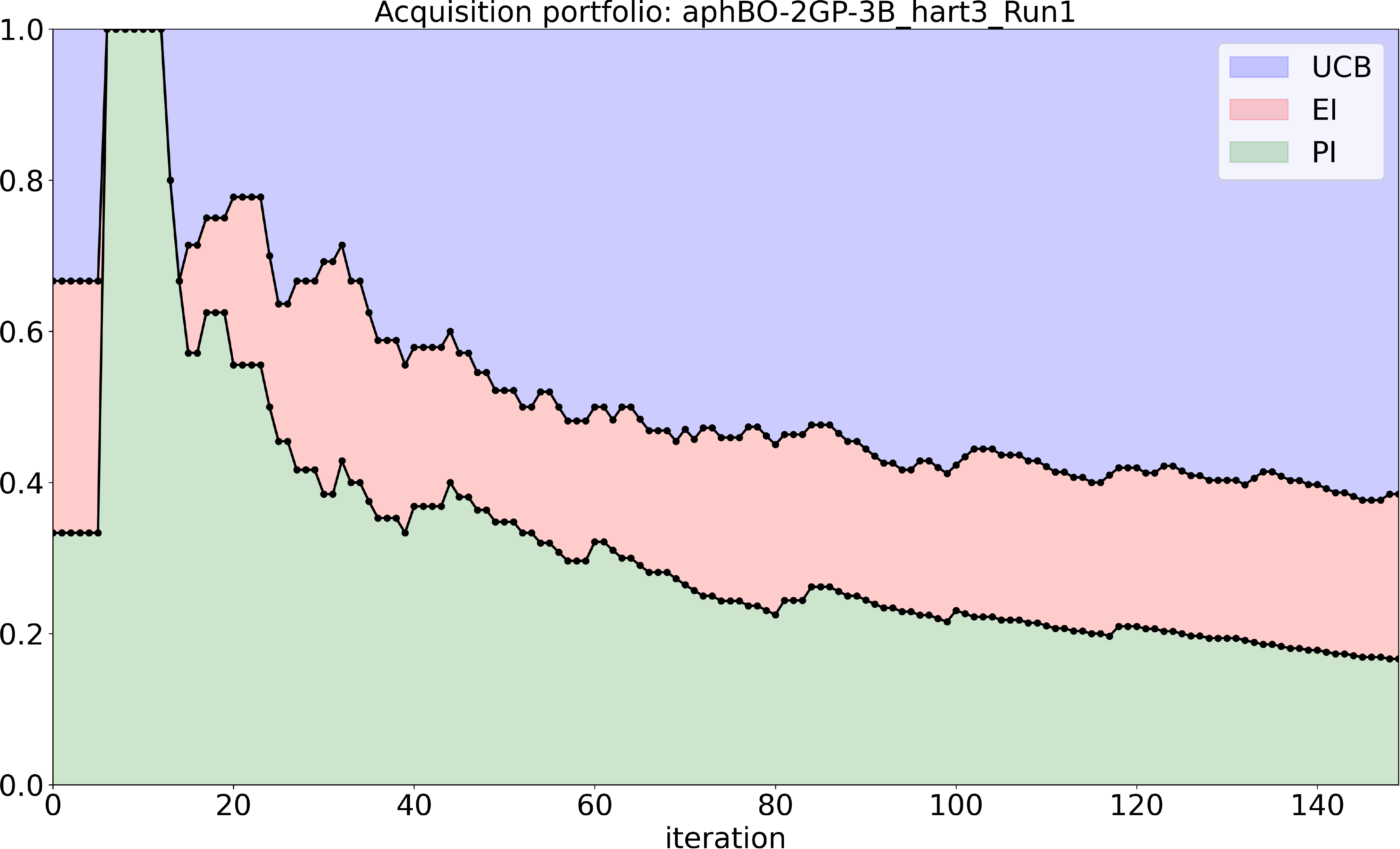}}
\hfill
\subcaptionbox{hart3: number of acquisition functions.
}
  [0.30\linewidth]{\includegraphics[width=0.30\textwidth, keepaspectratio]{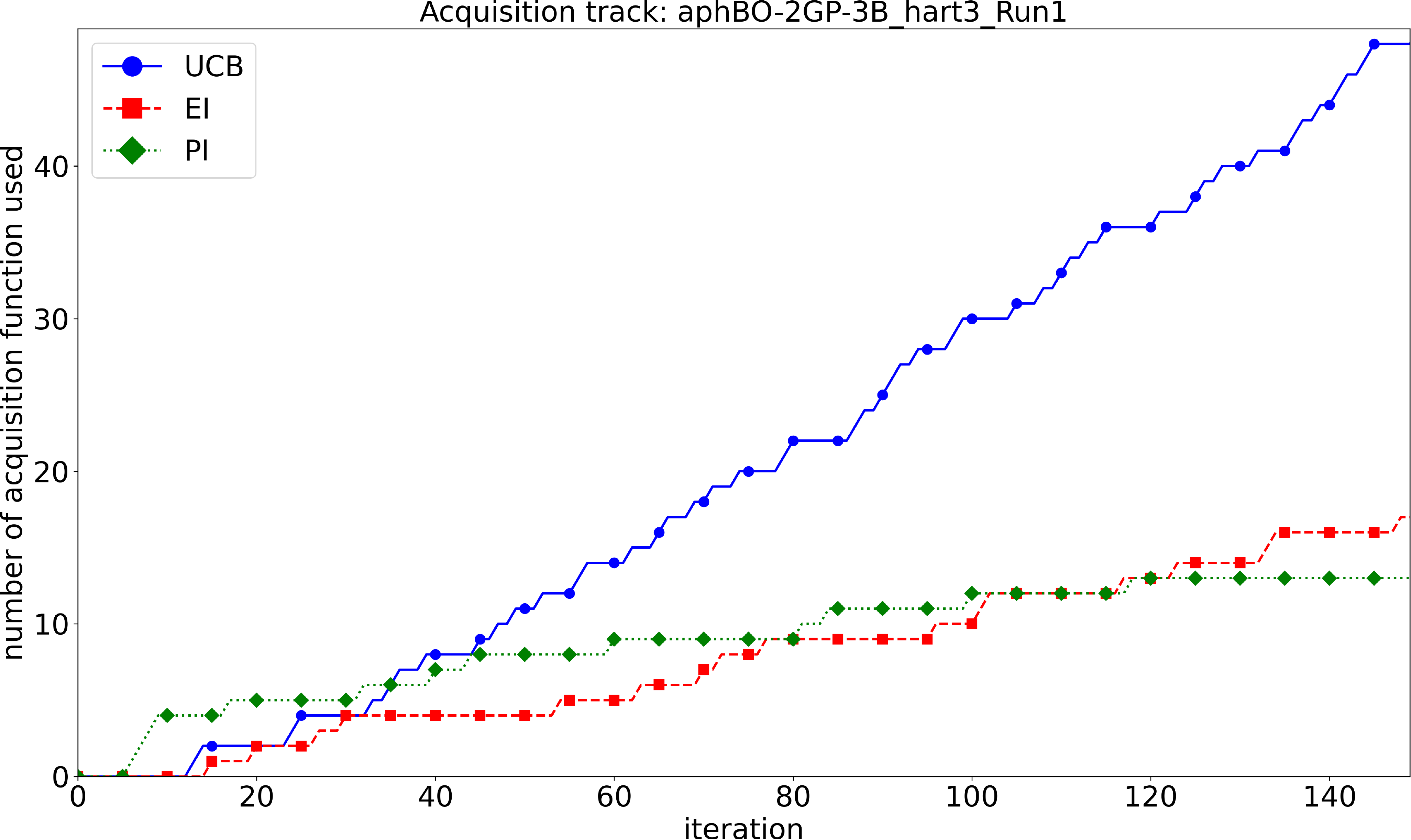}}
\hfill
\subcaptionbox{hart3: scheduler dashboard.
}
  [0.30\linewidth]{\includegraphics[width=0.30\textwidth, keepaspectratio]{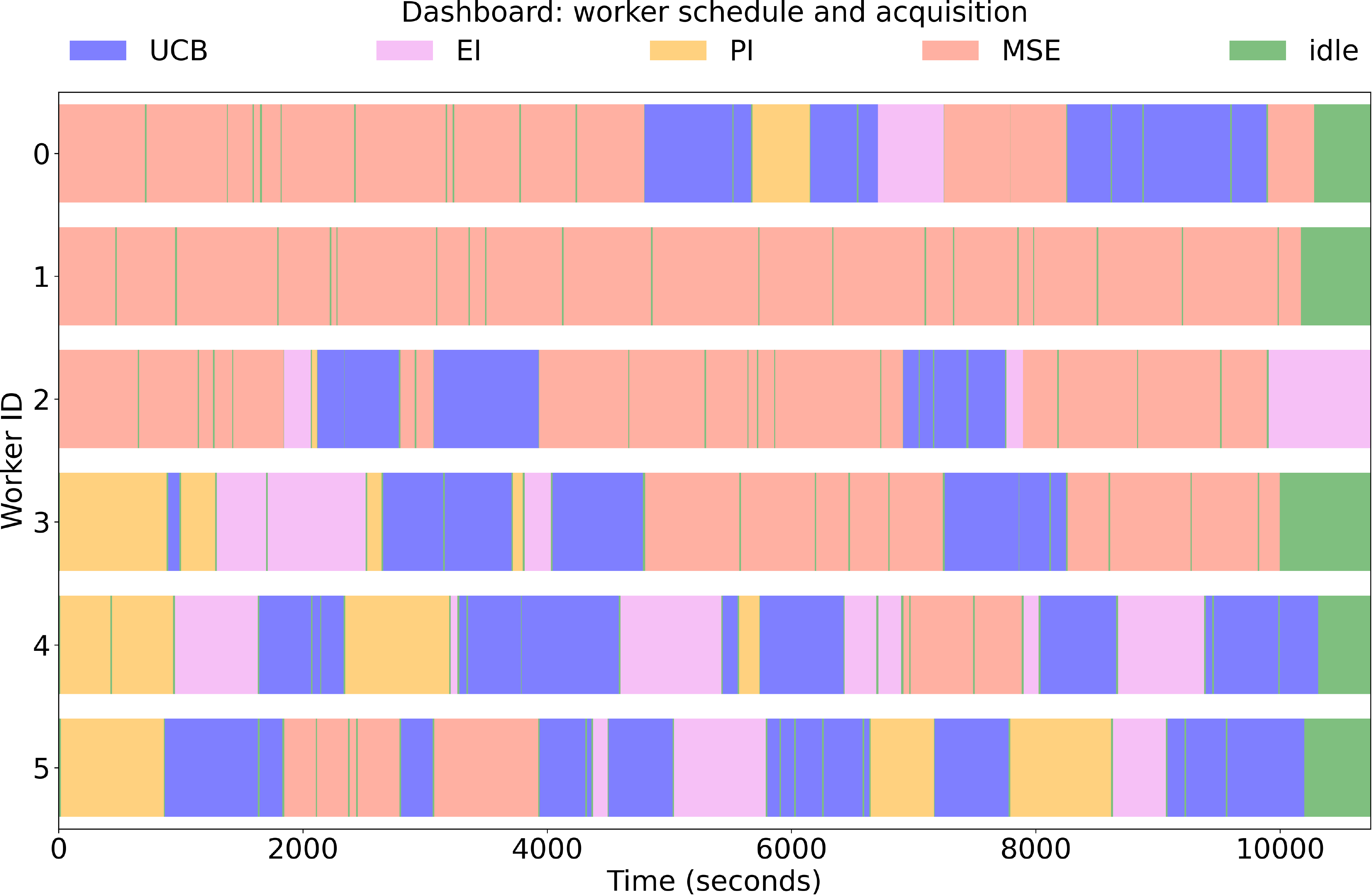}}
% \caption{Acquisition portfolio and scheduler.}
% \label{fig:portfolio_hart3}
% \end{figure}
\medskip
% \begin{figure}[!htbp]
\centering
\subcaptionbox{hart4: acquisition portfolio.
}
  [0.30\linewidth]{\includegraphics[width=0.30\textwidth, keepaspectratio]{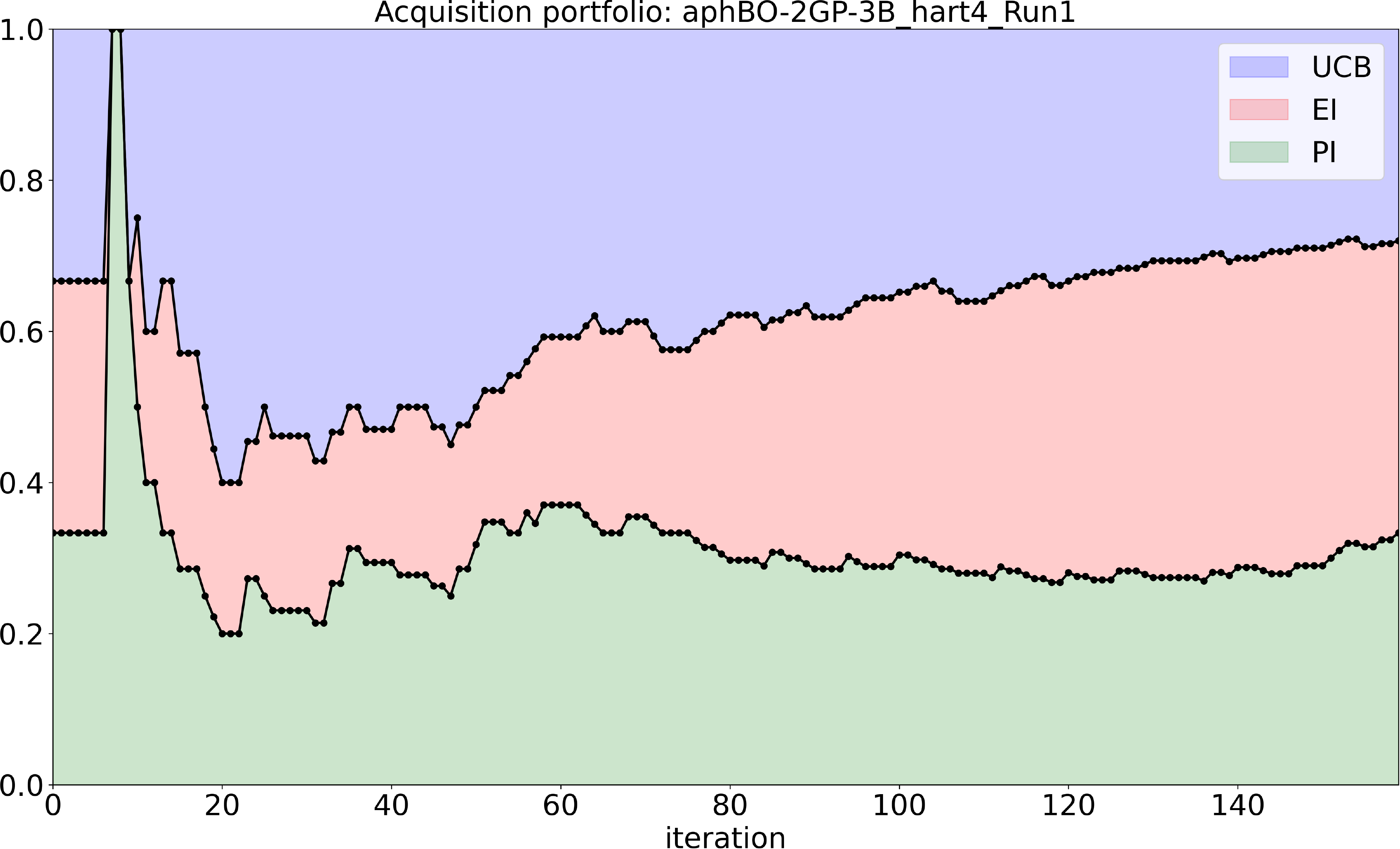}}
\hfill
\subcaptionbox{hart4: number of acquisition functions.
}
  [0.30\linewidth]{\includegraphics[width=0.30\textwidth, keepaspectratio]{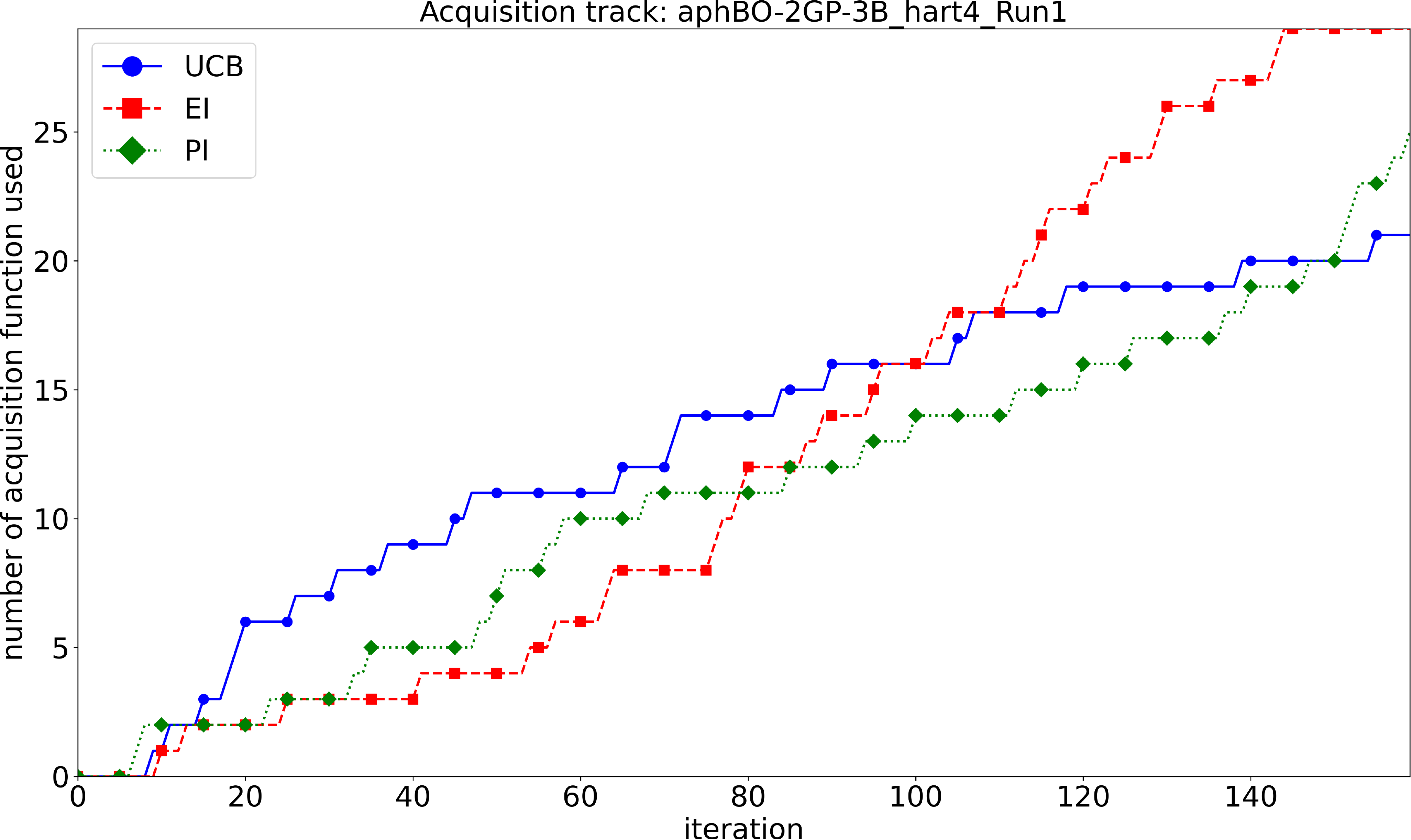}}
\hfill
\subcaptionbox{hart4: scheduler dashboard.
}
  [0.30\linewidth]{\includegraphics[width=0.30\textwidth, keepaspectratio]{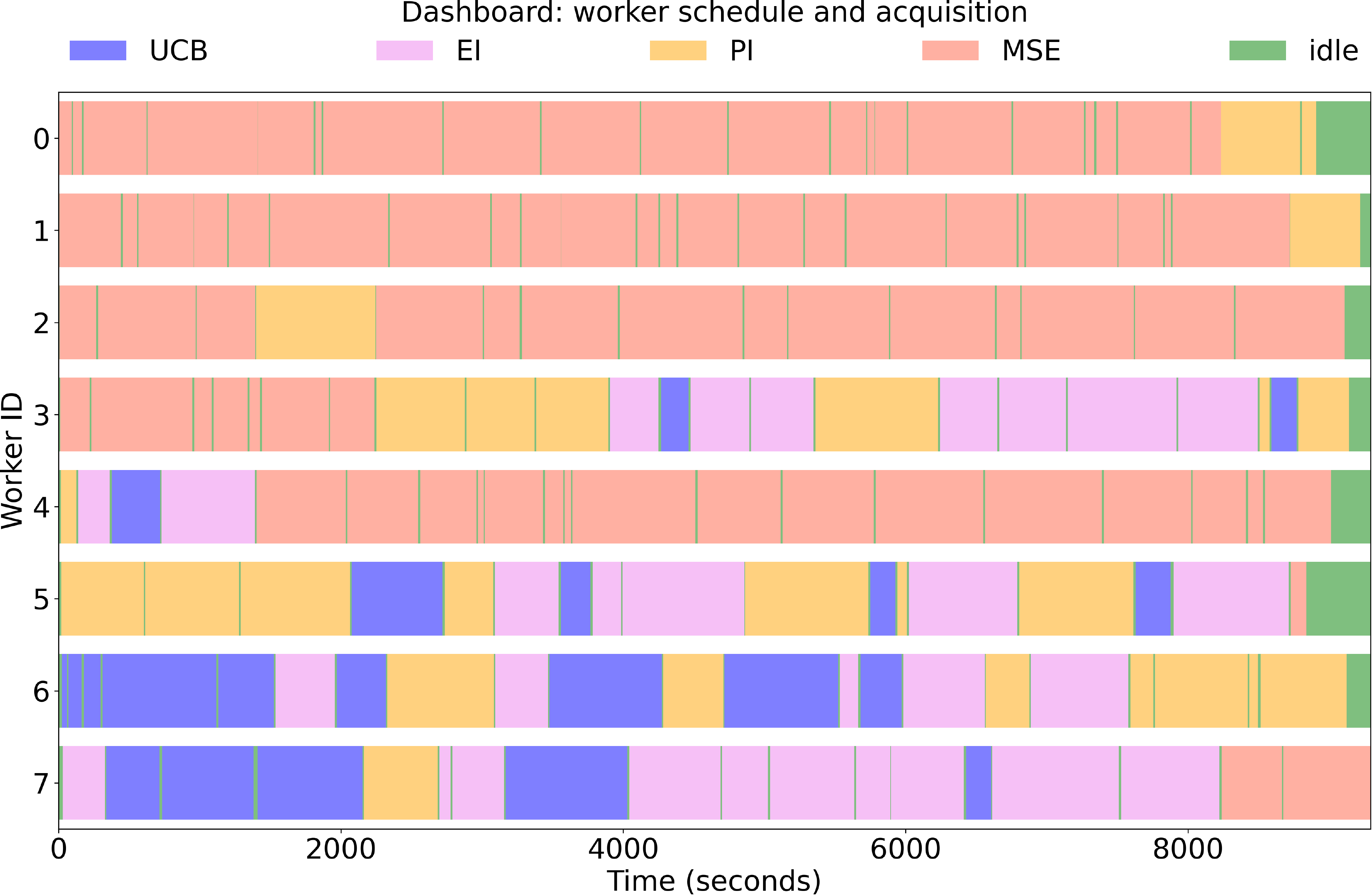}}
% \caption{Acquisition portfolio and scheduler.}
% \label{fig:portfolio_hart4}
% \end{figure}
\medskip
% \begin{figure}[!htbp]
\centering
\subcaptionbox{ackley: acquisition portfolio.
}
  [0.30\linewidth]{\includegraphics[width=0.30\textwidth, keepaspectratio]{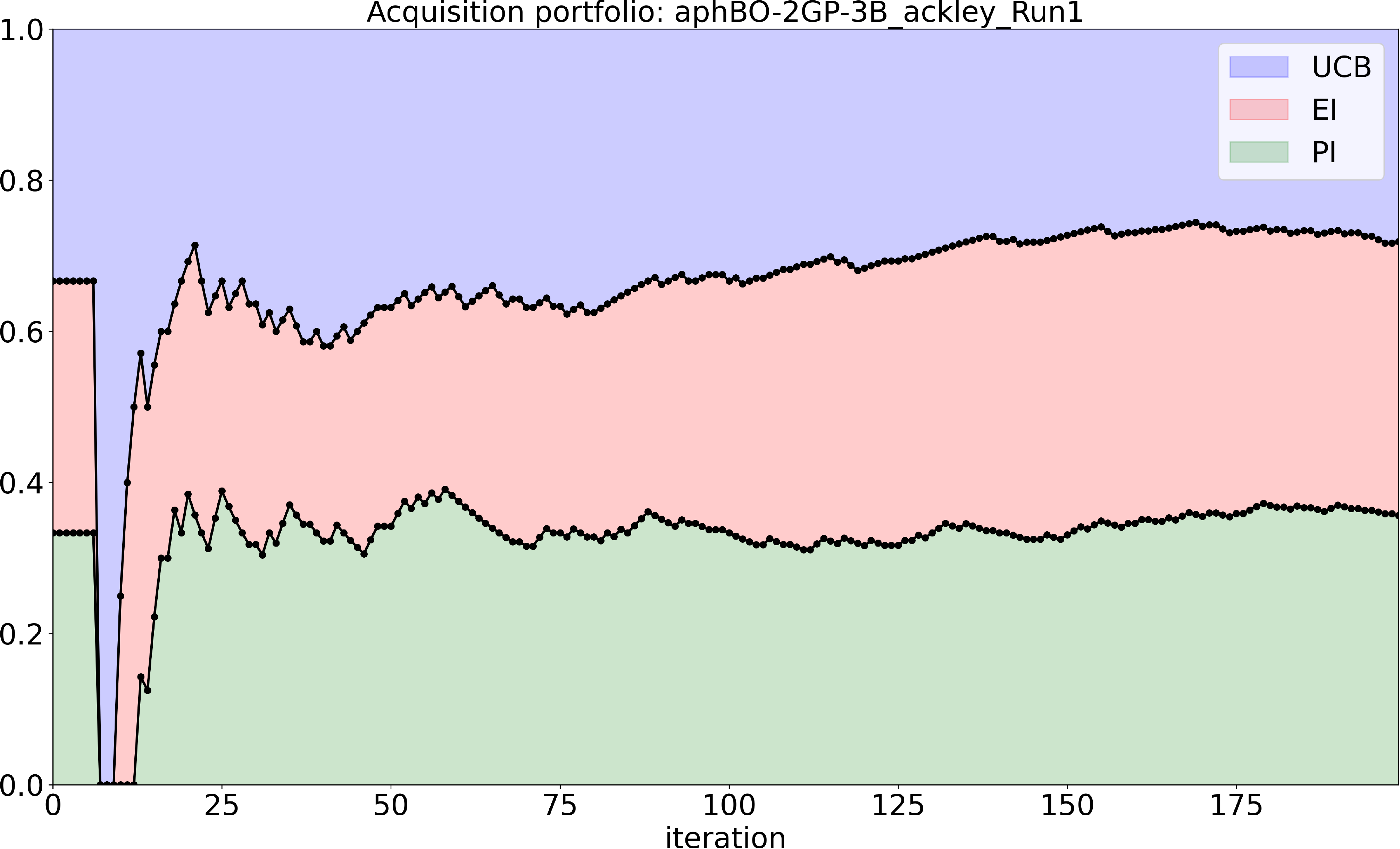}}
\hfill
\subcaptionbox{ackley: number of acquisition functions.
}
  [0.30\linewidth]{\includegraphics[width=0.30\textwidth, keepaspectratio]{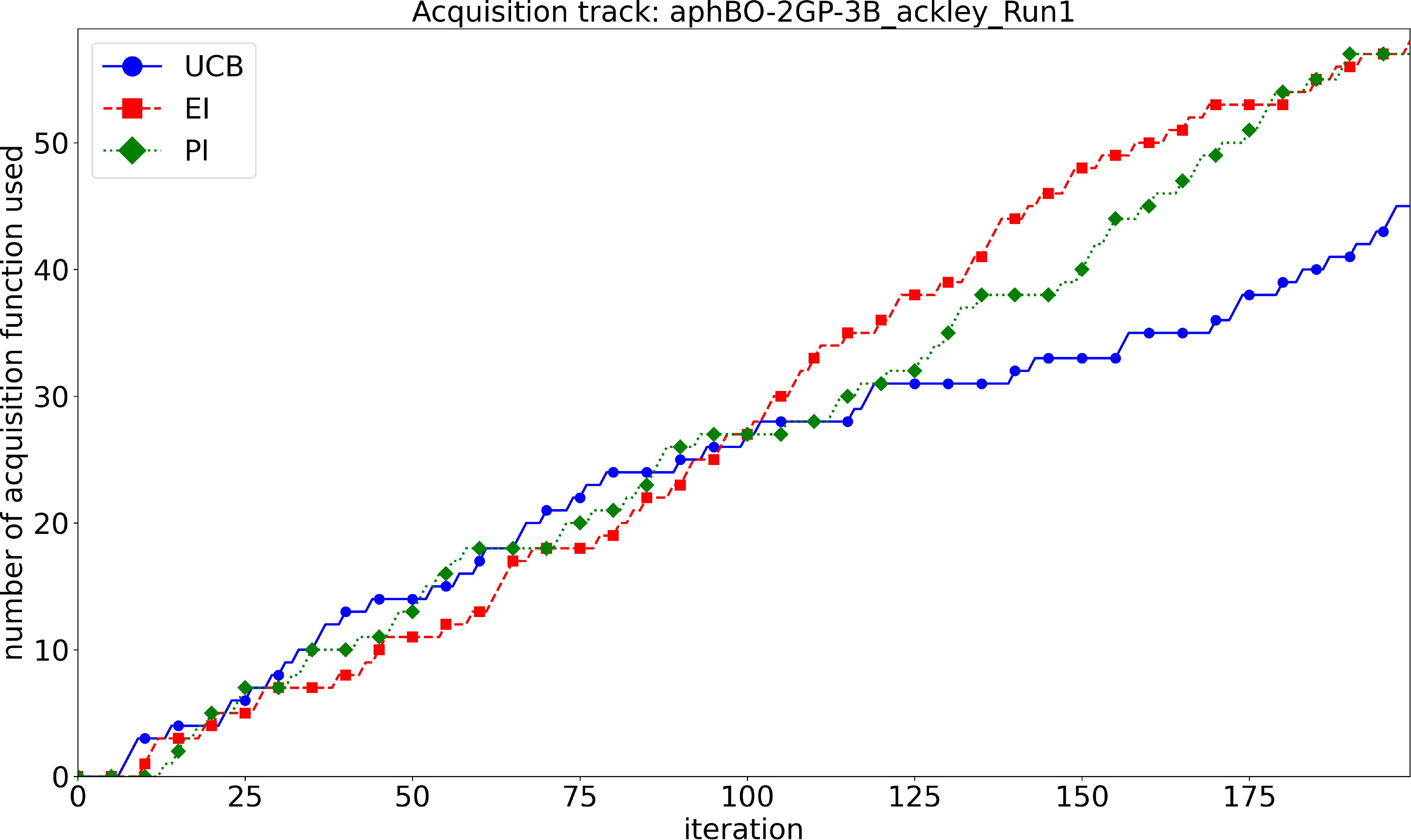}}
\hfill
\subcaptionbox{ackley: scheduler dashboard.
}
  [0.30\linewidth]{\includegraphics[width=0.30\textwidth, keepaspectratio]{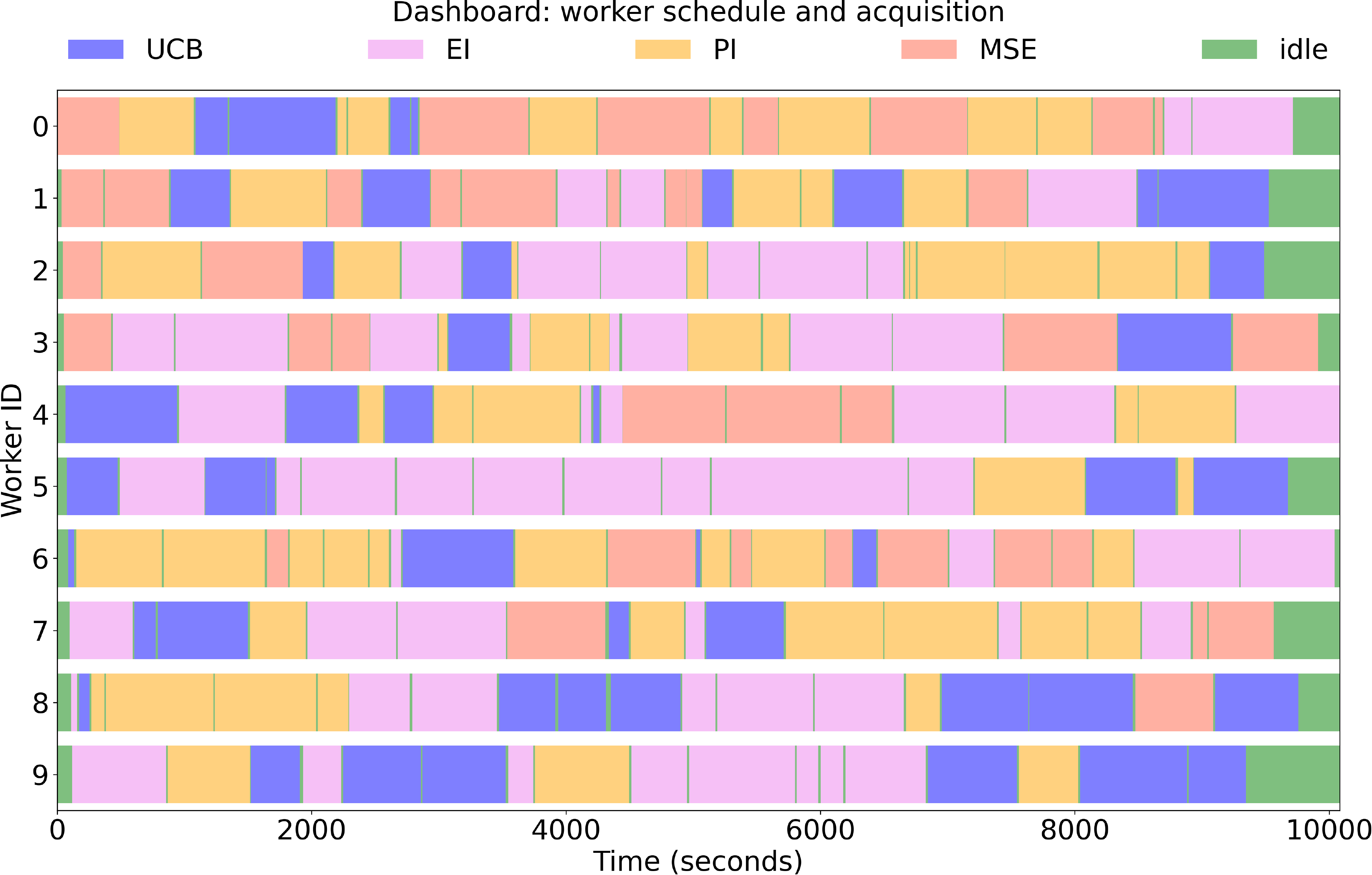}}
% \caption{Acquisition portfolio and scheduler.}
% \label{fig:portfolio_ackley}
\end{figure}
% \medskip
\begin{figure}[!htbp]\ContinuedFloat
\centering
\subcaptionbox{hart6: acquisition portfolio.
}
  [0.30\linewidth]{\includegraphics[width=0.30\textwidth, keepaspectratio]{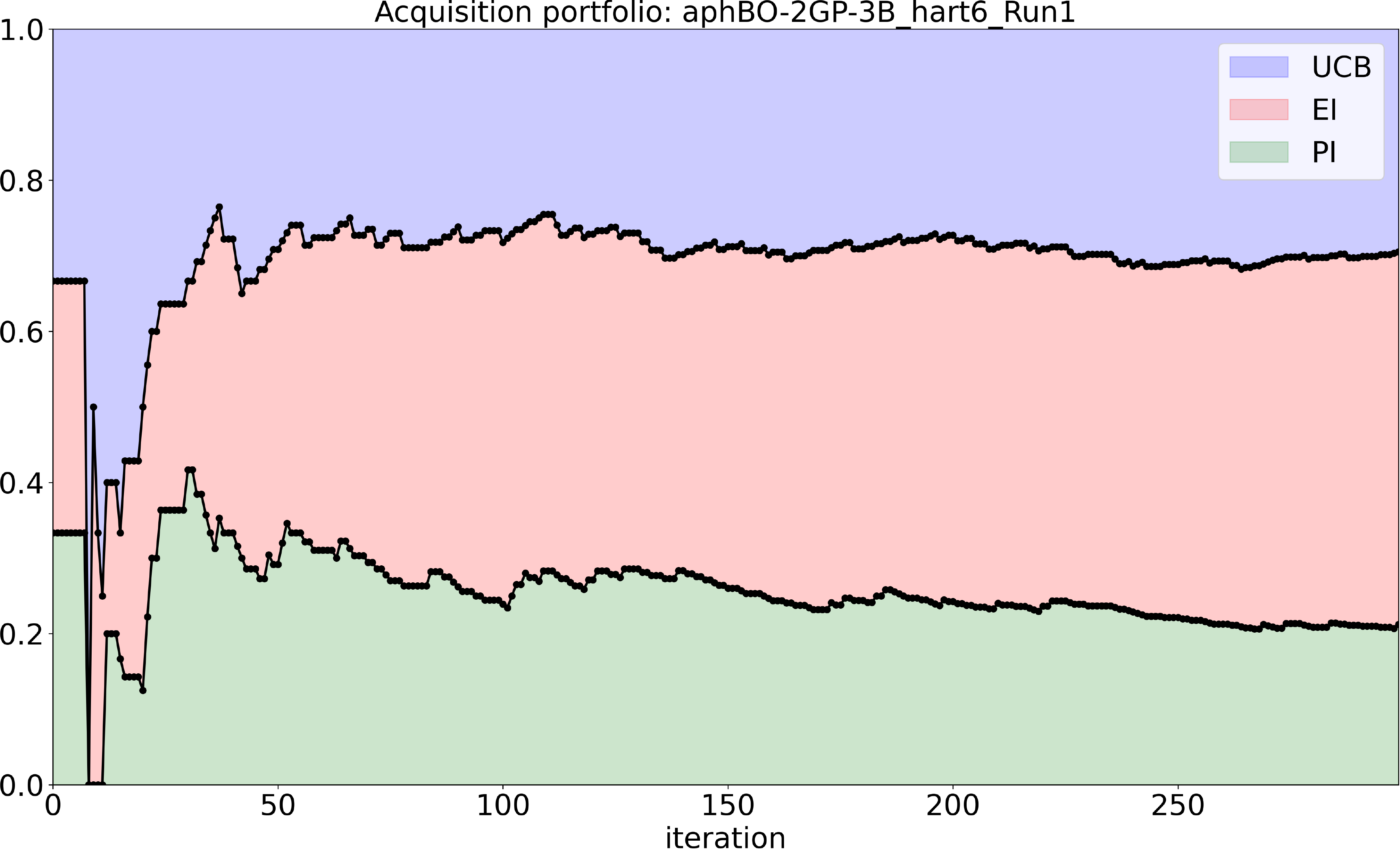}}
\hfill
\subcaptionbox{hart6: number of acquisition functions.
}
  [0.30\linewidth]{\includegraphics[width=0.30\textwidth, keepaspectratio]{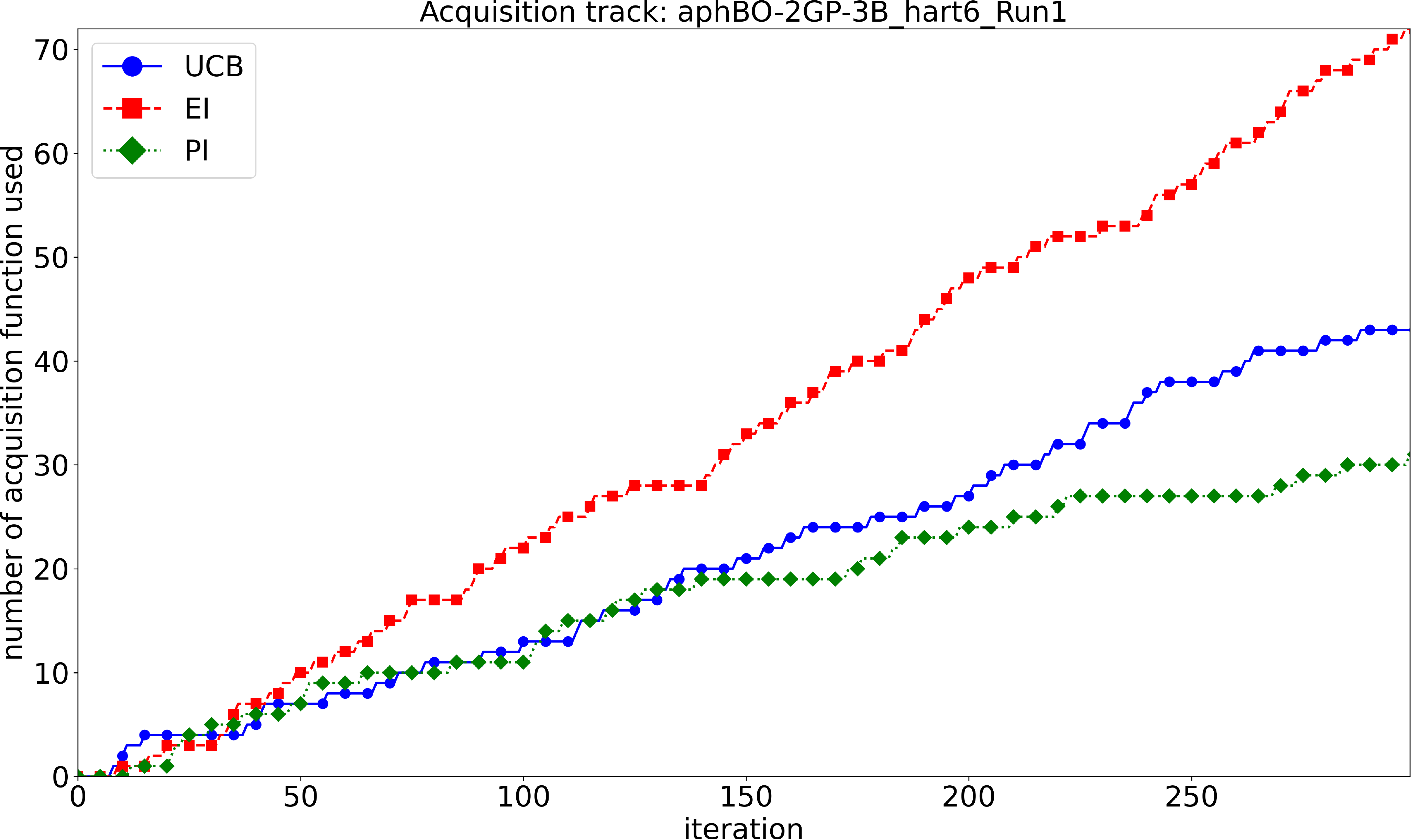}}
\hfill
\subcaptionbox{hart6: scheduler dashboard.
}
  [0.30\linewidth]{\includegraphics[width=0.30\textwidth, keepaspectratio]{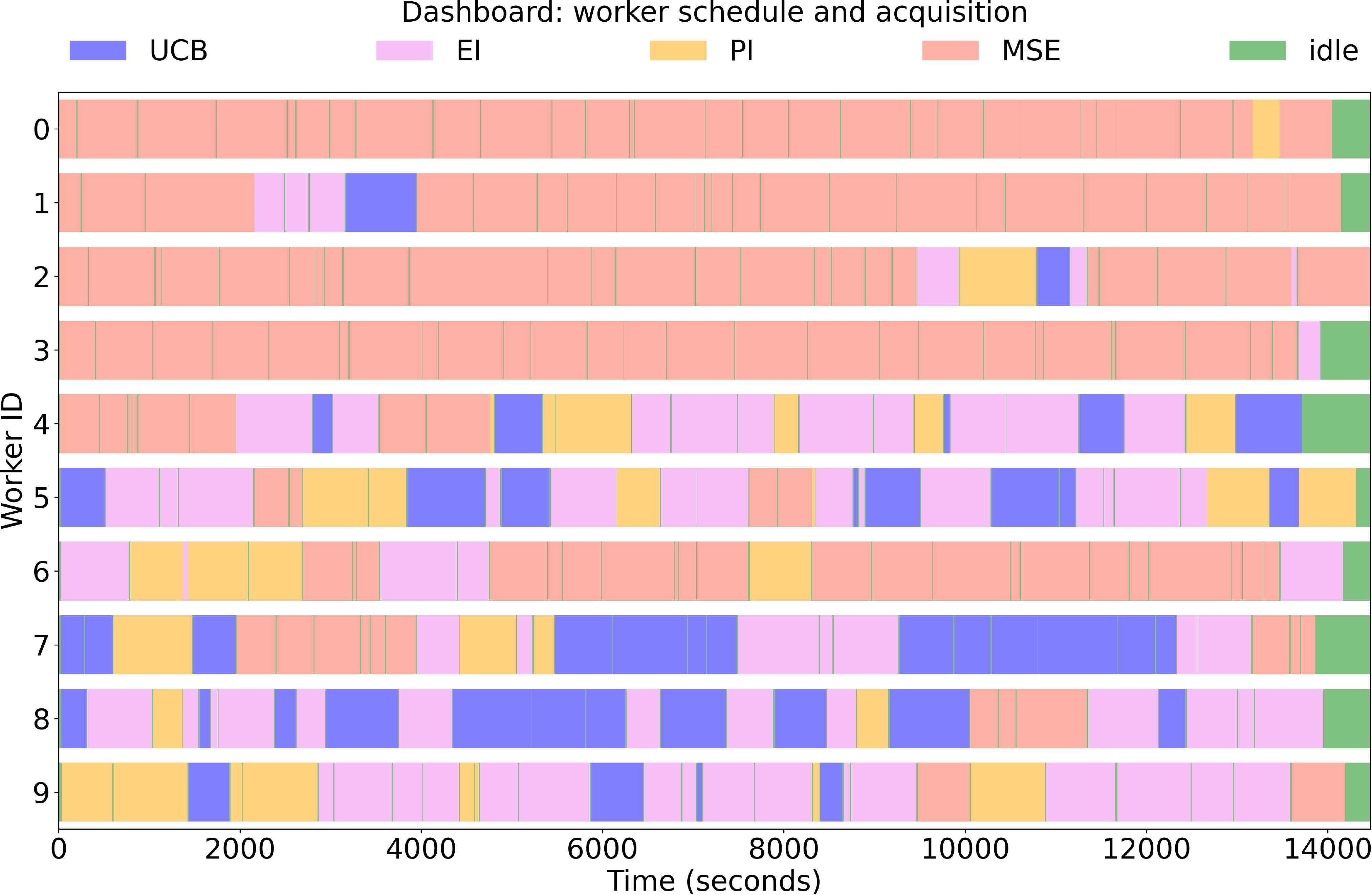}}
% \caption{Acquisition portfolio and scheduler.}
% \label{fig:portfolio_hart6}
% \end{figure}
\medskip
% \begin{figure}[!htbp]
\centering
\subcaptionbox{michal: acquisition portfolio.
}
  [0.30\linewidth]{\includegraphics[width=0.30\textwidth, keepaspectratio]{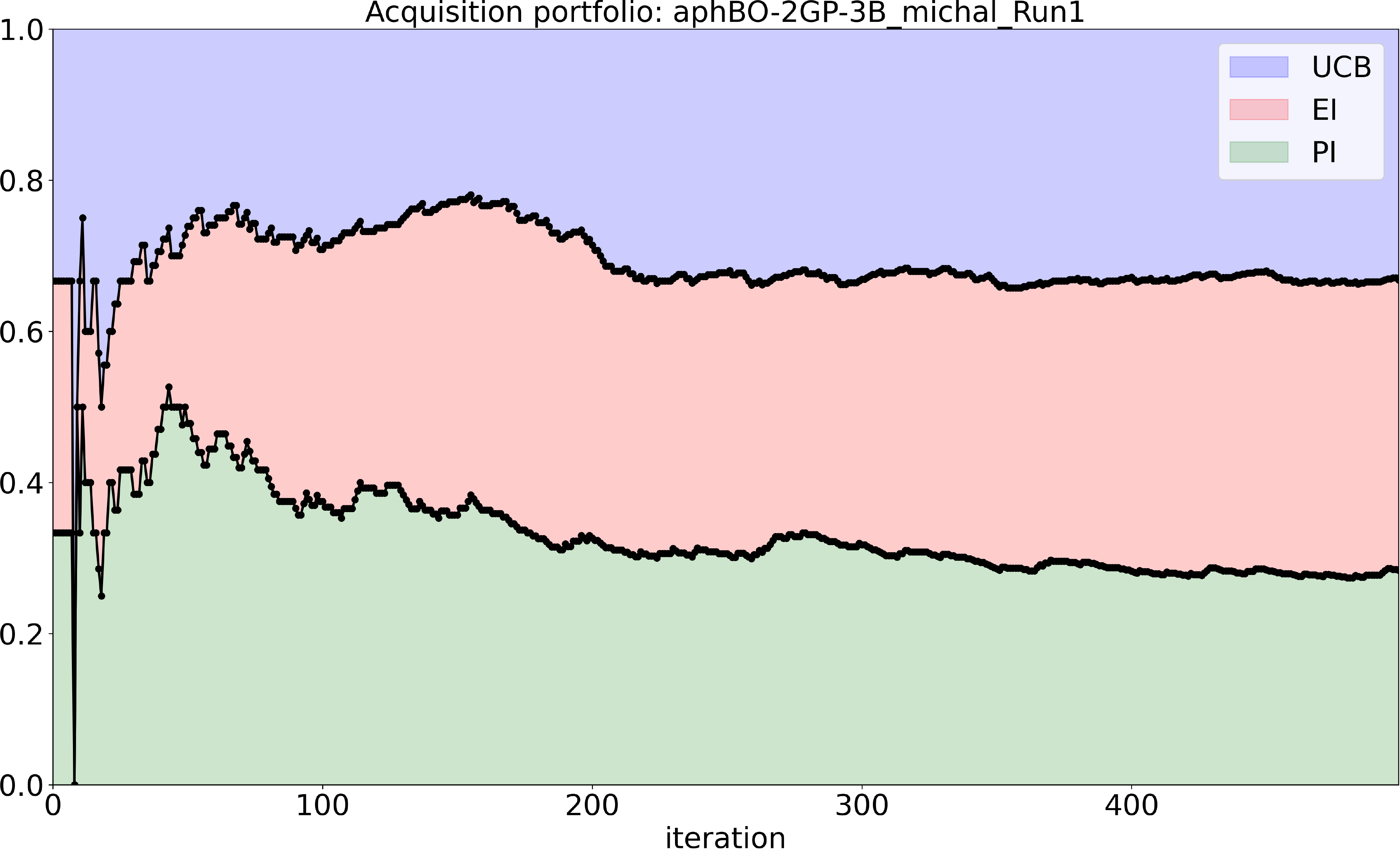}}
\hfill
\subcaptionbox{michal: number of acquisition functions.
}
  [0.30\linewidth]{\includegraphics[width=0.30\textwidth, keepaspectratio]{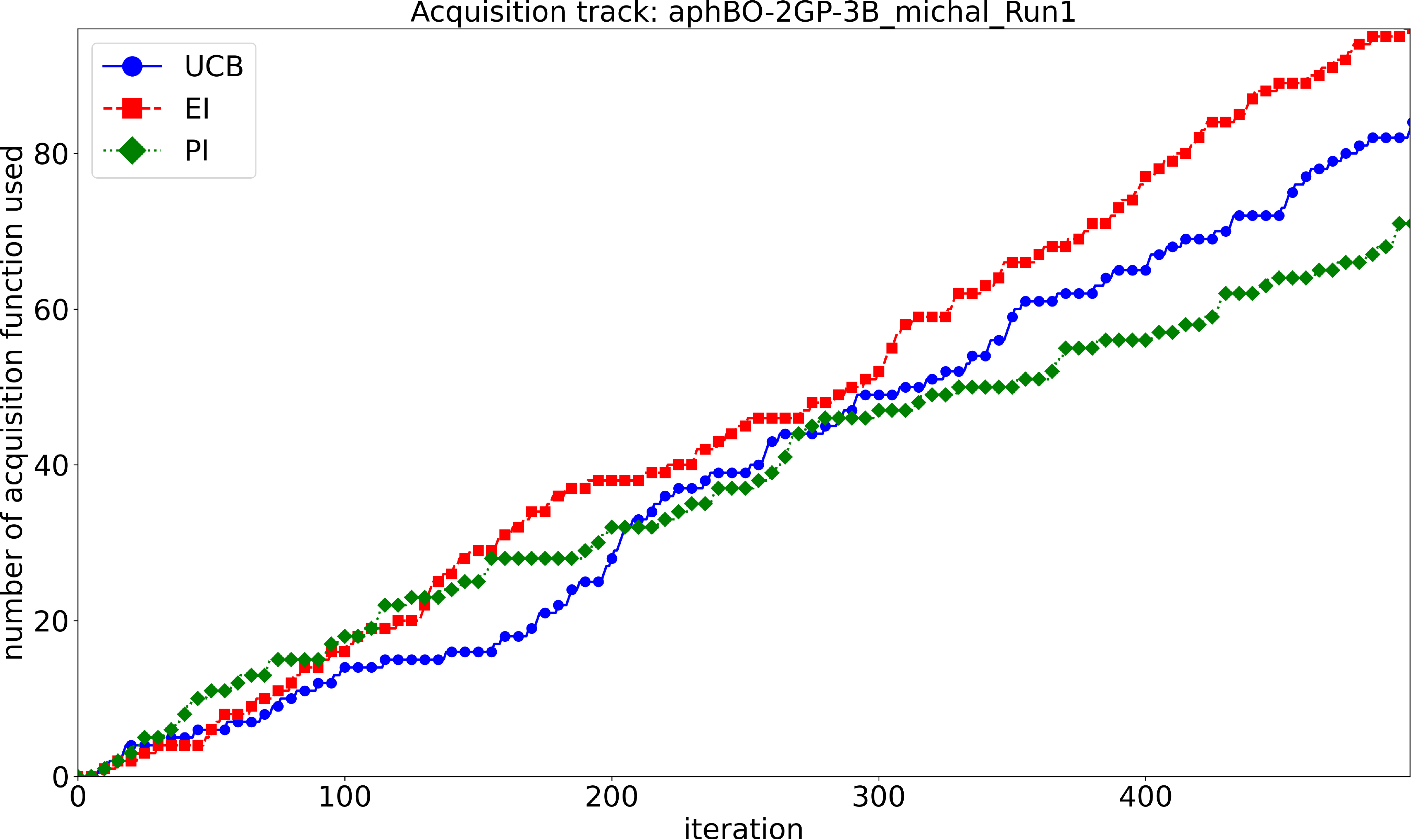}}
\hfill
\subcaptionbox{michal: scheduler dashboard.
}
  [0.30\linewidth]{\includegraphics[width=0.30\textwidth, keepaspectratio]{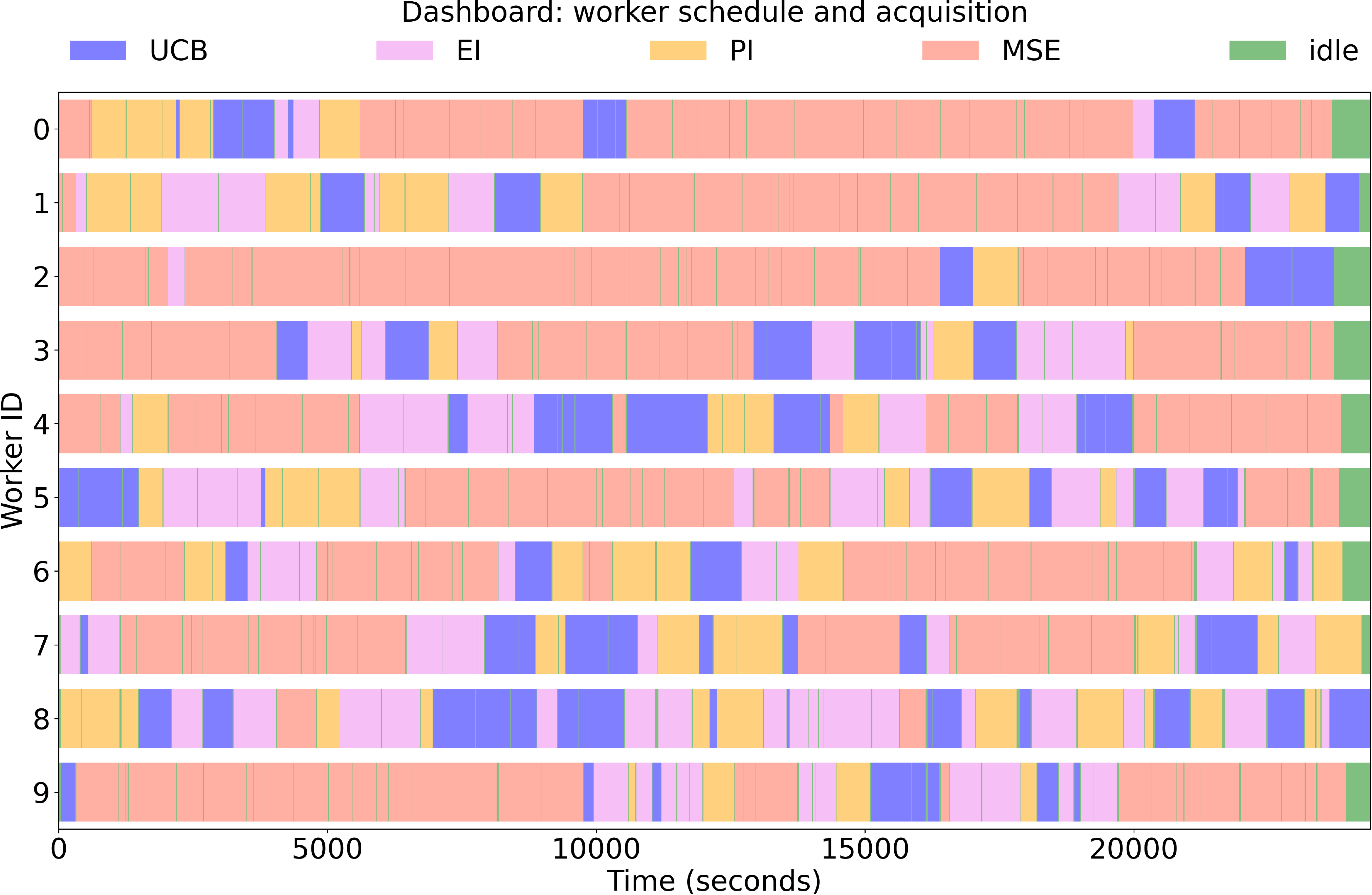}}
% \caption{Acquisition portfolio and scheduler.}
% \label{fig:portfolio_michal}
% \end{figure}
\medskip
% \begin{figure}[!htbp]
\centering
\subcaptionbox{perm0db: acquisition portfolio.
}
  [0.30\linewidth]{\includegraphics[width=0.30\textwidth, keepaspectratio]{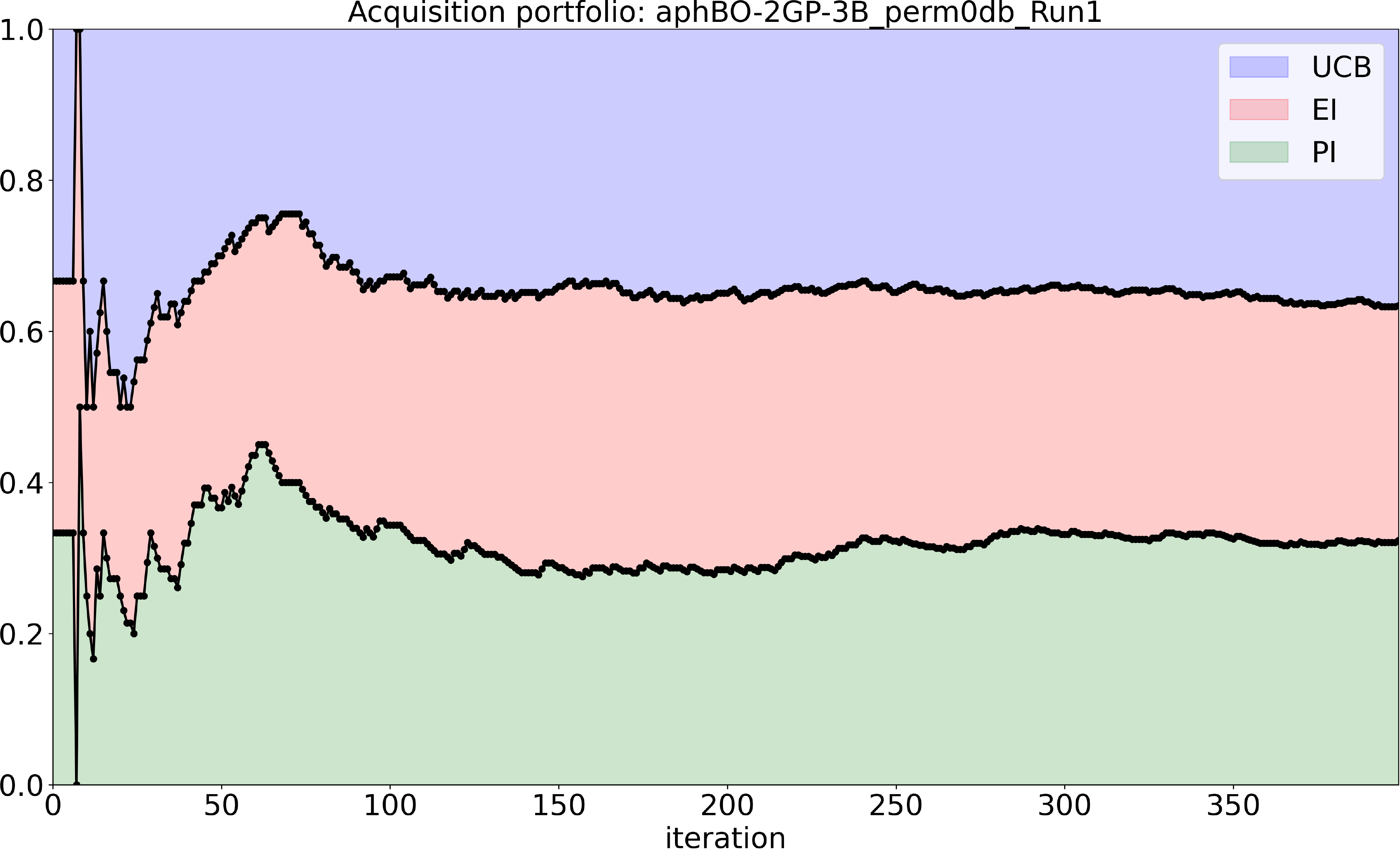}}
\hfill
\subcaptionbox{perm0db: number of acquisition functions.
}
  [0.30\linewidth]{\includegraphics[width=0.30\textwidth, keepaspectratio]{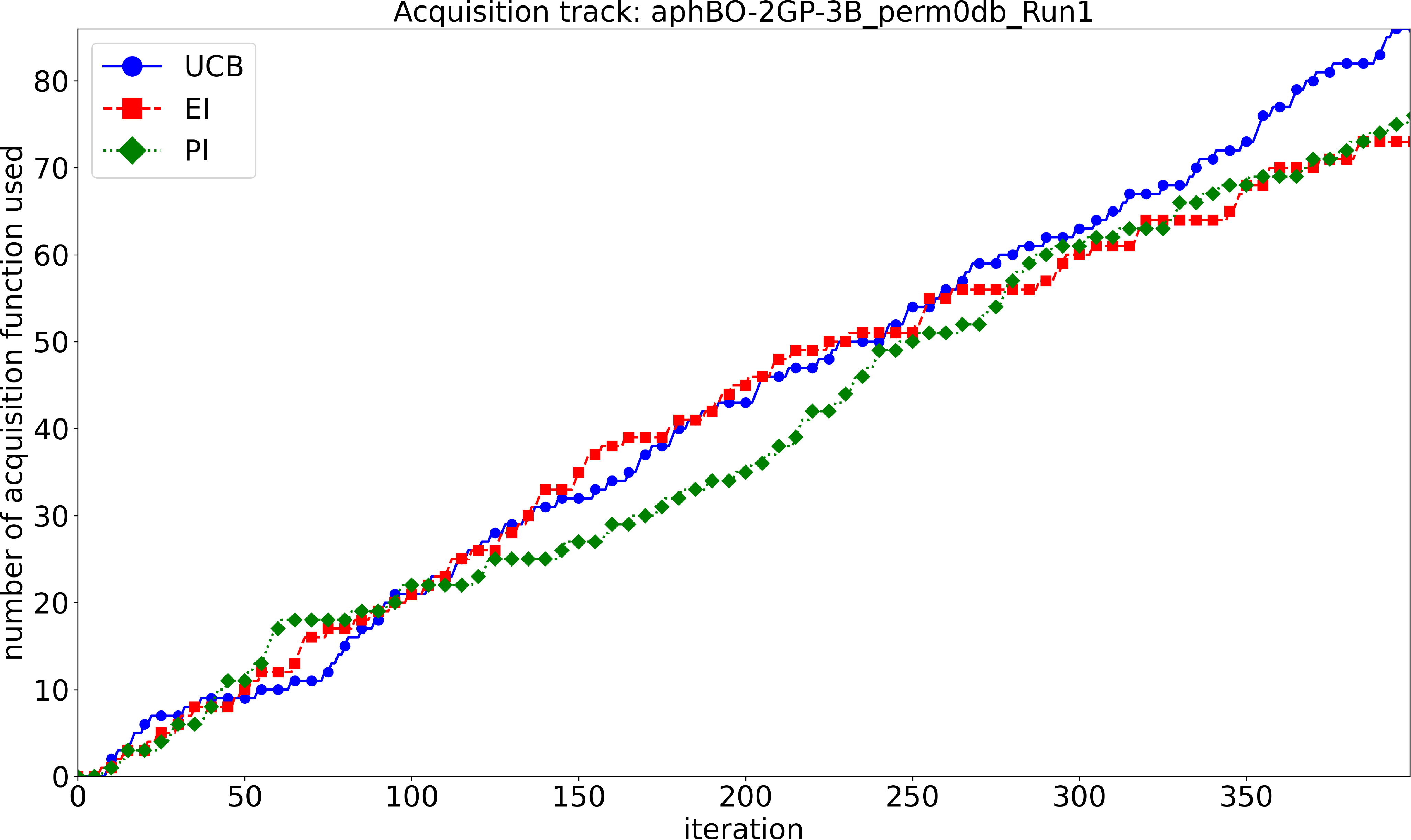}}
\hfill
\subcaptionbox{perm0db: scheduler dashboard.
}
  [0.30\linewidth]{\includegraphics[width=0.30\textwidth, keepaspectratio]{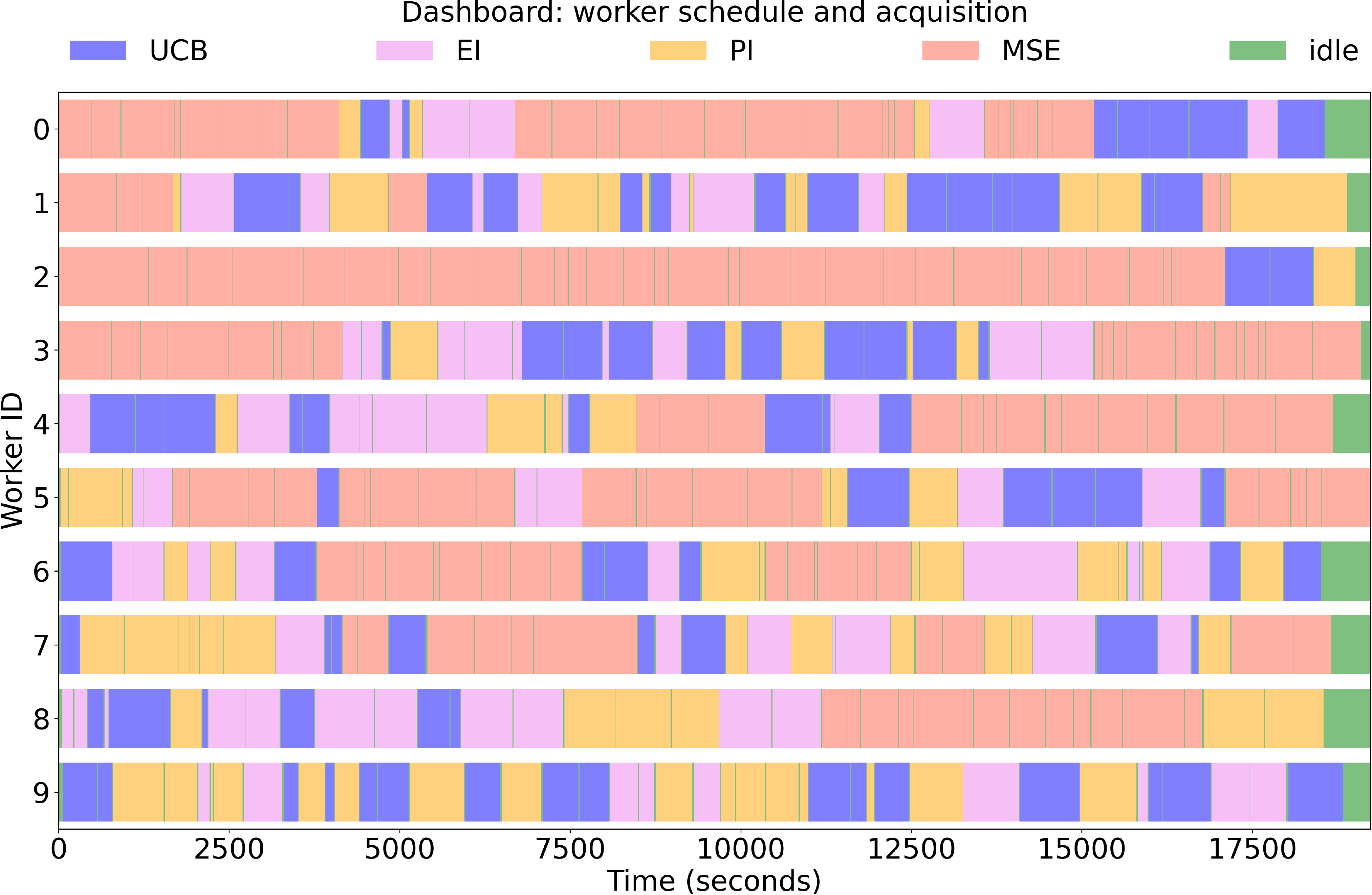}}
% \caption{Acquisition portfolio and scheduler.}
% \label{fig:portfolio_perm0db}
% \end{figure}
\medskip
% \begin{figure}[!htbp]
\centering
\subcaptionbox{rosen: acquisition portfolio.
}
  [0.30\linewidth]{\includegraphics[width=0.30\textwidth, keepaspectratio]{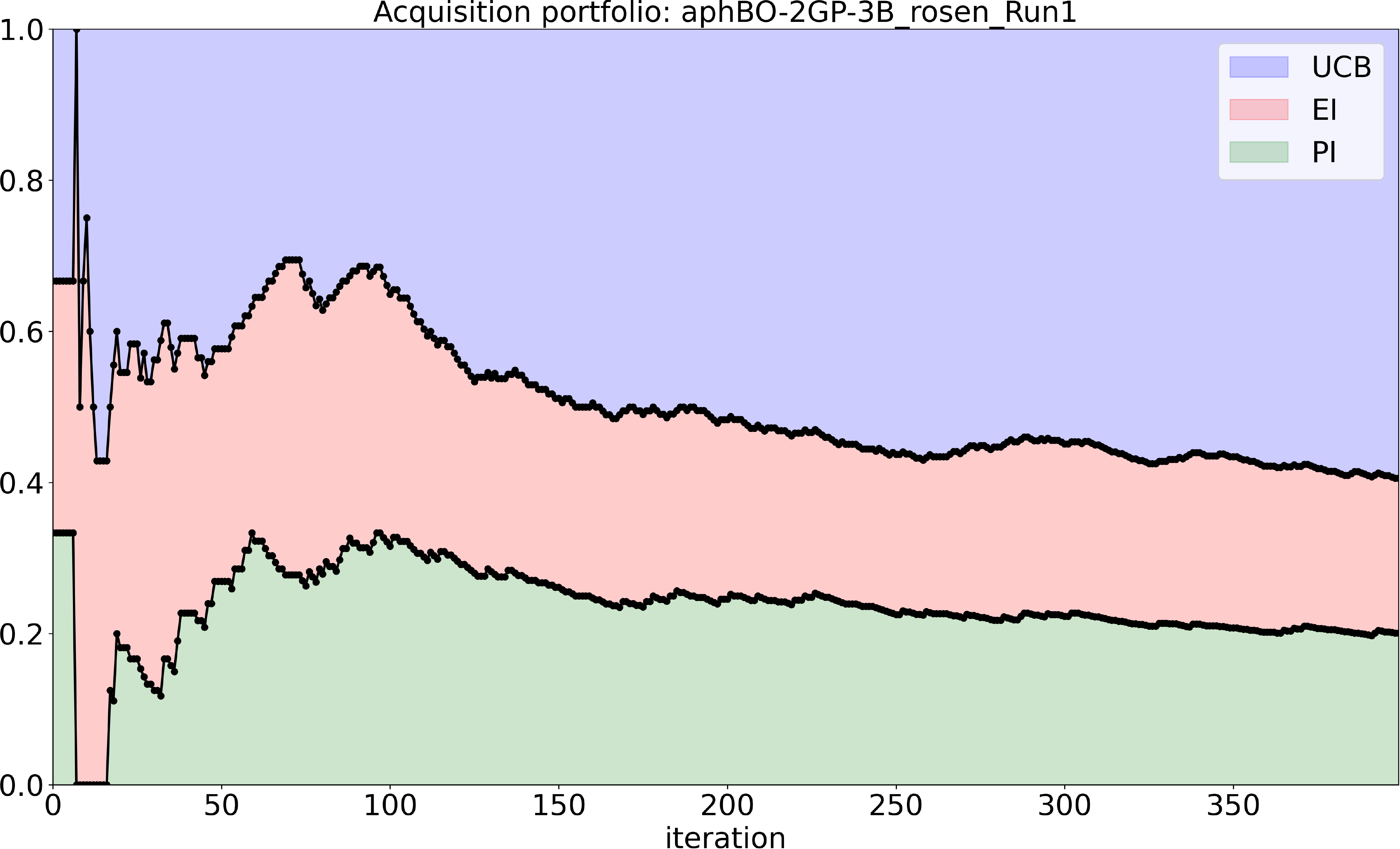}}
\hfill
\subcaptionbox{rosen: number of acquisition functions.
}
  [0.30\linewidth]{\includegraphics[width=0.30\textwidth, keepaspectratio]{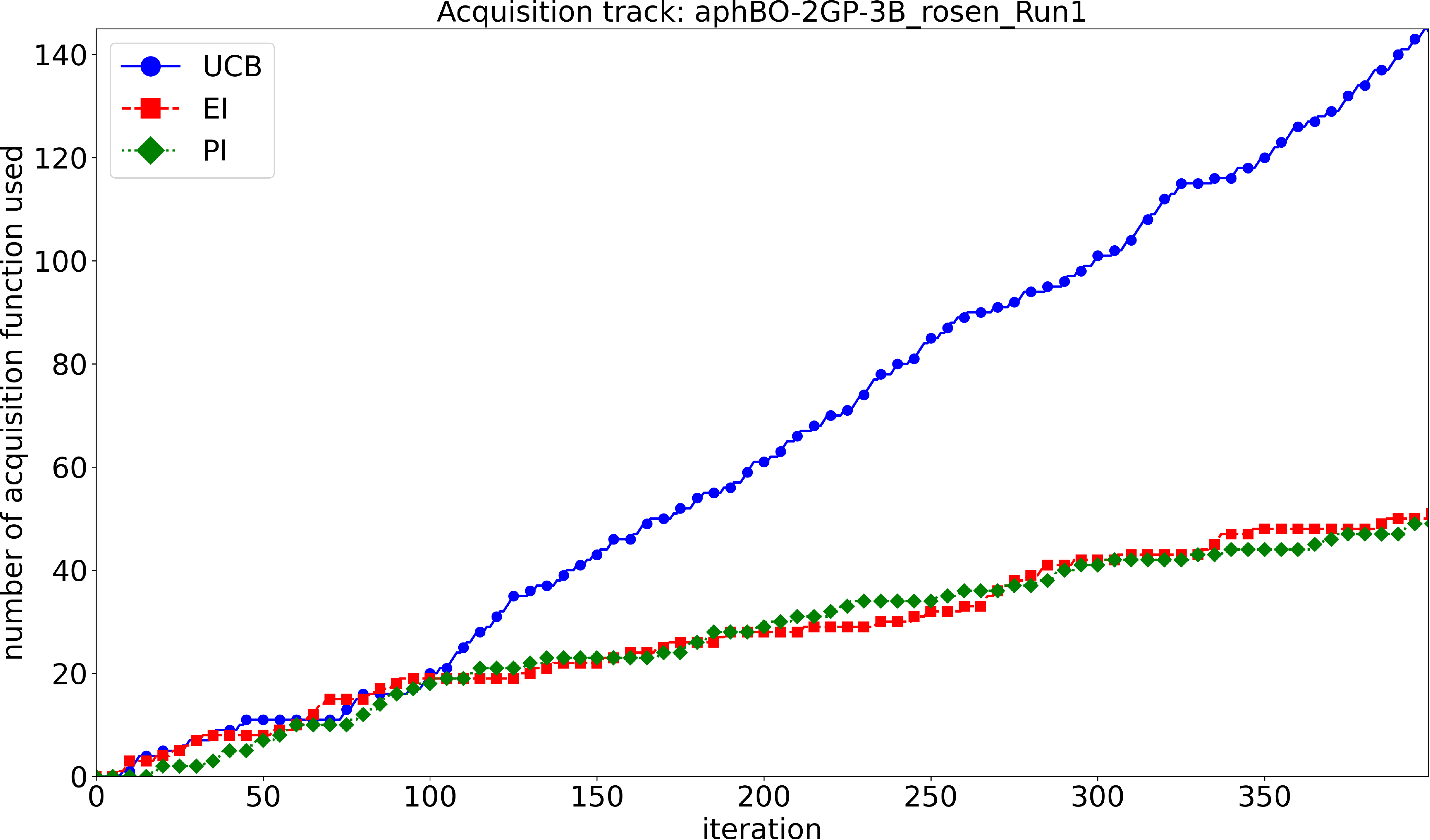}}
\hfill
\subcaptionbox{rosen: scheduler dashboard.
}
  [0.30\linewidth]{\includegraphics[width=0.30\textwidth, keepaspectratio]{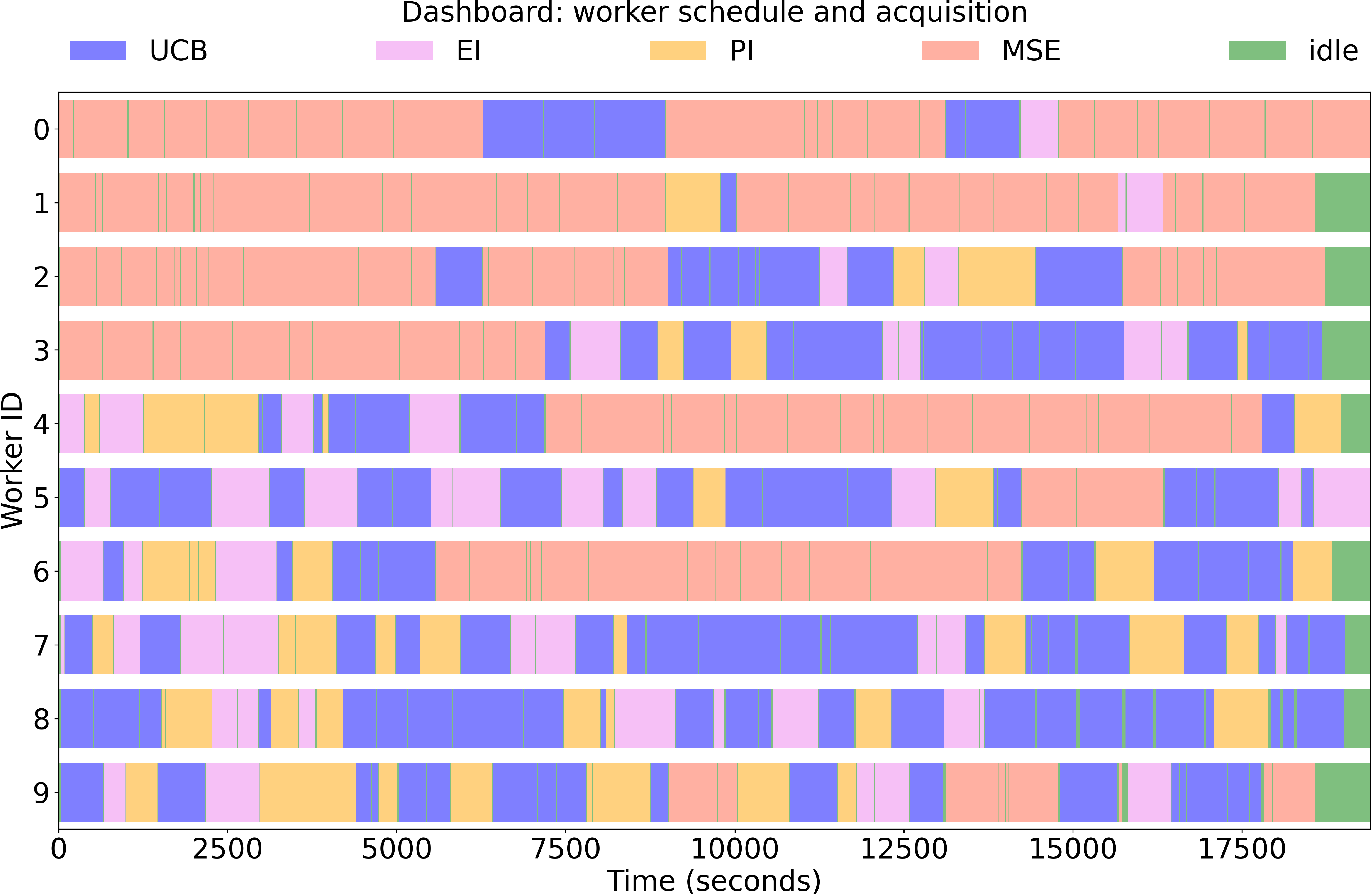}}
% \caption{Acquisition portfolio and scheduler.}
% \label{fig:portfolio_rosen}
% \end{figure}
\medskip
% \begin{figure}[!htbp]
\centering
\subcaptionbox{dixonpr: acquisition portfolio.
}
  [0.30\linewidth]{\includegraphics[width=0.30\textwidth, keepaspectratio]{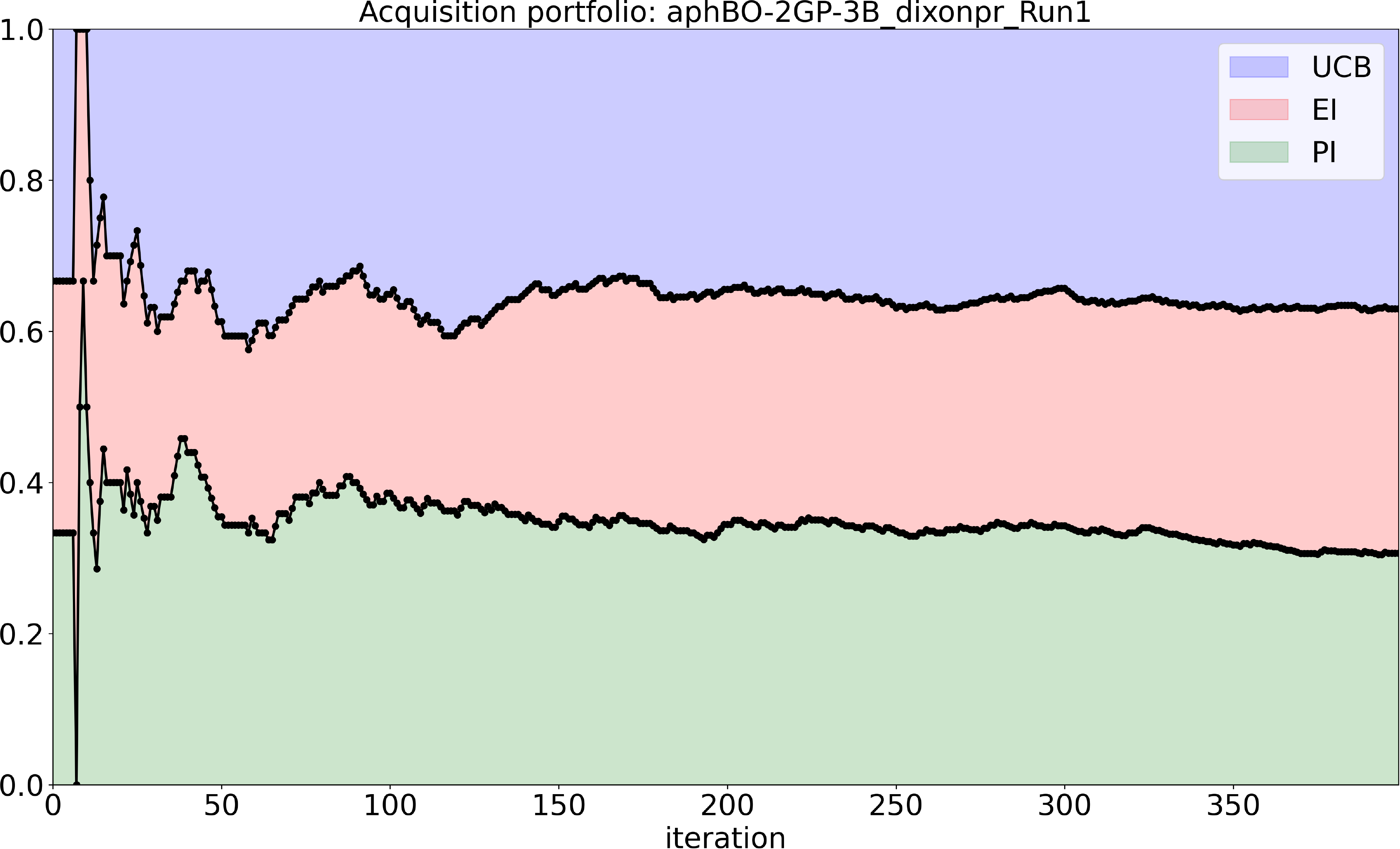}}
\hfill
\subcaptionbox{dixonpr: number of acquisition functions.
}
  [0.30\linewidth]{\includegraphics[width=0.30\textwidth, keepaspectratio]{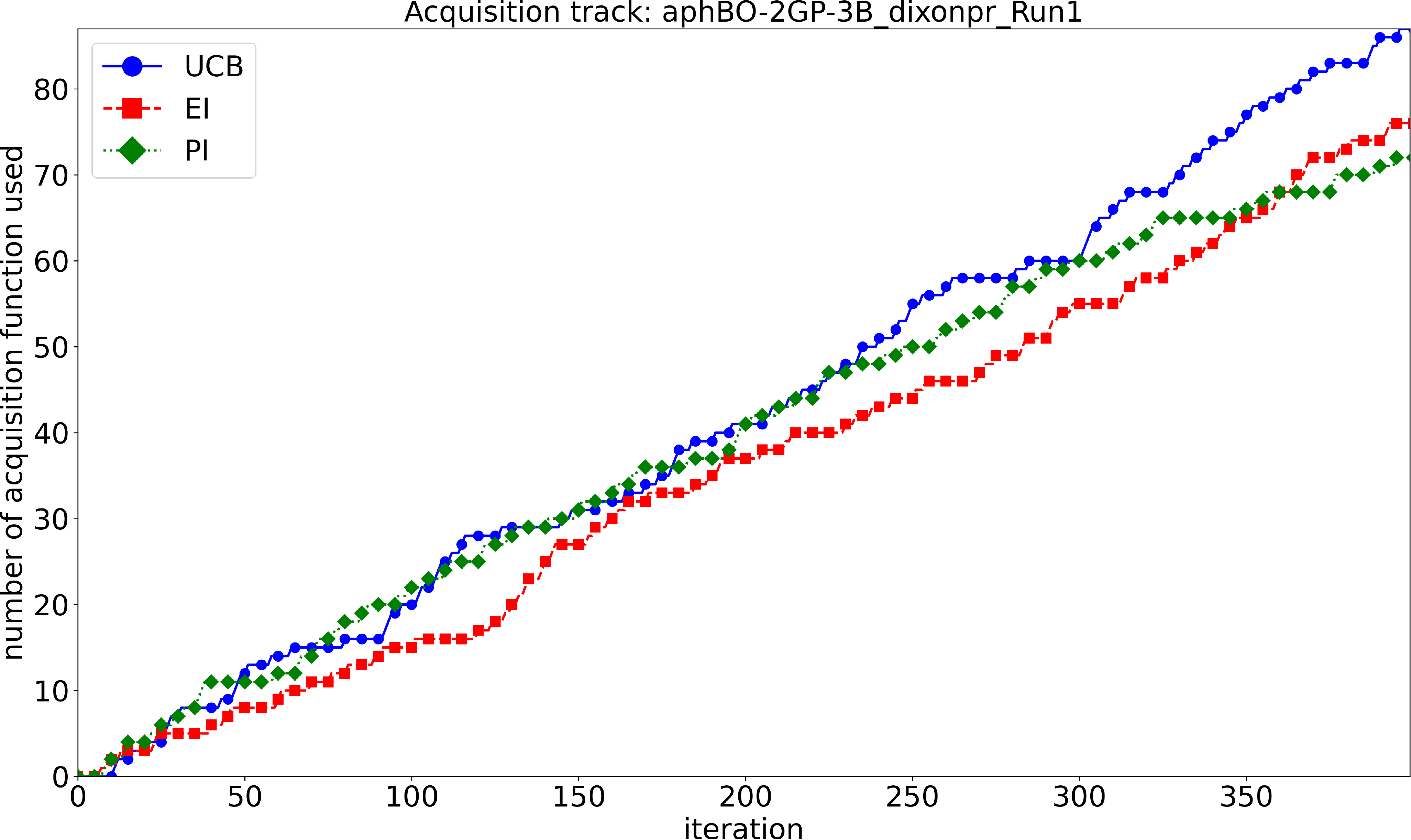}}
\hfill
\subcaptionbox{dixonpr: scheduler dashboard.
}
  [0.30\linewidth]{\includegraphics[width=0.30\textwidth, keepaspectratio]{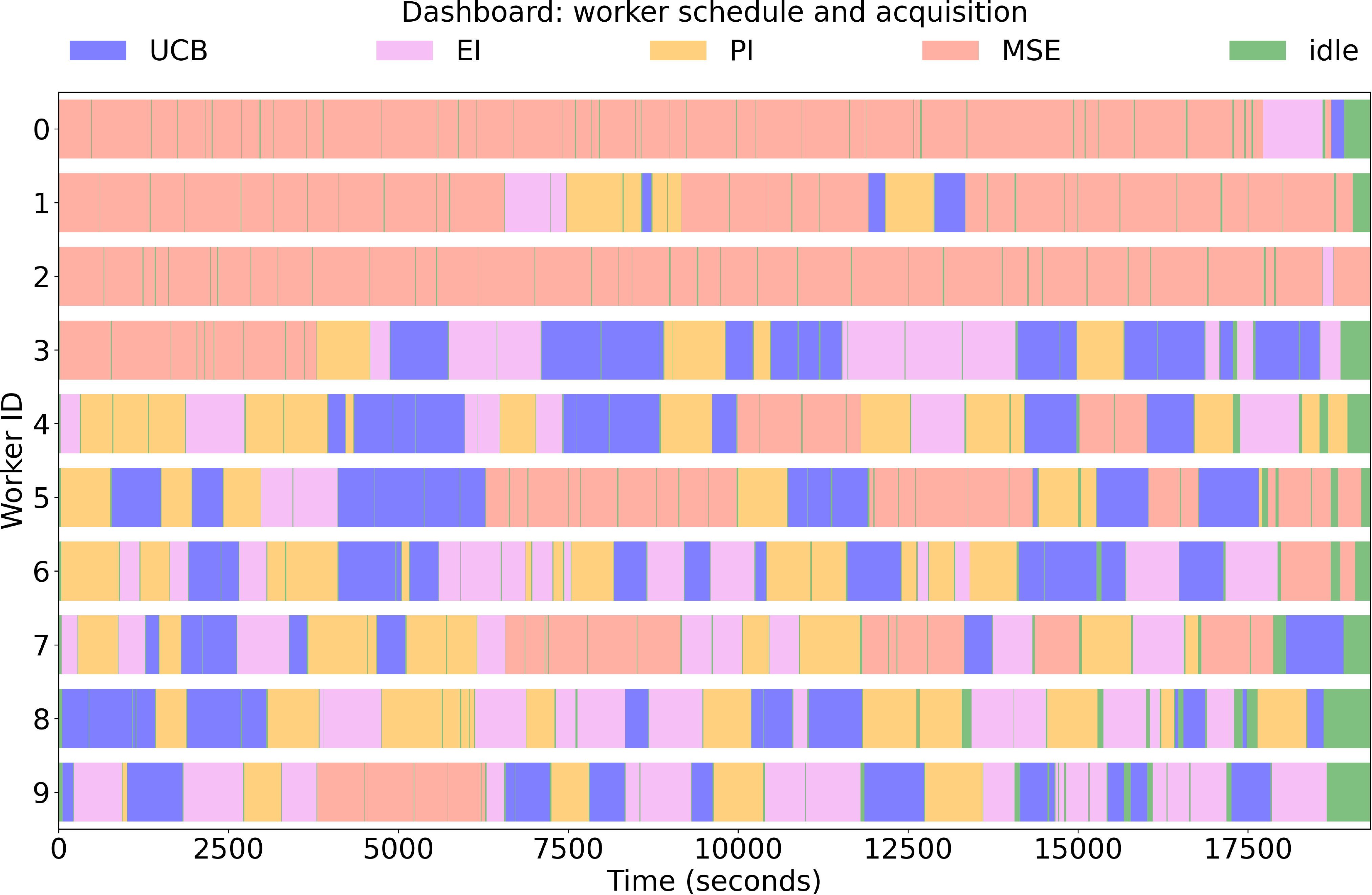}}
% \caption{Acquisition portfolio and scheduler.}
% \label{fig:portfolio_dixonpr}
\end{figure}

\begin{figure}[!htbp]\ContinuedFloat
\centering
\subcaptionbox{trid: acquisition portfolio.
}
  [0.30\linewidth]{\includegraphics[width=0.30\textwidth, keepaspectratio]{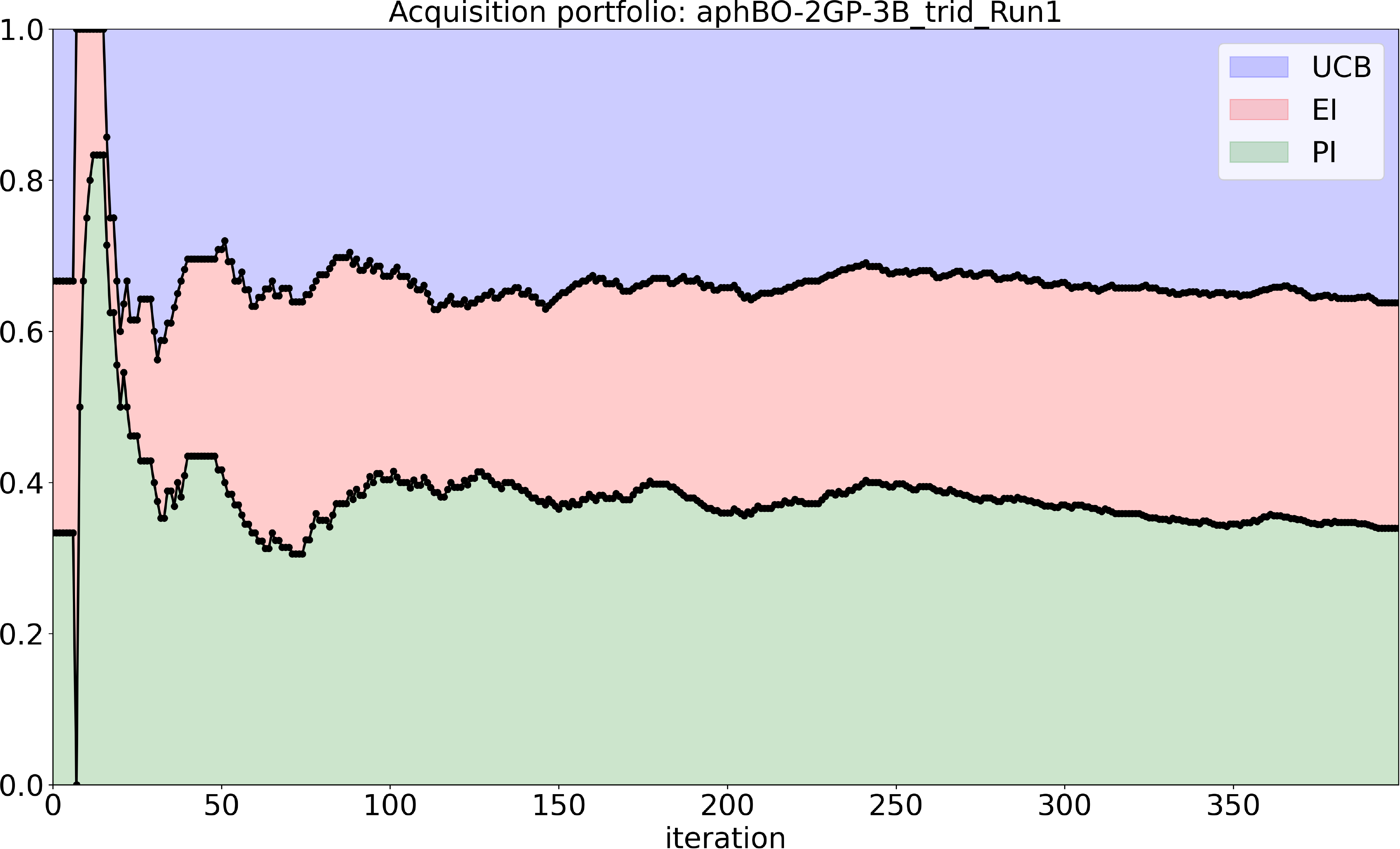}}
\hfill
\subcaptionbox{trid: number of acquisition functions.
}
  [0.30\linewidth]{\includegraphics[width=0.30\textwidth, keepaspectratio]{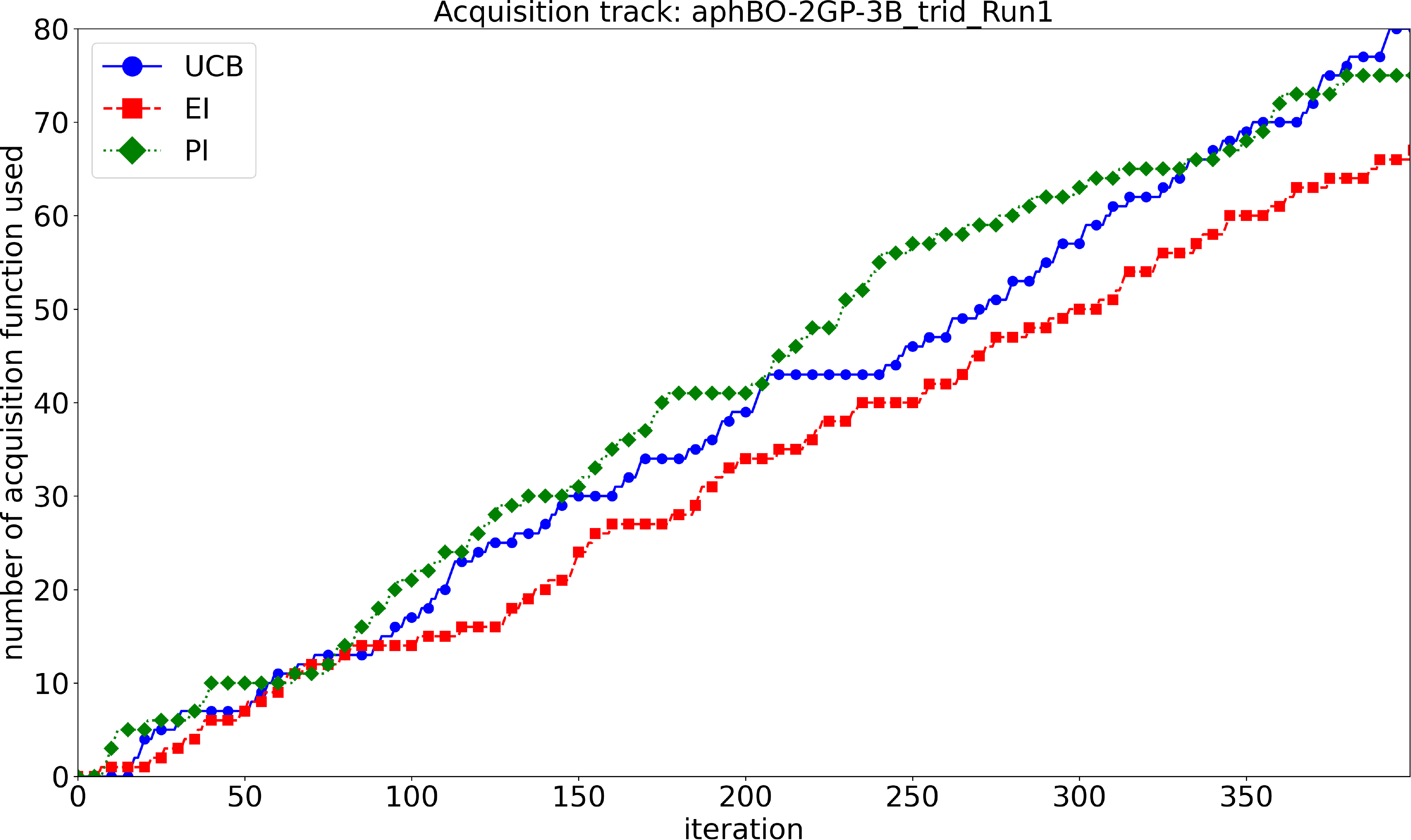}}
\hfill
\subcaptionbox{trid: scheduler dashboard.
}
  [0.30\linewidth]{\includegraphics[width=0.30\textwidth, keepaspectratio]{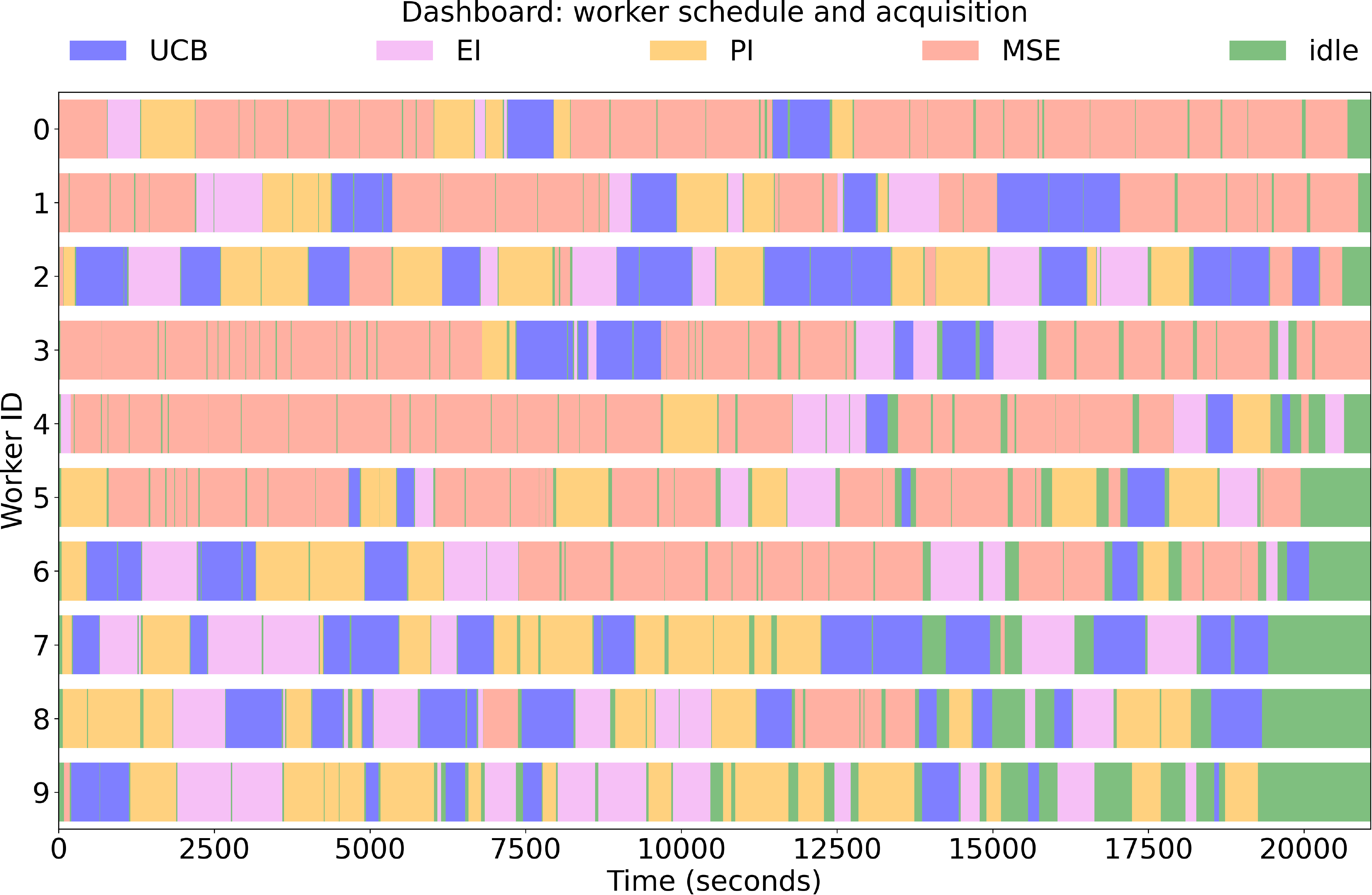}}
% \caption{Acquisition portfolio and scheduler.}
% \label{fig:portfolio_trid}
% \end{figure}
\medskip
% \begin{figure}[!htbp]
\centering
\subcaptionbox{sumsqu: acquisition portfolio.
}
  [0.30\linewidth]{\includegraphics[width=0.30\textwidth, keepaspectratio]{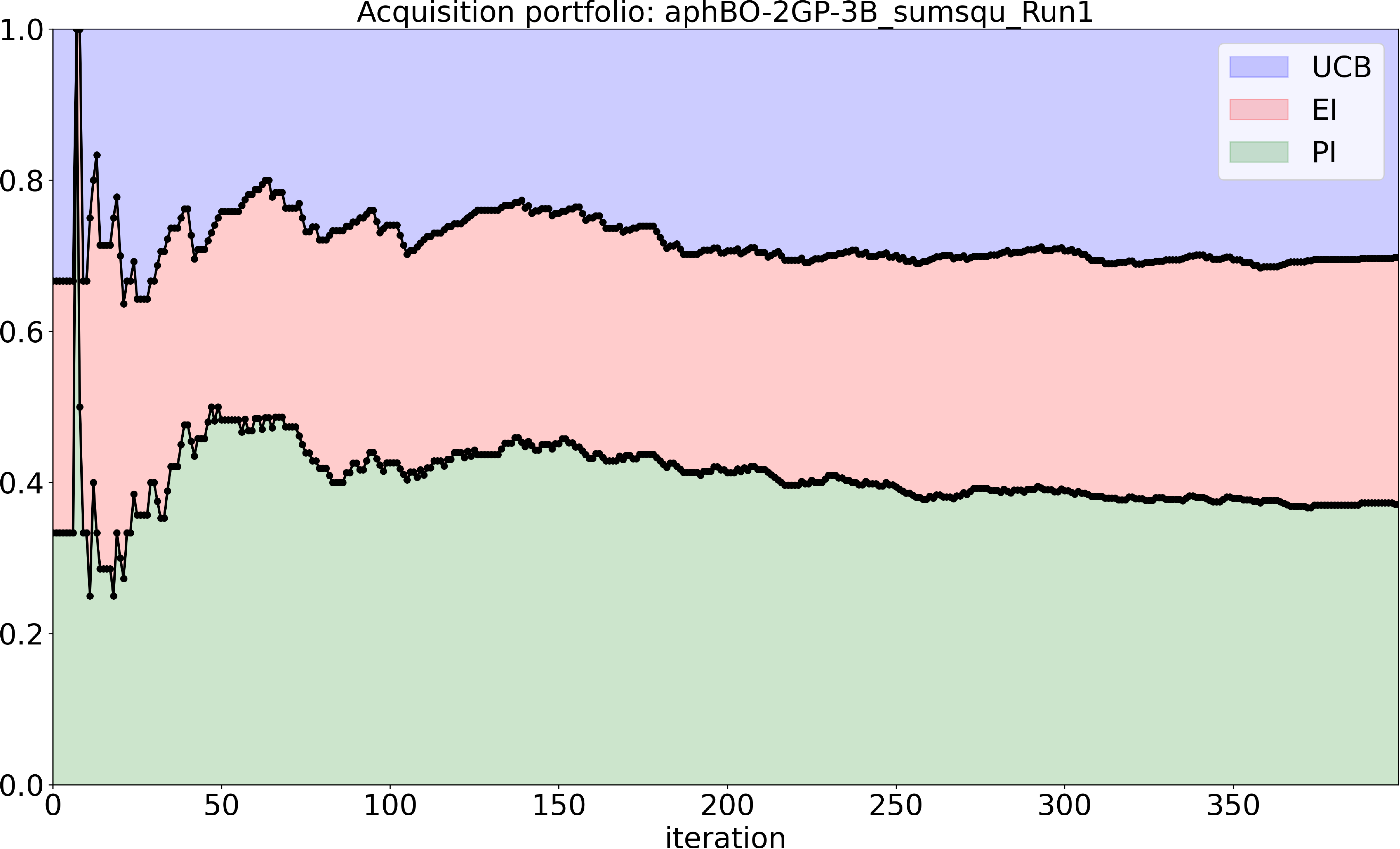}}
\hfill
\subcaptionbox{sumsqu: number of acquisition functions.
}
  [0.30\linewidth]{\includegraphics[width=0.30\textwidth, keepaspectratio]{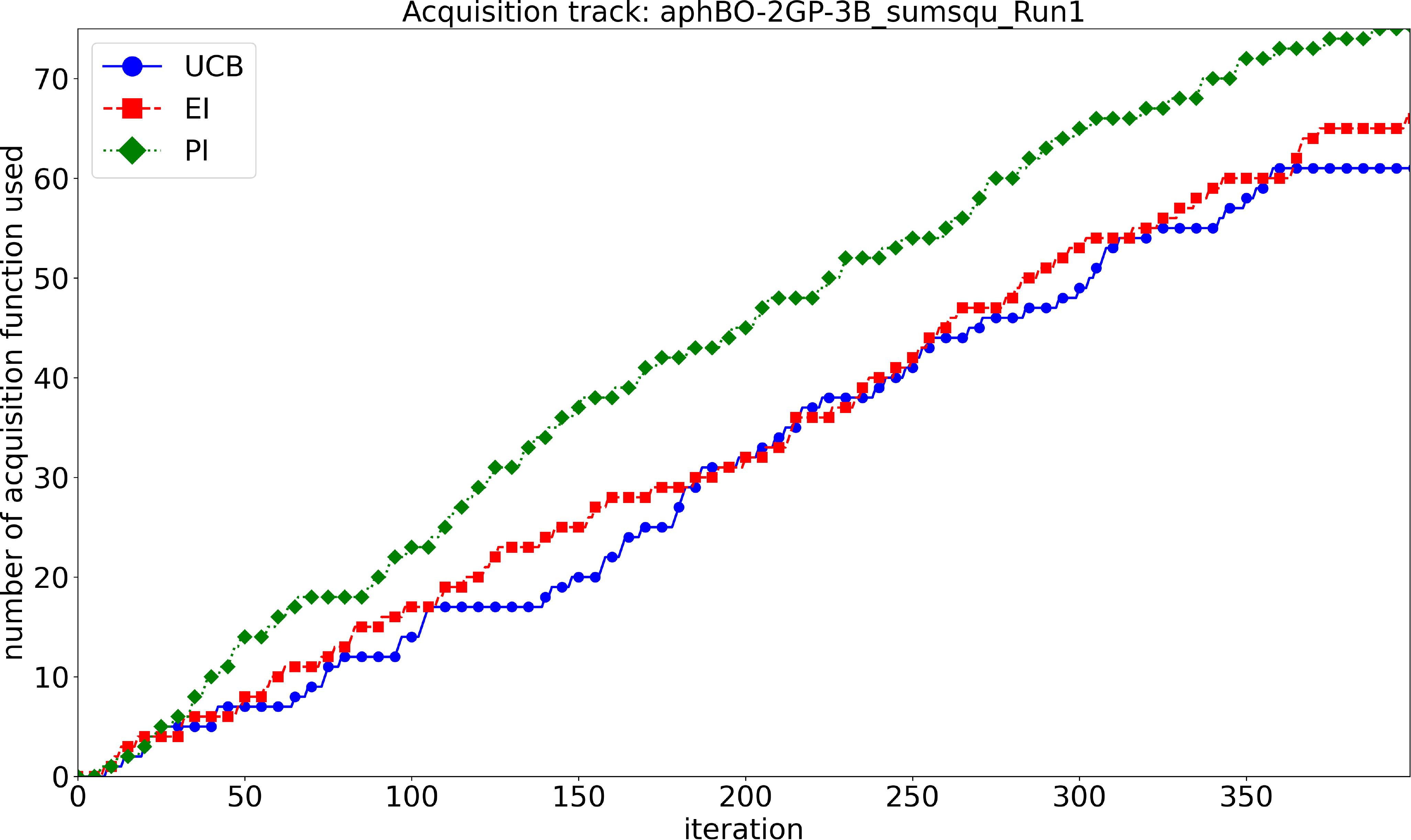}}
\hfill
\subcaptionbox{sumsqu: scheduler dashboard.
}
  [0.30\linewidth]{\includegraphics[width=0.30\textwidth, keepaspectratio]{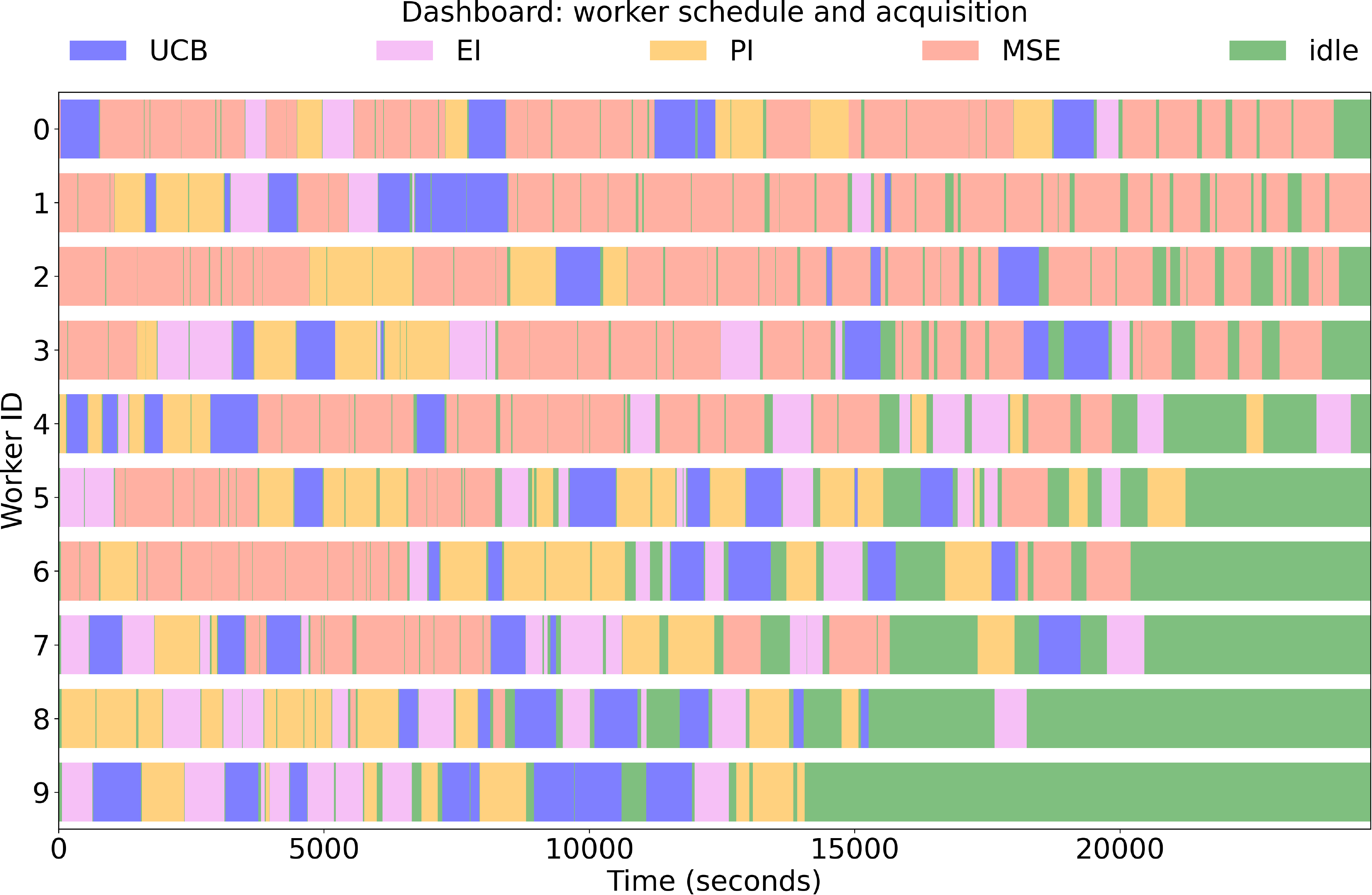}}
% \caption{Acquisition portfolio and scheduler.}
% \label{fig:portfolio_sumsqu}
% \end{figure}
\medskip
% \begin{figure}[!htbp]
\centering
\subcaptionbox{sumpow: acquisition portfolio.
}
  [0.30\linewidth]{\includegraphics[width=0.30\textwidth, keepaspectratio]{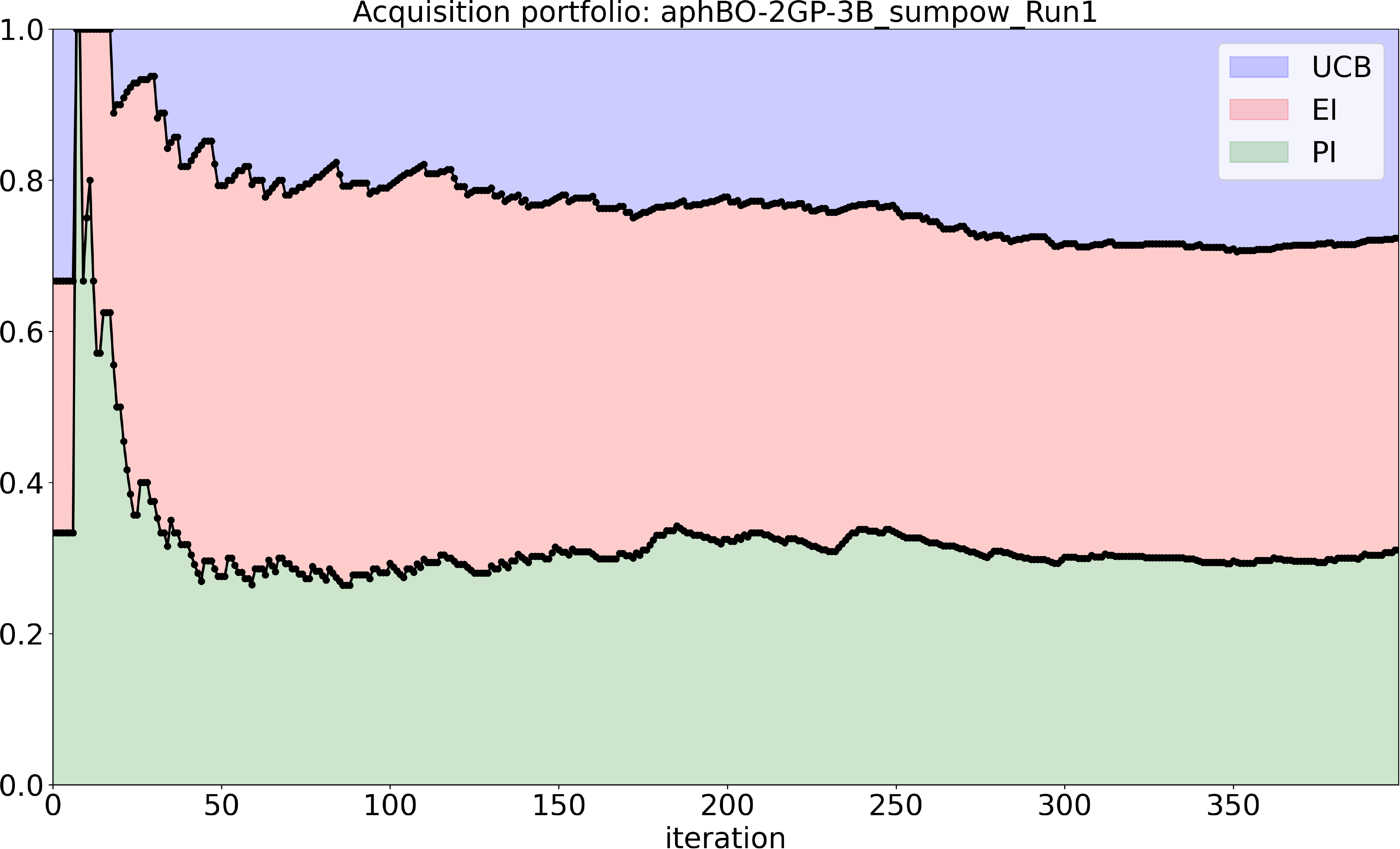}}
\hfill
\subcaptionbox{sumpow: number of acquisition functions.
}
  [0.30\linewidth]{\includegraphics[width=0.30\textwidth, keepaspectratio]{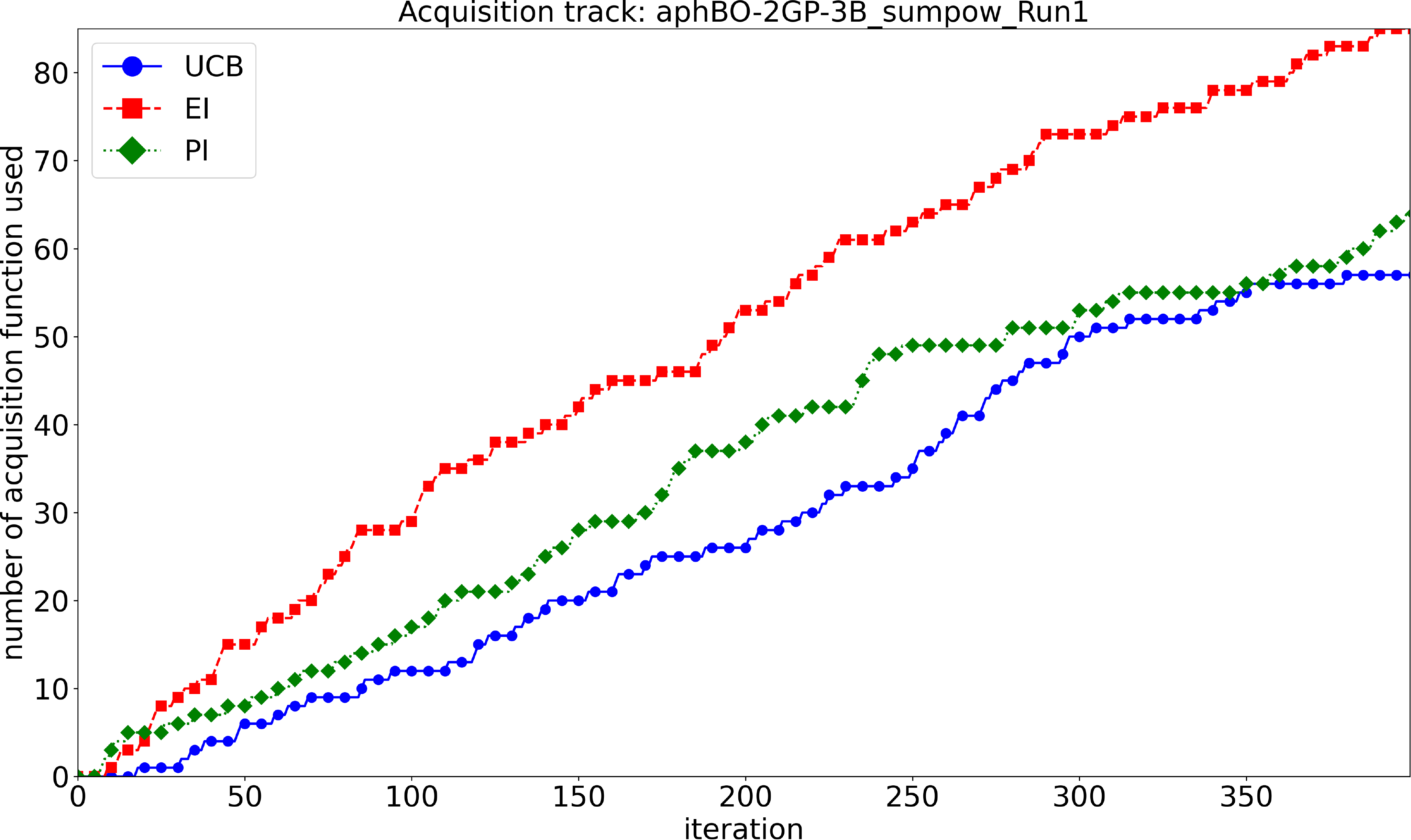}}
\hfill
\subcaptionbox{sumpow: scheduler dashboard.
}
  [0.30\linewidth]{\includegraphics[width=0.30\textwidth, keepaspectratio]{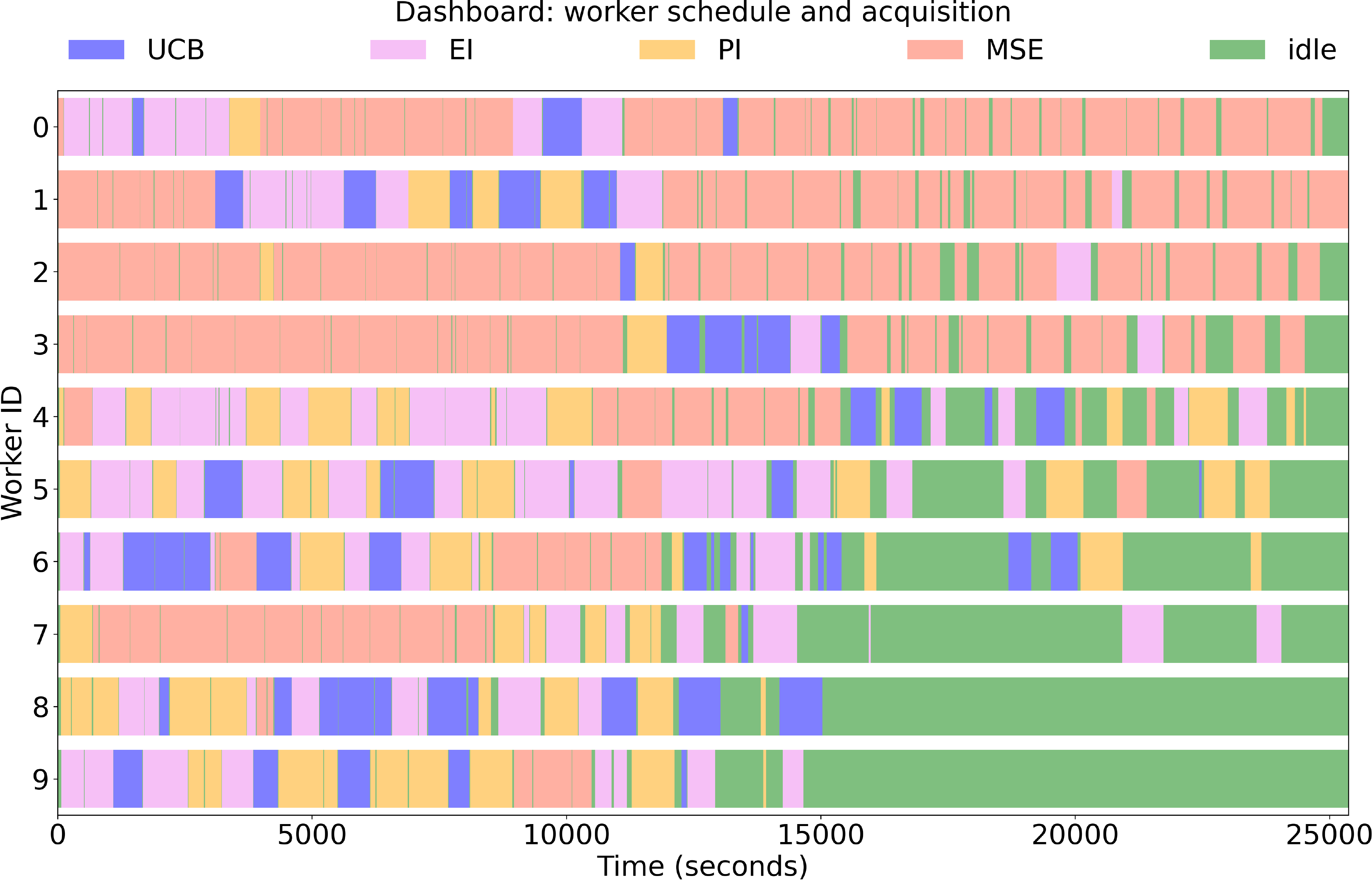}}
% \caption{Acquisition portfolio and scheduler.}
% \label{fig:portfolio_sumpow}
% \end{figure}
\medskip
% \begin{figure}[!htbp]
\centering
\subcaptionbox{spheref: acquisition portfolio.
}
  [0.30\linewidth]{\includegraphics[width=0.30\textwidth, keepaspectratio]{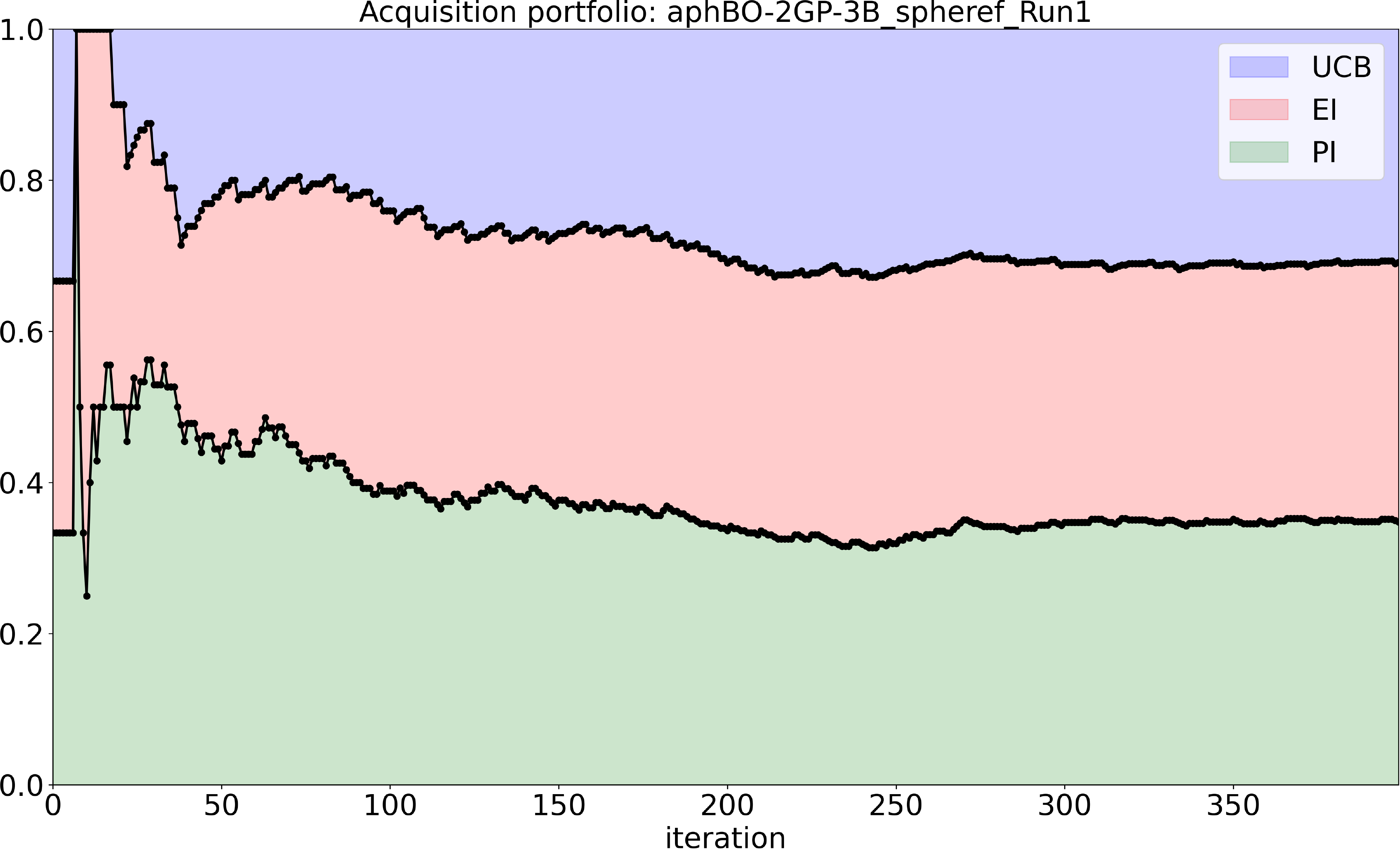}}
\hfill
\subcaptionbox{spheref: number of acquisition functions.
}
  [0.30\linewidth]{\includegraphics[width=0.30\textwidth, keepaspectratio]{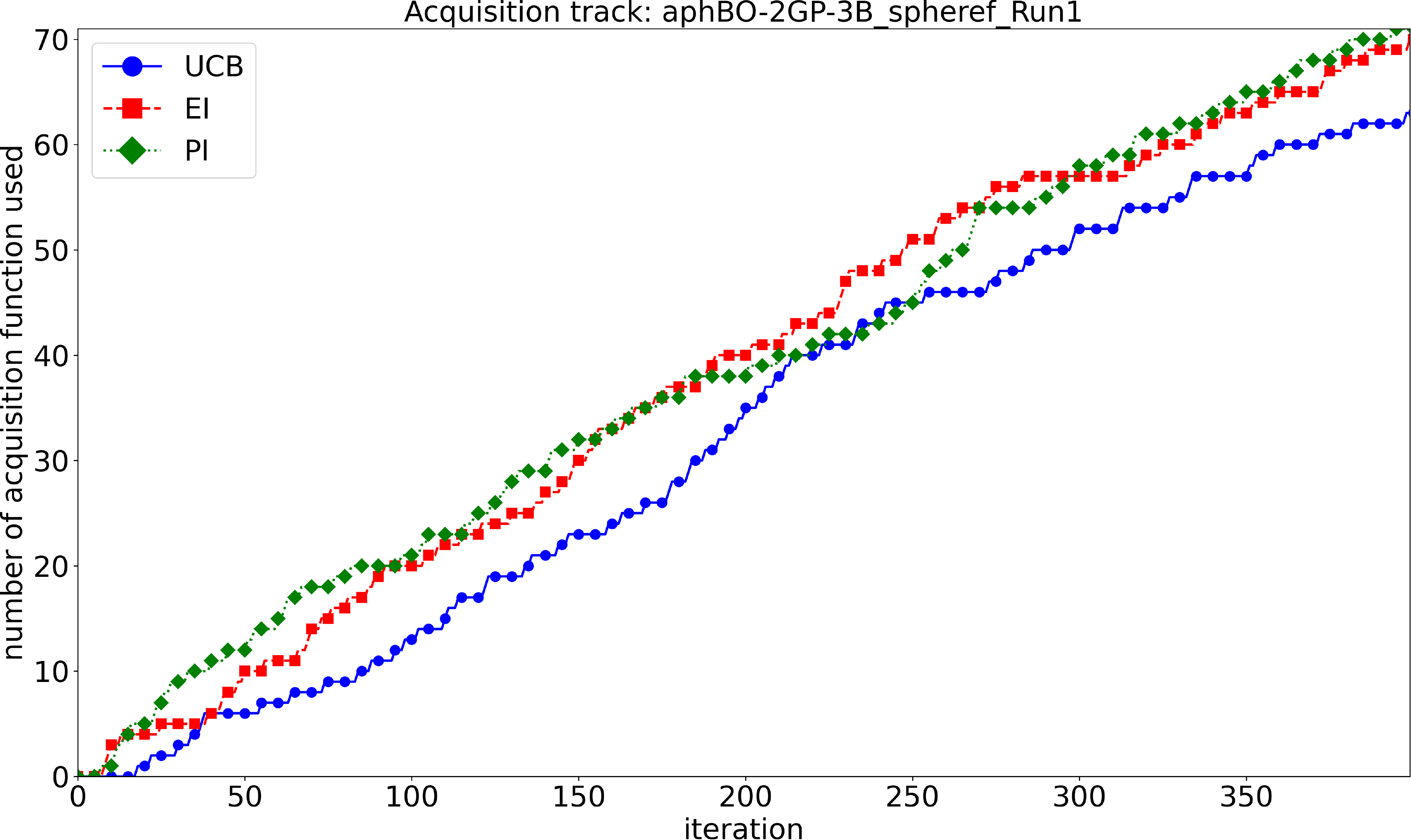}}
\hfill
\subcaptionbox{spheref: scheduler dashboard.
}
  [0.30\linewidth]{\includegraphics[width=0.30\textwidth, keepaspectratio]{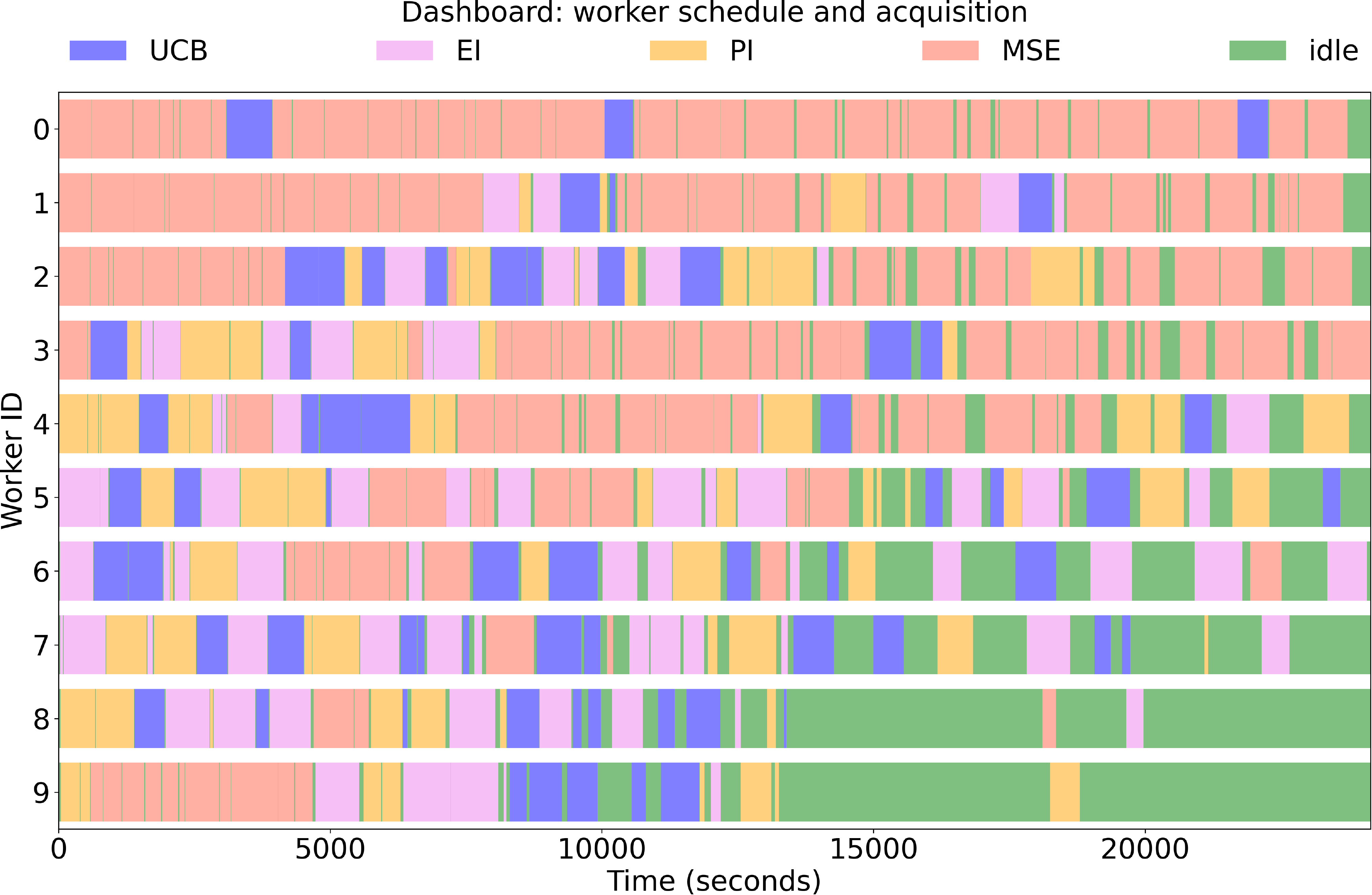}}
% \caption{Acquisition portfolio and scheduler.}
% \label{fig:portfolio_spheref}
% \end{figure}
\medskip
% \begin{figure}[!htbp]
\centering
\subcaptionbox{rothyp: acquisition portfolio.
}
  [0.30\linewidth]{\includegraphics[width=0.30\textwidth, keepaspectratio]{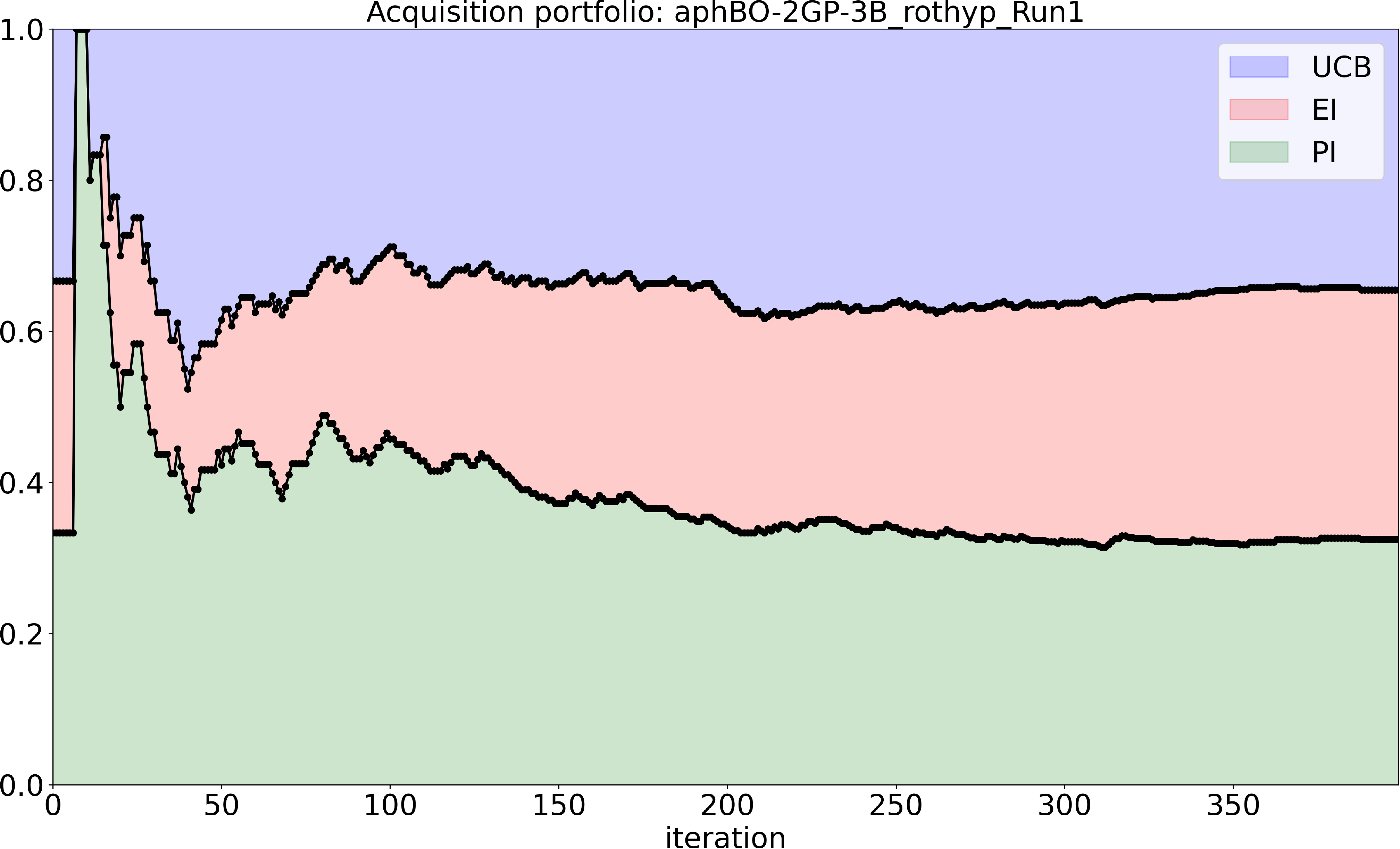}}
\hfill
\subcaptionbox{rothyp: number of acquisition functions.
}
  [0.30\linewidth]{\includegraphics[width=0.30\textwidth, keepaspectratio]{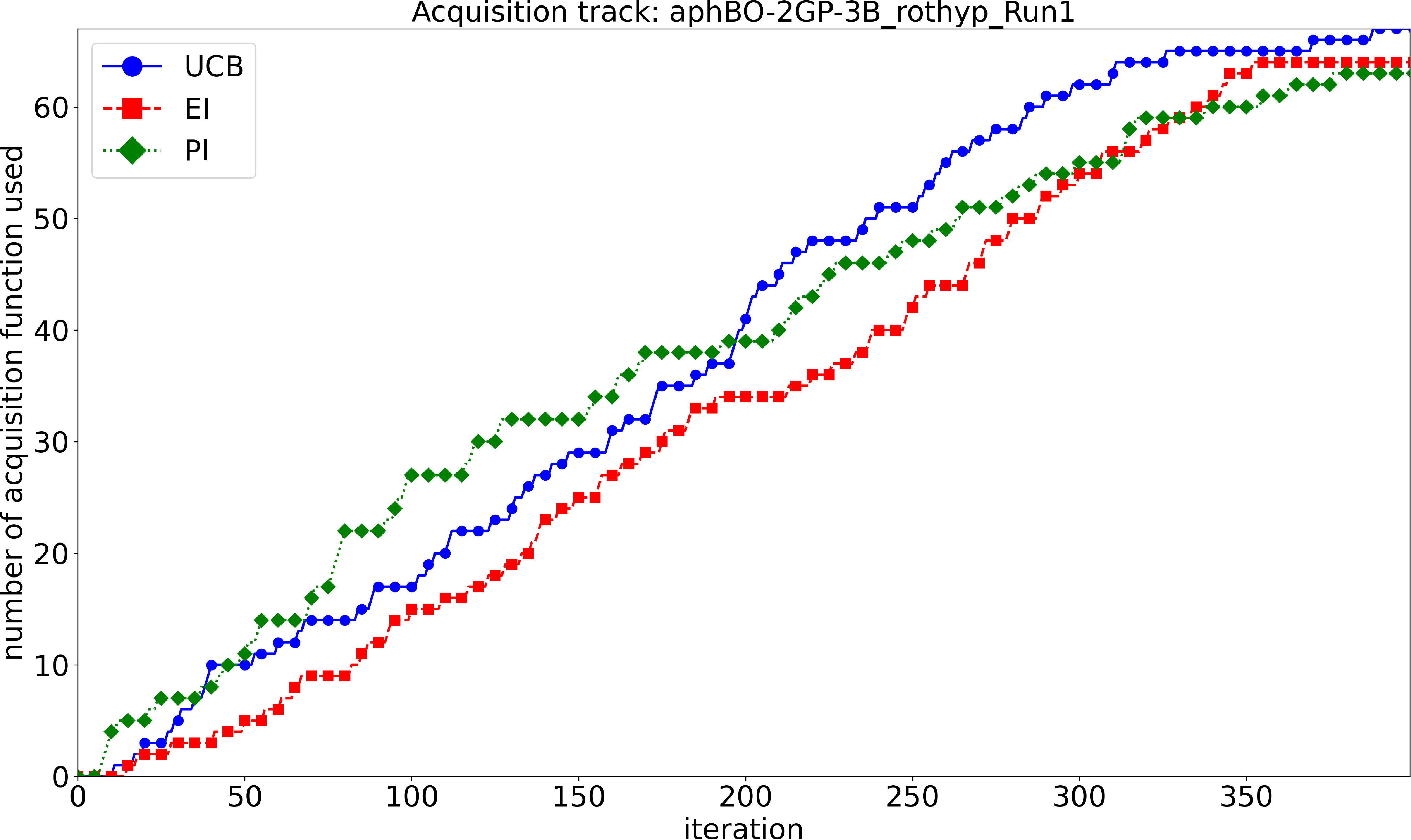}}
\hfill
\subcaptionbox{rothyp: scheduler dashboard.
}
  [0.30\linewidth]{\includegraphics[width=0.30\textwidth, keepaspectratio]{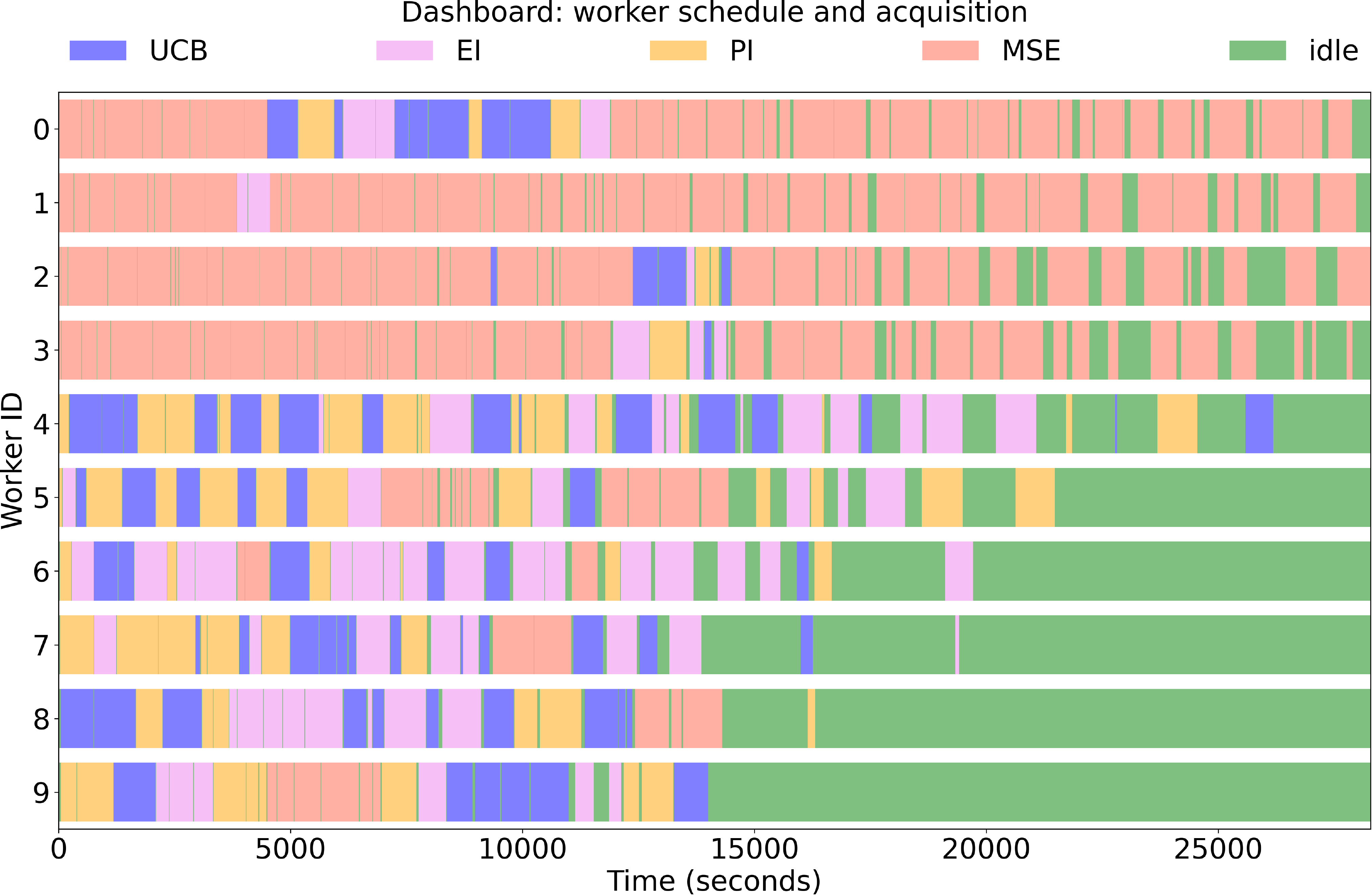}}
% \caption{Acquisition portfolio and scheduler.}
% \label{fig:portfolio_rothyp}
\end{figure}

\clearpage
\section{CFD results}
\label{app:CFDcomparison}

\begin{figure}[!htbp]
\centering
\subcaptionbox{Original design.
% \label{fig:original}
}
  [.45\linewidth]{\includegraphics[width=0.4001\textwidth, keepaspectratio]{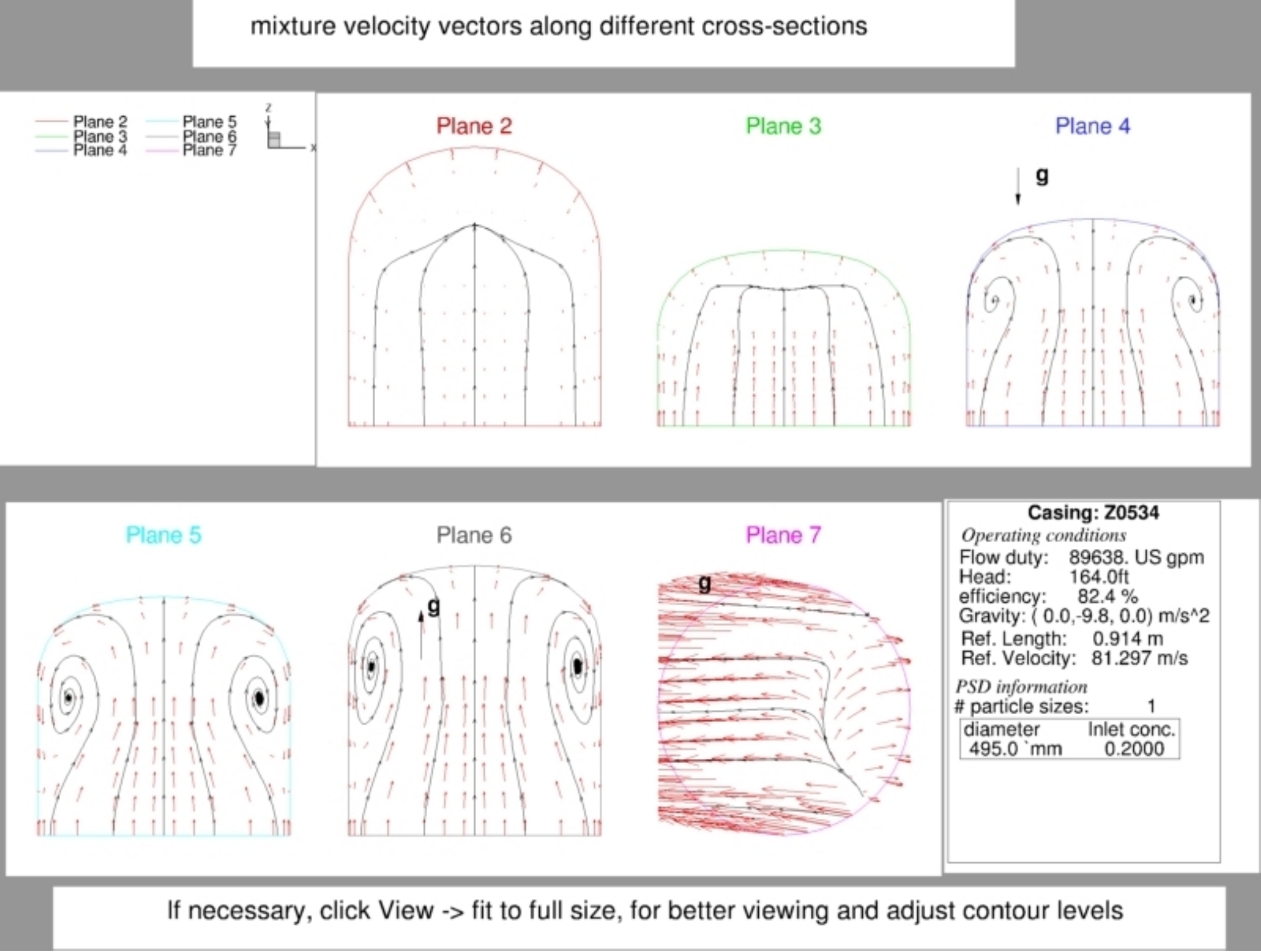}}
\hfill
\subcaptionbox{Optimal design.
% \label{fig:optimal}
}
  [.45\linewidth]{\includegraphics[width=0.4001\textwidth, keepaspectratio]{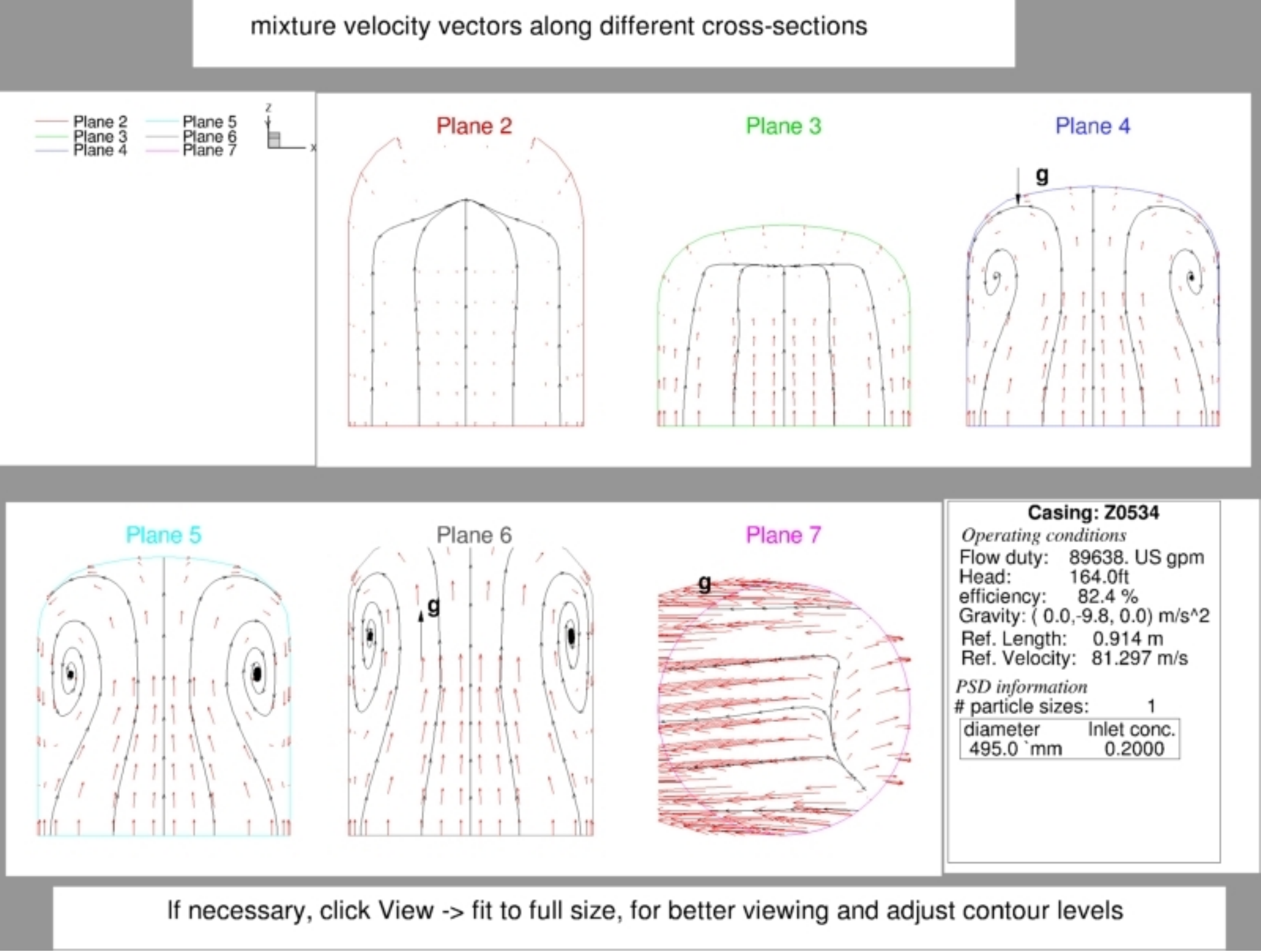}}

\centering
\subcaptionbox{Original design.
% \label{fig:original}
}
  [.45\linewidth]{\includegraphics[width=0.4001\textwidth, keepaspectratio]{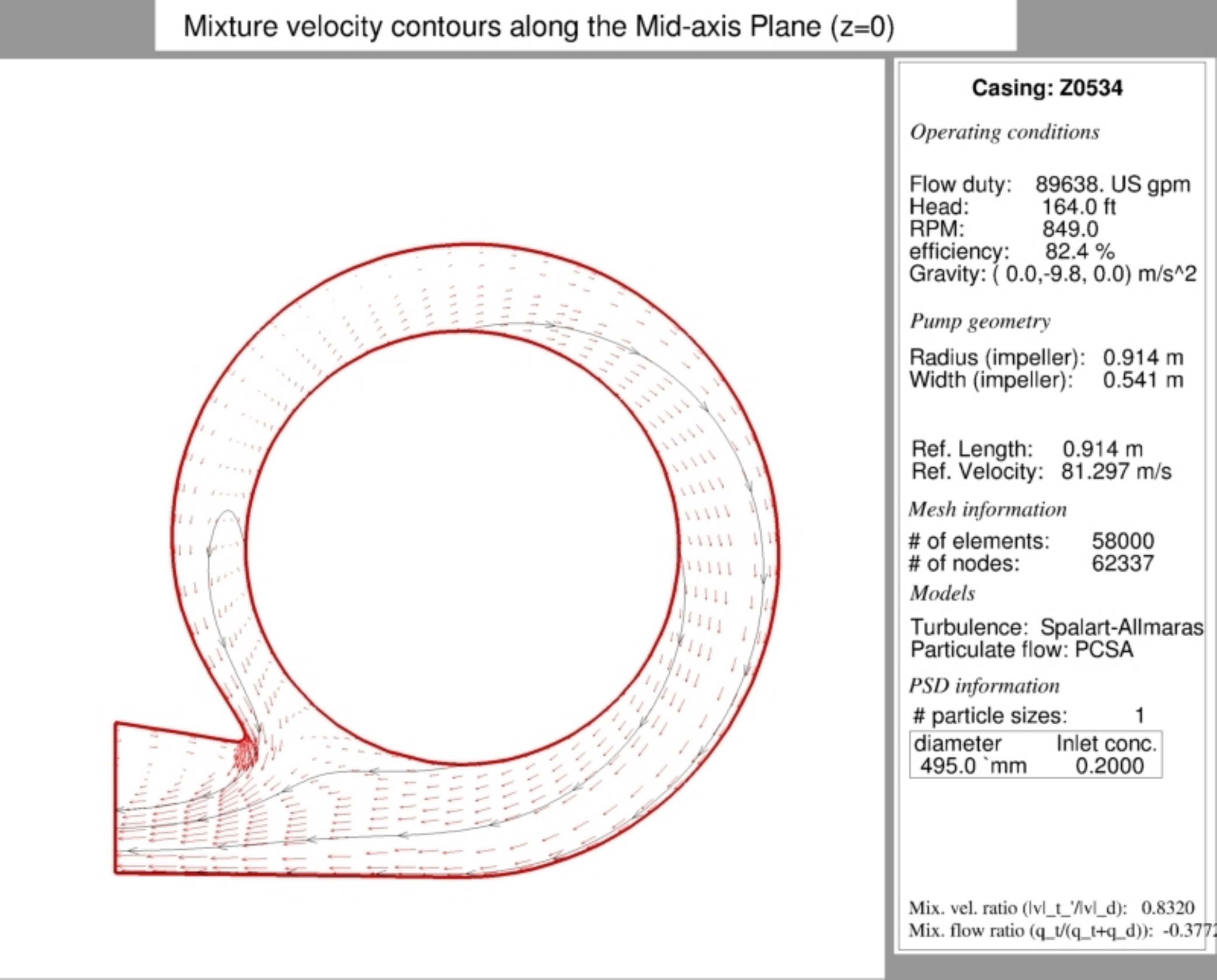}}
\hfill
\subcaptionbox{Optimal design.
% \label{fig:optimal}
}
  [.45\linewidth]{\includegraphics[width=0.4001\textwidth, keepaspectratio]{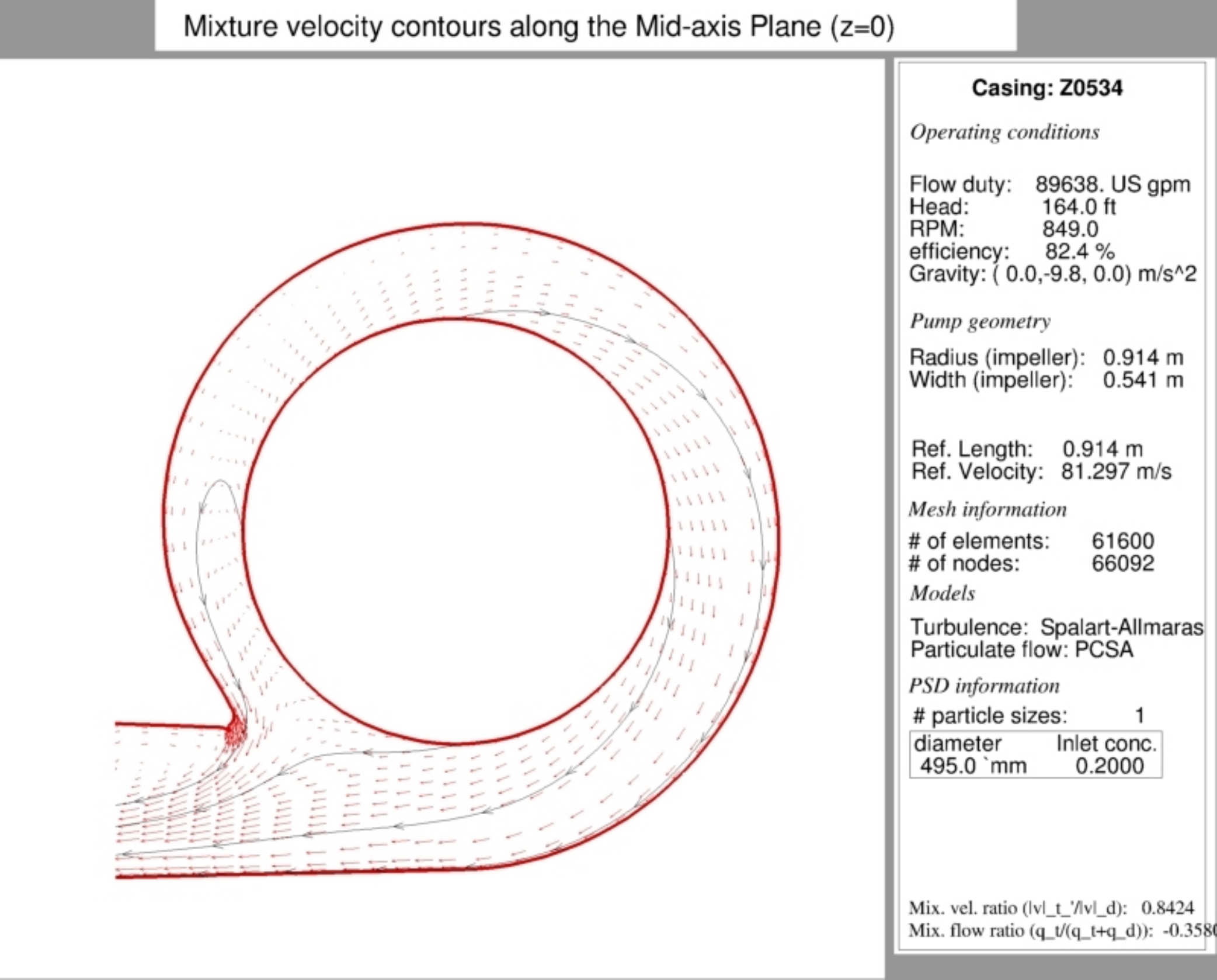}}
\caption{Comparison of mixture velocity.}
% \label{fig:comparison}
\end{figure}

\begin{figure}[!htbp]
\centering
\subcaptionbox{Original design.
% \label{fig:original}
}
  [.45\linewidth]{\includegraphics[width=0.4001\textwidth, keepaspectratio]{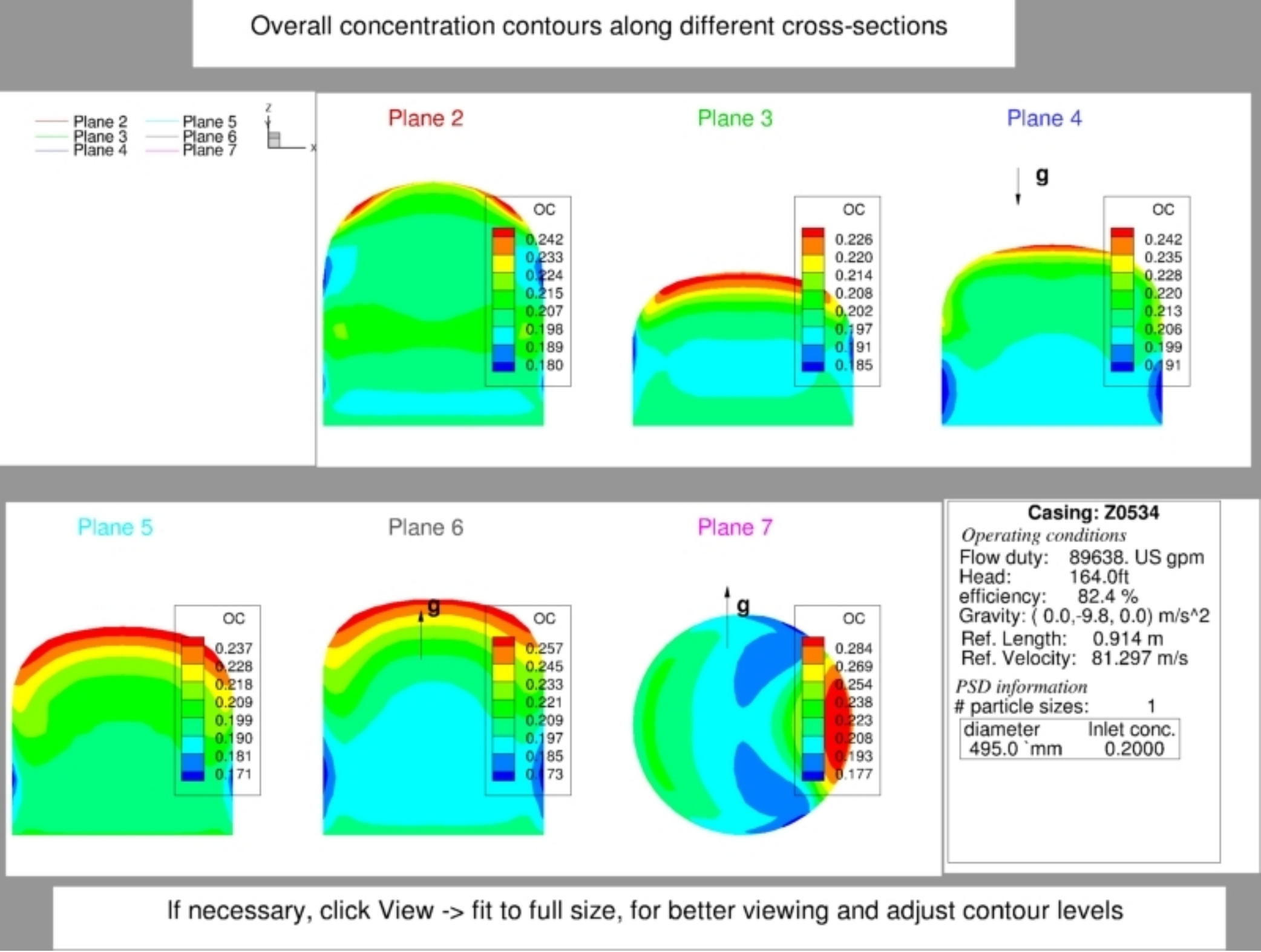}}
\hfill
\subcaptionbox{Optimal design.
% \label{fig:optimal}
}
  [.45\linewidth]{\includegraphics[width=0.4001\textwidth, keepaspectratio]{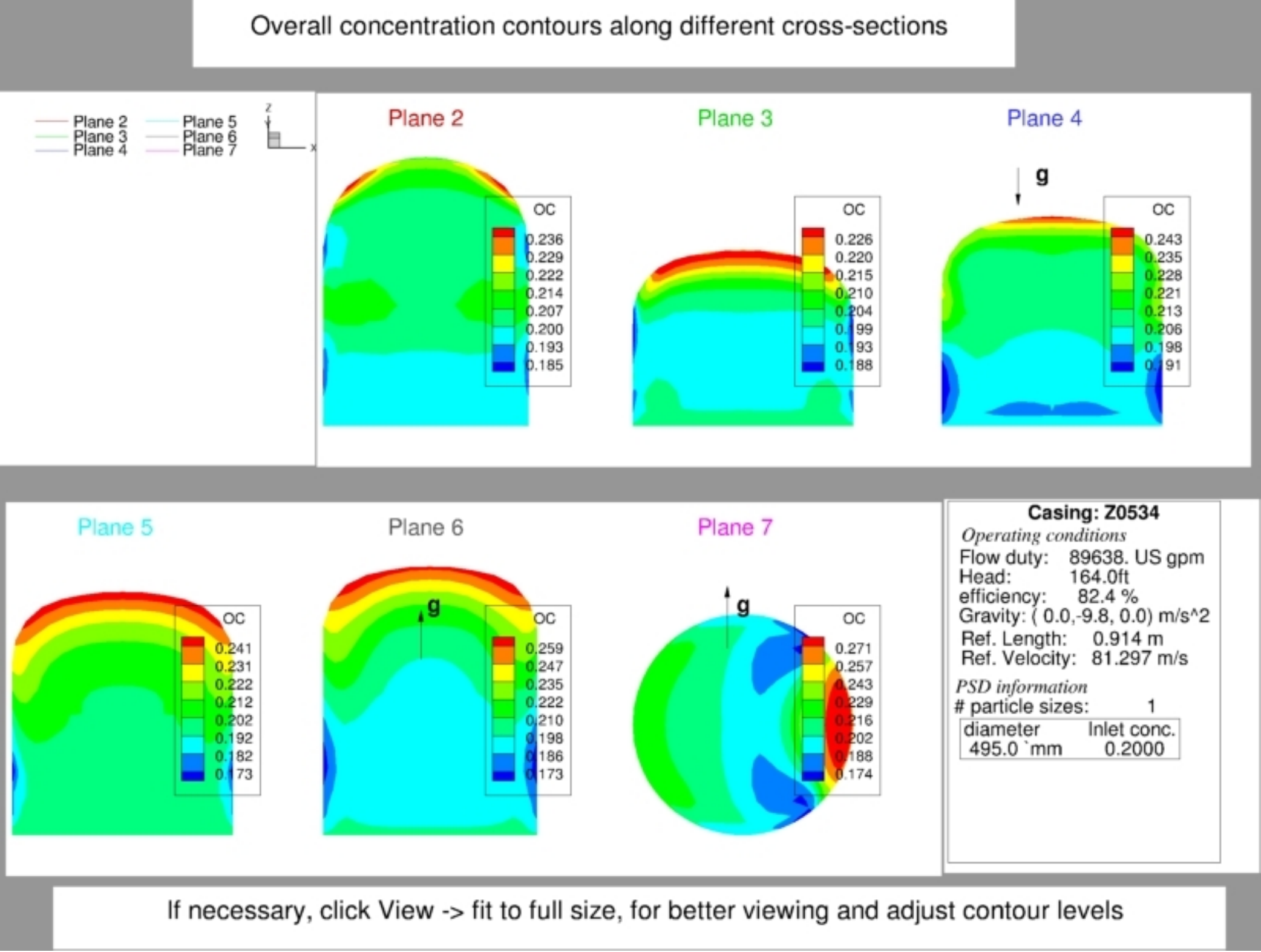}}

\centering
\subcaptionbox{Original design.
% \label{fig:original}
}
  [.45\linewidth]{\includegraphics[width=0.4001\textwidth, keepaspectratio]{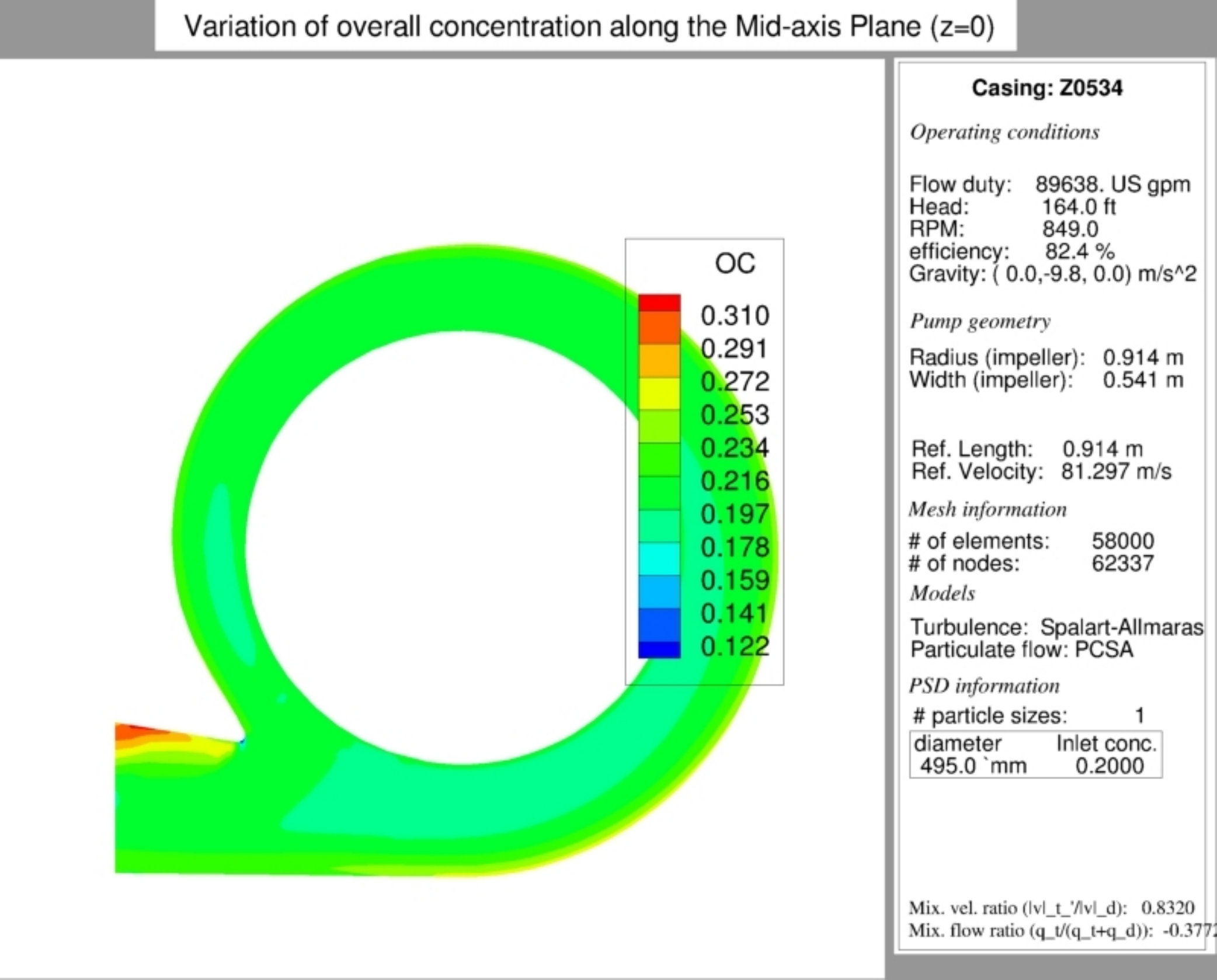}}
\hfill
\subcaptionbox{Optimal design.
% \label{fig:optimal}
}
  [.45\linewidth]{\includegraphics[width=0.4001\textwidth, keepaspectratio]{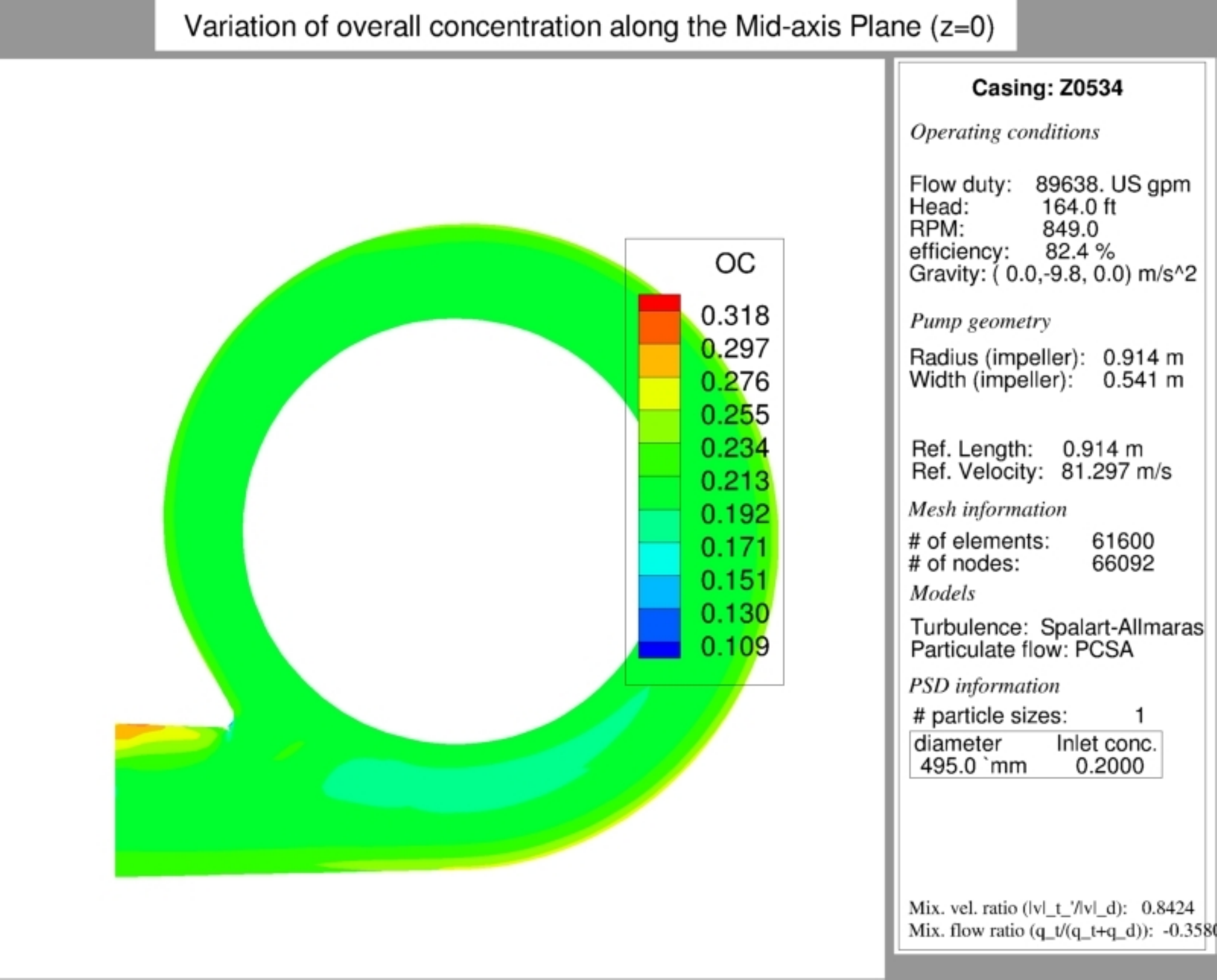}}
\caption{Comparison of overall concentration.}
% \label{fig:comparison}
\end{figure}

\begin{figure}[!htbp]
\centering
\subcaptionbox{Original design.
% \label{fig:original}
}
  [.45\linewidth]{\includegraphics[width=0.4001\textwidth, keepaspectratio]{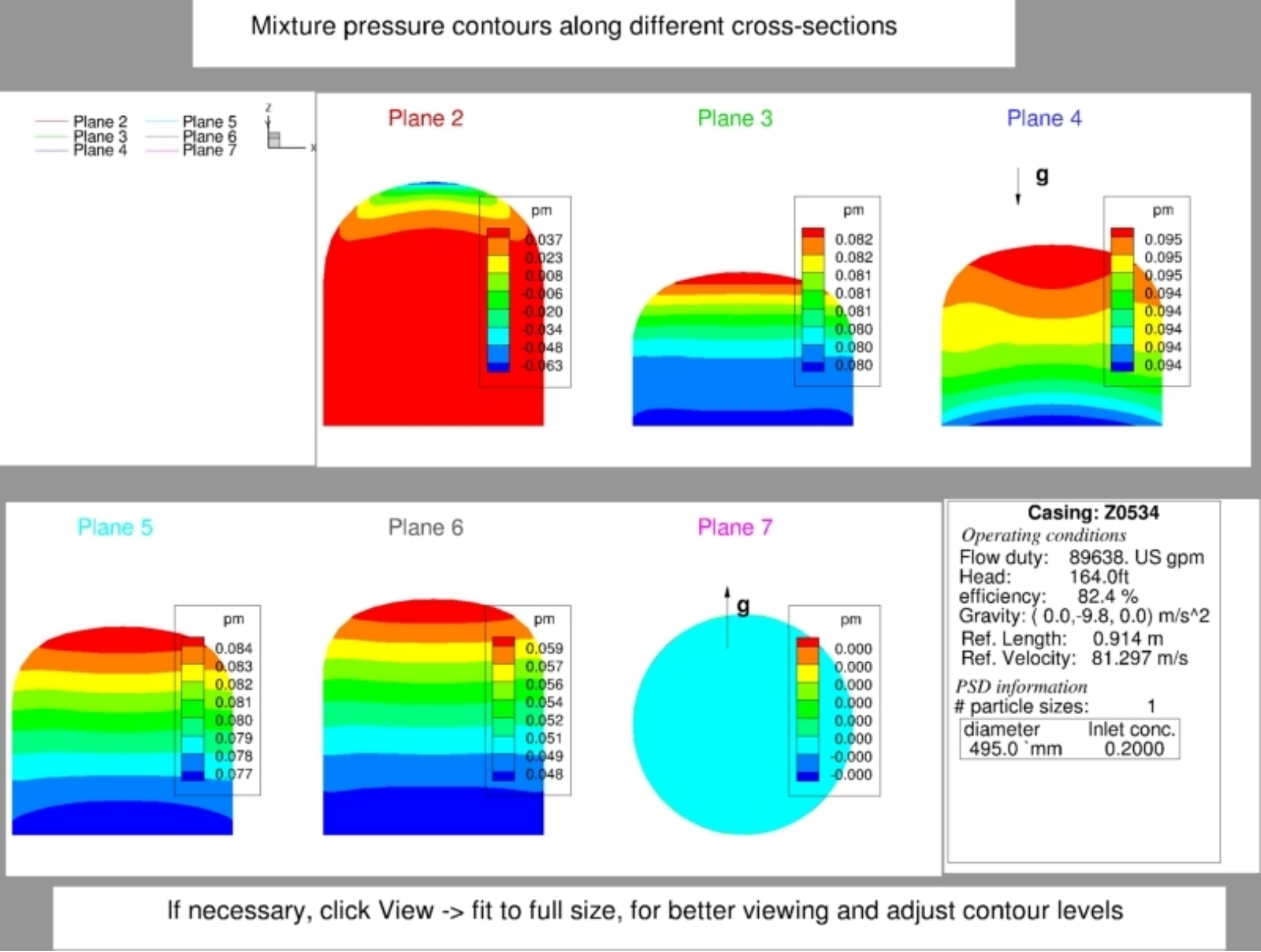}}
\hfill
\subcaptionbox{Optimal design.
% \label{fig:optimal}
}
  [.45\linewidth]{\includegraphics[width=0.4001\textwidth, keepaspectratio]{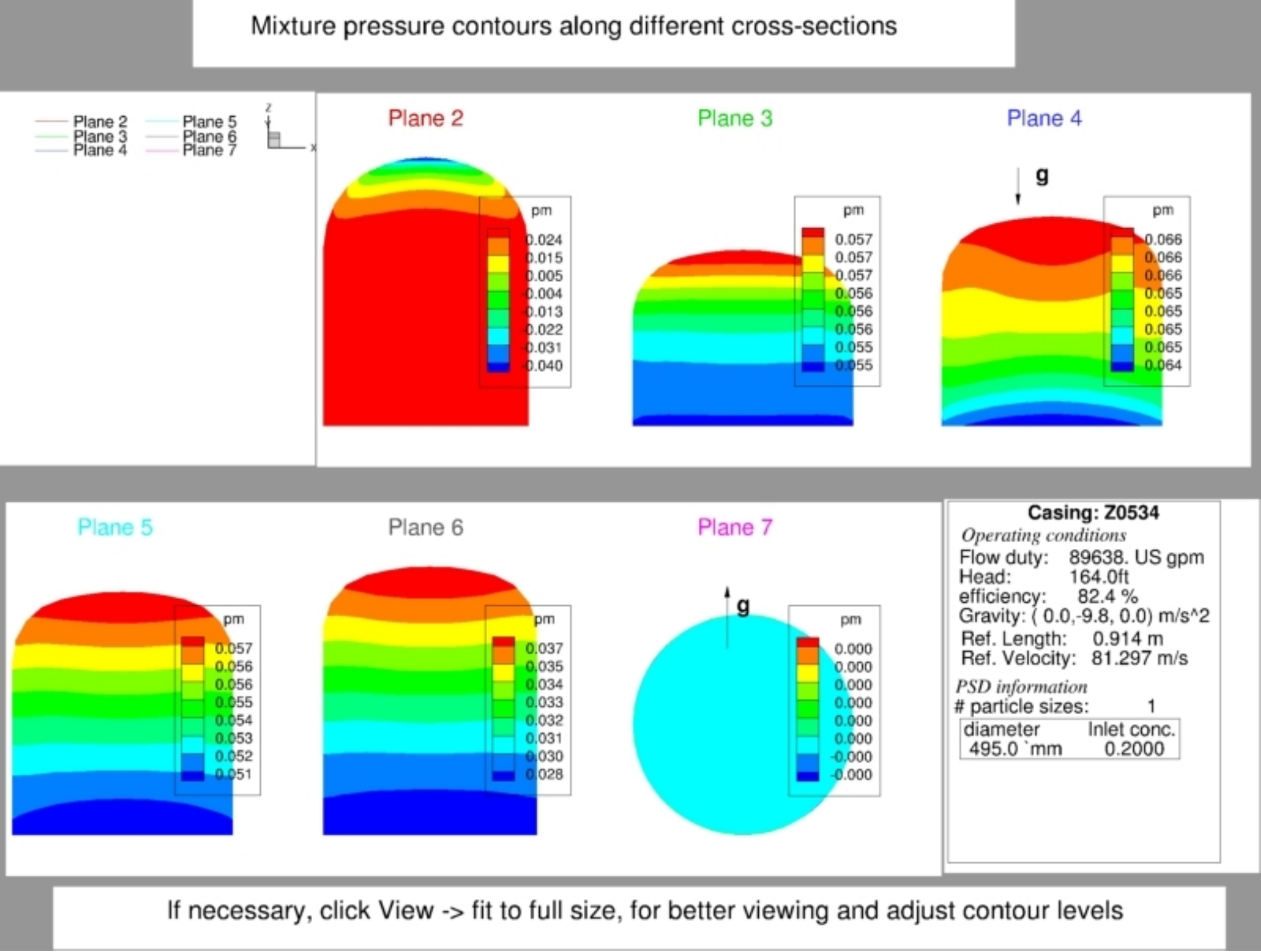}}

\centering
\subcaptionbox{Original design.
% \label{fig:original}
}
  [.45\linewidth]{\includegraphics[width=0.4001\textwidth, keepaspectratio]{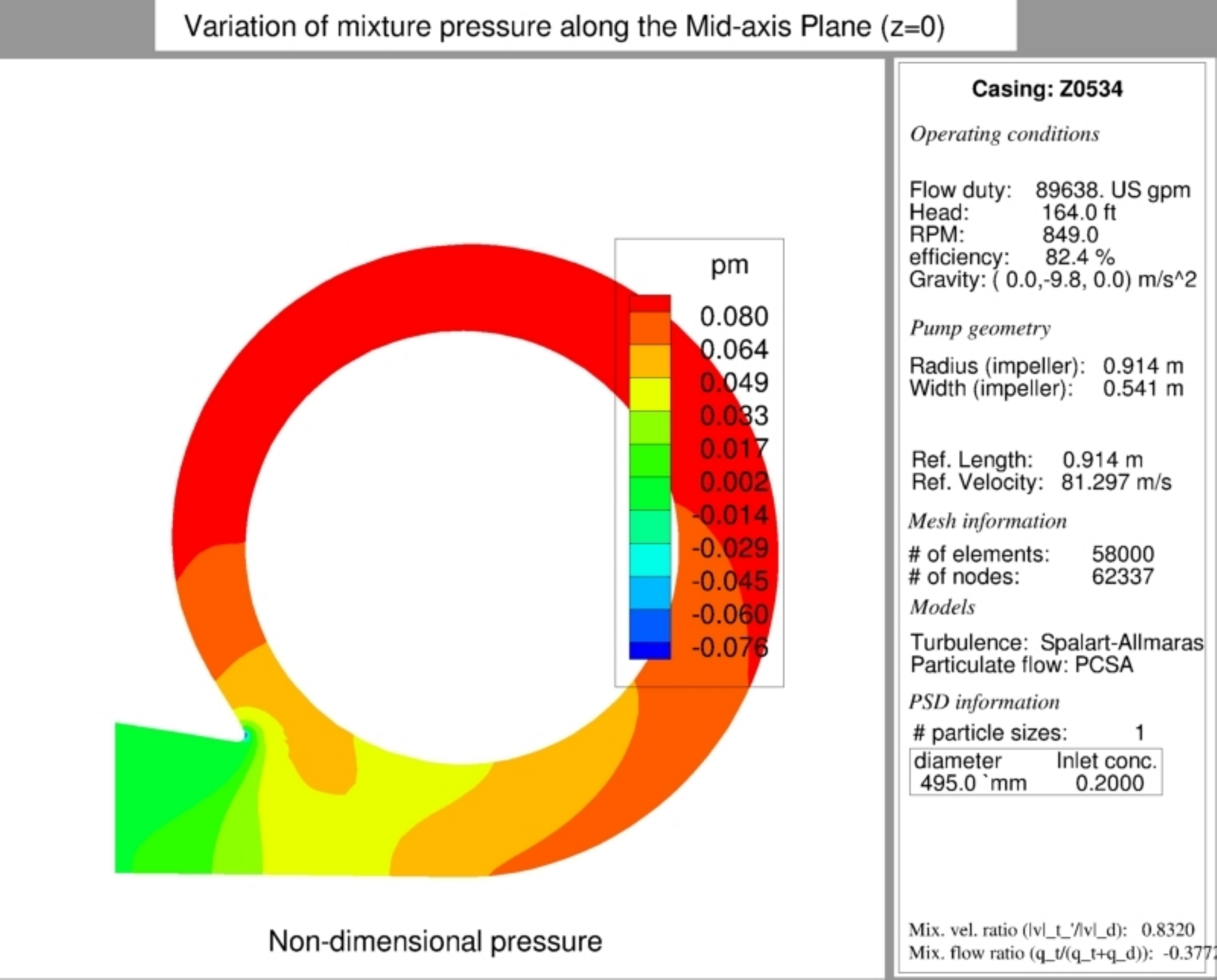}}
\hfill
\subcaptionbox{Optimal design.
% \label{fig:optimal}
}
  [.45\linewidth]{\includegraphics[width=0.4001\textwidth, keepaspectratio]{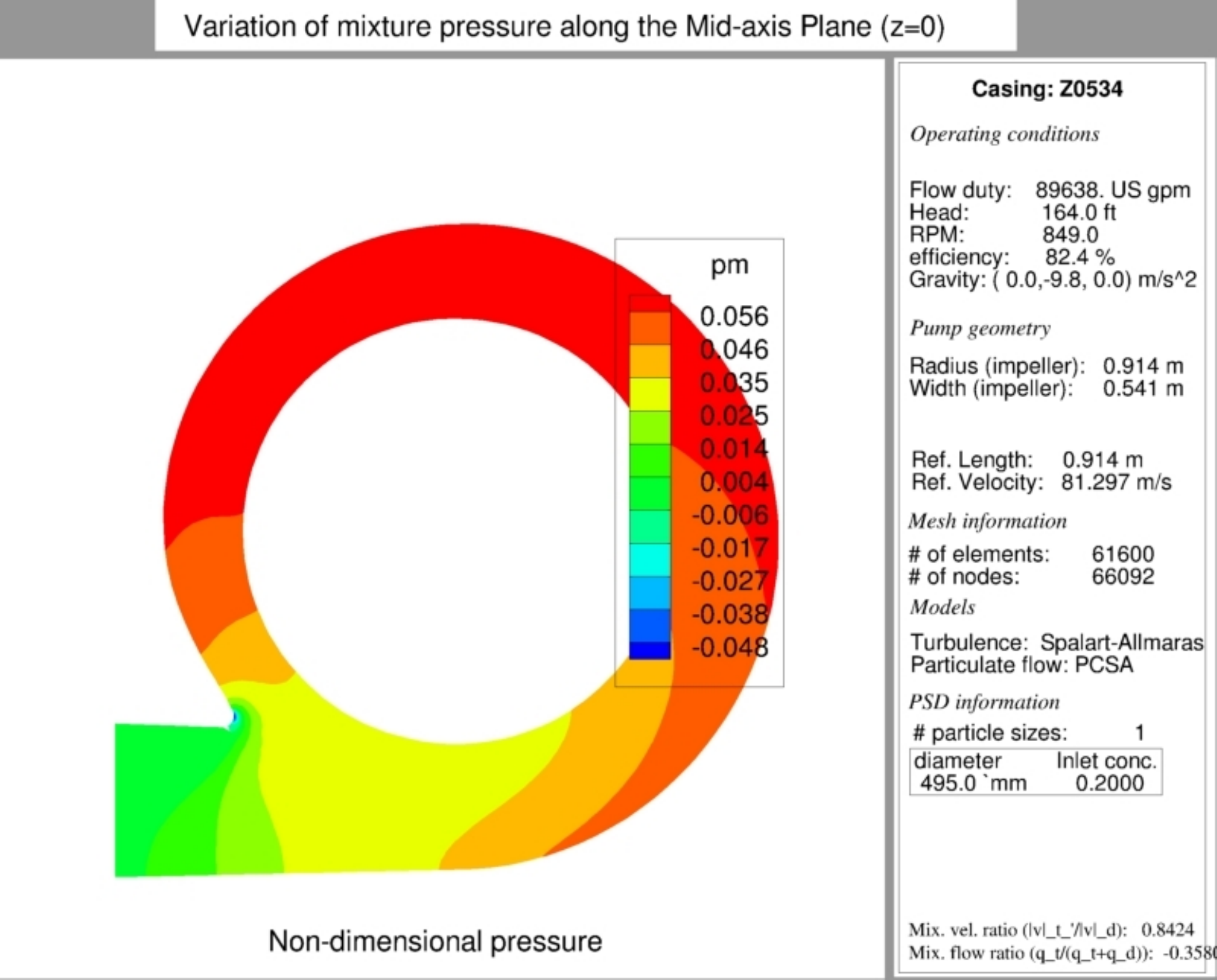}}
\caption{Comparison of pressure mixture.}
% \label{fig:comparison}
\end{figure}

\begin{figure}[!htbp]
\centering
\subcaptionbox{Original design.
% \label{fig:original}
}
  [.45\linewidth]{\includegraphics[width=0.4001\textwidth, keepaspectratio]{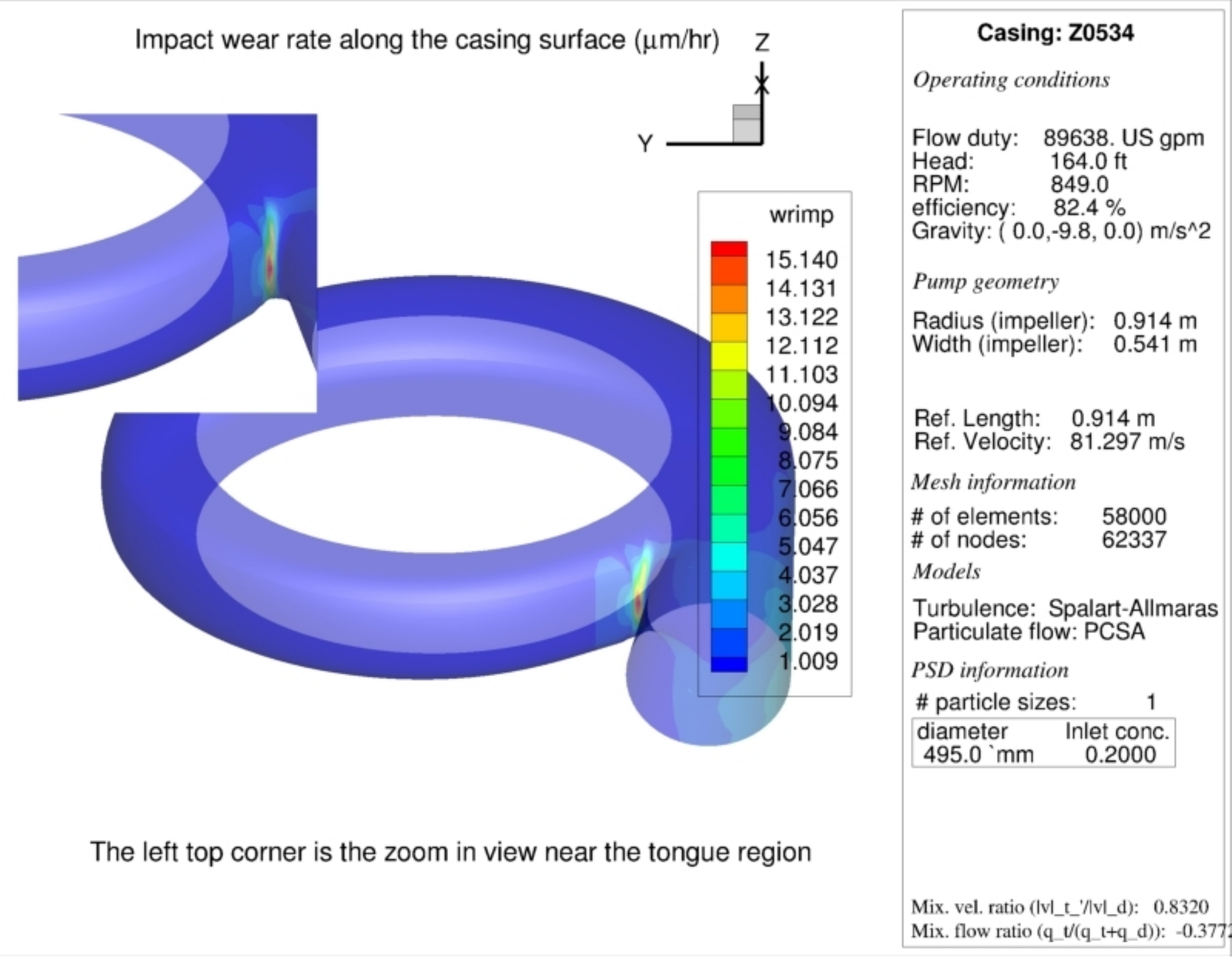}}
\hfill
\subcaptionbox{Optimal design.
% \label{fig:optimal}
}
  [.45\linewidth]{\includegraphics[width=0.4001\textwidth, keepaspectratio]{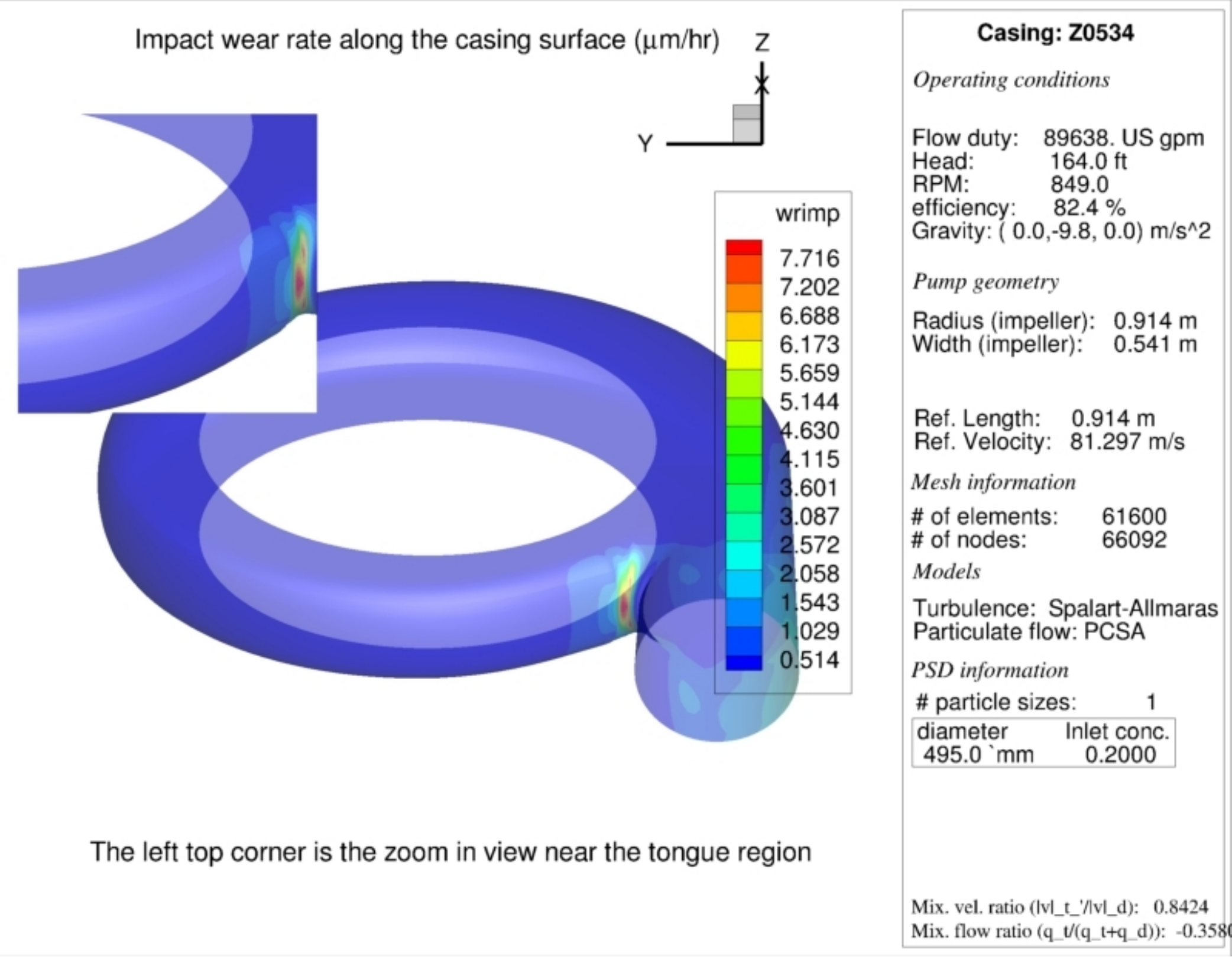}}

\centering
\subcaptionbox{Original design.
% \label{fig:original}
}
  [.45\linewidth]{\includegraphics[width=0.4001\textwidth, keepaspectratio]{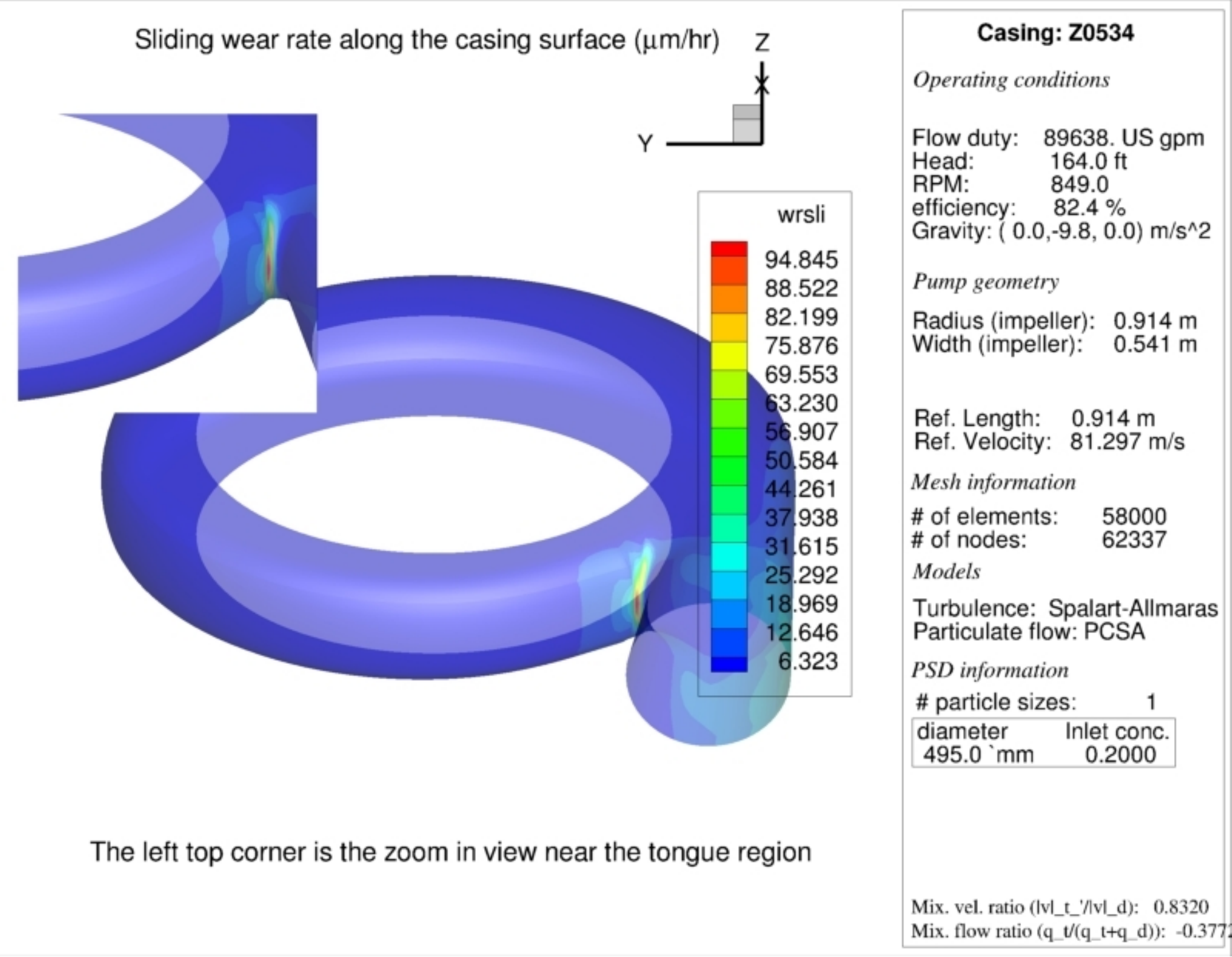}}
\hfill
\subcaptionbox{Optimal design.
% \label{fig:optimal}
}
  [.45\linewidth]{\includegraphics[width=0.4001\textwidth, keepaspectratio]{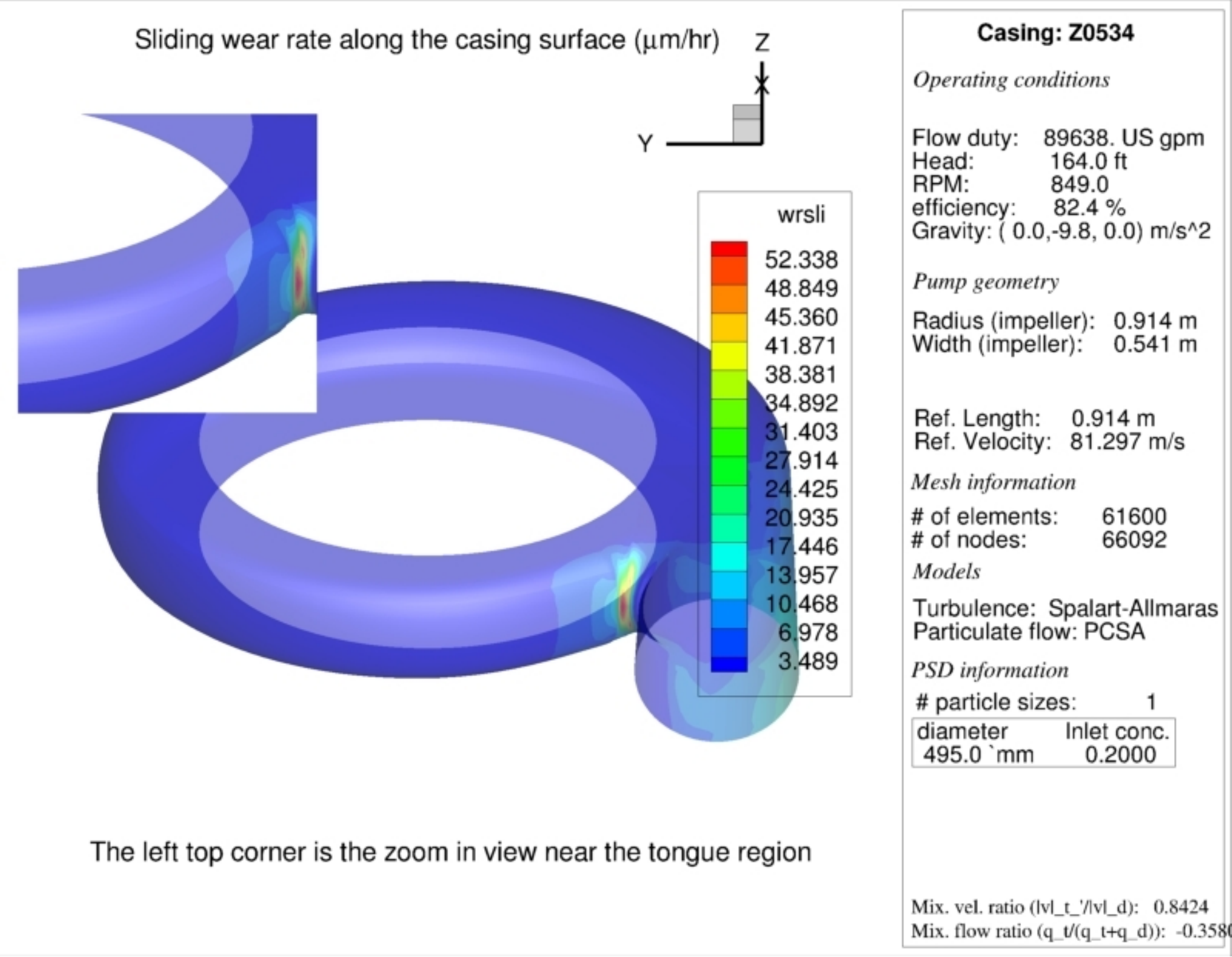}}

\centering
\subcaptionbox{Original design.
% \label{fig:original}
}
  [.45\linewidth]{\includegraphics[width=0.4001\textwidth, keepaspectratio]{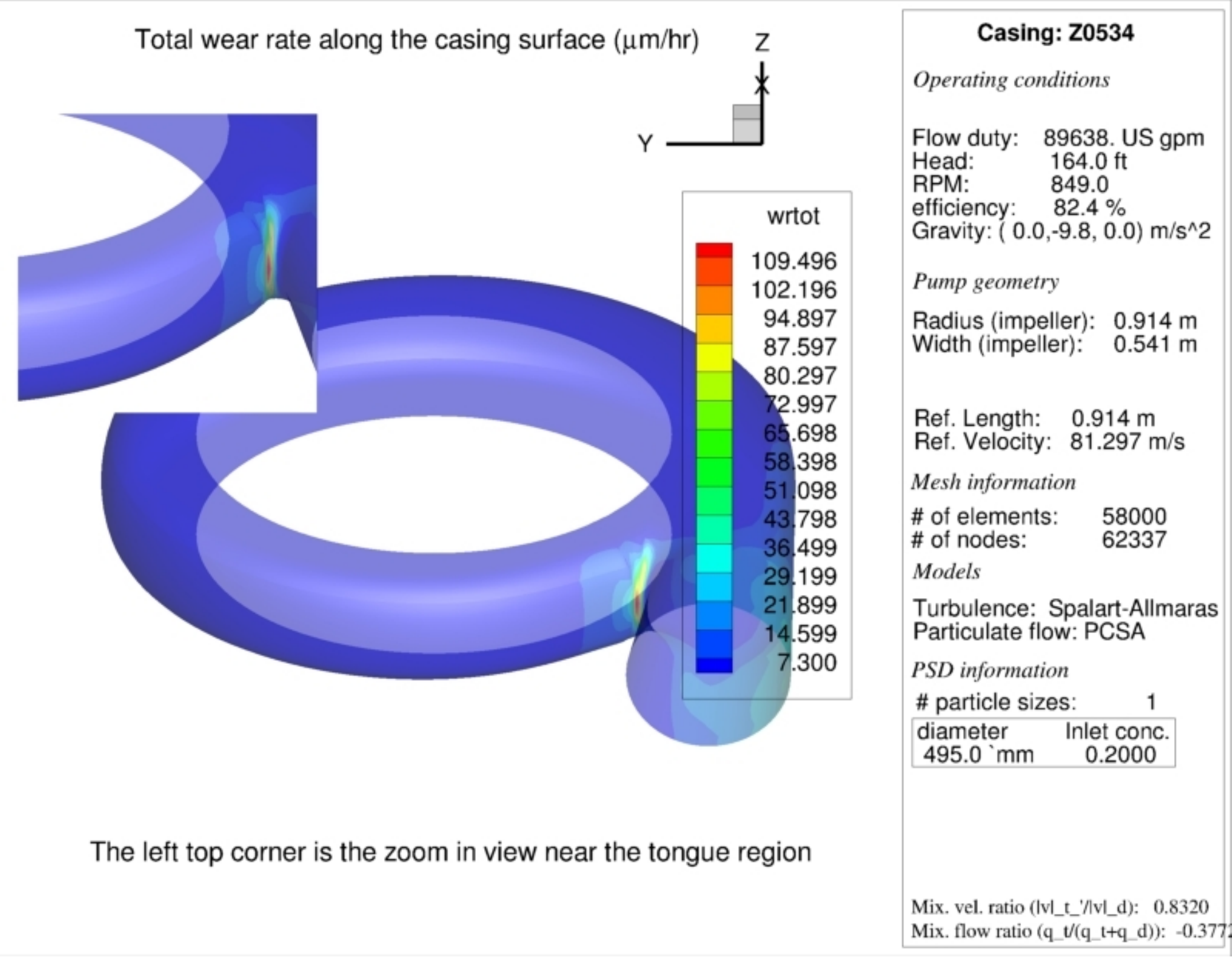}}
\hfill
\subcaptionbox{Optimal design.
% \label{fig:optimal}
}
  [.45\linewidth]{\includegraphics[width=0.4001\textwidth, keepaspectratio]{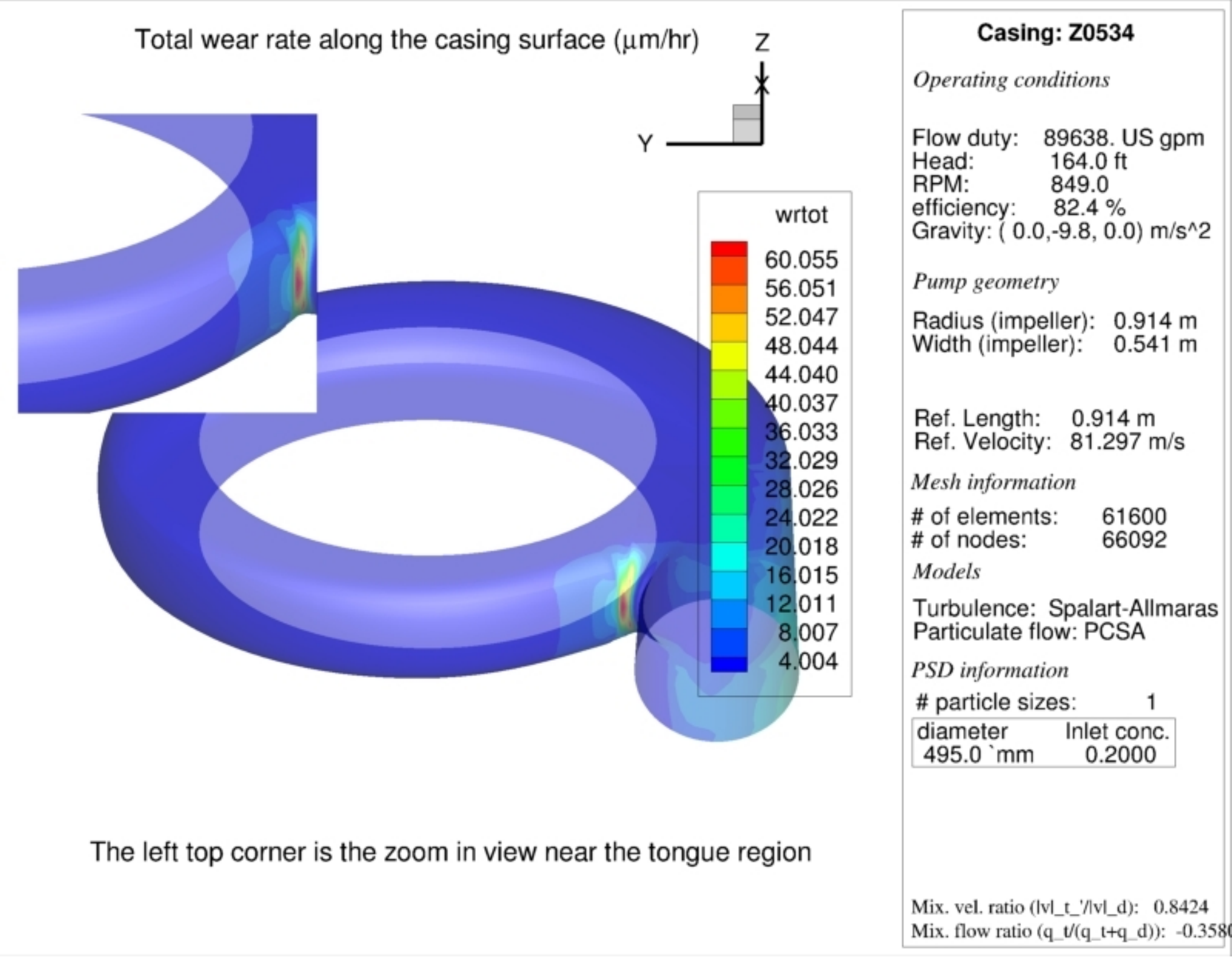}}
\caption{Comparison of wear in 3D.}
% \label{fig:comparison}
\end{figure}

\begin{figure}[!htbp]
\centering
\subcaptionbox{Original design.
% \label{fig:original}
}
  [.45\linewidth]{\includegraphics[width=0.4001\textwidth, keepaspectratio]{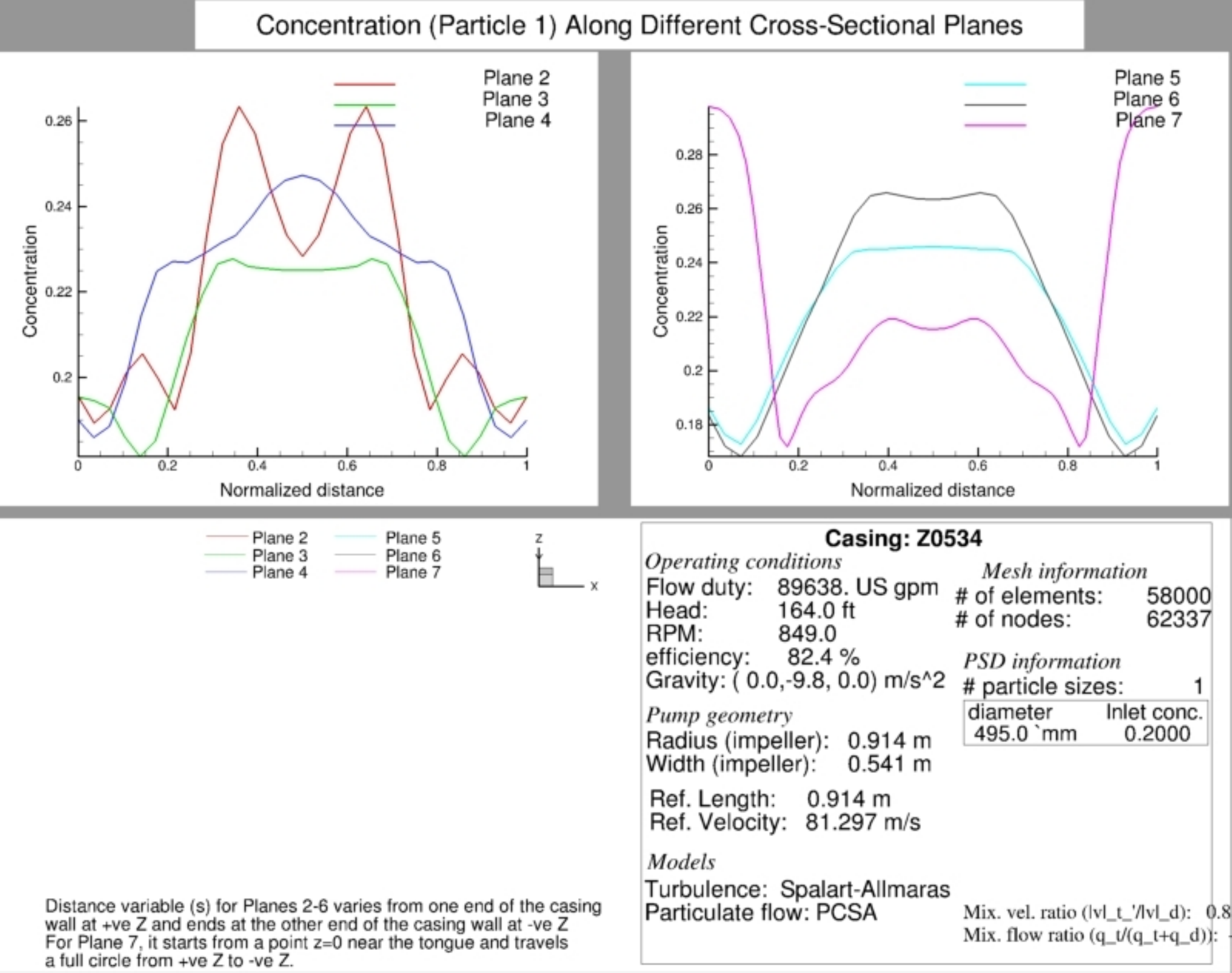}}
\hfill
\subcaptionbox{Optimal design.
% \label{fig:optimal}
}
  [.45\linewidth]{\includegraphics[width=0.4001\textwidth, keepaspectratio]{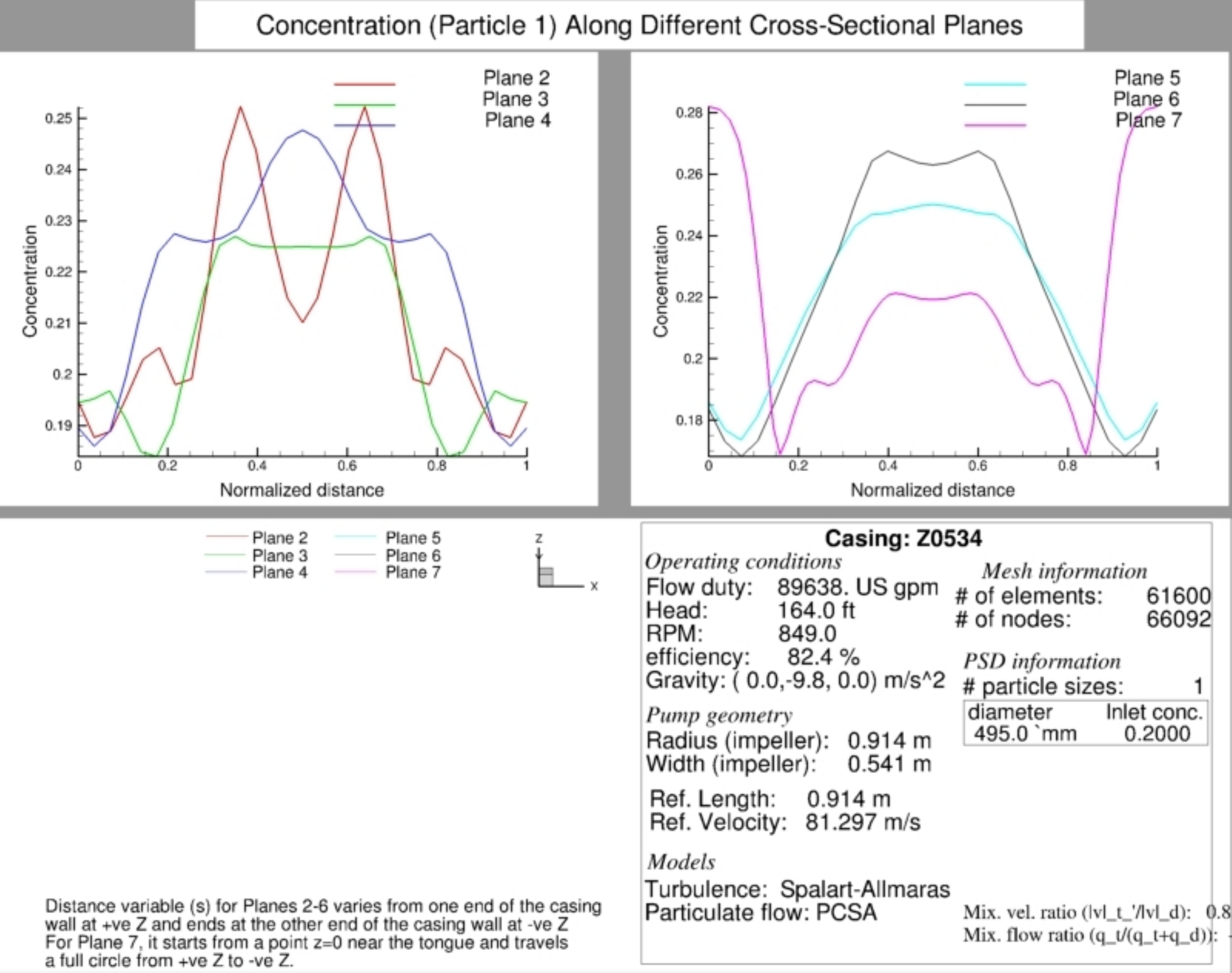}}

\centering
\subcaptionbox{Original design.
% \label{fig:original}
}
  [.45\linewidth]{\includegraphics[width=0.4001\textwidth, keepaspectratio]{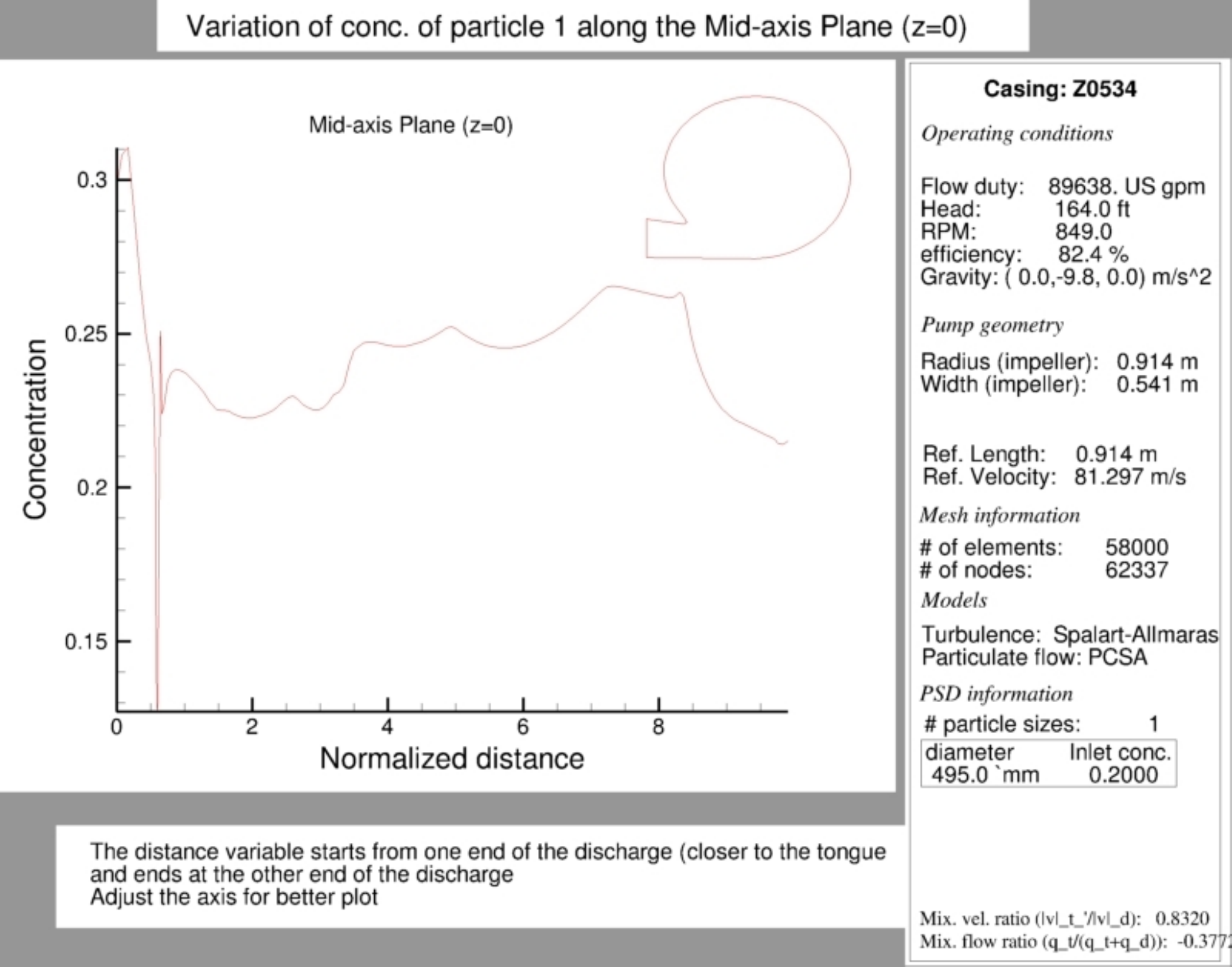}}
\hfill
\subcaptionbox{Optimal design.
% \label{fig:optimal}
}
  [.45\linewidth]{\includegraphics[width=0.4001\textwidth, keepaspectratio]{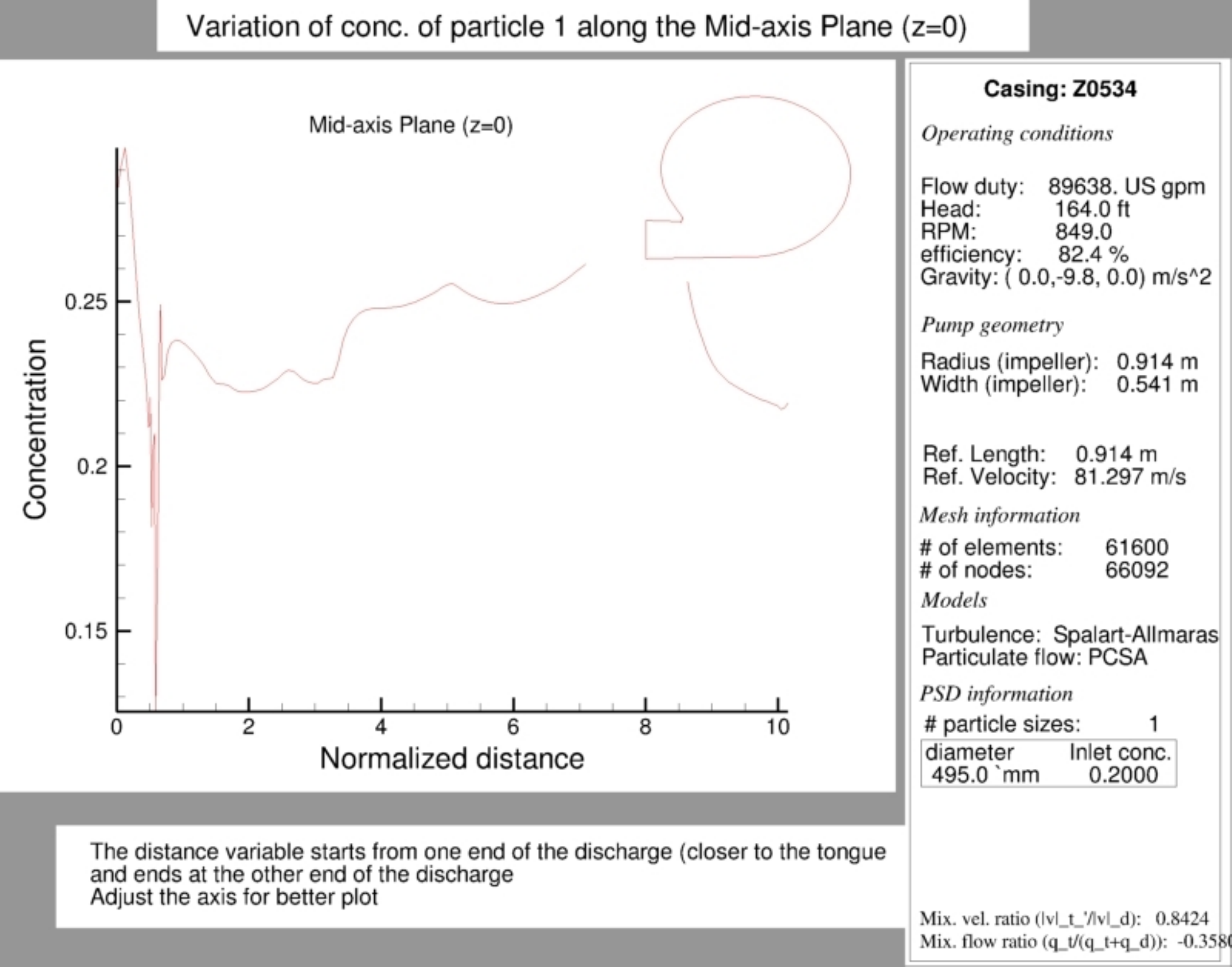}}

\centering
\subcaptionbox{Original design.
% \label{fig:original}
}
  [.45\linewidth]{\includegraphics[width=0.4001\textwidth, keepaspectratio]{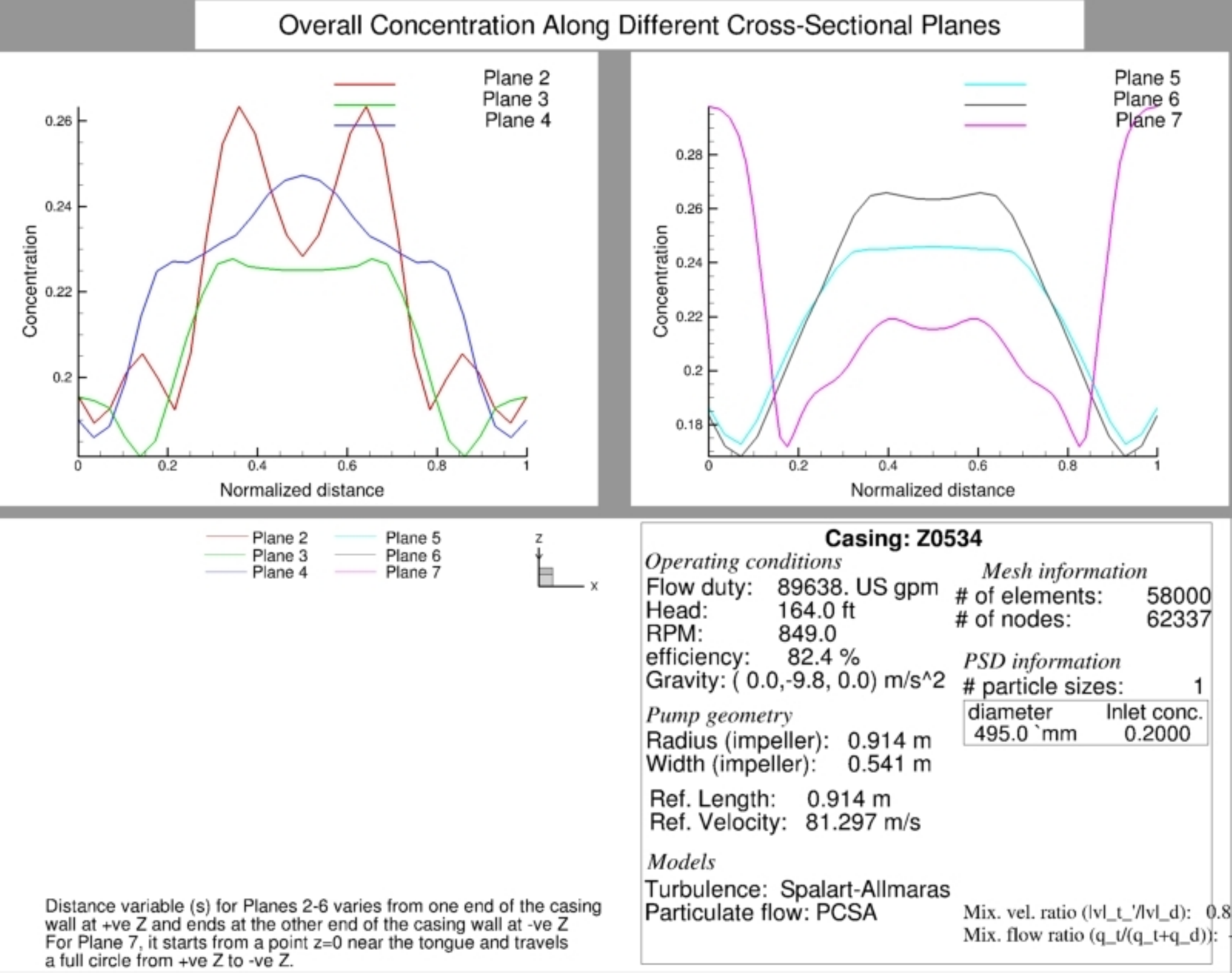}}
\hfill
\subcaptionbox{Optimal design.
% \label{fig:optimal}
}
  [.45\linewidth]{\includegraphics[width=0.4001\textwidth, keepaspectratio]{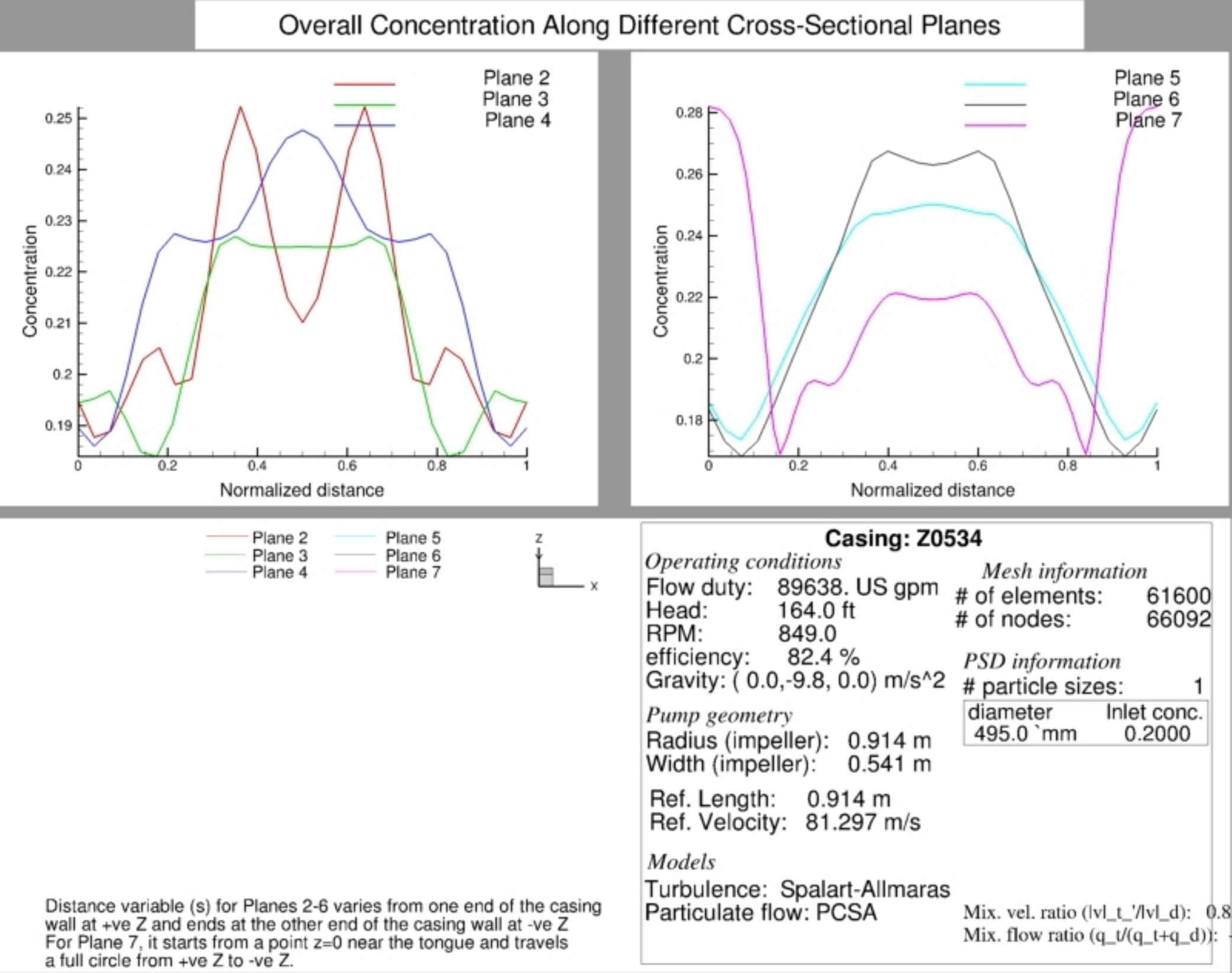}}

\caption{Comparison of concentration.}
% \label{fig:comparison}
\end{figure}

\begin{figure}[!htbp]
\centering
\subcaptionbox{Original design.
% \label{fig:original}
}
  [.45\linewidth]{\includegraphics[width=0.4001\textwidth, keepaspectratio]{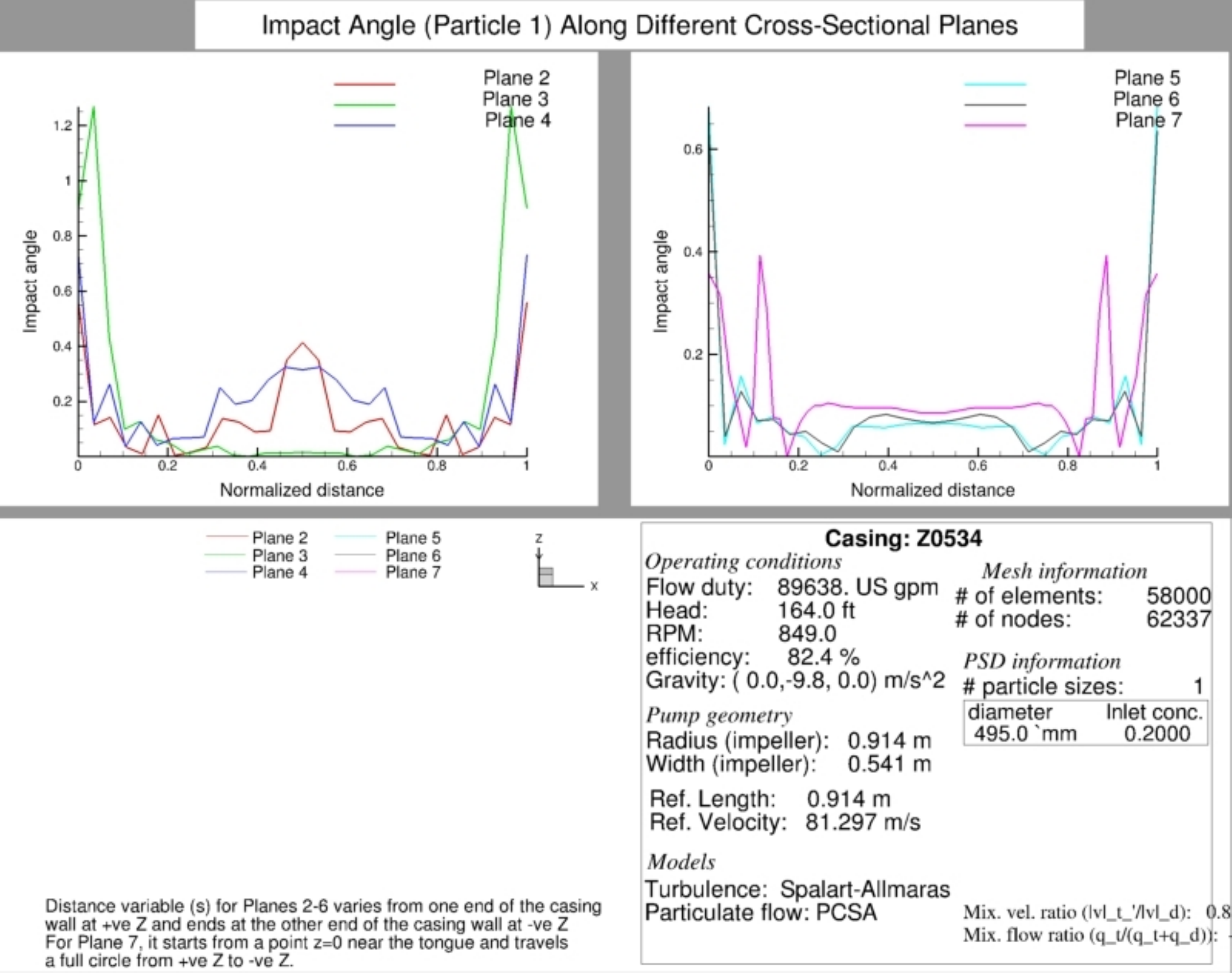}}
\hfill
\subcaptionbox{Optimal design.
% \label{fig:optimal}
}
  [.45\linewidth]{\includegraphics[width=0.4001\textwidth, keepaspectratio]{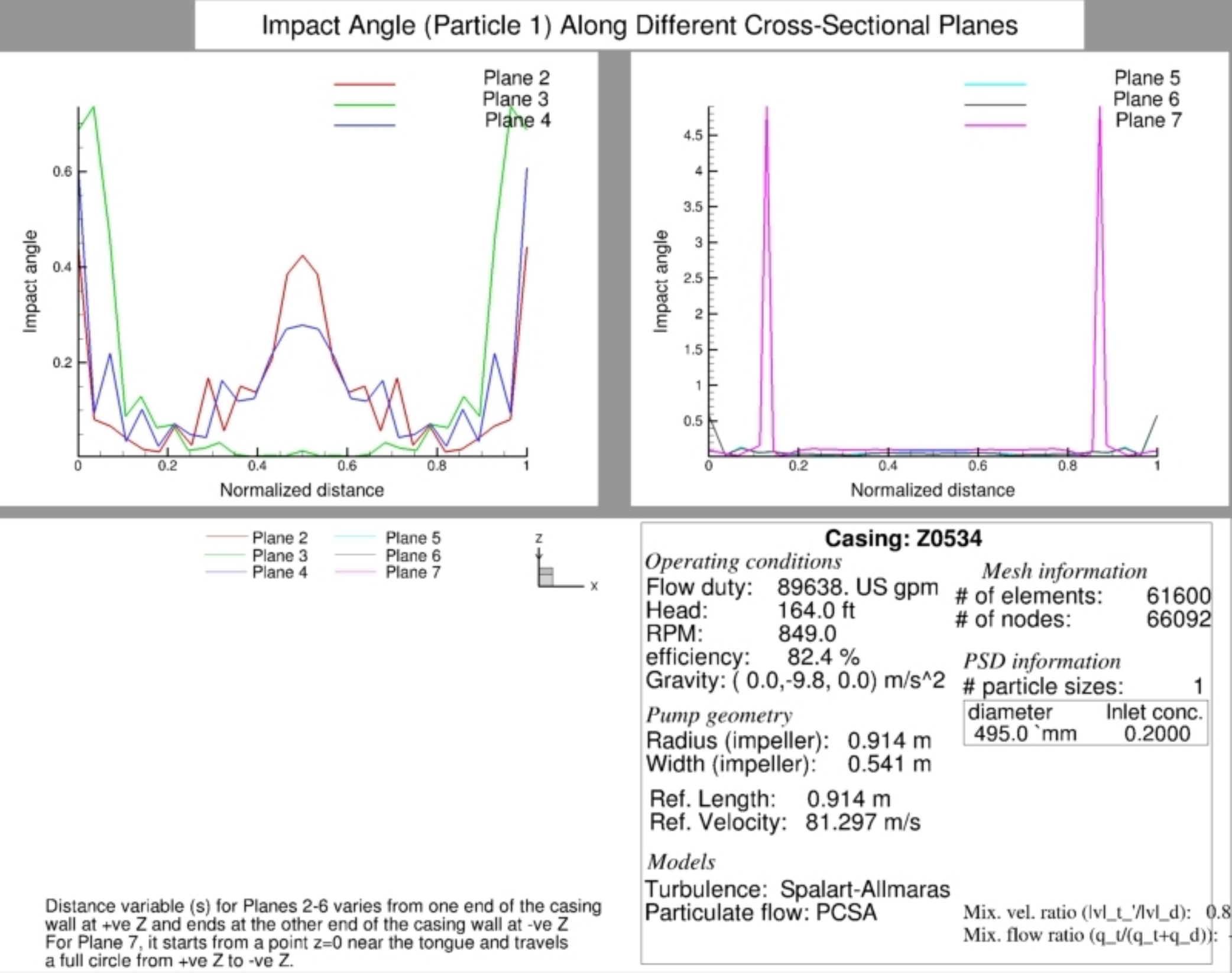}}

\centering
\subcaptionbox{Original design.
% \label{fig:original}
}
  [.45\linewidth]{\includegraphics[width=0.4001\textwidth, keepaspectratio]{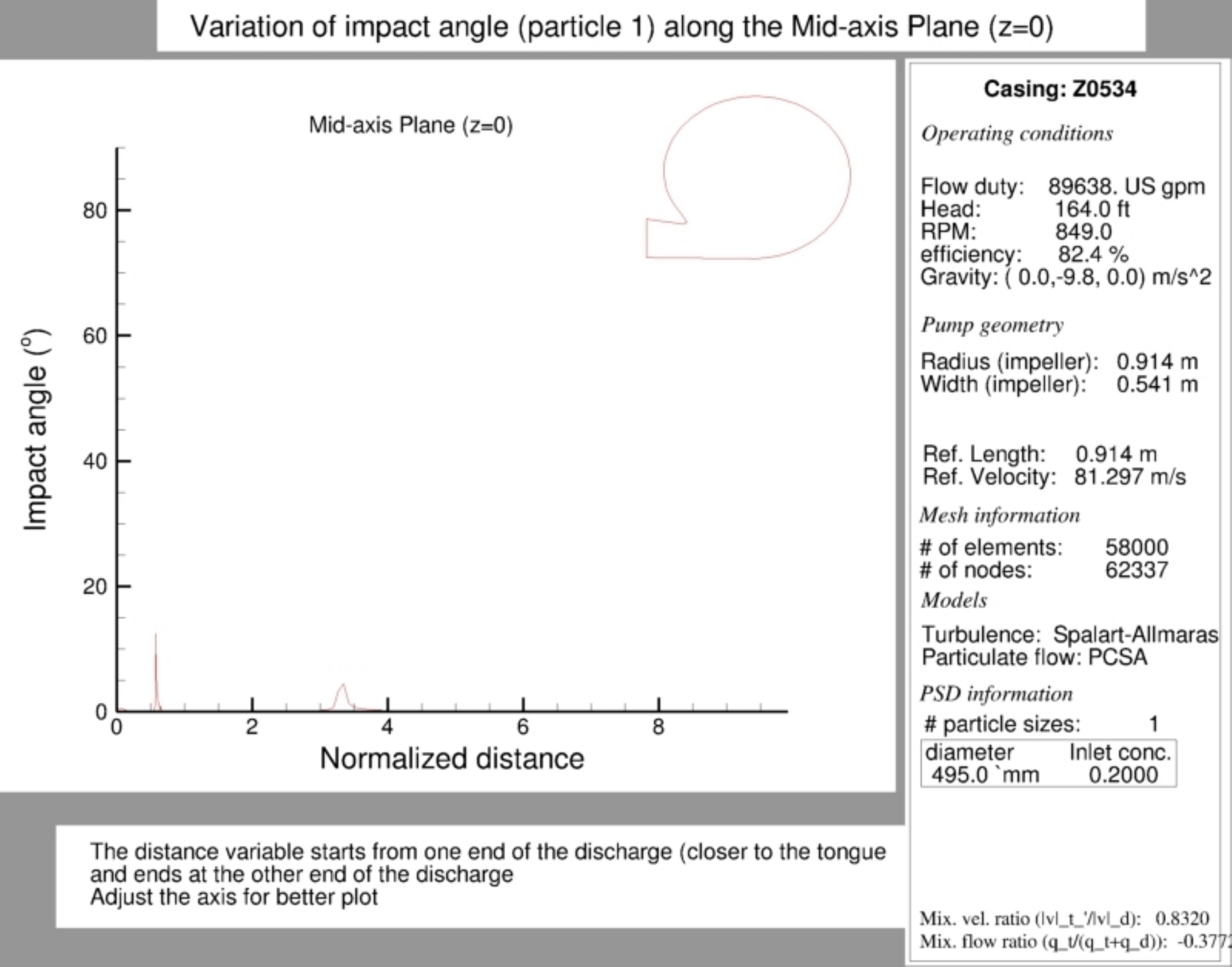}}
\hfill
\subcaptionbox{Optimal design.
% \label{fig:optimal}
}
  [.45\linewidth]{\includegraphics[width=0.4001\textwidth, keepaspectratio]{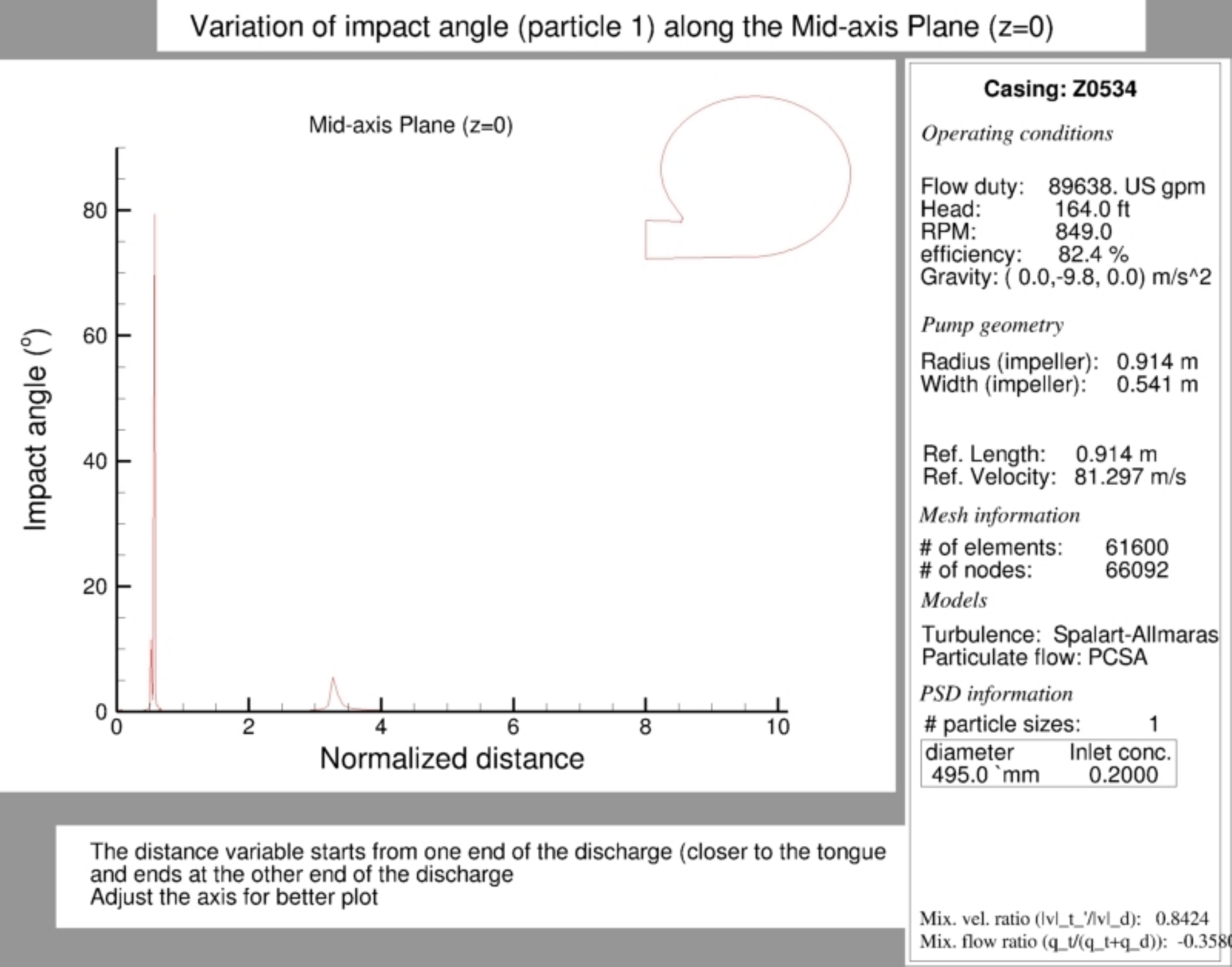}}
\caption{Comparison of impact angle.}
% \label{fig:comparison}
\end{figure}

\begin{figure}[!htbp]
\centering
\subcaptionbox{Original design.
% \label{fig:original}
}
  [.45\linewidth]{\includegraphics[width=0.4001\textwidth, keepaspectratio]{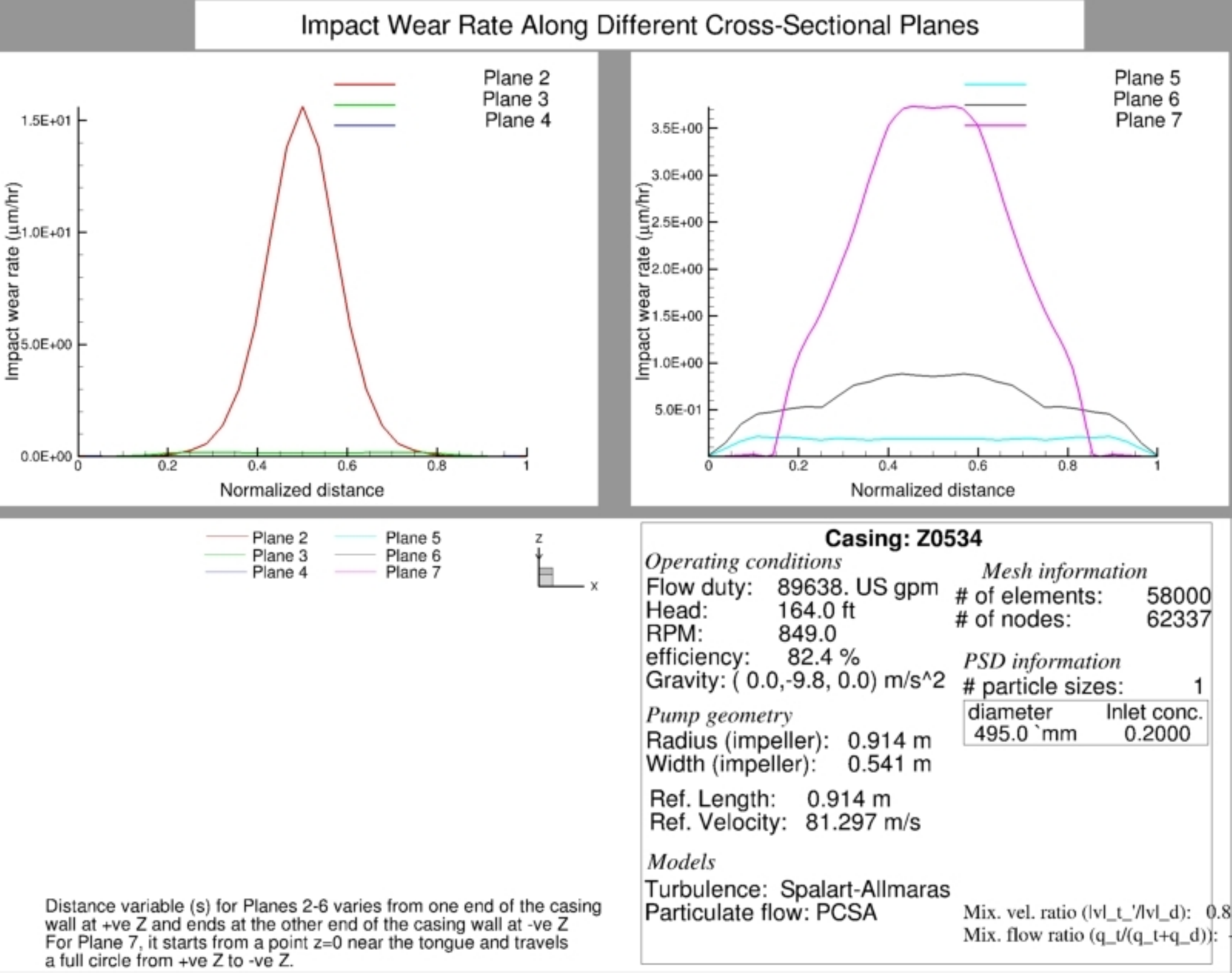}}
\hfill
\subcaptionbox{Optimal design.
% \label{fig:optimal}
}
  [.45\linewidth]{\includegraphics[width=0.4001\textwidth, keepaspectratio]{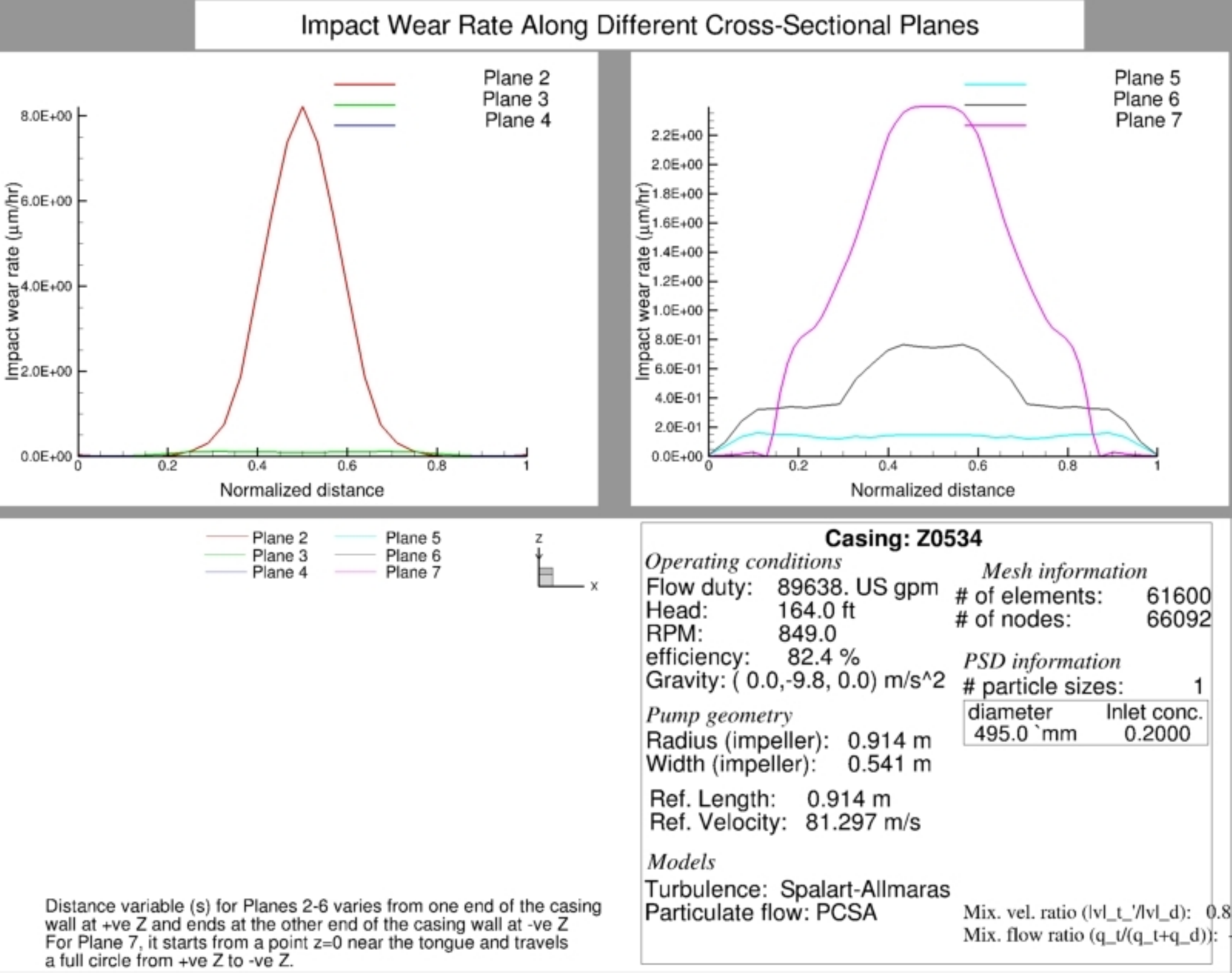}}

\centering
\subcaptionbox{Original design.
% \label{fig:original}
}
  [.45\linewidth]{\includegraphics[width=0.4001\textwidth, keepaspectratio]{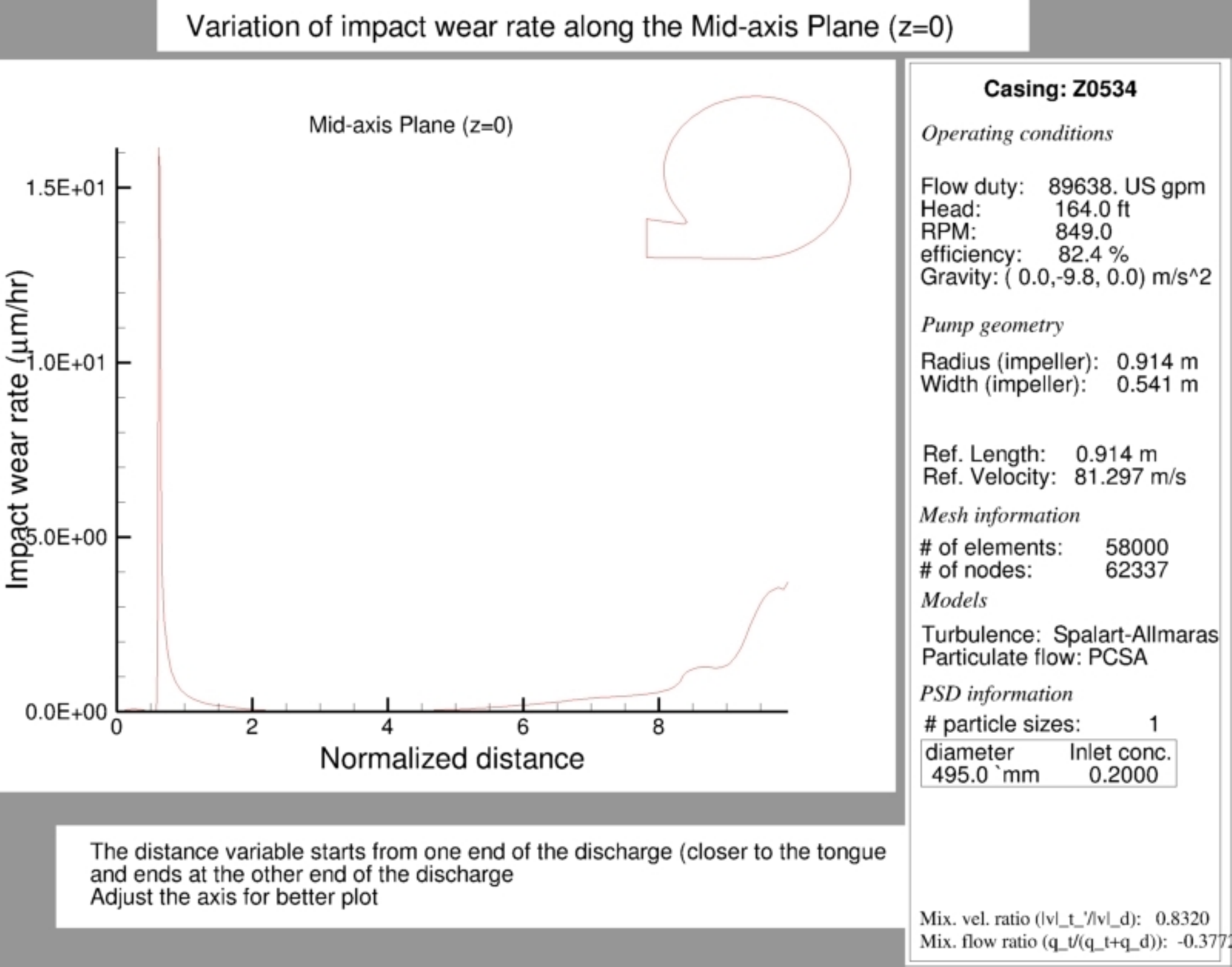}}
\hfill
\subcaptionbox{Optimal design.
% \label{fig:optimal}
}
  [.45\linewidth]{\includegraphics[width=0.4001\textwidth, keepaspectratio]{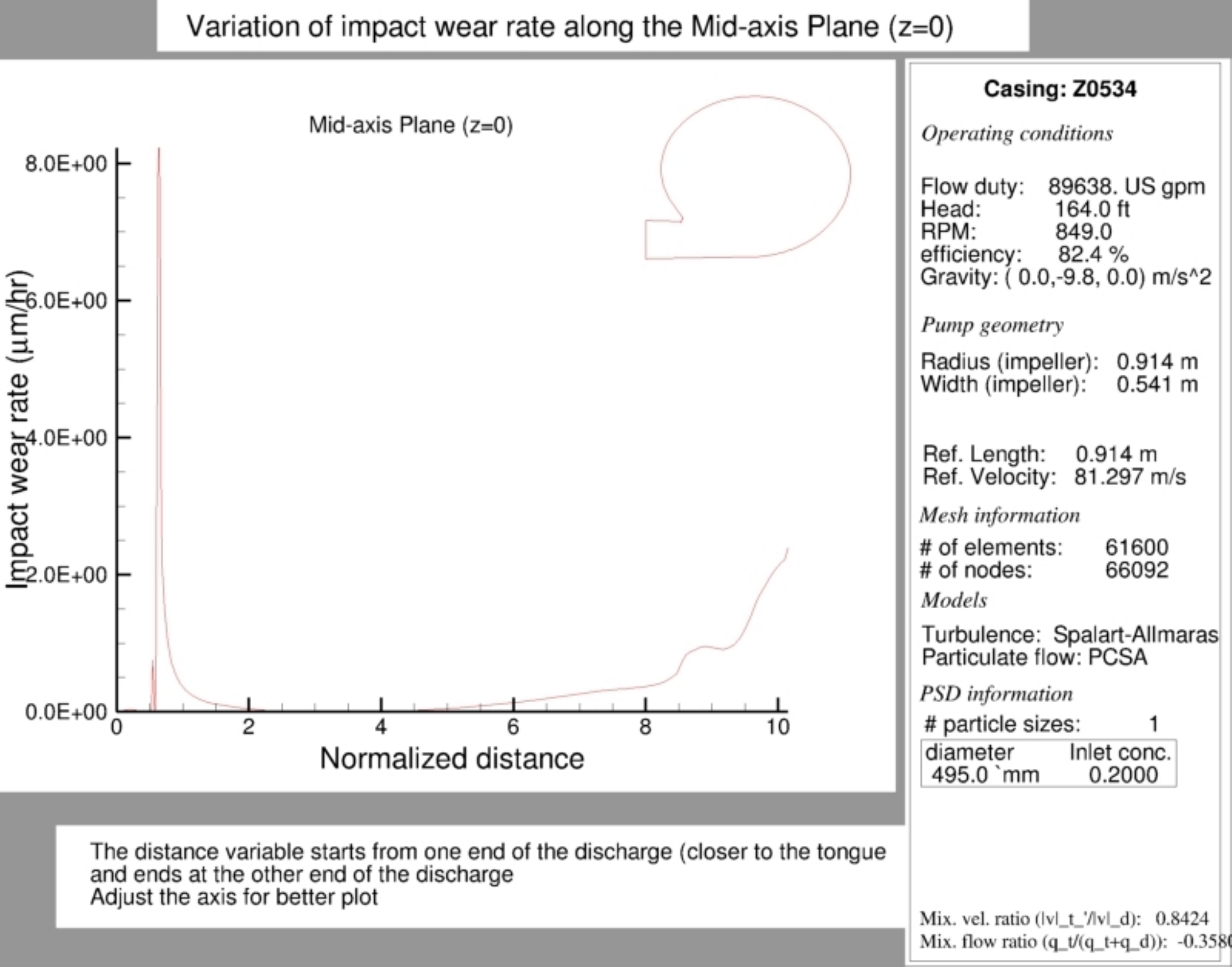}}
\caption{Comparison of impact wear.}
% \label{fig:comparison}
\end{figure}

\begin{figure}[!htbp]
\centering
\subcaptionbox{Original design.
% \label{fig:original}
}
  [.45\linewidth]{\includegraphics[width=0.4001\textwidth, keepaspectratio]{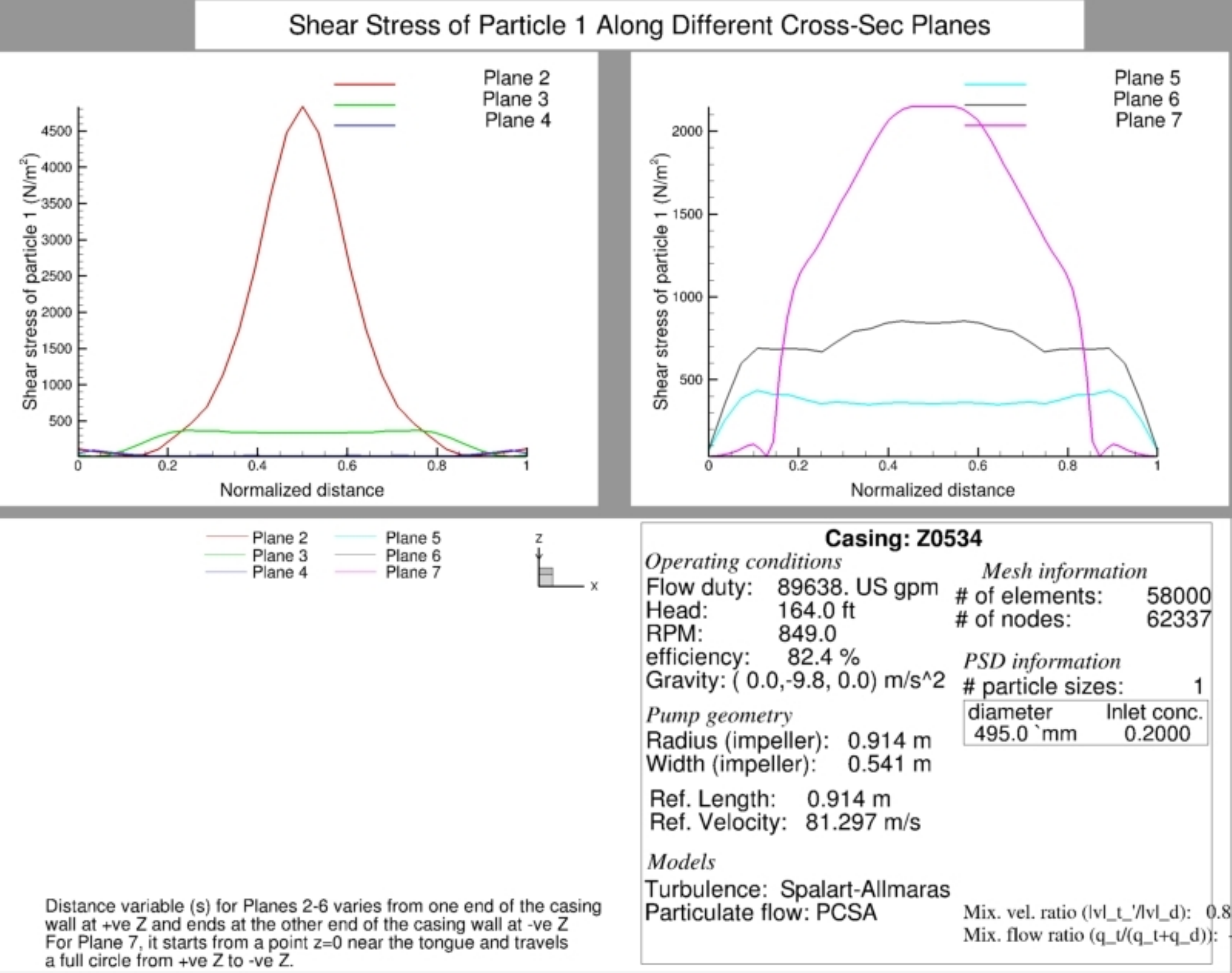}}
\hfill
\subcaptionbox{Optimal design.
% \label{fig:optimal}
}
  [.45\linewidth]{\includegraphics[width=0.4001\textwidth, keepaspectratio]{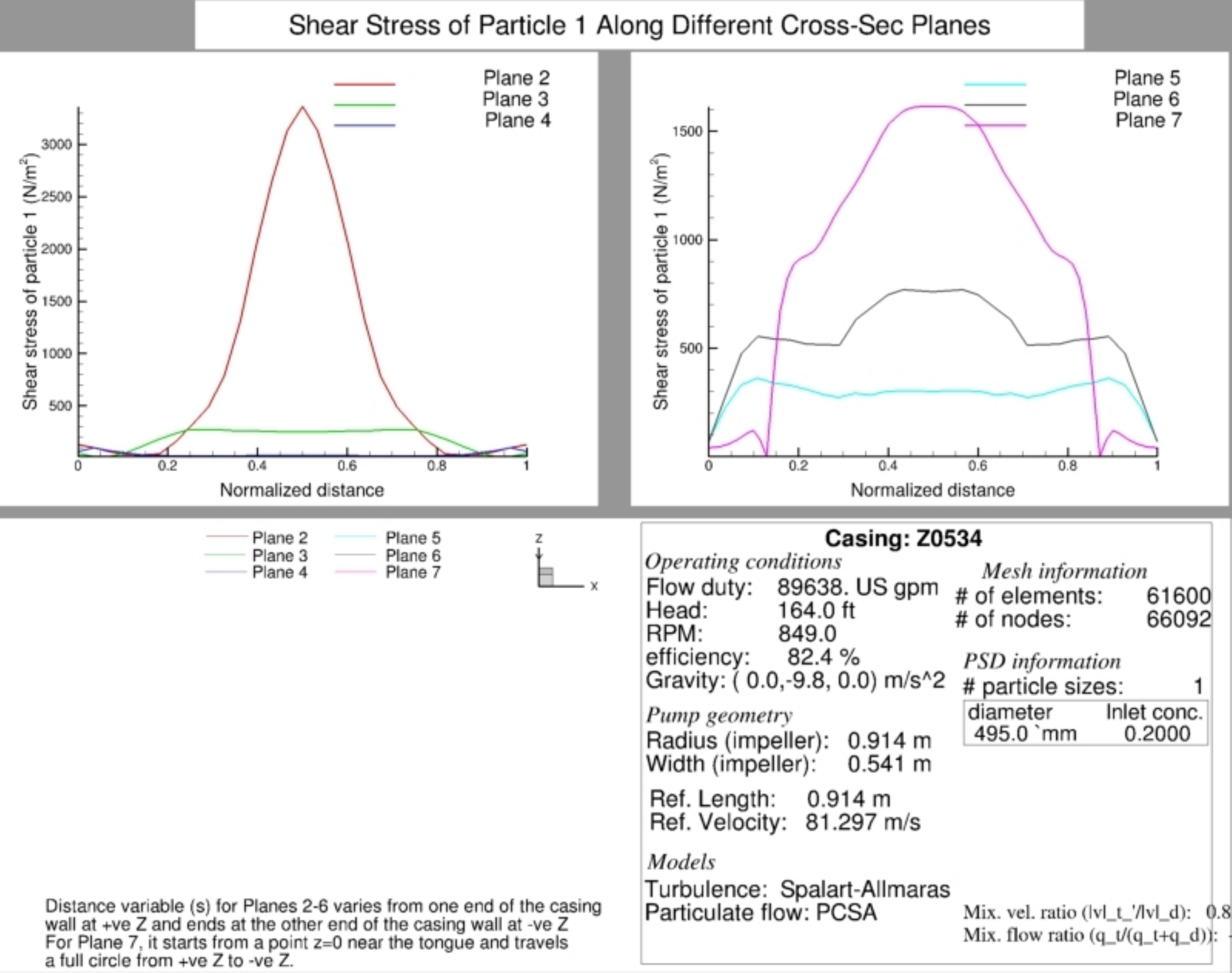}}

\centering
\subcaptionbox{Original design.
% \label{fig:original}
}
  [.45\linewidth]{\includegraphics[width=0.4001\textwidth, keepaspectratio]{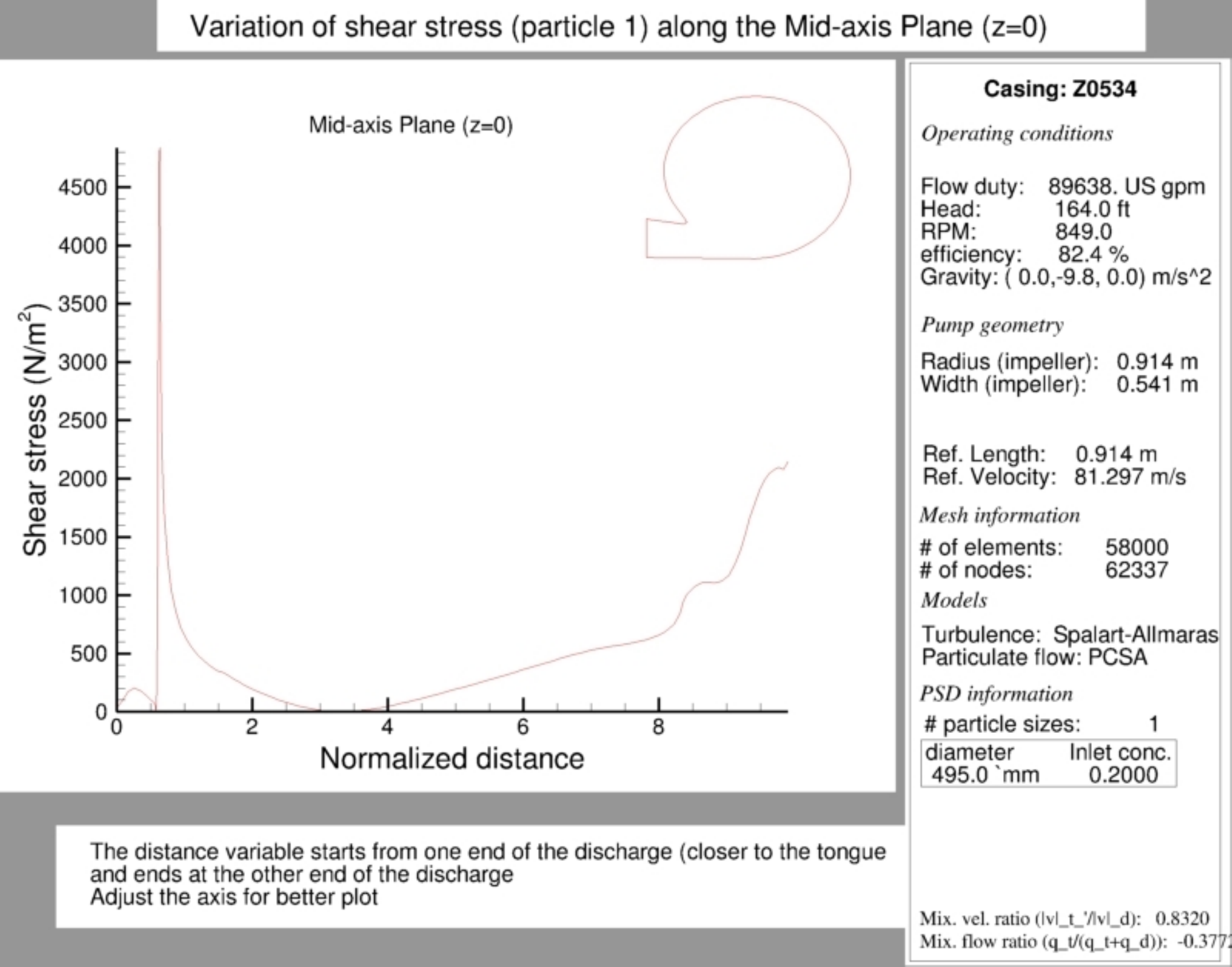}}
\hfill
\subcaptionbox{Optimal design.
% \label{fig:optimal}
}
  [.45\linewidth]{\includegraphics[width=0.4001\textwidth, keepaspectratio]{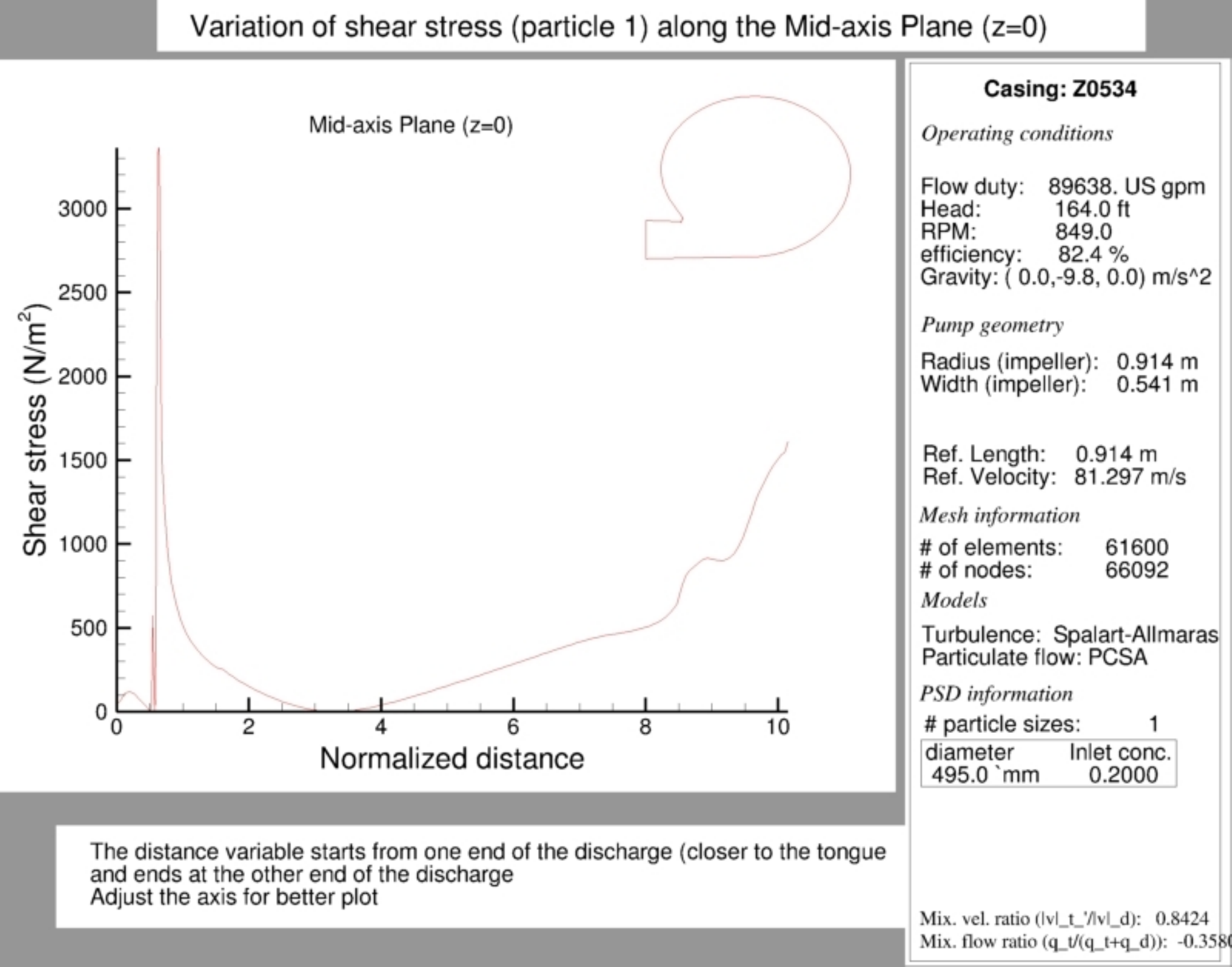}}
\caption{Comparison of shear stress.}
% \label{fig:comparison}
\end{figure}

\begin{figure}[!htbp]
\centering
\subcaptionbox{Original design.
% \label{fig:original}
}
  [.45\linewidth]{\includegraphics[width=0.4001\textwidth, keepaspectratio]{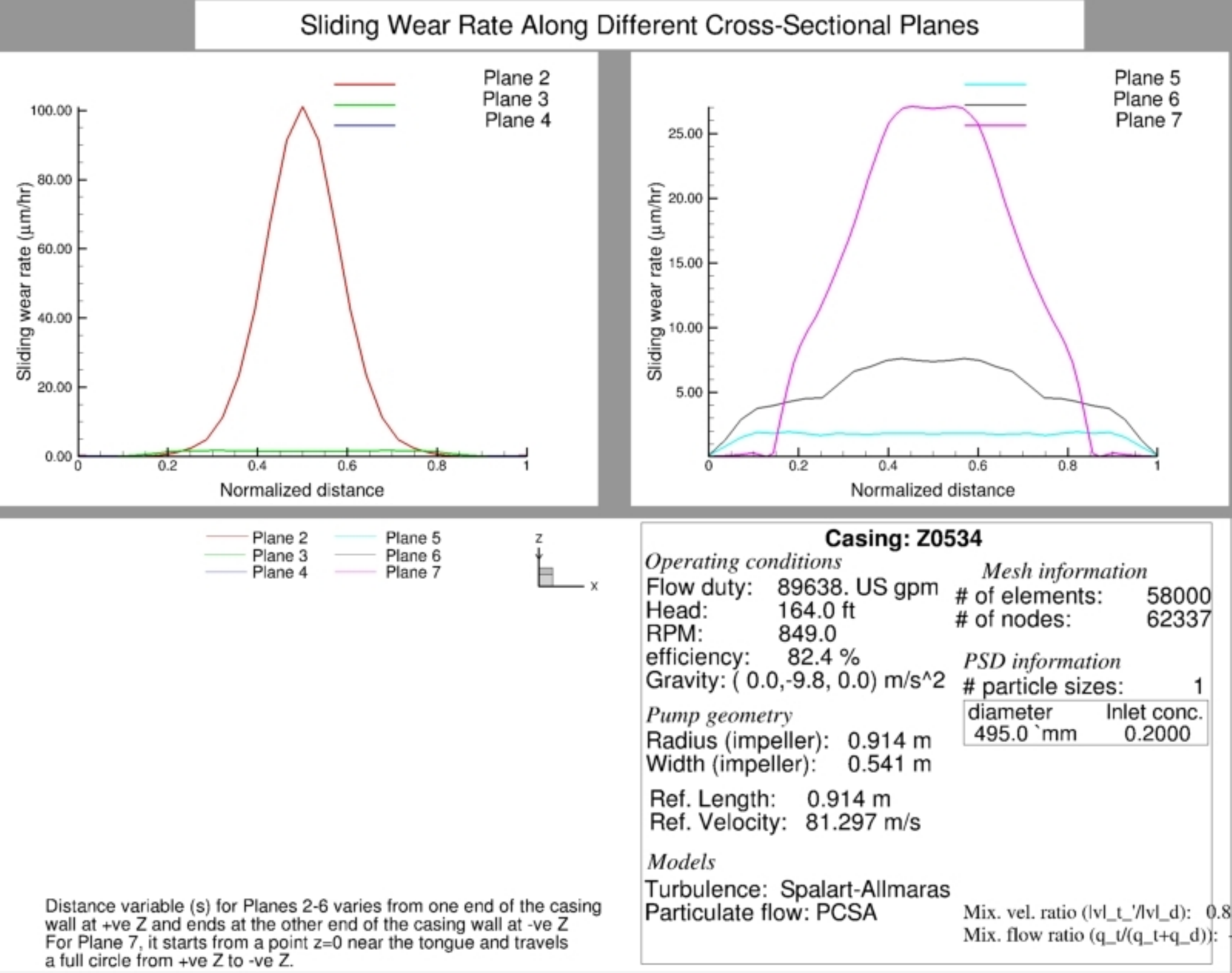}}
\hfill
\subcaptionbox{Optimal design.
% \label{fig:optimal}
}
  [.45\linewidth]{\includegraphics[width=0.4001\textwidth, keepaspectratio]{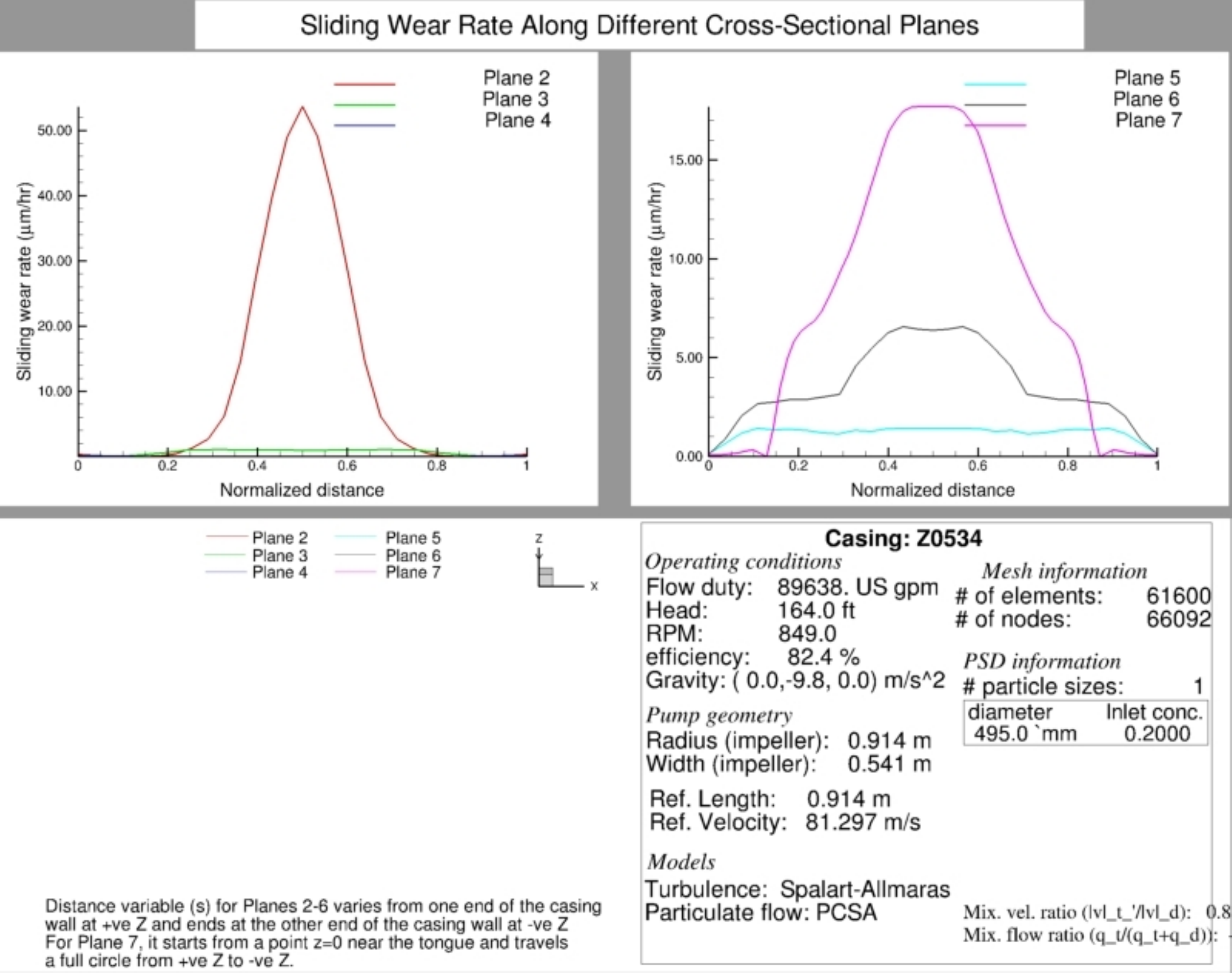}}

\centering
\subcaptionbox{Original design.
% \label{fig:original}
}
  [.45\linewidth]{\includegraphics[width=0.4001\textwidth, keepaspectratio]{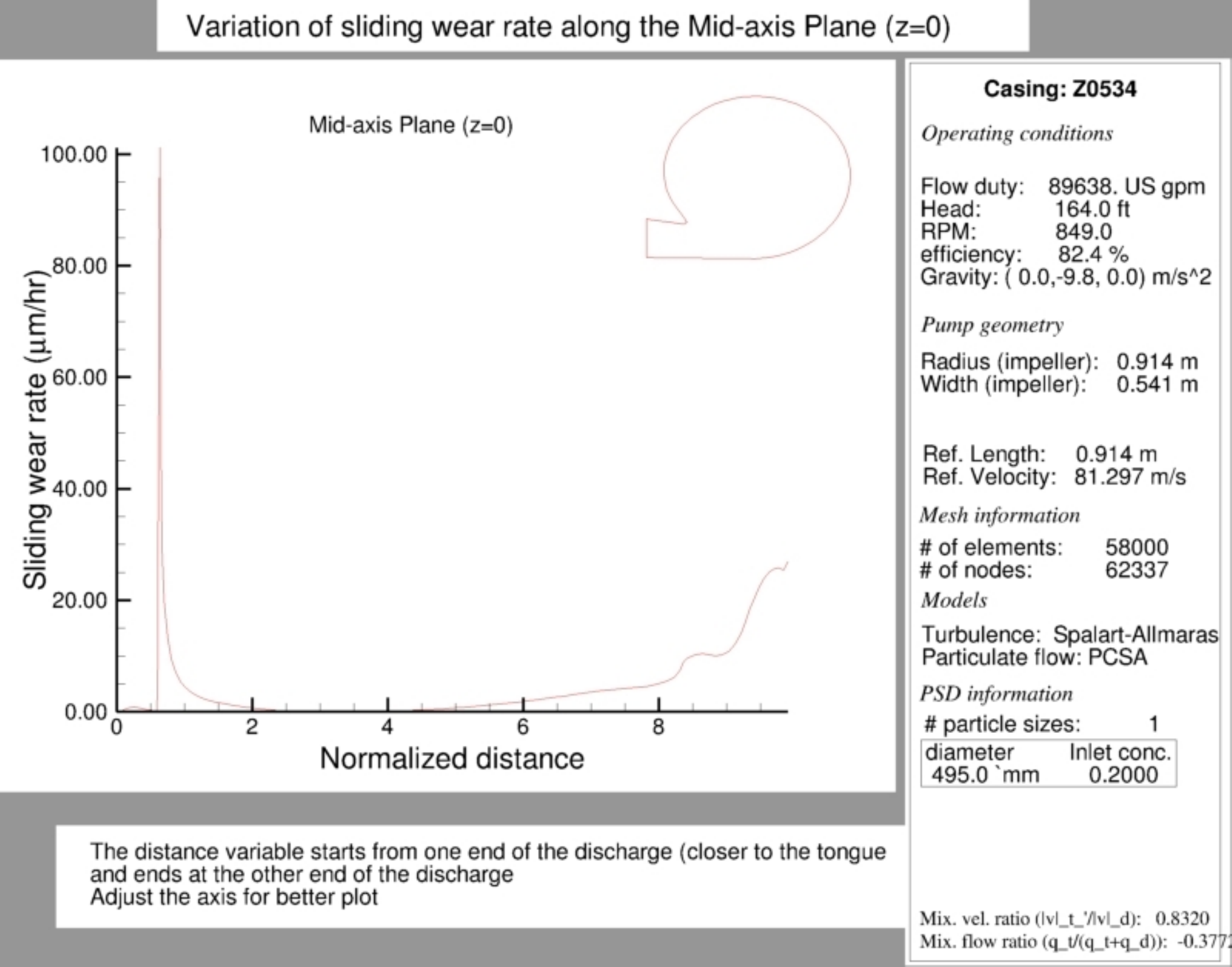}}
\hfill
\subcaptionbox{Optimal design.
% \label{fig:optimal}
}
  [.45\linewidth]{\includegraphics[width=0.4001\textwidth, keepaspectratio]{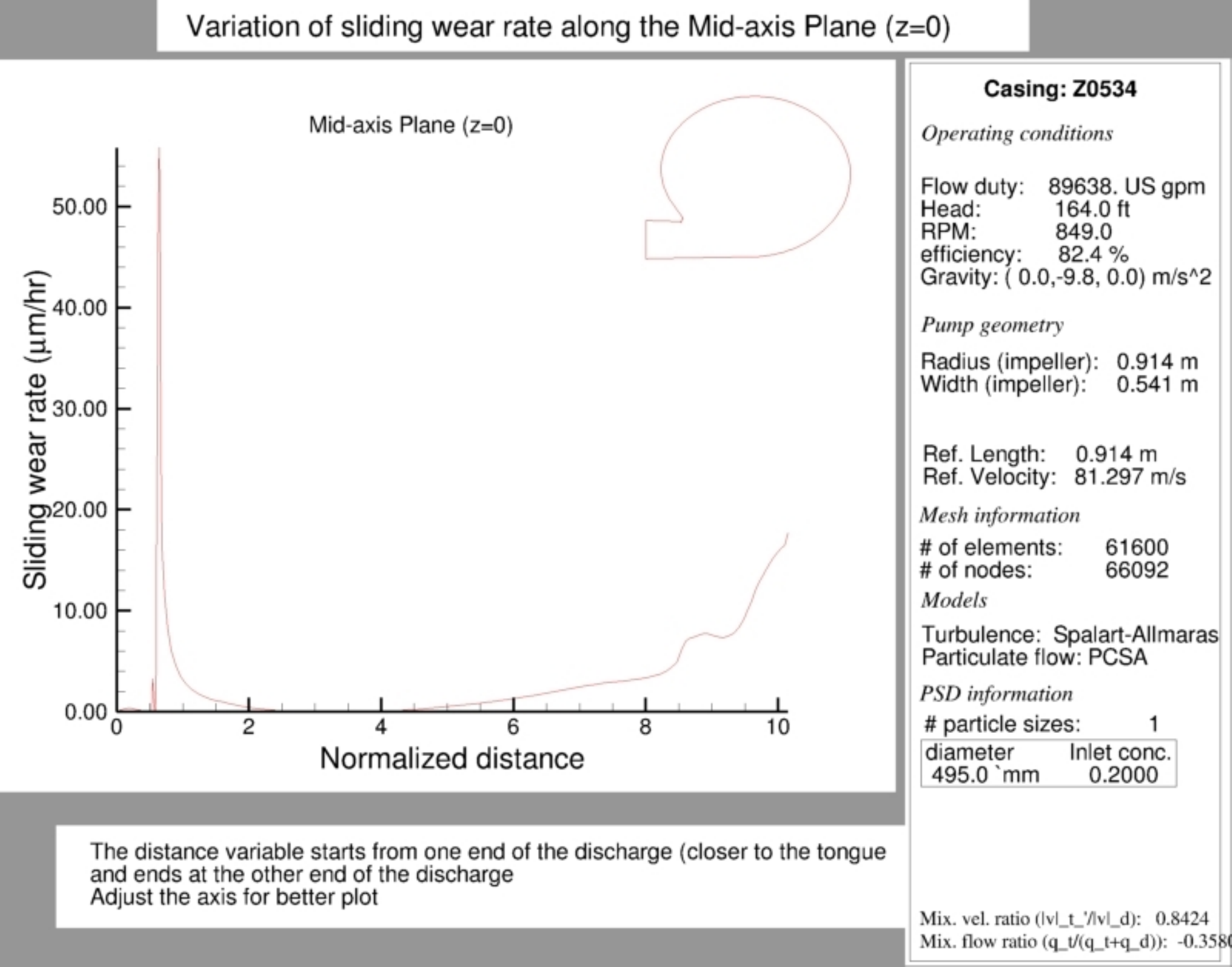}}
\caption{Comparison of sliding wear.}
% \label{fig:comparison}
\end{figure}

\begin{figure}[!htbp]
\centering
\subcaptionbox{Original design.
% \label{fig:original}
}
  [.45\linewidth]{\includegraphics[width=0.4001\textwidth, keepaspectratio]{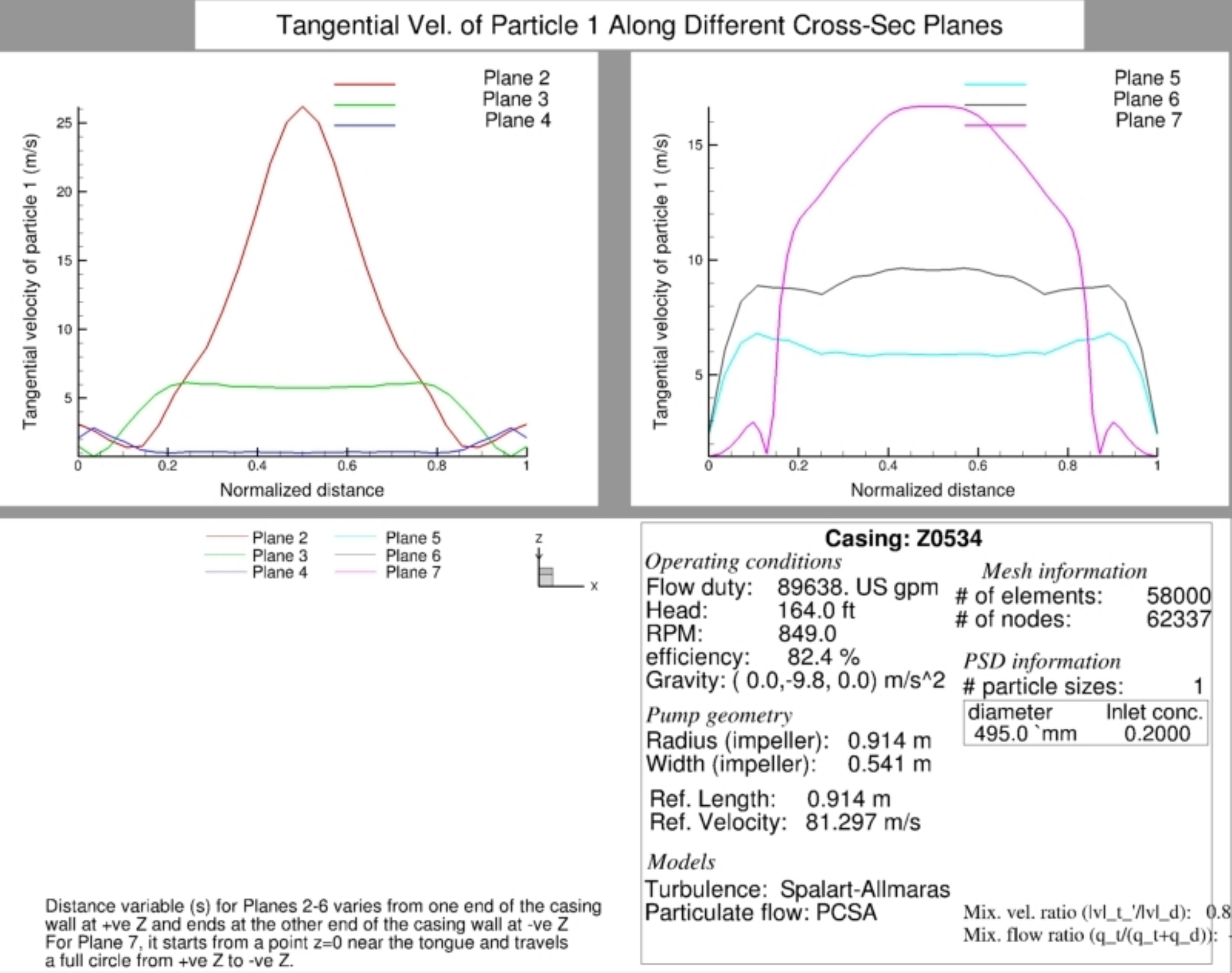}}
\hfill
\subcaptionbox{Optimal design.
% \label{fig:optimal}
}
  [.45\linewidth]{\includegraphics[width=0.4001\textwidth, keepaspectratio]{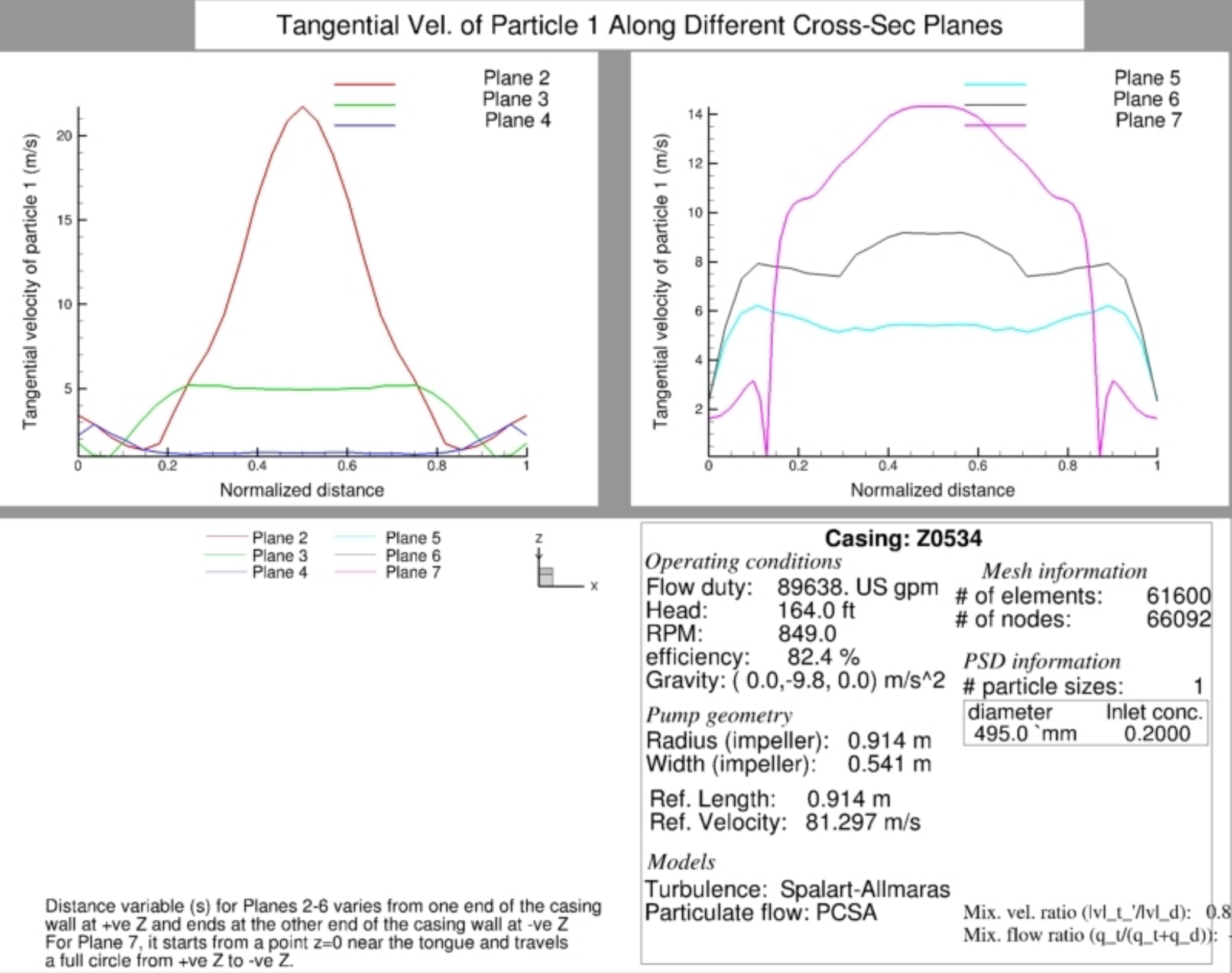}}

\centering
\subcaptionbox{Original design.
% \label{fig:original}
}
  [.45\linewidth]{\includegraphics[width=0.4001\textwidth, keepaspectratio]{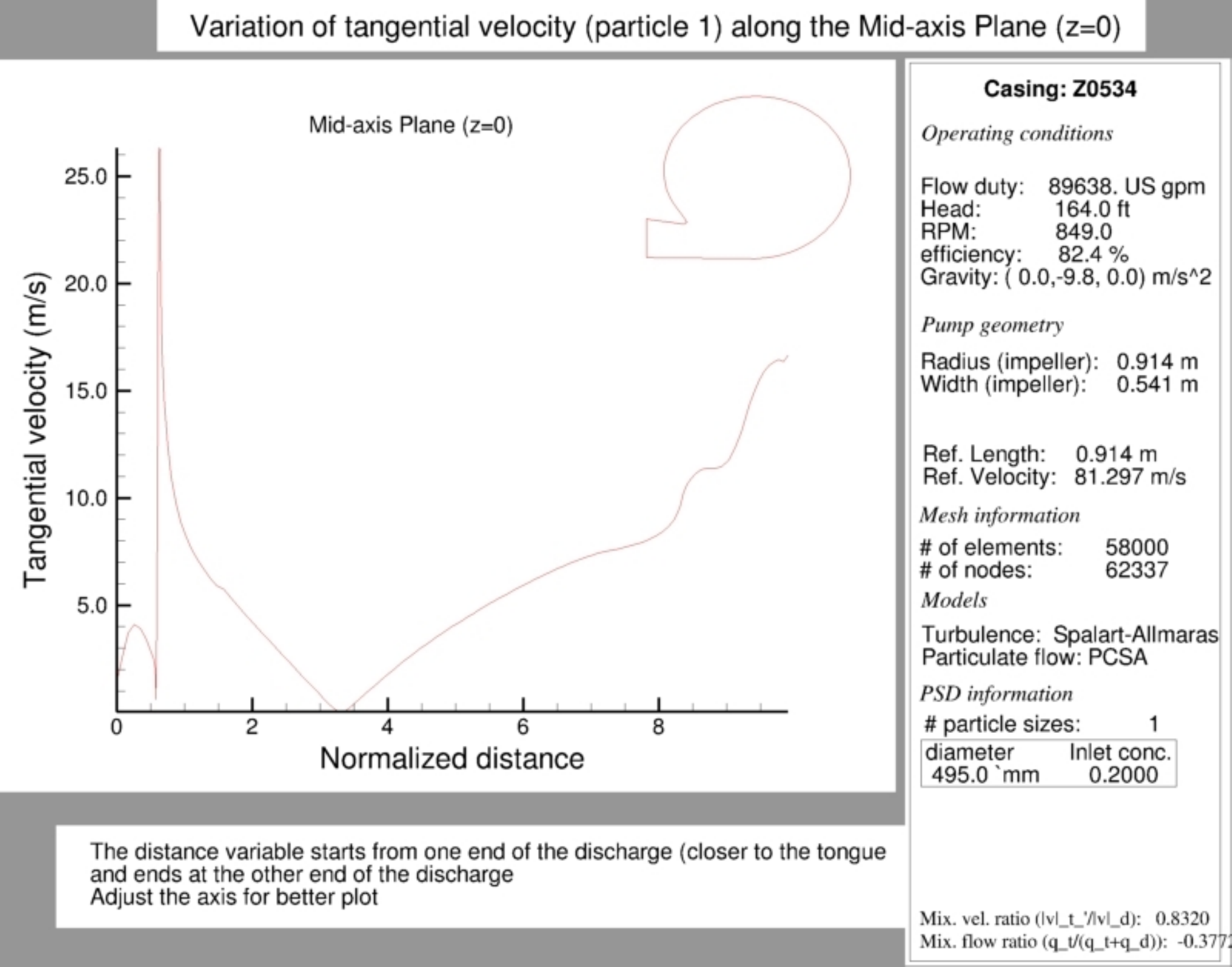}}
\hfill
\subcaptionbox{Optimal design.
% \label{fig:optimal}
}
  [.45\linewidth]{\includegraphics[width=0.4001\textwidth, keepaspectratio]{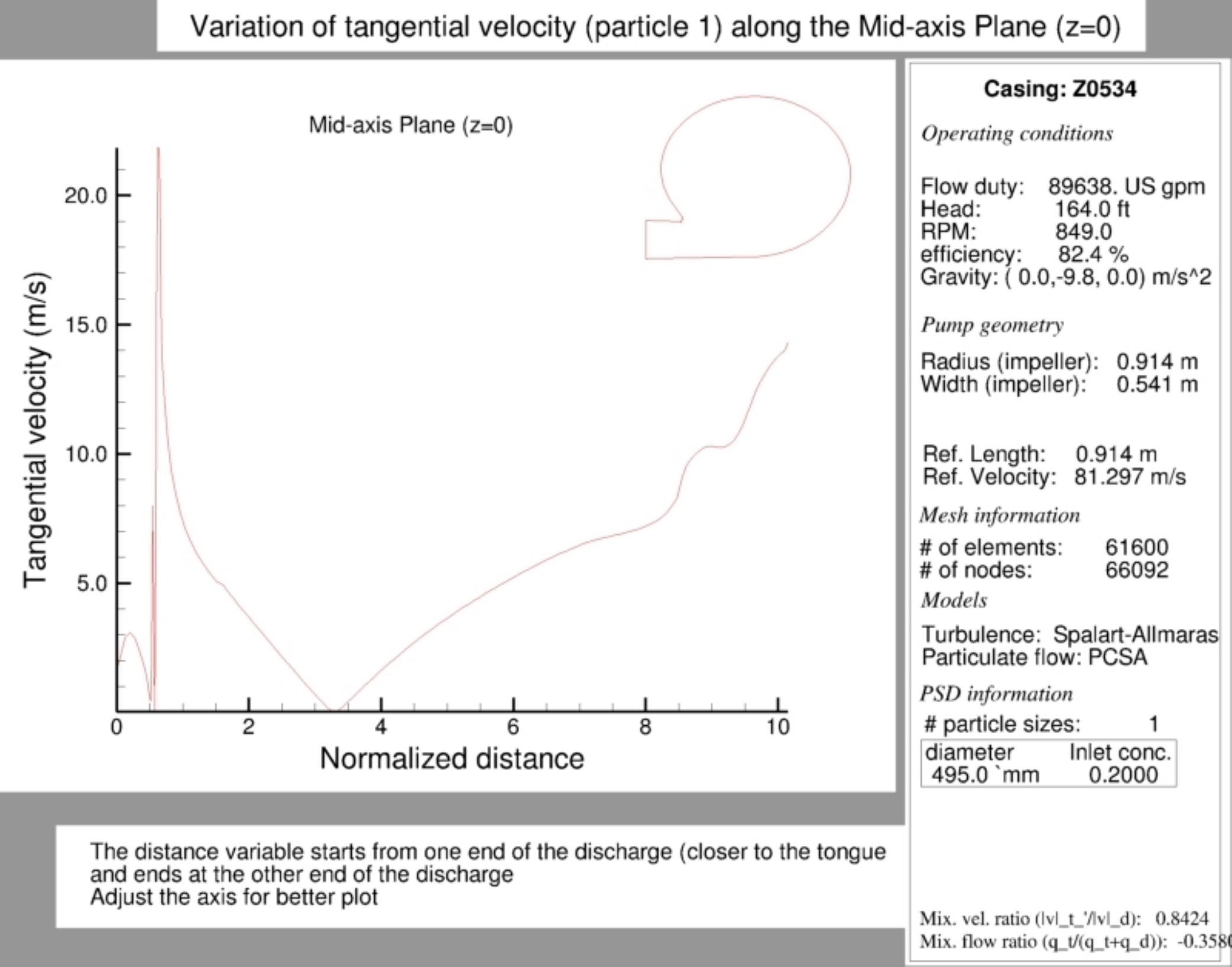}}
\caption{Comparison of tangential velocity.}
% \label{fig:comparison}
\end{figure}

\begin{figure}[!htbp]
\centering
\subcaptionbox{Original design.
% \label{fig:original}
}
  [.45\linewidth]{\includegraphics[width=0.4001\textwidth, keepaspectratio]{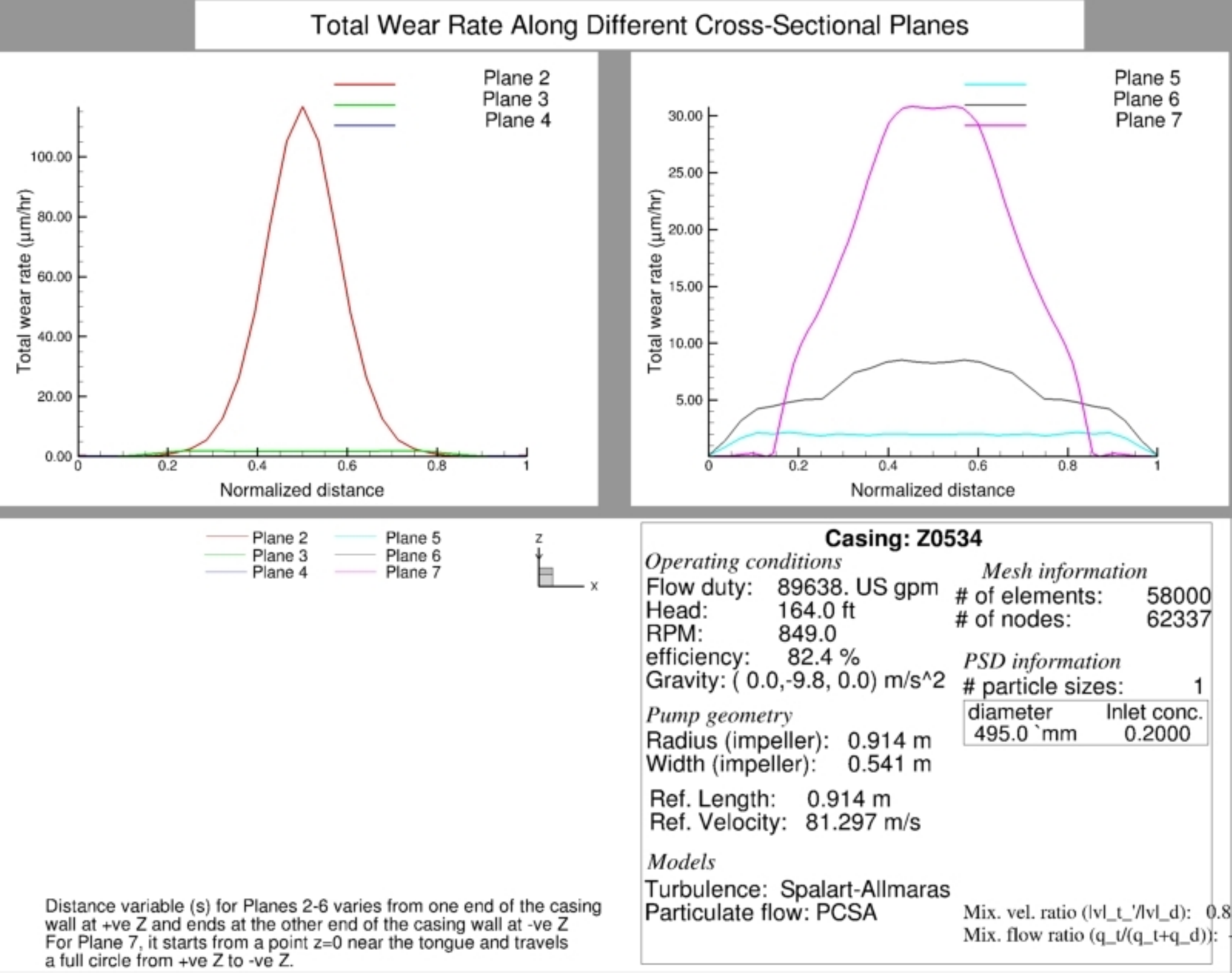}}
\hfill
\subcaptionbox{Optimal design.
% \label{fig:optimal}
}
  [.45\linewidth]{\includegraphics[width=0.4001\textwidth, keepaspectratio]{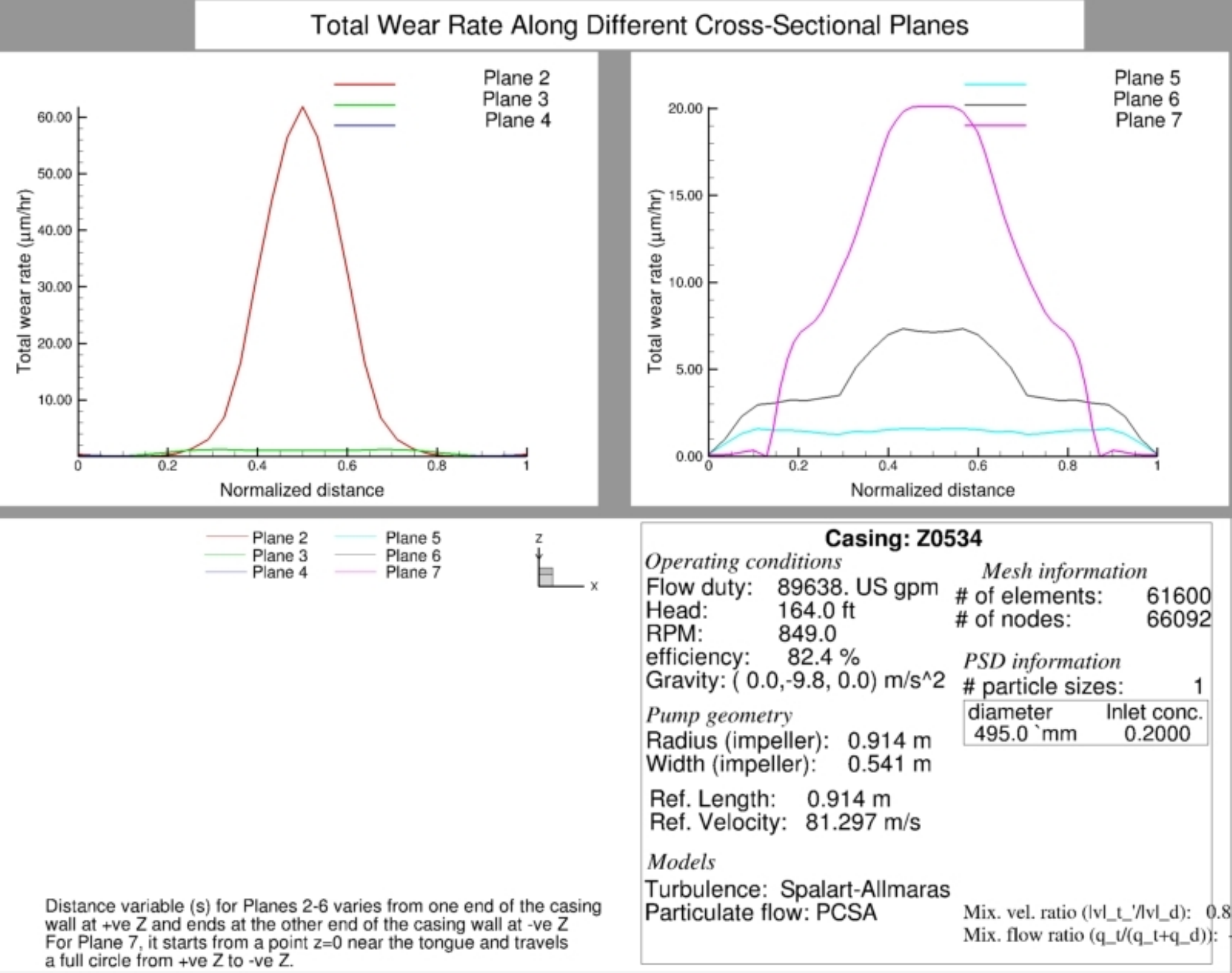}}

\centering
\subcaptionbox{Original design.
% \label{fig:original}
}
  [.45\linewidth]{\includegraphics[width=0.4001\textwidth, keepaspectratio]{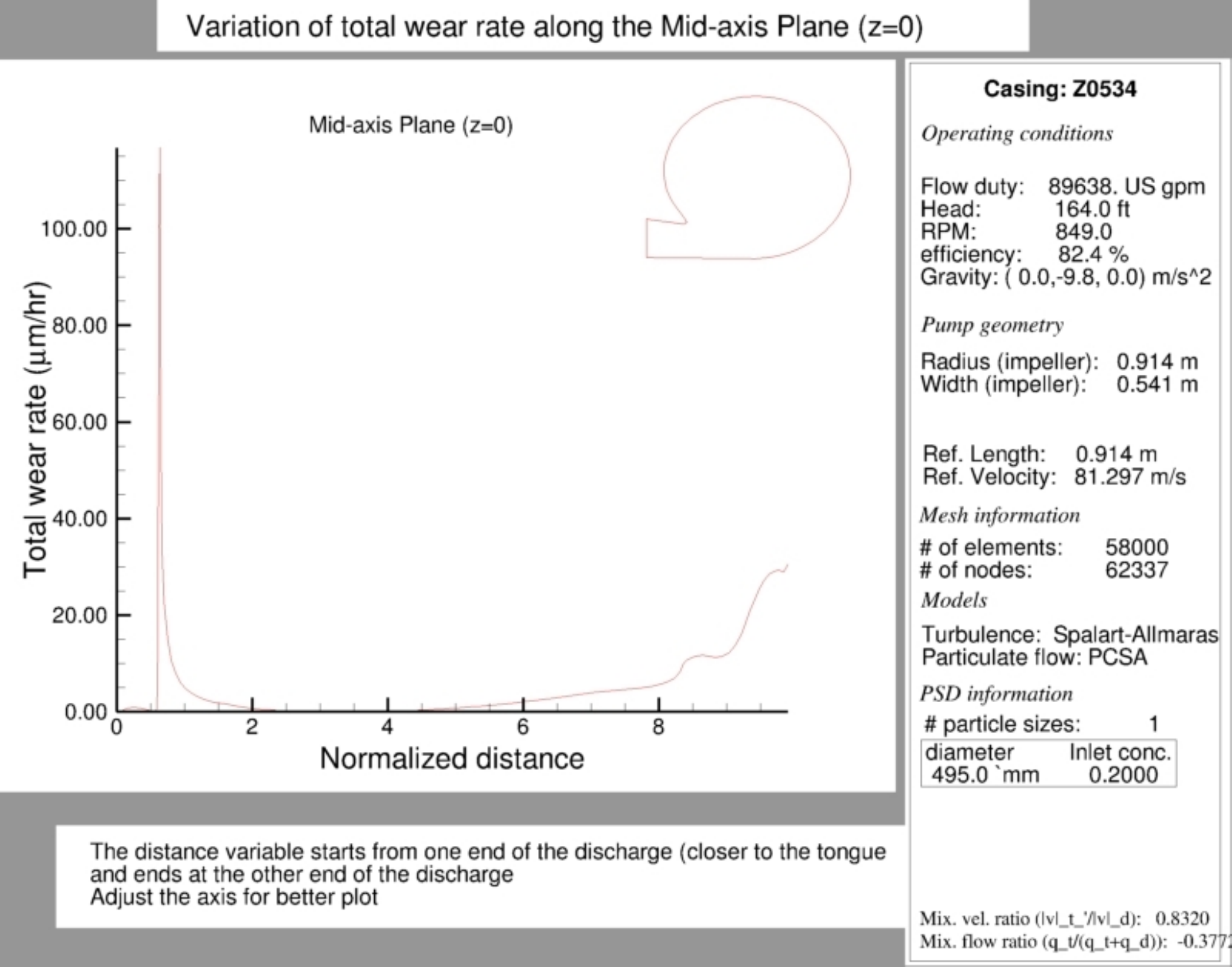}}
\hfill
\subcaptionbox{Optimal design.
% \label{fig:optimal}
}
  [.45\linewidth]{\includegraphics[width=0.4001\textwidth, keepaspectratio]{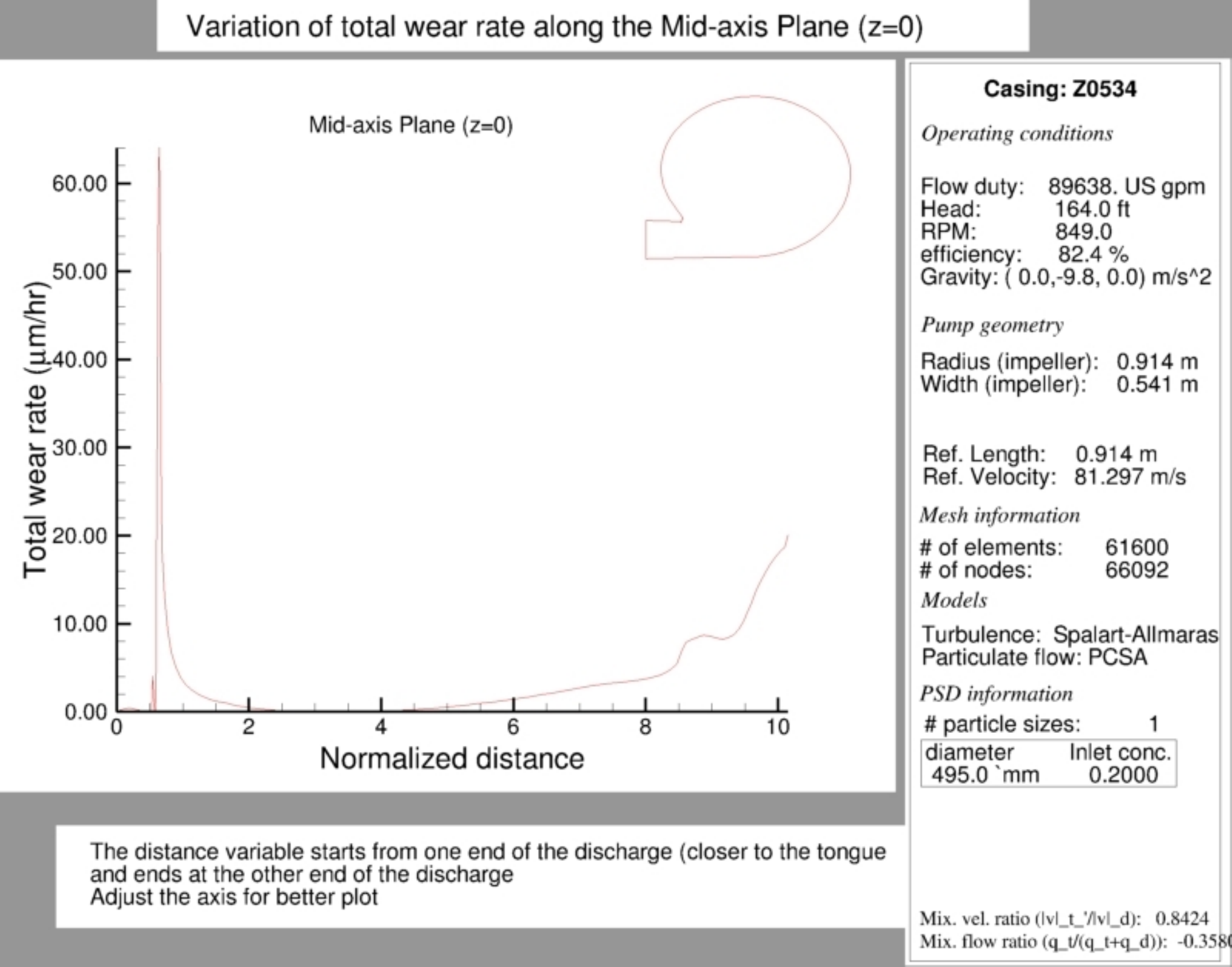}}
\caption{Comparison of total wear.}
% \label{fig:comparison}
\end{figure}

\begin{figure}[!htbp]
\centering
\subcaptionbox{Original design.
% \label{fig:original}
}
  [.45\linewidth]{\includegraphics[width=0.4001\textwidth, keepaspectratio]{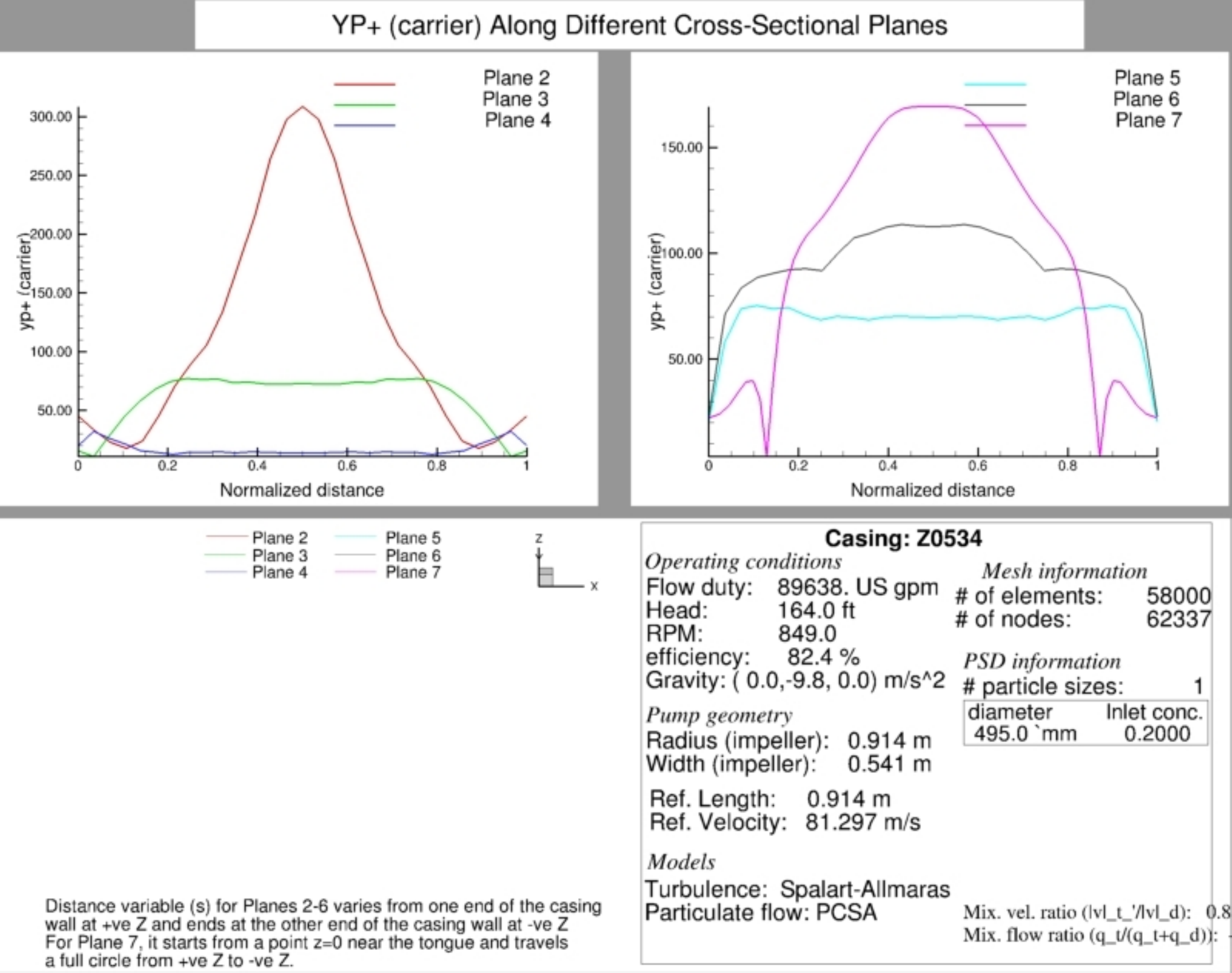}}
\hfill
\subcaptionbox{Optimal design.
% \label{fig:optimal}
}
  [.45\linewidth]{\includegraphics[width=0.4001\textwidth, keepaspectratio]{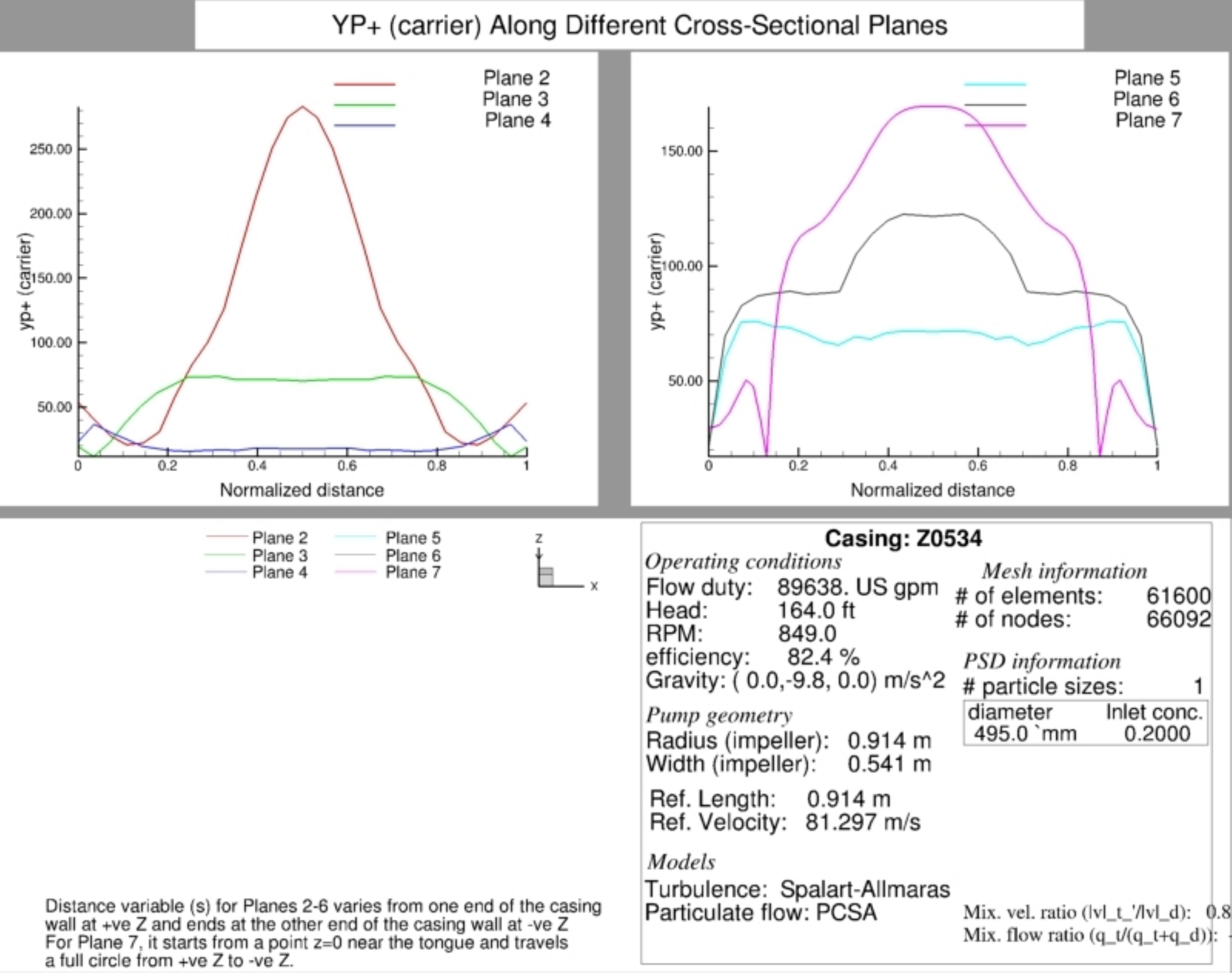}}

\centering
\subcaptionbox{Original design.
% \label{fig:original}
}
  [.45\linewidth]{\includegraphics[width=0.4001\textwidth, keepaspectratio]{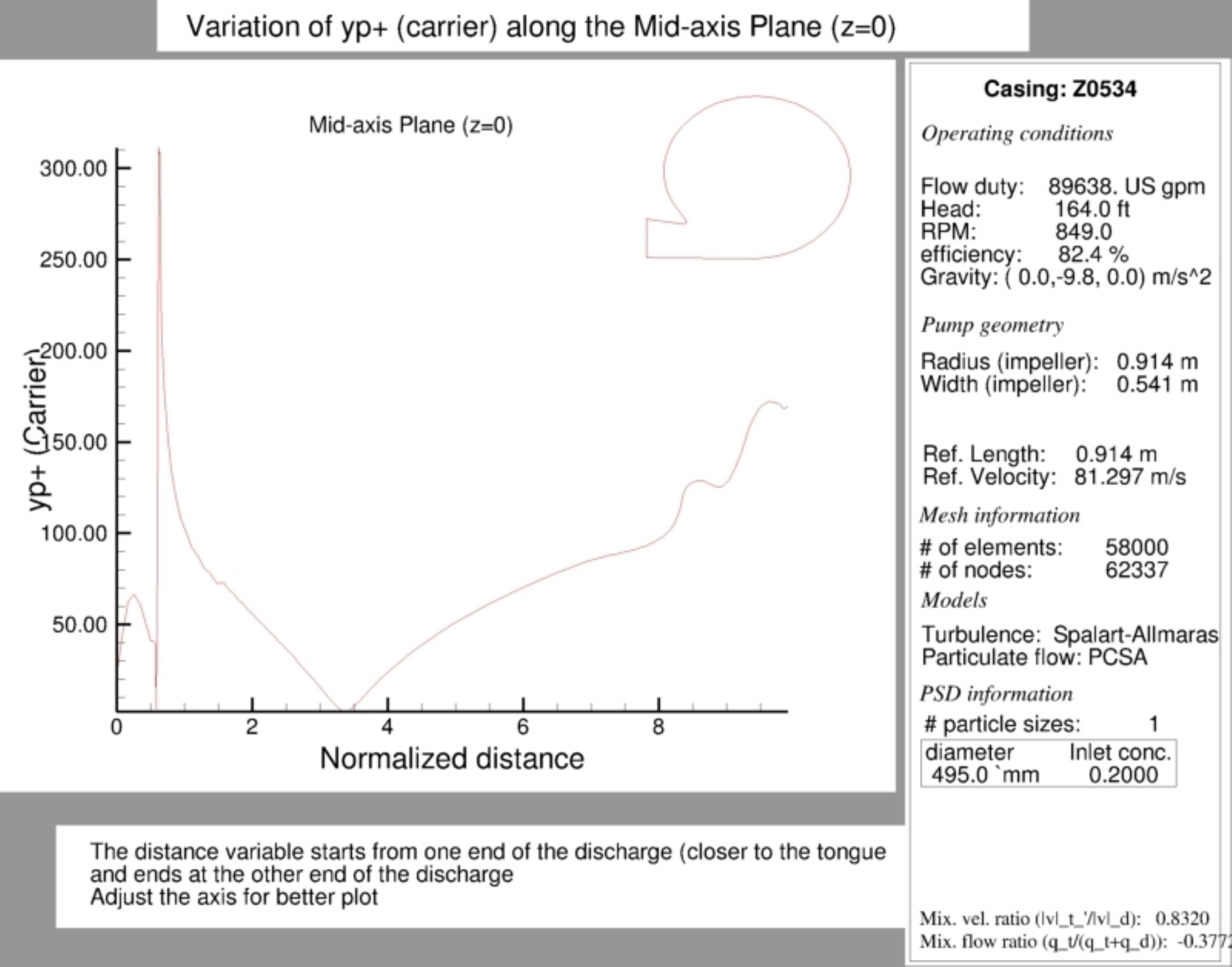}}
\hfill
\subcaptionbox{Optimal design.
% \label{fig:optimal}
}
  [.45\linewidth]{\includegraphics[width=0.4001\textwidth, keepaspectratio]{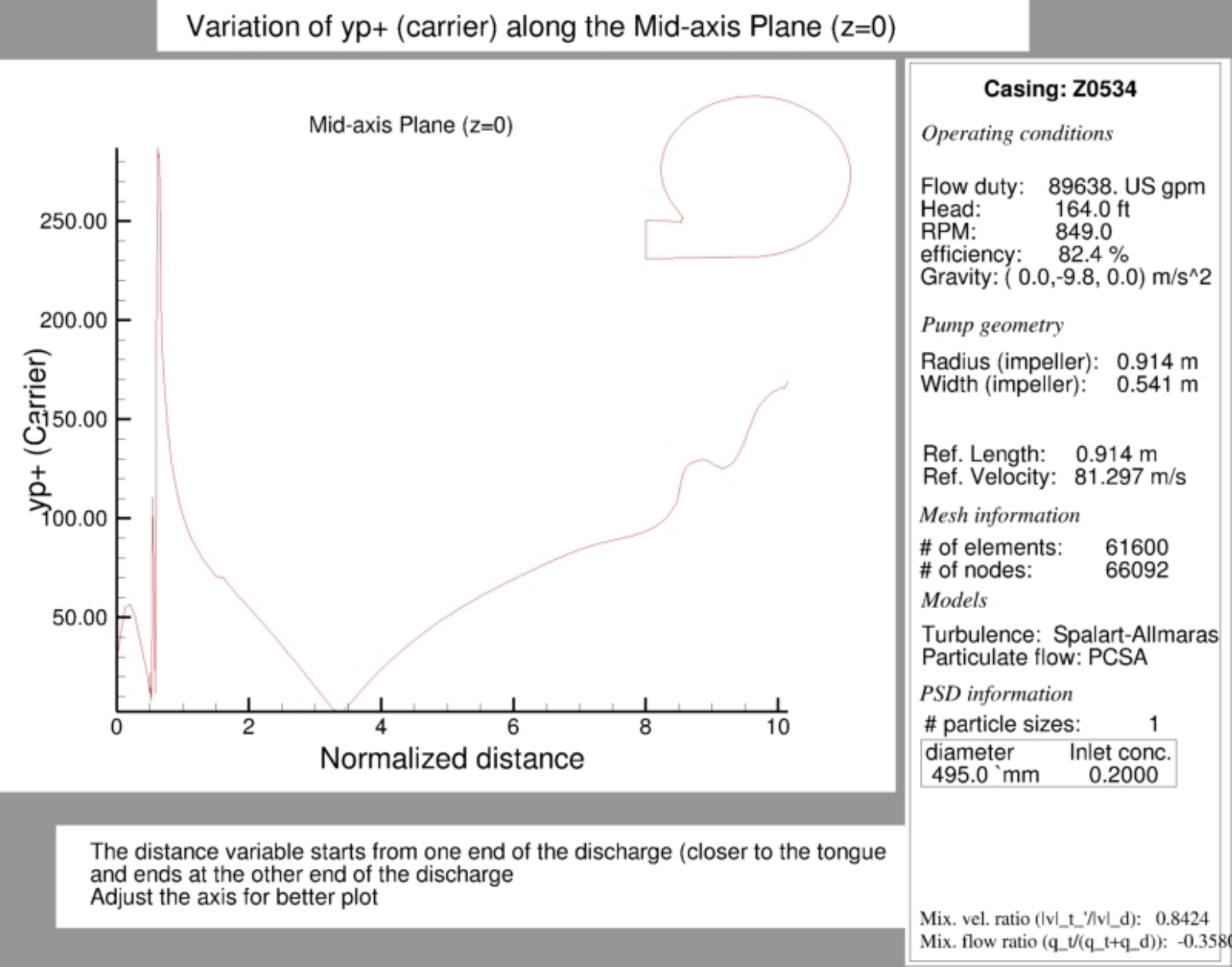}}

\centering
\subcaptionbox{Original design.
% \label{fig:original}
}
  [.45\linewidth]{\includegraphics[width=0.4001\textwidth, keepaspectratio]{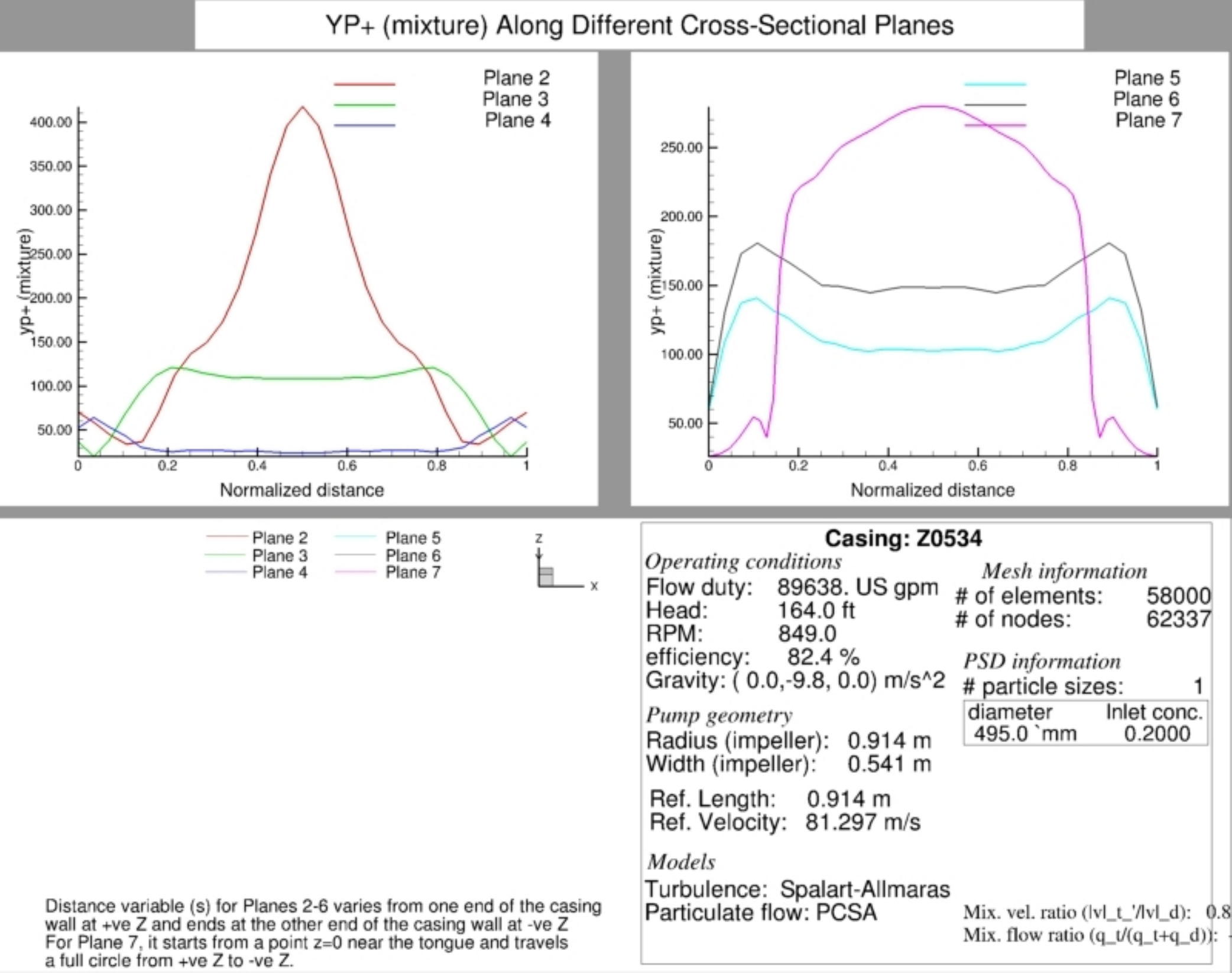}}
\hfill
\subcaptionbox{Optimal design.
% \label{fig:optimal}
}
  [.45\linewidth]{\includegraphics[width=0.4001\textwidth, keepaspectratio]{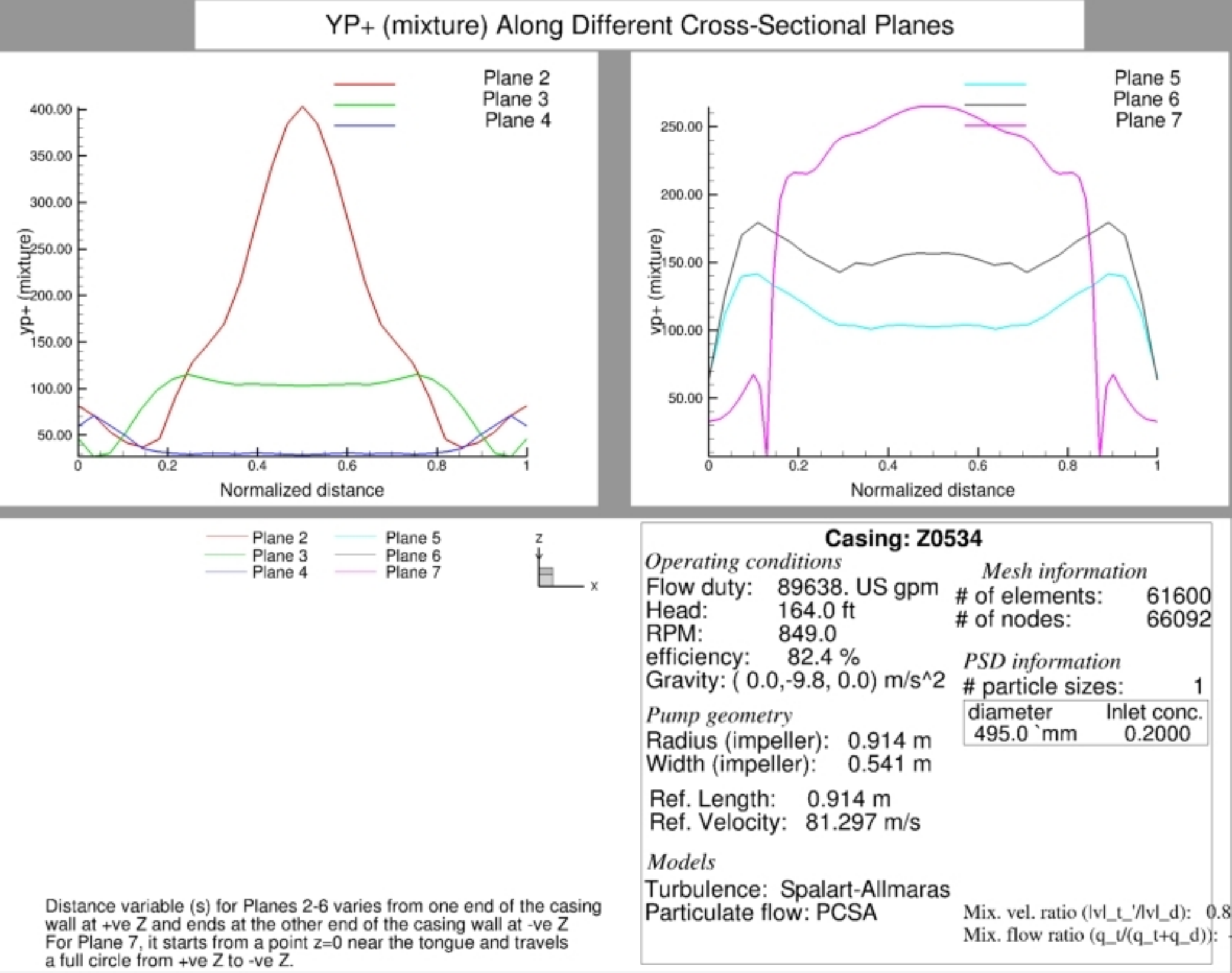}}
\caption{Comparison of $y^+$ value.}
% \label{fig:comparison}
\end{figure}

\begin{figure}[!htbp]
\centering
\subcaptionbox{Original design.
% \label{fig:original}
}
  [.45\linewidth]{\includegraphics[width=0.4001\textwidth, keepaspectratio]{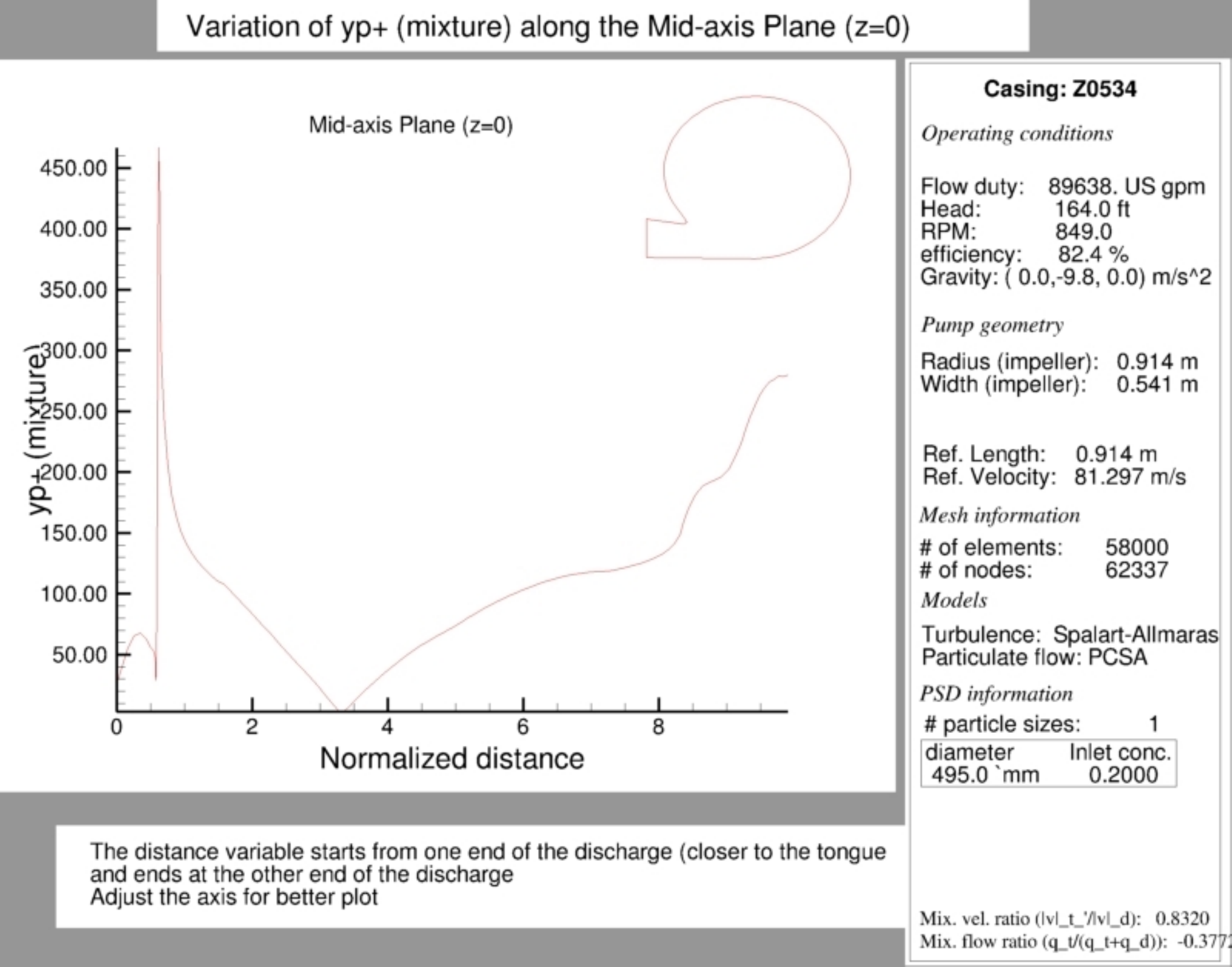}}
\hfill
\subcaptionbox{Optimal design.
% \label{fig:optimal}
}
  [.45\linewidth]{\includegraphics[width=0.4001\textwidth, keepaspectratio]{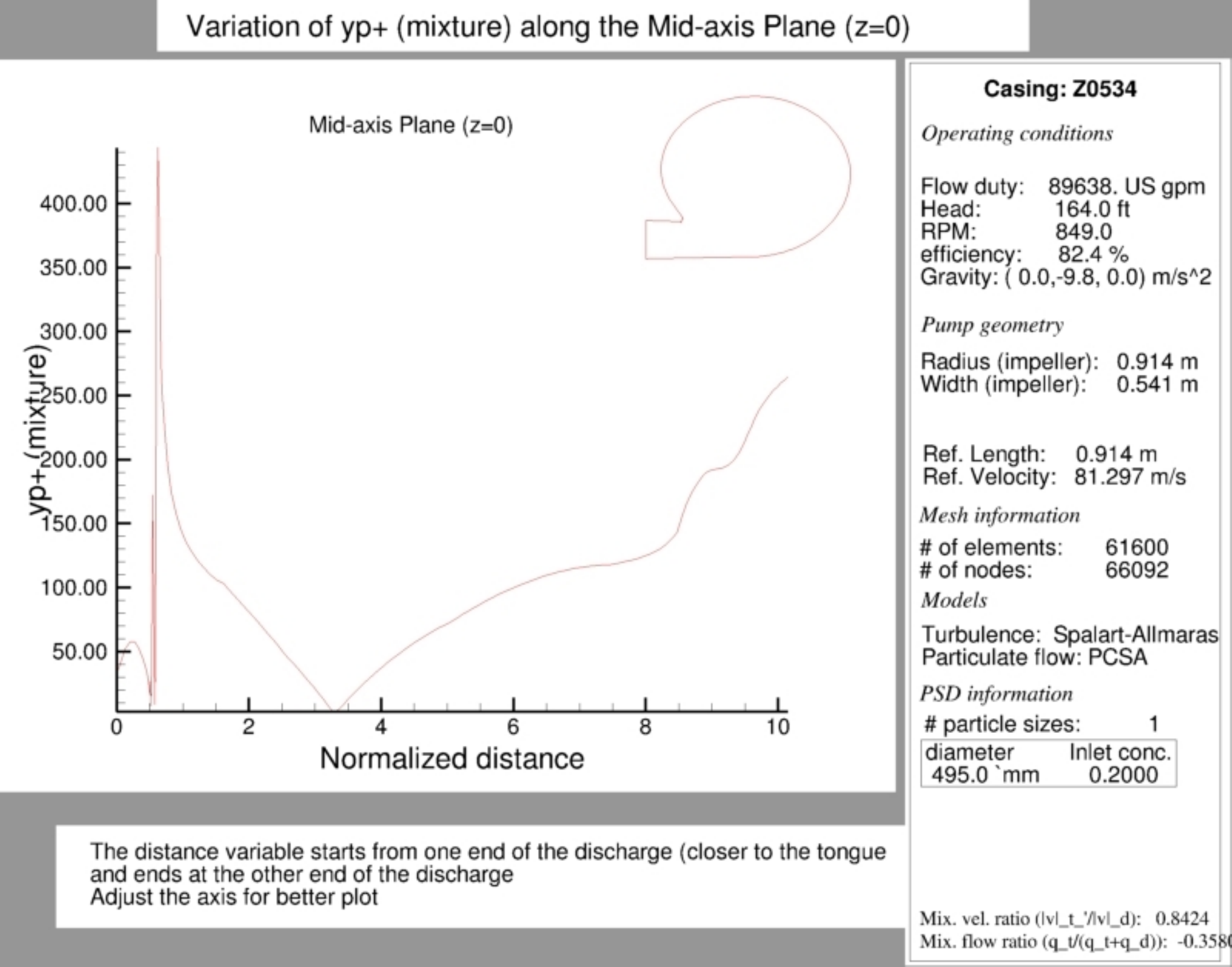}}

\centering
\subcaptionbox{Original design.
% \label{fig:original}
}
  [.45\linewidth]{\includegraphics[width=0.4001\textwidth, keepaspectratio]{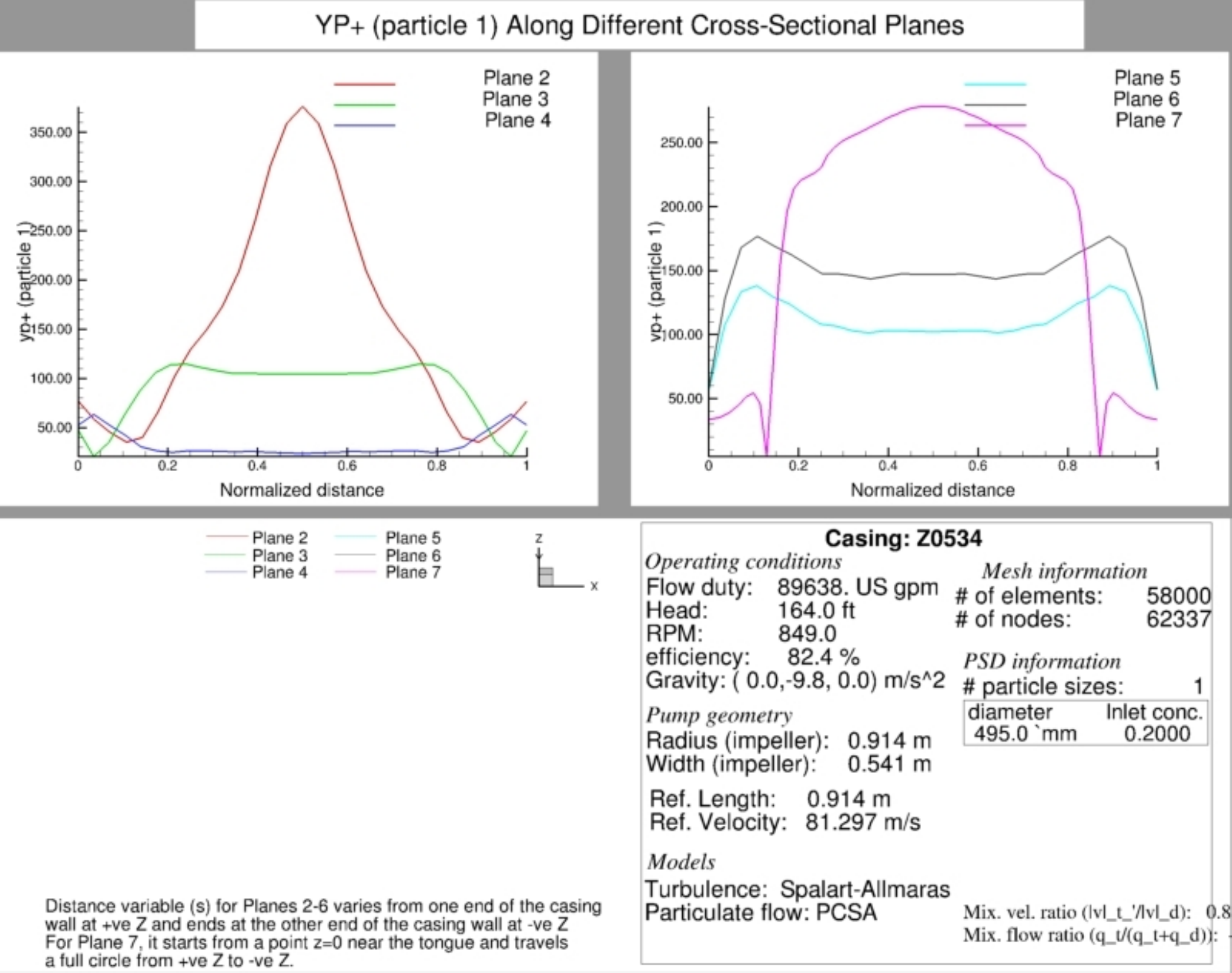}}
\hfill
\subcaptionbox{Optimal design.
% \label{fig:optimal}
}
  [.45\linewidth]{\includegraphics[width=0.4001\textwidth, keepaspectratio]{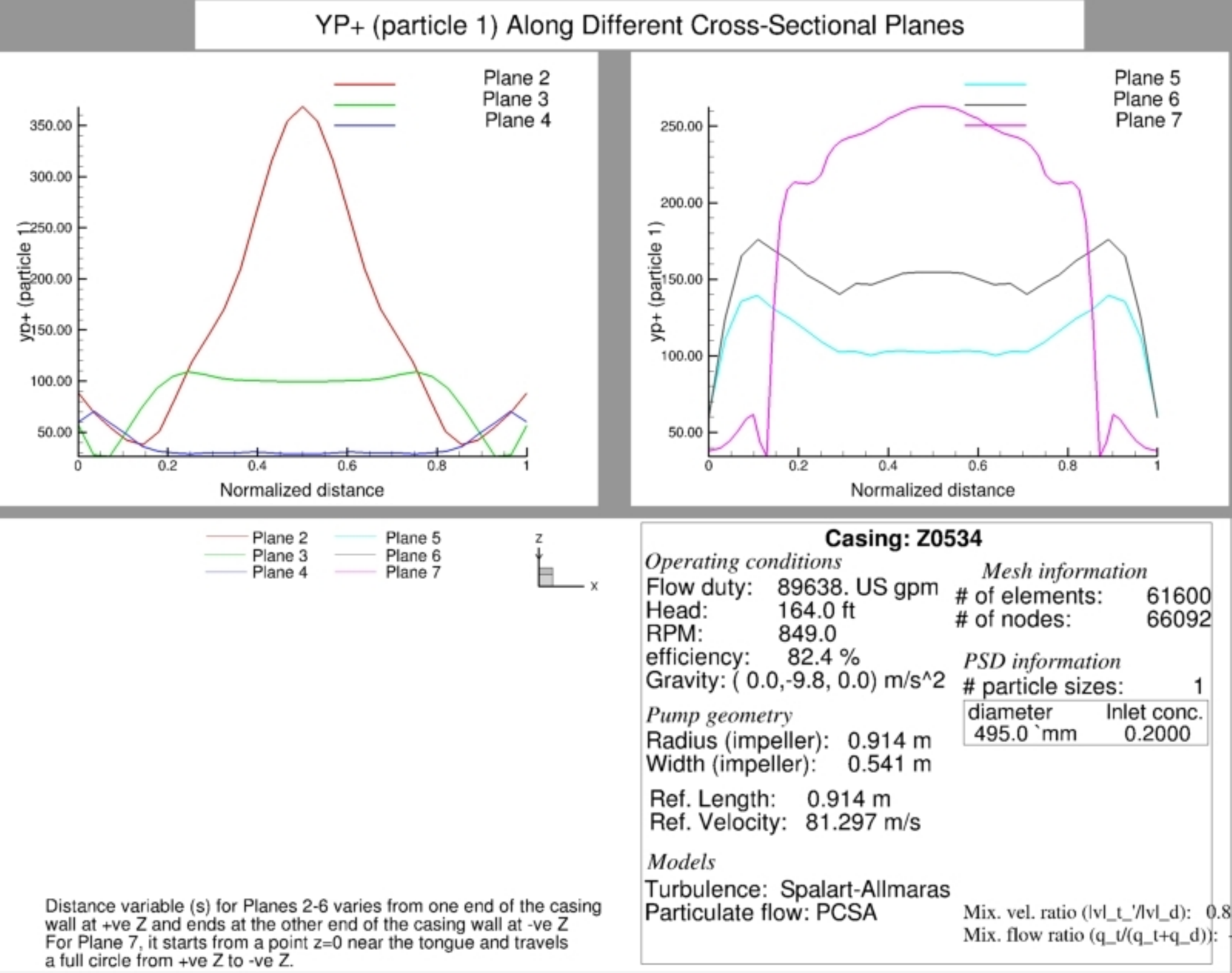}}

\centering
\subcaptionbox{Original design.
% \label{fig:original}
}
  [.45\linewidth]{\includegraphics[width=0.4001\textwidth, keepaspectratio]{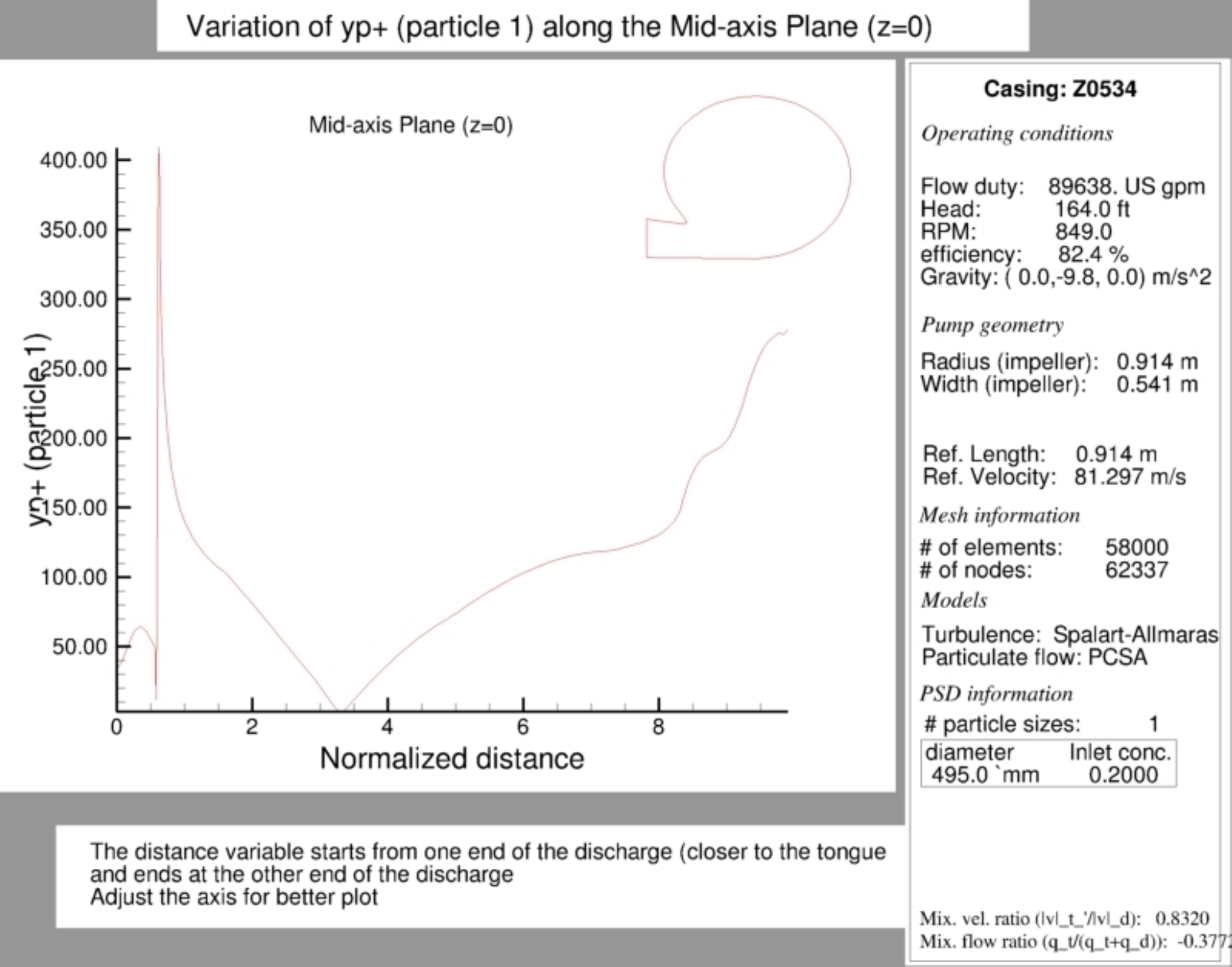}}
\hfill
\subcaptionbox{Optimal design.
% \label{fig:optimal}
}
  [.45\linewidth]{\includegraphics[width=0.4001\textwidth, keepaspectratio]{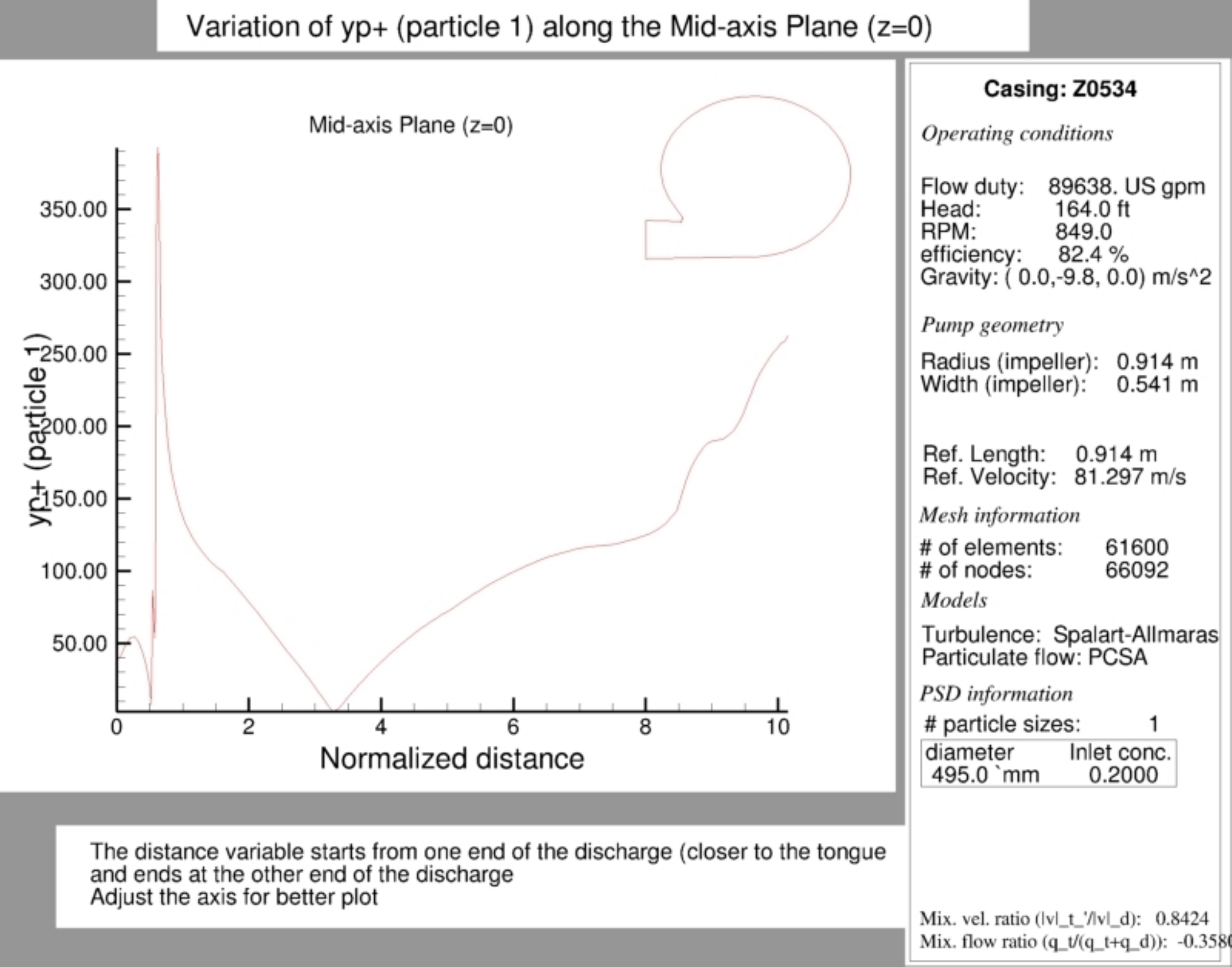}}
\caption{Comparison of $y^+$ value.}
% \label{fig:comparison}
\end{figure}

\end{document}